# Generalizing, Decoding, and Optimizing Support Vector Machine Classification

von Dipl. Math. Mario Michael Krell

Dissertation

zur Erlangung des Grades eines Doktors der
Naturwissenschaften
- Dr. rer. nat. -

*Dedicated to my parents*

**Abstract**

The classification of complex data usually requires the composition of processing steps. Here, a major challenge is the selection of optimal algorithms for preprocessing and classification (including parameterizations). Nowadays, parts of the optimization process are automized but expert knowledge and manual work are still required. We present three steps to face this process and ease the optimization. Namely, we take a theoretical view on classical classifiers, provide an approach to interpret the classifier together with the preprocessing, and integrate both into one framework which enables a semiautomatic optimization of the processing chain and which interfaces numerous algorithms.

First, we summarize the connections between support vector machine (SVM) variants and introduce a *generalized* model which shows that these variants are not to be taken separately but that they are highly connected. Due to the more general connection concepts, several further variants of the SVM can be generated including unary and online classifiers. The model improves the understanding of relationships between the variants. It can be used to improve teaching and to facilitate the choice and implementation of the classifiers. Often, knowledge about and implementations of one classifier can be transferred to the variants. Furthermore, the connections also reveal possible problems when applying some variants. So in certain situation, some variants should not be used or the preprocessing needs to prepare the data to fit to the used variant. Last but not least, it is partially possible to move with the help of parameters between the variants and let an optimization algorithm automatically choose the best model.

Having complex, high dimensional data and consequently a more complex processing chain as a concatenation of different algorithms, up to now it was nearly impossible to find out what happened in the classification process and which components of the original data were used. So in our second step, we introduce an approach called backtransformation. It enables a visualization of the complete processing chain in the input data space and thereby allows for a joint interpretation of preprocessing and classification to *decode* the decision process. The interpretation can be compared with expert knowledge to find out that the algorithm is working as expected, to generate new knowledge, or to find errors in the processing (e.g., usage of artifacts in the data).

The third step is meant for the practitioner and hence a bit more technical. We propose the signal processing and classification environment pySPACE which enables the systematic evaluation and comparison of algorithms. It makes the aforementioned approaches usable for the public. Different connected SVM models can be compared and the backtransformation can be applied to any processing chain due to a generic implementation. Furthermore, this open source software provides an interface for users, developers, and algorithms to *optimize* the processing chain for the data at hand including the preprocessing *as well as* the classification.

The benefits and properties of these three approaches (also in combination) are shown in different applications (e.g., handwritten digit recognition and classification of brain signals recorded with electroencephalography) in the respective chapters.

# Zusammenfassung

Die Klassifizierung komplexer Daten erfordert für gewöhnlich die Kombination von Verarbeitungsschritten. Hierbei ist die Auswahl optimaler Algorithmen zur Vorverarbeitung und Klassifikation (inlusive ihrer Parametrisierung) eine große Herausforderung. Teile dieses Optimierungsprozesses sind heutzutage schon automatisiert aber es sind immer noch Expertenwissen und Handarbeit notwendig. Wir stellen drei Möglichkeiten vor, um diesen Optimierungsprozess besser handhaben zu können. Dabei betrachten wir etablierte Klassifikatoren von der theoretischen Seite, stellen eine Möglichkeit zur Verfügung, den Klassifikator zusammen mit der Vorverarbeitung zu interpretieren, und wir integrieren beides in eine Software welche die semiautomatische Optimierung der Verarbeitungsketten ermöglicht und welche zahlreiche Verarbeitungsalgorithmen zur Verfügung stellt.

Im ersten Schritt, fassen wir die zahlreichen Varianten der Support Vector Machine (SVM) zusammen und führen ein verallgemeinerndes (generalizing) Modell ein, welches zeigt, dass diese Varianten nicht für sich allein stehen sondern dass sie sehr stark verbunden sind. Mit Hilfe der Betrachtung dieser Verbindungen ist es möglich weitere SVM-Varianten zu generieren wie zum Beispiel Online- und Einklassenklassifikatoren. Unser Model verbessert das Verständnis über die Zusammenhänge zwischen den Varianten. Es kann in der Lehre verwendet werden und um die Wahl und Implementierung eines Klassifikators zu vereinfachen. Oftmals können Erkenntnisse und Implementierungen von einem Klassifikator auf eine andere Variante übertragen werden. Desweiteren, können die entdeckten Verbindungen mögliche Probleme offenbaren, wenn man bestimmte Varianten anwenden möchte. In bestimmten Fällen sollten einige der Varianten nicht verwendet werden oder aber die restliche Verarbeitungskette müsste angepasst werden um mit dieser Variante verwendet werden zu können. Nicht zuletzt ist es teilweise möglich mit Hilfe von Parametern sich zwischen den verschiedenen Varianten zu bewegen und ein Optimierungsalgorithmus könnte dadurch die Bestimmung des besten Algorithmusses übernehmen.

Wenn man mit komplexen und hochdimensionalen Daten arbeitet, verwendet man oft auch komplexe Verarbeitungsketten. Bisher war es daher meist nicht möglich herauszufinden, welche Teile der Daten für den gesamten Klassifikationsprozess entscheidend sind. Um dies zu beheben, führen wir in unserem zweiten Schritt die "Backtransformation" (Rücktransformation) ein. Sie ermöglicht die Darstellung der kompletten Verarbeitungskette im Raum der Eingangsdaten und lässt damit eine gemeinsame Interpretation von Vorverarbeitung und Klassifikation zu, um den Entscheidungsprozess zu entschlüsseln (decode). Die anschließende Interpretation kann mit existierendem Expertenwissen abgeglichen werden um herauszufinden, ob sich die verwendete Verarbeitung erwartungsgemäß verhält. Sie kann auch zu neuen Erkenntnissen führen oder Fehler in der Verarbeitungskette aufdecken, wenn zum Beispiel sogenannte Artefakte in den Daten verwendet werden.

Der dritte Schritt ist für den Praktiker gedacht und daher etwas mehr technisch. Wir stellen unsere Signalverarbeitungs- und Klassifikationsumgebung pySPACE vor, welche die systematische Auswertung und den Vergleich von Verarbeitungsalgorithmen ermöglicht. Es stellt die zuvor genannten Ansätze der Öffentlichkeit zur Verfügung. Die verschiedenen, stark verbundenen SVM-Varianten können verglichen werden und die Backtransformation kann auf beliebige Verarbeitungsketten in pySPACE angewandt werden, dank einer generischen Implementierung. Desweiteren, stellt diese quelloffene Software eine Schnittstelle dar für Algorithmen, Entwickler und Benutzer um Vorverarbeitung und Klassifikation für die jeweils vorliegenden Daten zu optimieren.

Die Vorteile und Eigenschaften unserer drei Ansätze (auch in Kombination) werden in verschiedenen Anwendungen gezeigt, wie zum Beispiel der Handschrifterkennung oder der Klassifikation von Gehirnsignalen mit Hilfe der Elektroenzephalografie.

# Acknowledgements

First of all, I would like to thank all my teachers at school and university, especially Rosemarie Böhm, Armin Bochmann, Prof. Dr. Thomas Friedrich, Prof. Klaus Mohnke, Prof. Dr. Bernd Kummer, and Dr. Irmgard Kucharzewski. Without these people, I would have never become a mathematician.

Most importantly, I would like to thank my advisor and institute director Prof. Dr. Frank Kirchner and my project leader Dr. Elsa Andrea Kirchner. They gave me the opportunity to work at a great institute with great people at a great project. Usually project work means to be very restricted in the scientific work and there is not much space for own ideas. But in this case, I am grateful that I had much space for getting into machine learning, being creative, and developing my own ideas. Dr. Elsa Andrea Kirchner lead me into the very interesting and challenging area of processing electroencephalographic data and Prof. Dr. Frank Kirchner encouraged me to take a step back and to look at more general approaches which could also help robotics and to think about the bigger scientific problems and a longterm perspective. I also thank Prof. Dr. Christof Büskens for discussing the optimization perspective of this thesis with me.

For structuring this thesis, thinking more "scientific", and better thinking about how other persons perceive my written text and how to handle their criticism, Dr. Sirko Straube invested numerous "teaching lessons" and never lost patience. I very much appreciate that.

I would like to thank my friends and colleagues, David Feess and Anett Seeland. For getting into the basics of this thesis, I had a lot of support by David Feess who raised my first interest to make support vector machine variants easier to understand, to look into a sparse classifier trained on electroencephalographic data, and to improve the usability and availability of pySPACE. When needing someone for the discussion of any problem, Anett Seeland was always there and also solved a lot of programming problems for/with me.

I would like to thank my team leader of the team "Sustained Learning", Dr. Jan Hendrik Metzen. Together with Timo Duchrow he laid the foundations of pySPACE. He led numerous discussions of papers and algorithms and largely improved my critical view on research and possible flaws in analyses and gave me a lot of scientific advice. Hence, he also reviewed nearly all of my papers and helped a lot improving them.

I would also like to thank all my other coauthors, reviewers, and colleagues who supported my work, especially Lisa Senger, Hendrik Wöhrle, Johannes Teiwes, Marc Tabie, Foad Ghaderi, Su Kyoung Kim, Andrei Cristian Ignat, Elmar Berghöfer, Constantin Bergatt, Yohannes Kassahun, Bertold Bongardt, Renate Post-Gonzales, and Stephanie Vareillas.

Due to the influence of all these people on my thesis, I preferred using the more common plural "We" in this thesis and used the singular "I" only in very rare cases where I want to distinguish my contribution from the work of these people.

This work was supported by the *German Federal Ministry of Economics and Technology* (BMWi, grants FKZ 50 RA 1012 and FKZ 50 RA 1011). I would like to thank the funders. They had no role in study design, data collection and analysis, decision to publish, or preparation of this thesis.

A special thanks goes to all the external people who provided tools/help for this thesis, like the open source software developers, the free software developers, scientists, and the numerous people in the internet who ask questions and provide answers for programming problems (including LaTeX issues). Without these people neither pySPACE nor this thesis would exist. As outlined in this thesis, pySPACE is largely based on the open source software stack of Python (NumPy, SciPy, matplotlib, and MDP) and by wrapping scikit-learn a lot of other algorithms can be interfaced. Without the Mendeley software I would have lost the overview over the references and without LaTeX and the numerous additional packages I probably could not have created a readable document.

Last but not least, I would like to thank all my friends, choir conductors, and good music artists. Doing science requires people who cheer you up when things do not go well. To free one's mind, stay focused, or also for cheering up, music is a wonderful tool which accompanies me a lot.

# Contents

# Chapter 0

# Introduction

## 0.1 General Motivation

Humans are able to detect the animal in the wood, to separate lentils thrown into the ashes, to look for a needle in a haystack, to find the goal and the ball in a stadium, to spot a midge on the wall, .... In everyday life, humans and animals often have to base decisions on infrequent relevant stimuli with respect to frequent irrelevant ones. Humans and animals are experts for this situation due to selection mechanisms that have been extensively investigated, e.g., in the visual [Treue, 2003] and the auditory [McDermott, 2009] domain. In their book on signal detection theory, Macmillan and Creelman argue that this comparison of stimuli is the basic psychophysical process and that all judgements are of one stimulus relative to another [Macmillan and Creelman, 2005].[1]

In short, humans and animals are the experts for numerous classification tasks and their classification skills are important for their intelligence. It is a major challenge, to provide artificial systems like computers and robots with such a type of artificial intelligence to automatically discover patterns in data [Bishop, 2006]. Especially when striving for longterm autonomy of robots, such capabilities are needed (besides others) because a robot will certainly encounter new situations and should be able to map them to previous experience.

The focus of this manuscript will be on computer algorithms for classifying data into two categories (binary classification). Given some labeled data for a classifier, the difficulty is not to generate any appropriate model but the model should be generated quickly, provide a classification result quickly, be as simple as possible, and most importantly generalize well to so far unseen data.

There is a tremendous number of classification applications (e.g., terrain classification for robots [Hoepflinger et al., 2010, Filitchkin and Byl, 2012], image classifica-tion [LeCun et al., 1998, Golle, 2008, Le et al., 2012], color distinction for robot soccer [Röfer et al., 2011], email spam detection [Blanzieri and Bryl, 2009], and analysis of brain signals as input for intelligent man machine interfaces [Kirchner et al., 2013, Kirchner et al., 2014a, Kim and Kirchner, 2013]).

---

[1] This paragraph contains text snippets from [Straube and Krell, 2014] by Dr. Sirko Straube.

There is also a very large number of approaches to solve these problems. Often the original data (raw data) cannot be used by the classifier to build a model, but an additional preprocessing is required which transforms the raw data to so-called feature vectors which better describe the data, e.g., mean values, frequency power spectra, and amplitudes after a low pass filtering.[2] When dealing with classification tasks of complex data, the generation of meaningful features is a major issue. This is due to the fact that the data often consists of a superposition of a multitude of signals, together with dynamic and observational noise. Hence, the data processing usually requires the combination of different preprocessing steps in addition to a classifier. In fact, the generation of good features is usually more important than the actual classification algorithm [Domingos, 2012].[3]

Unfortunately, the challenge to define an appropriate processing of the data is so complicated, that expert knowledge is often required and that even with the help of this knowledge, the optimal processing might not be found due to the variety of possible choices of algorithms and parameterizations.[4] Testing every possible choice is completely impossible.

### The General Research Question

In this thesis, we present three related approaches to make this process easier. It is a small step into requiring less manual work and expert knowledge and automatizing this tuning process. It can be motivated by a general question. In this context, a machine learning expert might ask:

> "**How** shall I use **which** classifier (depending on the data at hand) and **what** features of my data does it rely on?"

The "which" refers to the variety of possible algorithms. Even after choosing the classifier, an implementation is required and the data needs to be preprocessed ("how") and after the processing the expert wants to know if the processing worked correctly and if it is even possible to learn something from it ("what").

---

[2] In Section 2.2.1 more examples will be given and it will be shown, how algorithms are combined for the feature generation to processing chains.

[3] Without an appropriate preprocessing, a classifier is not able to build a general model, which will give good results on unseen data.

[4] To distinguish model parameters of algorithms from the meta-parameters, which customize the algorithm, the latter are usually called hyperparameters.

Unfortunately, there is no fully satisfactory answer to the first part of this question according to the "no free lunch theorem" of optimization [Wolpert and Macready, 1997].[5] The answer to the second part depends on the complexity of the applied processing algorithms and might be very difficult to provide, especially when different algorithms are combined or adaptive or nonlinear algorithms are used.

Besides the *no free lunch theorem*, the difficulty of choosing the "right" classifier and answering the "which" is complicated by the dependence of the classifier on the preprocessing and the high number of existing algorithms.[6] Advantages of certain classifiers often depend on the application but also on the chosen way of tuning hyperparameters and implementing the algorithm (e.g., stop criterion for convergence). A common approach to compare classifiers is to have a benchmarking evaluation with a small subset of classifiers on a special choice of datasets. This can give a hint on the usefulness of certain classifiers for certain applications/datasets but does not provide a deeper understanding of the classifiers and how they relate to each other. A different approach is to clearly determine the relations between classifiers in order to facilitate the choice of an appropriate one. Unfortunately, only few connections between classifiers are known and, since they are spread all over the literature, it is quite difficult to conceive of them as a whole. Hence, summarizing the already known connections and deriving new ones is required to ease the choice of the classifier. This even holds for the numerous variants of the support vector machine (SVM). We will focus on that classifier, because it is very powerful and understanding the connection to its variants is already helpful. It is reasonable to pick a group which has a certain common ground, because it is impossible to connect all classifiers.

Additionally to looking at classifiers it is important to look at their input: the feature vectors, which are used as data for building the classifier (training samples). For finding the relevant features in the data, there are several algorithms in the context of feature selection [Guyon and Elisseeff, 2003, Saeys et al., 2007, Bolón-Canedo et al., 2012]. Even though these algorithms can improve classification accuracy and interpretability, they do not give information about the relevance of the features for the classifier finally used in a data processing chain. The answer to the question, "what features of my data does my classifier rely on", can be difficult to provide because of three issues. First, the classifier might have to be treated as a black box. Second, it might have nonlinear behavior, meaning that the relevance of certain features in the data is highly dependent on the sample which is classified. The third and most important point is, that the classifier is not applied to the raw data but pre-processed data. Hence, the classifier should *not* be regarded as a single algorithm, but instead the complete decision algorithm consisting of preprocessing algorithms *and* classifier and their interplay with the data need to be considered. For example, in the extreme case where a classifier is not even really required because the features are sufficiently good, it is important to look at the generation of the features to decode the decision algorithm.

---

[5] For our case, the theorem states that for every classification problem, where classifier $a$ is better than classifier $b$, there is a different problem where the opposite holds true.

[6] With a different preprocessing a different classifier might be appropriate, e.g., with a nonlinear instead of a linear model.

Last but not least, the question of "how" to apply the data processing is probably the most time consuming part of designing a good data processing chain. Performing hyperparameter optimization and large scale evaluations is cumbersome. A lot of time for programming and waiting for the results is required. Furthermore, when trying to reproduce results from other persons there is no access to the used implementations and the details of the evaluation scheme. The most complicated part might be to configure the processing for the needs of the concrete application and to generate optimal or at least useful features.

To fix all these problems completely is impossible but it is possible to tackle parts of them and go a step further towards a solution as outlined in the following section.

Despite this more general and abstract motivation, we will provide a more concrete motivation by an application in Section 0.4.

## 0.2 Objectives and Contributions

The main objective of this thesis is to ***provide (theoretical, practical, and technical) insights and tools for data scientists to simplify the design of the classification process.*** In contrast to other work, the goal is not to derive new algorithms or to tweak existing algorithms.

Here, a "classification process" also includes the complete evaluation process with the preprocessing, tuning of hyperparameters, and the analysis of results. Three subgoals can be identified, derived from the previously discussed question: *"How shall I use which classifier and what features of my data does it rely on?"*

❶ Theoretical aspect: Analyze the connections between SVM variants to derive a more "general" picture.

❷ Practical aspect: Construct an approach for decoding and interpreting the decision algorithm together with the preprocessing.

❸ Technical aspect: Implement a framework for better automatizing the process of optimizing the construction of an appropriate signal processing chain including a classifier.

Subgoal 1 targets the question of "which" classifier to use. The question of "what features of my data does it (the classifier) rely on" is covered by the second subgoal. The last subgoal requires us to answer the question of "how" to apply the classifier and supports Subgoal 1 by providing a platform to compare and analyze classifiers. It also supports Subgoal 2 as an interface for implementing it.

Note that this introduced numbering will also be used concerning the achievements of this thesis and the respective chapter numbers. Furthermore, it is important to note that there are connections between the goals, because the respective approaches can (and often have to) be combined. To face the three subgoals, the following approaches are taken.

**Contribution 1: Generalizing** Due to the ever-growing number of classification algorithms, it is difficult to decide which ones to consider for a given application. Knowledge about the relations between the classifiers facilitates the choice and implementation of classifiers. As such, instead of further specializing existing classifiers we take a unifying view. Considering only the variants of the classical support vector machine (C-SVM) [Vapnik, 2000, Cristianini and Shawe-Taylor, 2000, Müller et al., 2001, Schölkopf and Smola, 2002], some connections are already known but the knowledge about these connections is distributed over the literature.

We summarize these connections and introduce the following three general concepts building further intuitive connections between these classifiers.

The C-SVM belongs to the group of *batch learning* classifiers. These classifiers operate on the complete set of training data to build their model consuming large resources of memory and processing time. In contrast, *online learning* algorithms update their model with each single sample and, later on, forget the sample. They are very fast and memory efficient which is required for several applications but they usually perform less well. The *single iteration approach* describes a way to transfer batch learning to online learning classifiers. If the solution algorithm of the batch learning classifier is repeatedly iterated over the single training samples to update a linear classification function, an online learning algorithm can be generated by performing this update only once for each incoming sample.

The second concept, called *relative margin*, establishes a connection between the more geometrically motivated SVM and the regularized Fisher's discriminant (RFDA) coming from statistics.

The third concept, the *origin separation approach*, allows defining unary classifiers with the help of binary classifiers by taking the origin as a second class.[7]

---

[7] Unary classifiers use only one class for building a model but they are usually applied to binary classification problems, where the focus is to describe the more relevant class, or where not enough training samples are available from the second class to build a model.

Together with the existing more formal connection concepts (especially normal and squared loss, kernel functions, and normal and sparse regularization), these connections span the *complete space* of established SVM variants and additionally provide new not-yet discovered variants.

Knowing the *theory* of these novel connections simplifies the implementation of the algorithms and makes it possible to transfer extensions or modifications from one algorithm to the other connected ones. Thus, it enables to build a classifier that fits into the individual research aims. Furthermore, it simplifies teaching and getting to know these classifiers. Note that the connections are not to be taken separately but in most cases they can be combined.

**Contribution 2: Decoding**   Having the knowledge about the relations between classifiers is not always sufficient for choosing the best one. It is also important to *understand* the final processing model to find out what lies behind the data and to ensure that the classifier is not relying on artifacts (errors in the data). Existing approaches visualize the data and the single processing steps, but this might not be sufficient for a complete picture, especially when dimensionality reduction algorithms are used in the preprocessing. This is often the case for high-dimensional and noisy data. Hence, a representation of the entire processing chain including *both* classification and preprocessing is required. Our novel approach to calculate this representation is called backtransformation. It iteratively transforms the classification function back through the signal processing chain to generate a representation in the same format as the input data. This representation provides weights for each part of the data to tell which components are relevant for the complete processing and which parts are ignored. It can be directly visualized, when using classical data visualization approaches as they are for example used for image, electroencephalogram (EEG), and functional magnetic resonance imaging (fMRI) data. This *practical* contribution opens up the black box of the signal processing chain and can now be used to support the "close collaboration between machine learning experts and application domain ones" [Domingos, 2012, p. 85]. It can provide a deeper understanding of the processing and it can help to improve the processing and to generate new knowledge about the data. In some cases even new expert knowledge might be generated.

**Contribution 3: Optimizing**   For a generic implementation of the backtransformation an interface is required. Furthermore, it is still required to optimize the hyperparameters of the classifiers and the preprocessing for further improvement of the processing chain. Hence, it is necessary to have "an infrastructure that makes experimenting with many different learners, data sources, and learning problems easy and efficient" [Domingos, 2012, p. 85]. To solve this problem, we de-veloped the S̲ignal P̲rocessing A̲nd C̲lassification E̲nvironment written in Python (pySPACE) [Krell et al., 2013b]. It provides functionality for a systematic and automated comparison of numerous algorithms and (hyper-)parameterizations in a signal processing chain. Additionally, pySPACE enables the visualization of data, algorithms, and evaluation results in a common framework. With its large number of supporting features this software is unique and a major improvement to the existing open source software.

## 0.3 Structure

In this thesis, we present our steps to improve and automatize the process of designing a good processing chain for a classification problem (classifier connections, backtransformation, pySPACE). This thesis is structured as follows.

First, the different SVM variants are introduced including the known connections and in the following three more general concepts are introduced which connect them (Chapter 1). Second, the backtransformation concept is presented in Chapter 2. Third, the pySPACE framework, the more technical part of this thesis, and its use for optimization is shown in Chapter 3. All three main parts are also displayed in Figure 1 using the same numbering. Finally, a conclusion and an outlook is given in Chapter 4. In the appendix, all my publications are summarized. Furthermore, the appendix contains detailed proofs, information on the used data, and some configuration files used for the evaluations in the different chapters.

The related work and our proposed approaches are often highly connected and consequently presented separately in the respective chapters and not in an extra chapter about literature at the beginning of this thesis. Each approach integrates at least a part of the related work.

Even though the contributions of this thesis are separated into three chapters, they are still connected. For the evaluations in Chapter 1 and Chapter 2, the respective algorithms are integrated into pySPACE and the framework is used to perform the evaluations using the concepts described in Chapter 3. Furthermore, the backtransformation concept from Chapter 2 will be applied to the different classifiers from Chapter 1 and additional knowledge about the classifiers will be incorporated into a variant of the concept. Last but not least, all three parts should be combined to get the best result when analyzing data.

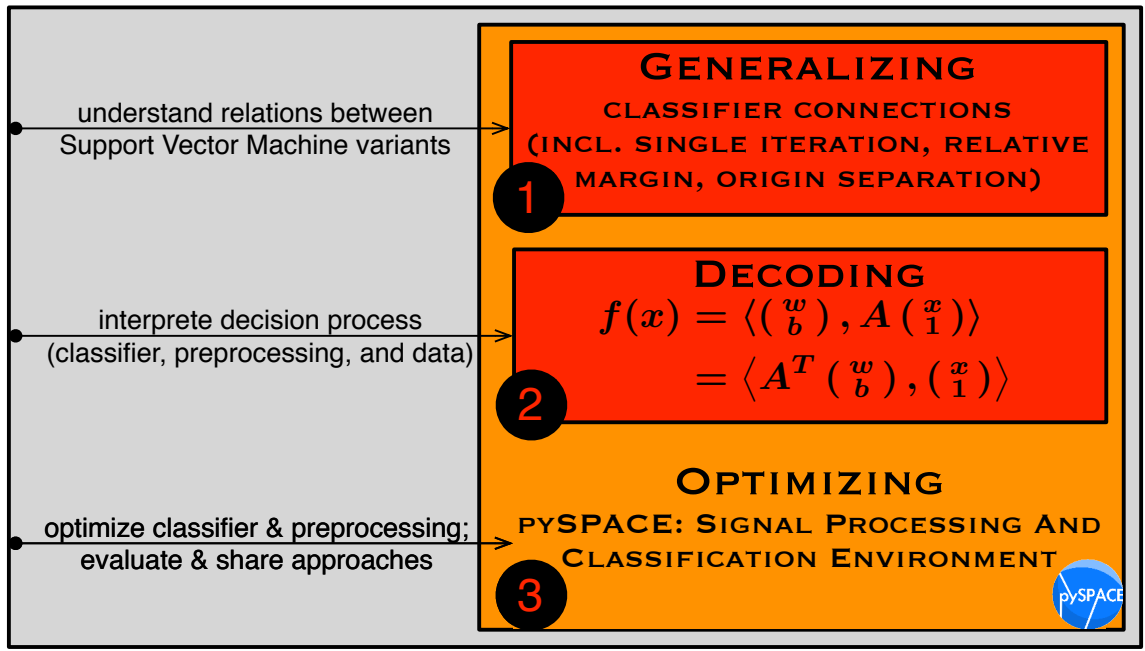

Figure 1: **Graphical abstract of this thesis.** The numbering is also used for the corresponding subgoals and respective chapters. The first part provides a more general picture of SVM variants by connecting them. The second part introduces the backtransformation concept to decode data processing chains. Finally, the third part presents our framework pySPACE which is an interface for optimizing signal processing chains. Furthermore, the previous two parts can be used and analyzed with this software.

## Disclaimer: Text Reuse

Single sentences but also entire paragraphs of this thesis are taken from my own publications without explicit quotation because I am the main author[8] or I contributed the used part to them.[9] Except for my summary paper [Krell et al., 2014c, see also Section 2.4.4], which is somehow scattered over some introductory parts, I explicitly mention these sources at the beginning of the respective chapters or sections where they are used. Often parts of these papers could be omitted by referring to other sections or they had to be adapted for consistency. On the other hand, additional information, additional experiments, the relation to the other parts of this thesis, or personal experiences are added.

---

[8]      [Krell et al., 2013b,      Krell et al., 2014a,      Krell et al., 2014c,      Krell and Straube, 2015, Krell and Wöhrle, 2014]

[9] [Feess et al., 2013, Straube and Krell, 2014]

**Notation**

In this thesis mostly the "standard notation" is used and it should be possible to infer the meaning from the context. Nevertheless, there is a list of acronyms and a list of used symbols at the end of this document. If some notation is unclear we refer to these lists. It will be directly mentioned, if the standard symbols are not used.

## 0.4 Application Perspective: P300 Detection

Even though the approaches derived in this thesis are very general and can be applied in numerous applications, they were originally developed with a concrete dataset/application in mind. We will first describe the general setting, continue with a description of the experiment which generated the data, and finally highlight the connection of the dataset to this thesis to provide an additional less abstract motivation.

### 0.4.1 General Background of the Dataset

Current brain-computer interfaces (BCIs) rely on machine learning techniques as the ones discussed in this thesis. They can be used to detect the P300 event-related potential (ERP)[10] for communication purposes (e.g., for P300 based spellers [Farwell and Donchin, 1988, Krusienski et al., 2006] or for controlling a virtual environment [Bayliss, 2003]), to detect interaction errors for automated correction [Ferrez and Millán, 2008, Kim and Kirchner, 2013], or to detect movement preparation or brain activity that is related to the imagination of movements for communication or control of technical devices [Bai et al., 2011, Blankertz et al., 2006, Kirchner et al., 2014b].

The P300 is not only used to implement active BCIs for communication and control but can furthermore be used more passively as it was investigated in the dataset described in the following. For example, in embedded brain reading (eBR) [Kirchner, 2014] the P300 is naturally evoked in case an operator detects and recognizes an important warning during interaction. Thus, the detection of the P300 is used to infer whether the operator will respond to the warning or not and to adapt the interaction interface with respect to the inferred upcoming behavior. A repetition of the warning by the interaction interface can be postponed in case a P300 is detected after a warning was presented since it can be inferred that the operator will respond to the warning. In case there is no P300 detected, the warning will be repeated instantly since it can be inferred that the operator did not detect and recognize the warning and will therefore not respond [Wöhrle and Kirchner, 2014]. Since in the explained case we are able to correlate the brain activity with the subject's behavior, the detected behavior can be used as a label to control for the correctness of the predicted brain states and hence to adapt the classifier by online learning to continuously improve classification performance [Wöhrle et al., 2013b] (Section 1.2).

---

[10] This is a special signal in the measurement of electrical activity along the scalp (electroencephalogram). The name refers to a *positive* peak at the parietal region which occurs roughly 300 ms (or with a larger latency) after the presentation of a rare but important visual stimulus (see also Section 0.4.2).

The previous description was created with the help of Dr. Elsa Andrea Kirchner, who headed the experiments for the dataset. The following rather short dataset description is adapted from [Feess et al., 2013] where the data was used to compare different sensor selection algorithms. A very detailed description of the experiment and related experiments is provided in [Kirchner et al., 2013].

## 0.4.2   Description of the Dataset

The data described in this section has been acquired from a BCI system that belongs to the class of *passive* BCIs: the purpose is the gathering of information about the user's mental state rather than a voluntary control of a system [Zander and Kothe, 2011, Kirchner, 2014]. Therefore, no deliberate participation of the subject is required.

The goal of the system is to identify whether the subject distinctively perceived certain rare *target* stimuli among a large number of unimportant *standard* stimuli. It is expected that the *targets* in such scenarios elicit an ERP called P300 whereas the *standards* do not [Courchesne et al., 1977].

Five subjects participated in the experiment and carried out two sessions on different days each. A session consisted of five *runs* with 720 *standard* and 120 *target* stimuli per run. EEG data were recorded at 1 kHz with an actiCAP EEG system (Brain Products GmbH, Munich, Germany) from 62 channels following the extended 10–20 layout. (This system usually uses 64 channels. Electrodes TP7 and TP8 were used for electromyogram (EMG) measurements and are excluded here.)[11]

The data was recorded in the Labyrinth Oddball scenario (see Figure 2), a testbed for the use of passive BCIs in robotic telemanipulation. In this scenario, participants were instructed to play a simulated ball labyrinth game, which was presented through a head-mounted display. The insets in the photograph show the labyrinth board as seen by the subject. While playing, one of two types of visual stimuli was displayed every 1 second with a jitter of $\pm 100$ ms. The corners arranged around the board represent these stimuli. As can be seen, the difference in the *standard* and *target* stimuli is rather subtle: in the first case the top and bottom corners are slightly larger and in the latter the left and right corners are larger. The subjects were in-structed to ignore the *standard* stimuli and to press a button as a reaction to the rare *target* stimuli.

---

[11] The electrode layout with 64 electrodes is depicted in Figure C.6.

Both *standard* and *target* stimuli elicit a visual potential as seen in the averaged time series in Figure 2 (strong negative peak at around 200 ms after the stimuli). Additionally, *target* stimuli induce a positive ERP, the P300, with maximum amplitude around 600 ms after stimulus at electrode Pz. It is assumed that the P300 is evoked by rare, relevant stimuli that are recognized, and cognitively evaluated by the subject.

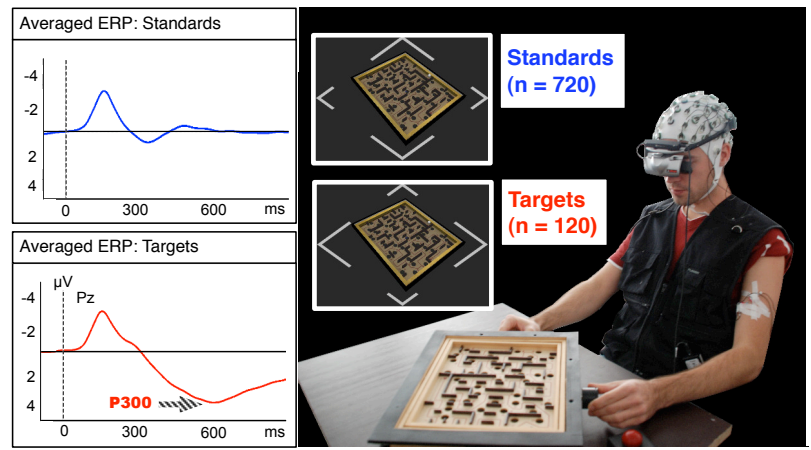

Figure 2: **Labyrinth Oddball:** The subject plays a physical simulation of a ball labyrinth game. He has to respond to rare *target* stimuli by pressing a buzzer and ignore the more frequent *standard* stimuli. The insets show the shape of the stimuli, which can be distinguished by the length of the edges. The graphs to the left depict the event-related potentials (ERPs) evoked by both stimulus types at electrode Pz. Both stimuli elicit an early negative potential attributed to visual processing, but only *targets* evoke an additional strong, positive potential around 600 ms after the stimulus. Visualization and description taken from [Feess et al., 2013].

The BCI only needs to passively monitor whether the operator of the labyrinth game correctly recognized and distinguished these stimuli. There is an objective affirmation of the successful stimulus recognition, because a button has to be pressed, whenever a *target* is recognized. No feedback is given to the user.

### 0.4.3 Relevance for this Thesis

Even though this data is not (yet) open source, it was used in this thesis for several reasons as listed in the following.

- It provides numerous datasets to have a comparison of algorithms.[12]
- EEG data classification is a very challenging task where the applied signal processing chain usually performs much better than the human.

---

[12] Up to 50 datasets/recordings, depending on the evaluation scheme.

- The data has a very bad signal to noise ratio. Thus it is a challenge to optimize the processing chain.
- The data was recorded in a controlled and somehow artificial scenario but in fact it targets a much more promising application of a BCI where the humans intentions are monitored with the help of the EEG during a teleoperation task with many robots where robots act more autonomously. This task can be very challenging and the monitoring can be used to avoid malfunction in the interface. When analyzing the P300 data and tuning processing chains, we kept this more complex application in mind.
- The dataset was the motivation for all findings in this thesis as described in the following.
- The aforementioned more practical application requires online learning to integrate new training data for performance improvement and to account for the different types of drifts in the data (see Section 1.2).
- Support vector machine and Fisher's discriminant were common classifiers on this type of data [Krusienski et al., 2006] (see Section 1.3).
- Depending on the application, which uses the P300, it might be very difficult to acquire data from a second class and consequently a classifier is of interest, which only works with one class (see Section 1.4). Altogether, a more *general* model of classification algorithms and their properties and connections is helpful here.
- There is always the danger of relying on artifacts (e.g., muscle artifacts, eye movement) and there is an interest from neurobiology to *decode* the processing chain, which is built to classify the P300 (see Chapter 2). For the given dataset, we could show that eye artifacts are not relevant.
- Finding a good processing chain for such demanding data requires a lot of hyperparameter *optimization* and comparison of different algorithms. Furthermore, it is useful to exchange processing chains between scientist to find flaws, communicate problems and approaches, and help each other improving the processing (see Chapter 3).
- There is an interest in using as few sensors and time points for the processing to make the set up easier and the processing faster (see Section 3.4.3). To derive such selection algorithm it can be helpful to combine the tools and insights, derived in this thesis.

A reference processing chain for this data is depicted in Figure 3.4. In the evaluations in this thesis, only the difference to this processing scheme is reported. The processing chain assumes that the data has already been segmented in samples of one second length after the target or standard stimulus.

# Chapter 1

# Generalizing: Classifier Connections

## Contents

The aim of this chapter is to summarize known and novel connections between SVM variants to derive a more general view on this group of classifiers. This shall facilitate the choice of the classifier given certain data or applications at hand.

Given some data-value pairs $(x_j, y_j)$ with $x_j \in \mathbb{R}^m$ and $j \in \{1, \dots, n\}$ a common task is to find a function $F$ which maps $F(x_j) = y_j$ as good as possible and which should also perform well on unseen data. If $y_j$ is from a continuous space like $\mathbb{R}$, an algorithm deriving such a function $f$ is called *regression* algorithm. If $y_j$ is from a discrete domain, the algorithm is a classifier. In this thesis, we will focus on the case of binary classification with $y_j \in \{-1, +1\}$. In most cases, linear classifiers will be used with $f(x) = \langle w, x_j \rangle + b$, where $w$ is the classification vector and $b$ the offset. To finally map the classification function to a decision ($\{-1, +1\}$), the signum function is applied ($F(x) = \text{sgn}(f(x))$).

The classical support vector machine (C-SVM) [Vapnik, 2000, Cristianini and Shawe-Taylor, 2000, Müller et al., 2001, Schölkopf and Smola, 2002] is a well-established binary classifier.[1]  Good generalization properties, efficient implementations, and powerful extensions like the kernel trick or possible sparsity properties (explained in Section 1.1), make the SVM attractive for numerous variants and applications [LeCun et al., 1998, Guyon et al., 2002, Lal et al., 2004, LaConte et al., 2005, Golle, 2008, Tam et al., 2011, Filitchkin and Byl, 2012, Kirchner et al., 2013]. The most important variants are

- $\nu$ support vector machine ($\nu$-SVM) [Schölkopf et al., 2000, Section 1.1.1.3],
- support vector regression (SVR) [Vapnik, 2000, Smola and Schölkopf, 2004, Section 1.1.1.4],
- least squares support vector machine (LS-SVM) [Van Gestel et al., 2002, Section 1.1.2],
- relative margin machine (RMM) [Shivaswamy and Jebara, 2010, Krell et al., 2014a, Section 1.1.4 and 1.3],
- passive-aggressive algorithm (PAA) [Crammer et al., 2006, Section 1.1.5, 1.1.6.2, and 1.2.4],
- support vector data description (SVDD) [Tax and Duin, 2004, Section 1.1.6.1],
- and classical one-class support vector machine ($\nu$oc-SVM) [Schölkopf et al., 2001b, Section 1.1.6.3].
- Furthermore, regularized Fisher's discriminant (RFDA) can be seen as a SVM variant, too [Mika, 2003, Krell et al., 2014a, Section 1.1.3 and 1.3].

In the literature these algorithms are usually treated as distinct classifiers. This also holds for the evaluations. Some connections between these classifiers are known but scattered erratically over the large body of literature. First, in Section 1.1 the models and these connections will be summarized. In the following, general concepts for a unifying view are proposed to further connect these classifiers and ease the process of choosing a fitting classifier:

- The *single iteration* concept creates online learning classifiers like PAA from batch learning classifiers to save computational resources (Section 1.2).

- The *relative margin* concept intuitively connects SVM, SVR, RMM, LS-SVM, and RFDA (Section 1.3).

- The *origin separation* concept transforms binary to unary classifiers like $\nu$oc-SVM for outlier detection or to data description (Section 1.4).

---

[1] The $C$ in the abbreviation probably refers to the hyperparameter $C$ in the classifier definition and is used to distinguish it from other related classifiers (see also Section 1.1.1).

By combining these three approaches, a large number of additional variants can be generated (see Fig. 1.1). In Section 1.5 the connections between the aforementioned classifiers will be summarized and possible scenarios explained where the knowledge of the connections is helpful (e.g., implementation, application, and teaching).

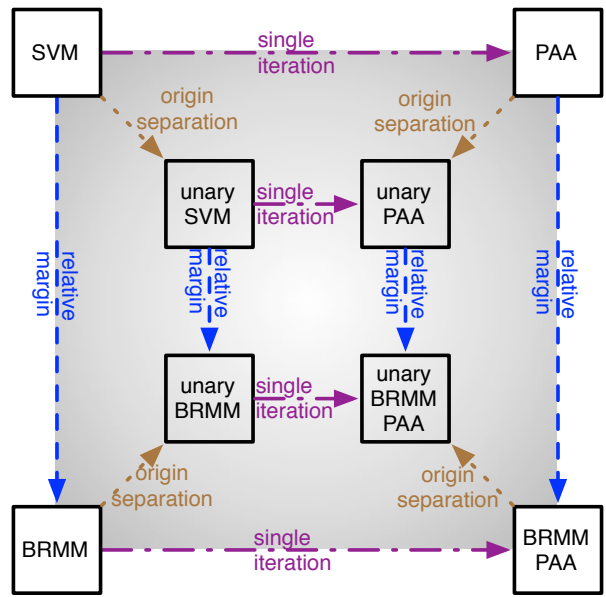

Figure 1.1: **3D-Cube of our novel connections (commutative diagram).** Combining the approaches, introduced in Chapter 1: relative margin (vertical arrows) to generate the balanced relative margin machine (BRMM) which is the connection to the regularized Fisher's discriminant (RFDA), single iteration (horizontal arrows) to generate online classifiers like the passive-aggressive algorithm (PAA), and the origin separation (diagonal arrows) to generate unary classifiers from binary ones. Each approach is associated with one dimension of the cube and going along one edge means to apply or remove the respective approach from the classifier at the edge.

## 1.1   Support Vector Machine and Related Methods

In this section, we introduce all the aforementioned SVM variants including some basic concepts and known connections which are pure parameter mappings and no general concepts. For further reading, we refer to the large corpus of books about SVMs. Readers who are familiar with the basics of support vector machines and its variants can continue with the next section.

The models will be required in the following sections which introduce three general concepts to connect them. Putting everything together in Section 1.5, we will show that the SVM variants, introduced in this section, are all highly connected.

### 1.1.1   Support Vector Machine

In a nutshell, the principle of the C-SVMs is to construct two parallel hyperplanes with maximum distance such that the samples belonging to different classes are separated by the space between these hyperplanes (see also Figure 1.2). Such space between the planes is usually called margin—or *inner margin* in our context.

Commonly, only the Euclidean norm $\left( \|v\|_2 = \sqrt{\sum v_i^2} \right)$ is used for measuring the distance between points but it is also possible to use an arbitrary $p$-norm $\left( \|v\|_p = \sqrt[p]{\sum v_i^p} \right)$ with $p \in [1, \infty]$.[2]

For getting the distance between two parallel hyperplanes or a point and a hyperplane instead, the respective dual $p'$-norm has to be used with $\frac{1}{p} + \frac{1}{p'} = 1$ [Mangasarian, 1999]. If $p = \infty$, $p'$ is defined to be 1. Having the two hyperplanes $H_{+1}$ and $H_{-1}$ with

$$H_z = \{x | \langle w, x \rangle + b = z\} \; , \tag{1.1}$$

their distance is equal to $\frac{2}{\|w\|_{p'}}$ according to Mangasarian. (In case of the Euclidean norm, this effect is also known from the Hesse normal form.) The resulting model reads as:

**Method 1** (Maximum Margin Separation)**.**

$$\begin{aligned} \max_{w,b} \quad & \frac{1}{\|w\|_{p'}} \\ s.t. \quad & y_j(\langle w, x_j \rangle + b) \geq 1 \quad \forall j : 1 \leq j \leq n. \end{aligned} \tag{1.2}$$

For new data, the respective linear classification function is:

$$f(x) = \langle w, x_j \rangle + b. \tag{1.3}$$

---

[2] Due to convergence properties it holds $\|v\|_\infty := \max |v_i|$.

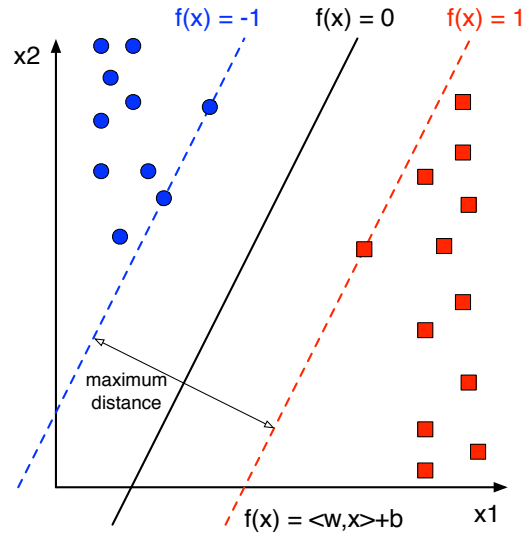

Figure 1.2: **Support vector machine scheme.** The blue dots are training samples with $y = -1$ and the red dot with $y = +1$ respectively. Displayed are the three parallel hyperplanes $H_{+1}$, $H_0$, and $H_{-1}$.

To get a mapping to the class labels $-1$ and $+1$ we use the decision function

$$F(x) = y(x) = \text{sgn}(f(x)) := \begin{cases} +1 & \text{if } f(x) > 0, \\ -1 & \text{otherwise.} \end{cases} \tag{1.4}$$

For better solvability, Method 1 is reformulated to an equivalent one. The fraction is inverted, and the respective minimization problem is solved instead. Furthermore, the root is omitted to simplify the optimization process and further calculations. An additional scaling factor is added for better looking formulas when solving the optimization problem. These superficial modifications do not change the optimal solution. The resulting reformulated model reads:

**Method 2** (Hard Margin Separation Support Vector Machine)**.**

$$\begin{aligned}
\min_{w,b} \quad & \tfrac{1}{p'} \left\| w \right\|_{p'}^{p'} \\
s.t. \quad & y_j(\langle w, x_j \rangle + b) \geq 1 \quad \forall j : 1 \leq j \leq n.
\end{aligned} \tag{1.5}$$

Since strict separation margins are prone to overfitting or do not exist at all, some disturbance in the form of samples penetrating the margin is allowed denoted with the error variable $t_j$.

When speaking of the C-SVM, normally the Euclidean norm is used ($p = p' = 2$):

**Method 3** (L1–Support Vector Machine)**.**

$$\begin{aligned}
\min_{w,b,t} \quad & \tfrac{1}{2}\,\|w\|_2^2 + C\sum t_j \\
s.t. \quad & y_j(\langle w, x_j\rangle + b) \;\geq 1 - t_j \quad \forall j : 1 \leq j \leq n, \\
& t_j \;\geq 0 \qquad\quad \forall j : 1 \leq j \leq n.
\end{aligned} \qquad (1.6)$$

The hyperparameter $C$ defines the compromise between the width of the margin $\left(\text{regularization term } \tfrac{1}{2}\,\|w\|_2^2\right)$ and the amount of samples lying in or on the wrong side of the margin $(t_j > 0)$.[3] This principle is called *soft margin*, because the margins defined by the two hyperplanes $H_{+1}$ and $H_{-1}$ can be violated by some samples (see also Figure 1.3). In the final solution of the optimization problem only these samples and the samples on the two hyperplanes are relevant and provide the SVM with its name.

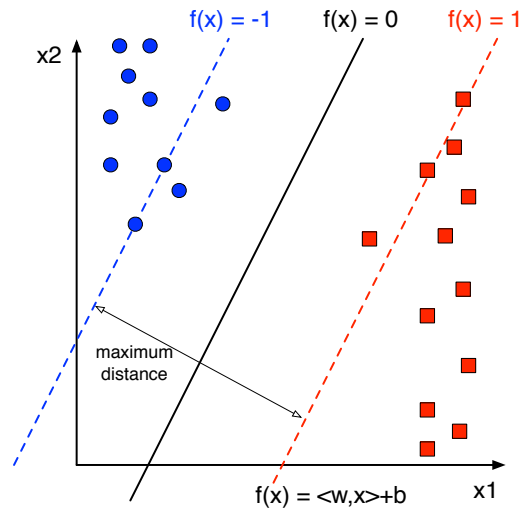

Figure 1.3: **Soft margin support vector machine scheme.** In contrast to Figure 1.2, some samples are on the wrong side of the hyperplanes within the margin.

**Definition 1** (Support Vector)**.** *The vectors defining the margin, i.e., those data examples $x_j$ where $t_j > 0$ or where identity holds in the first inequality constraint (Method 3), are the* support vectors*.*

The L1 in the method name of the SVM (Method 3) refers to the loss term $\|t\|_1 = \sum t_j$ for $t_j \geq 0$ in the target function. A L2 variant that uses $\|t\|_2^2$ instead was suggested but is rarely used in applications, especially when kernels are used. When using kernels, it is important to have as few support vectors as possible and the L2 variant often has much more support vectors [Mangasarian and Kou, 2007] (see also Section 1.1.1.2).

---

[3] $C$ is called regularization constant, cost factor, or complexity.

#### 1.1.1.1 Lagrange Duality

For deriving solution algorithms for the optimization problem of the C-SVM and for the introduction of kernels—presently one of the most important research topics in SVM theory— it is useful to apply duality theory[4] from optimization, e.g., [Burges, 1998]. Solving the dual instead of the primal (original) optimization problem can be easier in some cases and if certain requirements are fulfilled, then the solutions are connected via the Karush-Kuhn-Tucker (KKT) conditions, e.g., [Boyd and Vandenberghe, 2004]. Finally, duality theory enables necessary optimality conditions, which can be used to solve the optimization problems. Even though the following calculations will be only performed for the C-SVM, the concepts also apply for numerous variants and the respective calculations are similar (as partially shown in the appendix).

To avoid a degenerated optimization problem, it is required to check if at least one point fulfills all restrictions (feasible point), if there is a solution of the optimization problem, and if the problem can be "locally linearized", i.e., fulfills a constrained qualification, e.g., [Boyd and Vandenberghe, 2004]. These points are usually ignored in the SVM literature probably because they seem obvious from the geometrical perspective. Nevertheless, they are the basis of most optimization algorithms for SVMs.

**Theorem 1** (The C-SVM Model is well defined)**.** *The C-SVM optimization problem has feasible points and a solution always exists, if there is at least one sample for each class. Additionally when using the hard margin the sets of the two classes need to be strictly separable. Furthermore, Slater's constraint qualification is fulfilled.*[5]

The question of how to determine the solution is a main topic of Section 1.2. The benefit of this theorem is twofold. We proofed that the model is well defined and that we can apply Lagrange duality. The advantage of Lagrange duality for Method 3 is a reformulation of the optimization problem, which is easier to solve and which allows replacing the original norm by much more complex distance measures (called kernel trick). This advantage does not hold for the variants based on other norms ($p \neq 2$).

For obtaining the dual optimization problem, first of all the respective Lagrange function has to be determined. For this, dual variables are introduced for every inequality ($\alpha_j, \gamma_j$) and the inequalities are rewritten to have the form $g(w, b, t) \leq 0$. The Lagrange function is the target function plus the sum of the reformulated inequality functions weighted with the dual variables:

$$L_1(w, b, t, \alpha, \gamma) = \frac{1}{2} \|w\|_2^2 + \sum C_j t_j - \sum \alpha_j (y_j(\langle w, x_j \rangle + b) - 1 + t_j) - \sum \gamma_j t_j. \quad (1.7)$$

---

[4] This should not be mixed up with the previously mentioned duality of the norms. The dual optimization problem can be seen as an alternative/additional view on the original optimization problem.

[5] The proof is given in Appendix B.1. Other constraint qualifications do exist, but Slater's was most easy to check for the given *convex* optimization problem.

For the L2–SVM this yields:

$$L_2(w, b, t, \alpha) = \frac{1}{2} \|w\|_2^2 + \sum C_j t_j^2 - \sum \alpha_j (y_j (\langle w, x_j \rangle + b) - 1 + t_j). \qquad (1.8)$$

To consider the label or the time for the weighting of errors, $C$ has been chosen sample dependent ($C_j$).

With a case study, it can be shown that the original optimization is equivalent to optimizing:

$$\min_{w, b, t} \sup_{\alpha \geq 0, \gamma \geq 0} L_1(w, b, t, \alpha, \gamma). \qquad (1.9)$$

Infeasible points in the original optimization problem get a value of infinity due to usage of the supremum and for the feasible points the original target function value is obtained. According to Theorem 1, the optimization problem has a solution and Slater's constraint qualification is fulfilled. Consequently the duality theorem can be applied [Burges, 1998]. It states that we can exchange minimization and "supremization" and that the solutions for both problems are the same:

$$\min_{w, b, t} \sup_{\alpha \geq 0, \gamma \geq 0} L_q(w, b, t, \alpha, \gamma) = \max_{\alpha \geq 0, \gamma \geq 0} \min_{w, b, t} L_q(w, b, t, \alpha, \gamma), \, q \in \{1, 2\}. \qquad (1.10)$$

The advantages of the new resulting optimization problem, called dual optimization problem, are twofold. First, the inner part is an unconstrained optimization problem which can be analytically solved. Second, the remaining constraints are much easier to handle than the constraints in the original (primal) optimization problem.

For simplifying the dual optimization problem, the minimization problem is solved by calculating the derivatives of the Lagrange function for the primal variables and setting them to zero. This is the standard solution approach for unconstrained optimization.

$$\frac{\partial L_k}{\partial w} = w - \sum_j \alpha_j y_j x_j, \, \frac{\partial L_k}{\partial b} = - \sum_j \alpha_j y_j, \, \frac{\partial L_1}{\partial t_j} = C_j - \alpha_j - \gamma_j, \, \frac{\partial L_2}{\partial t_j} = 2 t_j C_j - \alpha_j. \quad (1.11)$$

The most important resulting equations are

$$w = \sum_j \alpha_j y_j x_j, \qquad (1.12)$$

which gives a direct relation between $w$ and $\alpha$, and

$$\sum_j \alpha_j y_j = 0, \qquad (1.13)$$

which is a linear restriction on the optimal $\alpha$. For the L1 variant, the equation

$$C_j - \alpha_j = \gamma_j \tag{1.14}$$

results in the side effect that $\gamma_j$ can be omitted in the optimization problem and $\alpha_j$ gets the upper bound $C_j$ instead, due to the constraint $\gamma \geq 0$. Finally, substituting the equations for optimality into $L_q$ and multiplying the target function with $-1$ to obtain a minimization problem results in the following theorem:

**Theorem 2** (Dual L1– and L2–SVM Formulations)**.** *The term*

$$\min_{C_j \geq \alpha_j \geq 0, \sum \alpha_j y_j = 0} \frac{1}{2} \sum_{i,j} \alpha_i \alpha_j y_i y_j \langle x_i, x_j \rangle - \sum_j \alpha_j \tag{1.15}$$

*is the dual optimization problem for the L1–SVM and*

$$\min_{\alpha_j \geq 0, \sum \alpha_j y_j = 0} \frac{1}{2} \sum_{i,j} \alpha_i \alpha_j y_i y_j \langle x_i, x_j \rangle - \sum_j \alpha_j + \frac{1}{4} \sum_j \frac{\alpha_j^2}{C_j} \tag{1.16}$$

*for the L2–SVM.*

The dual of the hard margin SVM is given in Theorem 18 and for the L2 variant a more detailed calculation is provided in Appendix B.1.3.

In the dual formulation, only the pairwise scalar products of training samples are required. This is exploited in the kernel trick (Section 1.1.1.2). Note that only in the L1 case there is an upper bound on the dual variables. Furthermore, when looking more detailed into the calculations we realize that the additional equation in the dual feasibility constraints is a result of $b$ being a free variable which is not minimized in the target function. These observations will be again relevant in Section 1.2.

The $\alpha_j$ are connected to the primal problem via Equation (1.12) but also via the complementary slackness equations [Boyd and Vandenberghe, 2004]:

$$\alpha_j > 0 \Rightarrow y_j(\langle w, x_j \rangle + b) \leq 1. \tag{1.17}$$

Consequently, only samples on the margin or on the wrong side of the margin contribute to the classification function according to Equation (1.12). All the other samples are irrelevant and do not "support" the decision function. Hence the name.

For the L1 case, it additionally holds

$$t_j > 0 \Rightarrow \gamma = 0 \Rightarrow \alpha_j = C_j. \tag{1.18}$$

This immediately tells us that every sample which is on the wrong side of the margin gets the maximum weight assigned and that every $x_j$ with $\alpha_j > 0$ is a support vector.

Sometimes $\alpha_j > 0$ is used instead to define the term *support vector*.

So a specialty of the SVM is that only a fraction of the data is needed to describe the final solution. Interestingly, $w$ could be split into the difference of two prototypes where each corresponds to one class:

$$w = \sum_{j:y_j=1,\alpha_j>0} \alpha_j x_j - \sum_{j:y_j=-1,\alpha_j>0} \alpha_j x_j = w_{+1} - w_{-1}. \tag{1.19}$$

So especially for the L1 case, the prototypes are in its core only the average of the samples of one class which are difficult to distinguish from samples of the other class. In the L2 case, it is a weighted average. When looking at the implementation details in Section 1.2, it turns out that weights are higher if it is more difficult to distinguish the sample from the opposite class.

Additionally to implementation aspects in Section 1.2, the results of this section will be also used in the following to introduce kernels. Here, the weighted average of samples is not used anymore. It is replaced by a weighted sum of functions.

### 1.1.1.2 Loss Functions, Regularization Terms, and Kernels

This section introduces three important concepts in the context of SVMs which are also used in other areas of machine learning like regression, dimensionality reduction, and classification (without SVM variants). They are already a first set of (known but loose) connections in the form of parameter mappings between SVM variants. They will be repeatedly referred to in the other sections.

**Loss Functions**    First, we will have a closer look at the $t_j$ in the C-SVM models. Instead of using $t_j$, it is also possible to omit the side constraints by replacing $t_j$ in the target function of the model with the function

$$\max\left\{0, 1 - y_j(\langle w, x_j \rangle + b)\right\}. \tag{1.20}$$

The underlying function $l(t) = \max\left\{0, 1 - ys\right\}$ is called *hinge loss*, where $y \in \{-1, +1\}$ is the class label and $s$ is the classification score. The respective squared function for the L2–SVM is called *squared hinge loss*. In case of the hard margin SVM a $t_j$ or a respective replacing function could be introduced by defining

$$t_j = \begin{cases} \infty & \text{if } 1 - y_j(\langle w, x_j \rangle + b) > 0, \\ 0 & \text{else.} \end{cases} \tag{1.21}$$

**Definition 2** (Loss Function). *The term summing up the model errors $t_j^q$ is called loss term (sometimes also empirical error). The respective function defining the error of the* *algorithm model in relation to a single sample is called loss function.*

There are several ways of choosing the loss function, each resulting in a new classifier. A (not complete) list of existing loss functions is given in Table 1.1. For some of them, there is a corresponding underlying density model [Smola, 1998]. Three choices have already been introduced and many more will be used in the following sections.

| name | function |
|---|---|
| hinge loss | $\max\{0, \xi\}$ |
| squared hinge loss | $\max\{0, \xi\}^2$ |
| Laplacian loss | $\lvert\xi\rvert$ |
| Gaussian loss | $\frac{1}{2}\xi^2$ |
| $\epsilon$ insensitive loss | $\max\{0, \xi - \epsilon, -\xi - \epsilon\}$ |
| Huber's robust loss | $\begin{cases} \frac{1}{2\sigma}\xi^2 & \text{if } \lvert\xi\rvert \le \sigma, \\ \lvert\xi\rvert - \sigma\frac{1}{2} & \text{if } \lvert\xi\rvert > \sigma \end{cases}$ |
| polynomial loss | $\frac{1}{p}\lvert\xi\rvert^p$ |
| piecewise polynomial loss | $\begin{cases} \frac{1}{p\sigma^{p-1}}\lvert\xi\rvert^p & \text{if } \lvert\xi\rvert \le \sigma, \\ \lvert\xi\rvert - \sigma\frac{p-1}{p} & \text{if } \lvert\xi\rvert > \sigma \end{cases}$ |
| LUM loss [Liu et al., 2011] | $\begin{cases} 1 - \xi & \text{if } \xi < \frac{c}{1+c}, \\ \frac{1}{1+c}\left(\frac{a}{(1+c)\xi - c + a}\right)^a & \text{if } \xi \ge \frac{c}{1+c} \end{cases}$ |
| $0 - 1$ loss | $\begin{cases} 0 & \text{if } \xi \ge -1, \\ 1 & \text{if } \xi < -1 \end{cases}$ |
| logistic loss | $\log(1 + \exp(\xi + 1))$ |

**Table 1.1: Loss functions with** $\xi := 1 - ys$. $y \in \{-1, +1\}$ is the class label and $s$ is the classification score. For some functions additional hyperparameters are used ($\sigma$, $p$, $c$, $a$, $\epsilon$).

**Regularization Terms**  If for a classification algorithm only the loss function were minimized, chances are high that it will overfit to the given data. This means that it might perfectly match the given training data but might not generalize well and that it will perform worse on the testing data. To avoid such behavior, often a *regularization term* is used — like $\frac{1}{p'}\lVert w\rVert_{p'}^{p'}$ in the C-SVM definition. Sometimes, this term is also called *prior probability* [Mika et al., 2001]. The target function of the respective algorithm is always the weighted sum of loss term and regularization term.

An advantage of using $\frac{1}{2}\lVert w\rVert_2^2$ as regularization function is its differentiability and strong convexity. When using a convex regularization and loss function, every local optimum is also a global one. Furthermore, together with the convexity of the optimization problem the strong convexity results in the effect that there is always a unique $w$ solving the optimization problem [Boyd and Vandenberghe, 2004]. This does not hold for 1-norm regularization ($p' = 1$, $\|w\|_1 = \sum |w_i|$) where there could be more than one optimal solution. $p' \in \{1, 2\}$ are the most common choices for regularization. The most common case is $p' = 2$, due to its intuitiveness and its nice properties in the duality theory setting (see Section 1.1.1.1 and Section 1.2). The advantage of the regularization with $p' = 1$ is its tendency to sparse solutions.[6] Some more information about this behavior is given in Section 1.3.3.4. If $w$ can be split into vectors $w^{(1)}, \ldots, w^{(k)}$ the terms $\|w\|_{1,2} := \sum \left\|w^{(i)}\right\|_2$ and $\|w\|_{1,\infty} := \sum \left\|w^{(i)}\right\|_\infty$ are sometimes used to induce grouped sparsity [Bach et al., 2012]. This means, that the classifier tends to completely set some components $w^{(i)}$ to zero vectors.

**Kernels**  In Theorem 2 it could be shown in the case of $p' = 2$ that the C-SVM problem can be reformulated to only work on the pairwise scalar products of the training data samples $x_j$ and not the single samples anymore. This is used in the *kernel trick*, where the scalar product is replaced by a *kernel function* $k$. This results in a nonlinear separation of the data which is very advantageous, if the data is not linearly separable. The respective classification function becomes

$$f(x) = b + \sum_{i=1}^{n} \alpha_j k(x, x_i). \tag{1.22}$$

The most common kernel functions are displayed in Table 1.2. For some applications like text or graph classification, a kernel function is directly applied to the data without the intermediate step of creating a feature vector.

| name | function |
|---|---|
| linear kernel | $\langle x_i, x_j \rangle$ |
| polynomial kernel | $(\gamma \langle x_i, x_j \rangle + b)^d$ |
| sigmoid kernel | $\tanh(\gamma \langle x_i, x_j \rangle + b)$ |
| Gaussian kernel (RBF) | $\exp\left(-\frac{\|x_i - x_j\|_2^2}{2\sigma^2}\right)$ |
| Laplacian kernel | $\exp\left(-\frac{\|x_i - x_j\|_2}{\sigma}\right)$ |

Table 1.2: **Kernel functions** applied to the input $(x_i, x_j)$. The other variables in the functions are additional hyperparameters to customize/tune the kernel for the respective application.

The use of the kernel can be compared with the effect of lifting the data to a higher dimensional space before applying the separation algorithm. Instead of defin-ing the lifting, only the kernel function has to be defined. As a direct mathematical consequence, a kernel function is usually required to be a symmetric, positive semidefinite, and continuous function. These requirements are also called Mercer conditions. By furthermore restricting this function to a compact space (e.g., each vector component is only allowed to be in a bounded and closed interval) the Mercer theorem can be applied.

---

[6] More components of $w$ are mapped to zero.

**Theorem 3** (Mercer Theorem [Mercer, 1909]). *Let $X$ be a compact set and $k : X \times X \to \mathbb{R}$ be a symmetric, positive semi-definite, and continuous function. Then there exists an orthonormal basis $e_i \in L^2(X)$ and non-negative eigenvalues $\lambda_i$ such that*

$$k(s,t) = \sum_{j=1}^{\infty} \lambda_j e_j(s) e_j(t) \,. \tag{1.23}$$

Now using $\Phi = \mathrm{diag}(\sqrt{\lambda}) \, (e_1, e_2, \ldots)$, we get

$$k(a,b) = \langle \Phi(a), \Phi(b) \rangle \; \forall a,b. \tag{1.24}$$

Consequently, $\Phi$ is a mapping in a high dimensional space, were the standard scalar product is used. The proof of the Mercer Theorem also gives a rule on how to construct the basis. This rule uses the derivatives of the kernel function. Hence, for the linear and polynomial kernel a finite basis can be constructed but especially for the Gaussian kernel, which is also called radial basis function (RBF) kernel, only a mapping into an infinite dimensional space is possible because the derivative of the exponential function never vanishes.

Instead of the previous argument using the dual optimization problem, the following *representer theorem* is also used in the literature to introduce kernels.

**Theorem 4** (Nonparametric Representer Theorem [Schölkopf et al., 2001a]). *Suppose we are given a nonempty set $X$, a positive definite real-valued kernel $k$ on $X \times X$, a training sample $(x_1, y_1), \ldots, (x_n, y_n) \in X \times \mathbb{R}$, a strictly monotonically increasing real-valued function $g$ on $[0, \infty)$, an arbitrary cost function $c : (X \times \mathbb{R}^2)^n \to \mathbb{R} \cup \{\infty\}$, and a class of functions*

$$F = \left\{ f \in \mathbb{R}^X \, \middle| \, f(\cdot) = \sum_{i=1}^{\infty} \beta_i k(\cdot, z_i), \beta_i \in \mathbb{R}, z_i \in X, \|f\| < \infty \right\}. \tag{1.25}$$

*Here, $\|\cdot\|$ is the norm in the reproducing kernel Hilbert space (RHKS) $H_k$ associated with $k$, i.e. for any $z_i \in X, \beta_i \in \mathbb{R} \; (i \in \mathbb{N})$,*

$$\left\| \sum_{i=1}^{\infty} \beta_i k(\cdot, z_i) \right\|^2 = \sum_{i=1}^{\infty} \sum_{j=1}^{\infty} \beta_i \beta_j k(z_i, z_j). \tag{1.26}$$

*Then any $f \in F$ minimizing the regularized risk functional*

$$c((x_1, y_1, f(x_1)), \ldots, (x_n, y_2, f(x_n))) + g(\|f\|) \qquad (1.27)$$

*admits a representation of the form*

$$f(\cdot) = \sum_{i=1}^{n} \alpha_j k(\cdot, x_i). \qquad (1.28)$$

For the C-SVM, the decision function $f$ is optimized in contrast to the classification vector $w$. The cost function $c$ is used for the loss term and $g$ is used for the regularization term $\frac{1}{2}\|f\|^2$. The theorem states that $f$ can be replaced in the optimization problem with a finite sum using only the training samples. The result is the same as the previous approach for introducing the kernel.

No matter which way is chosen to introduce the kernel, the *kernel trick* can be applied to most of the SVM variants with 2-norm regularization introduced in the following except the online passive aggressive algorithm, because it does not keep the samples in memory.

Even after building its model, the SVM has to keep the training data (only the support vectors) for the classification function when using nonlinear kernels. In this case the size of the solution (usually) grows with the size of the training data. Such a type of model is called non-parametric model. In contrast, when using the linear kernel the SVM provides a parametric model of the data with the parameters $w$ and $b$, because the number of parameters is independent from the size of the training data. The usage of linear and RBF kernel is not unrelated but there is an interesting connection.

**Theorem 5** (RBF kernel generalizes linear kernel for the C-SVM)**.** *According to [Keerthi and Lin, 2003], the linear C-SVM with the regularization parameter $C'$ is the limit of a C-SVM with RBF kernel and hyperparameters $\sigma^2 \to \infty$ and $C = C'\sigma^2$.*

This theorem was used by [Keerthi and Lin, 2003] to suggest a hyperparameter optimization algorithm, which first determines the optimal linear classifier and then uses the relation $C = C'\sigma^2$ to reduce the space of hyperparameters to be tested for the RBF kernel classifier. It could be also used into the other direction. If the optimal $C$ and $\sigma^2$ become too large, the linear classifier with $C' = \frac{C}{\sigma^2}$ could be considered instead. Consequently, the connection between the two variants can be directly used to speed up the hyperparameter optimization and also to somehow optimize the choice of the best variant.

### 1.1.1.3 $\nu$-**Support Vector Machine**

The hyperparameter $C$ in the L1-SVM model influences the number of support vectors but this influence cannot be mathematically specified.[7] The $\nu$ support vector machine ($\nu$-SVM) has been introduced with a different parametrization of the C-SVM to be able to provide a lower bound on the number of support vectors in relation to the number of training samples [Schölkopf et al., 2000, Crisp and Burges, 2000]:

**Method 4** ($\nu$-Support Vector Machine ($\nu$-SVM))**.**

$$\begin{aligned} \min_{w,t,\rho,b} \quad & \tfrac{1}{2}\|w\|_2^2 - \nu\rho + \tfrac{1}{n}\sum t_j \\ \text{s.t.} \quad & y_j\left(\langle w,x_j\rangle + b\right) \geq \rho - t_j \text{ and } t_j \geq 0 \ \forall j : 1 \leq j \leq n \,. \end{aligned} \tag{1.29}$$

The additional hyperparameter $\nu \in [0,1]$ which replaces the $C \in (0,\infty)$ is the reason for the name of the algorithm. The original restriction $\rho' \geq 0$ is omitted for simplicity as suggested in [Crisp and Burges, 2000]. For the problem to be feasible,

$$\nu \leq \frac{\min\left\{\sum\limits_{y_j=+1} y_j, -\sum\limits_{y_j=-1} y_j\right\}}{n} \tag{1.30}$$

has to hold [Crisp and Burges, 2000]. Similar to the calculations in Section 1.1.1.1 the dual optimization can be derived (Theorem 19):

$$\begin{aligned} \min_{\alpha} \quad & \tfrac{1}{2}\sum_{i,j}\alpha_i\alpha_j y_i y_j \langle x_i,x_j\rangle \\ \text{s.t.} \quad & \tfrac{1}{n} \geq \alpha_j \geq 0 \qquad \qquad \forall j : 1 \leq j \leq n, \\ & \sum_j \alpha_j y_j = 0, \sum_j \alpha_j = \nu \,. \end{aligned} \tag{1.31}$$

Due to the restrictions, $\nu$ defines the minimum percentage of support vectors used from the training data. If there is no $\alpha \in \left(0, \tfrac{1}{n}\right)$, then $\nu$ is the exact percentage of support vectors and not just a bound.

**Theorem 6** (Equivalence between C-SVM and $\nu$-SVM)**.** *If $\alpha(C)$ is an optimal solution for the dual of the C-SVM with hyperparameter $C > 0$, the $\nu$-SVM has the same solution with $\nu = \frac{1}{Cl}\sum \alpha_i(C)$ despite a scaling with $Cl$.*

*On the other hand, if $\rho$ is part of the optimal solution of a $\nu$-SVM with $\nu > 0$ and a negative objective value, by choosing $C = \frac{1}{\rho l}$ the C-SVM provides the same (scaled) optimal solution.*

The proof and further details on this theorem can be found in [Chang and Lin, 2001].

---

[7] Especially since this parametrization largely depends on the scaling/normalization of the data.

### 1.1.1.4  Support Vector Regression

Parallel to the C-SVM, the support vector regression (SVR) has been developed
[Vapnik, 2000, Smola and Schölkopf, 2004]. In the literature, the name *Support Vec-*
*tor Machine* is sometimes also used for the SVR. For a better distinction, the name
Support Vector Classifier (SVC) is sometimes used for the C-SVM.  As the name
indicates, SVR is a regression algorithm ($y_j \in \mathbb{R}$) and not a classifier.  The formal
definition is:

**Method 5** (L1–Support Vector Regression (SVR))**.**

$$
\begin{aligned}
\min_{w,b,t} \quad & \tfrac{1}{2}\|w\|_2^2 + C\sum s_j + C\sum t_j \\
s.t. \quad & \epsilon + s_j \geq \langle w, x_j \rangle + b - y_j \;\; \geq -\epsilon - t_j \quad \forall j : 1 \leq j \leq n \\
& \qquad\qquad\qquad s_j, t_j \;\; \geq 0 \qquad\qquad \forall j : 1 \leq j \leq n.
\end{aligned}
\tag{1.32}
$$

The additional hyperparameter $\epsilon$ defines a region, where errors are allowed. Due
to this $\epsilon$-tube the SVR tends to have few support vectors which are at the border
or outside of this tube.  Using a squared loss or "hard margin" loss, other regular-
ization, or kernels works for this algorithm as well as for the C-SVM.  According to
[Smola and Schölkopf, 2004], the dual optimization problem is

$$
\begin{aligned}
\min_{\alpha,\beta} \quad & \tfrac{1}{2}\sum_{i,j}(\alpha_i - \beta_i)(\alpha - \beta)\langle x_i, x_j\rangle - \sum_j y_j(\alpha_j - \beta_j) + \epsilon\sum_j (\alpha_j + \beta_j) \\
s.t. \quad & 0 \leq \alpha_j \leq C \quad \forall j : 1 \leq j \leq n \\
& 0 \leq \beta_j \leq C \quad \forall j : 1 \leq j \leq n \\
& \sum(\alpha_j - \beta_j) = 0.
\end{aligned}
\tag{1.33}
$$

It is connected to the primal optimization problem via

$$
w = \sum_j (\alpha_j - \beta_j) = 0.
\tag{1.34}
$$

Apart from the underlying theory from statistical learning (using regularization,
loss term, and kernels) and the fact that the solution also only depends on a subset
of samples called support vectors, there seems to be no direct intuitive connection
between SVR and C-SVM. Nevertheless, the existence of a parameter mapping could
be proven by [Pontil et al., 1999] in the following theorem:

**Theorem 7** (Connection between SVR and C-SVM)**.** *Suppose the classification prob-*
*lem of the C-SVM in Method 3 is solved with regularization parameter $C$ and the*
*optimal solution is found to be $(w, b)$.  Then, there exists a value $a \in (0, 1)$ such that*
*$\forall \epsilon \in [a, 1)$, if the problem of the SVR in Method 5 is solved with regularization param-*
*eter $(1 - \epsilon)C$, the optimal solution will be $(1 - \epsilon)(w, b)$.*

*Proof.* The proof by [Pontil et al., 1999] will not be repeated, here. Instead, in Section 1.3 this theorem will become immediately clear by introducing a third classifier (balanced relative margin machine) which is connected intuitively to SVR *and* C-SVM. As a consequence the choice of $a$ will be geometrically motivated. $\qquad\square$

As already demanded by [Pontil et al., 1999], there is also a $\nu$-SVR similar to the $\nu$-SVM in Section 1.1.1.3 [Schölkopf et al., 2000, Smola and Schölkopf, 2004].

**Method 6** ($\nu$-Support Vector Regression)**.**

$$\begin{aligned}
\min_{w,b,t} \quad & \tfrac{1}{2}\|w\|_2^2 + C\left(n\nu\epsilon + \sum s_j + \sum t_j\right) \\
s.t. \quad & \epsilon + s_j \geq \langle w, x_j \rangle + b - y_j \geq -\epsilon - t_j \quad \forall j : 1 \leq j \leq n \\
& s_j, t_j \geq 0 \qquad \forall j : 1 \leq j \leq n.
\end{aligned} \tag{1.35}$$

$\nu$-SVR and SVR are connected similar to $\nu$-SVM and C-SVM [Chang and Lin, 2002]. Interestingly, this time $\nu$ is not replacing $C$. Instead, it is indirectly replacing $\epsilon$ which is now a model parameter and not a hyperparameter anymore. $\nu$ provides a weighting for the automatic selection of $\epsilon$. This is reasonable, because in the SVR model, $\epsilon$ has the main influence on the number of support vectors.[8] In contrast to the $\nu$-SVM model, the proof for the existence of solutions for the C-SVM can be directly transferred to the $\nu$-SVR.

We recently suggested a novel "variant" of the SVR for creating a regression of the upper/lower bound of a mapping with randomized real-valued output [Fabisch et al., 2015]. It is called positive upper boundary support vector estimation (PUBSVE). Further details are provided in Appendix B.5.

## 1.1.2 Least Squares Support Vector Machine

Using the *Gaussian loss* (which is also called least squares error) *in the SVM model* instead of the (squared) hinge loss directly results in the least squares support vector machine (LS-SVM) [Suykens and Vandewalle, 1999]. This change in the loss function is substantial and results in a very different classifier:[9]

**Method 7** (Least Squares Support Vector Machine LS-SVM)**.**

$$\begin{aligned}
\min_{w,b,t} \quad & \tfrac{1}{2}\|w\|_2^2 + \tfrac{C}{2}\sum t_j^2 \\
s.t. \quad & y_j(\langle w, x_j \rangle + b) = 1 - t_j \quad \forall j : 1 \leq j \leq n.
\end{aligned} \tag{1.36}$$

Note that this classifier is the exact counterpart to *ridge regression*[10]

---

[8] A smaller $\epsilon$-tube where model errors are allowed, result in more errors and each of the corresponding samples is a support vector.

[9] The difference will become clear in Section 1.3.

[10] More details are provided in Appendix B.2.1.

[Hoerl and Kennard, 1970, Saunders et al., 1998]. The motivation of this classifier was to solve a "set of linear equations, instead of quadratic programming for classical SVM's" [Suykens and Vandewalle, 1999, p. 1]. This comes at the prize of using all samples for the solution (except the ones with $x_j \in H_{y_j}$) in contrast to having few support vectors. Consequently, the method might be disadvantageous when working with kernels on large datasets, because the kernel function needs to be applied to every training sample and the new incoming sample which shall be classified.

For solving the optimization problem of the classifier the use of Lagrange multiplier is not necessary, but it enables the use of kernels and the comparability with the C-SVM. The application of it can be justified analogously to Theorem 1. The respective Lagrange function is

$$L(w, b, t, \alpha) = \frac{1}{2} \|w\|_2^2 + \frac{C}{2} \sum t_j^2 - \sum \alpha_j (y_j(\langle w, x_j \rangle + b) - 1 + t_j). \tag{1.37}$$

In contrast to the formulation of the L2–SVM, it holds $\alpha_j \in \mathbb{R}$. Setting the derivative of $L$ to zero results in the equations:

$$w = \sum_j \alpha_j y_j x_j \,, \tag{1.38}$$

$$0 = \sum_j \alpha_j y_j \,, \tag{1.39}$$

$$t_j = \frac{\alpha_j}{C} \qquad\qquad \forall j : 1 \le j \le n \,, \tag{1.40}$$

$$1 = y_j(\langle w, x_j \rangle + b) + t_j \qquad\qquad \forall j : 1 \le j \le n \,, \tag{1.41}$$

which are sufficient for solving the problem [Suykens and Vandewalle, 1999]. Substituting the first and third equation into the fourth equation and introducing a kernel function $k$ reduces the set of equations to a set of $(n + 1)$ equations with $(n + 1)$ variables:

$$0 = \sum_j \alpha_j y_j, \tag{1.42}$$

$$1 = y_j b + \frac{\alpha_j}{C} + \sum_i \alpha_i y_i y_j k(x_i, x_j) \qquad\qquad \forall j : 1 \le j \le n. \tag{1.43}$$

With a very large $n$ this set of equations might become too difficult to solve and a special quadratic programming approach might be required as suggested for the C-SVM (see Section 1.2), which does not require to compute and store all $k(x_i, x_j)$.

### 1.1.3 Regularized Fisher's Discriminant

The LS-SVM is also closely connected to the regularized Fisher's discriminant as outlined in this section. In Section 1.3 we will show that are a special cases of a more general classifier.

Originally, the Fisher's discriminant (FDA) is defined as the optimal vector $w$ that maximizes the ratio of variance between the classes and the variance within the classes after applying the linear classification function:

$$w \in \arg\max_a \frac{(a^T(\mu_2 - \mu_1))^2}{a^T(\Sigma_2 + \Sigma_1)a}\,. \tag{1.44}$$

Here, $\mu_i$ and $\Sigma_i$ are the mean and variance of the training data from class $i$, respectively. We can see that every positive scaling of $w$ is a solution, too. Further, in terms of the linear classification functions $f(x) = \langle w, x \rangle + b$, the definition of the FDA does not impose any constraints on the choice of the offset $b$. These ambiguities are the reason why different reformulations of the original problem can be found in the literature. For a good comparison with the C-SVM we need the following equivalent definition [Van Gestel et al., 2002, Mika, 2003]:

$$\min_{w,b} \sum_{j=1}^n \left(\langle w, x_j \rangle + b - y_j\right)^2 \tag{1.45}$$

where the offset $b$ is integrated into the optimization and where a scaled $w$ is not a solution anymore.. This method is also called Minimum Squared Error method or Least Squares method. In [Duda et al., 2001] a similar model was derived but with a fixed offset.

For normal distributed data with equal covariance matrices for both classes but different mean, the FDA is known to be the Bayes optimal classifier [Mika et al., 2001]. Motivated by the concept of Bayesian priors, Mika suggests to have an additional regularization term in the target function [Mika, 2003]:

**Method 8** (Regularized Fisher's Discriminant (RFDA))**.**

$$\begin{aligned}\min_{w,b,t} \quad &Reg(w,b) + C\,\|t\|_2^2 \\ s.t. \quad &\langle w, x_j \rangle + b = y_j + t_j \,\forall j : 1 \leq j \leq n.\end{aligned} \tag{1.46}$$

Here the variable $t$ is used to describe the loss with the help of restrictions as in the C-SVM model. For the RFDA this is not necessary but it will help us for the comparison with other methods.

**Theorem 8** (Equivalence of LS-SVM and RFDA)**.** *Using $Reg(w,b) = \frac{1}{2}\,\|w\|_2^2$ as regularization results in the* least squares support vector machine*.*

*Proof.*  Direct consequence of the definitions.                                    □

Mika also suggests to introduce kernels for the kernel Fisher discriminant with regularization (KFD) by replacing $w$ with $\sum_j \alpha_j y_j x_j$ ($\alpha \in \mathbb{R}$) and by replacing the resulting scalar products with a kernel function. For the regularization Mika suggests to apply a regularization directly on $\alpha$. Using $\|\alpha\|_1$ for example as regularization term results in sparse solutions in the kernel space [Mika et al., 2001]. A similar approach was also mentioned for SVMs [Mangasarian and Kou, 2007].[11]

In [Mika, 2003] it is also mentioned that non-Gaussian distribution assumptions result in other loss terms. Further choices like Laplacian loss (for Laplacian noise) will be examined in Section 1.3.

### 1.1.4   Relative Margin Machine

The following classifier is the basis of a novel classifier which generalizes most of the already introduced classifiers (see Section 1.3).

The relative margin machine (RMM) from [Shivaswamy and Jebara, 2010] extended the C-SVM by an additional *outer margin* that accounts for the spread of the data and adds a data dependent regularization:

**Method 9** (Relative Margin Machine (RMM))**.**

$$
\begin{aligned}
\min_{w,b,t} \quad & \tfrac{1}{2}\|w\|_2^2 + C\sum t_j \\
s.t. \quad & y_j(\langle w, x_j\rangle + b) \geq 1 - t_j \quad \forall j : 1 \leq j \leq n \\
& \tfrac{1}{2}(\langle w, x_j\rangle + b)^2 \leq \tfrac{R^2}{2} \quad \forall j : 1 \leq j \leq n \\
& t_j \geq 0 \quad \forall j : 1 \leq j \leq n.
\end{aligned}
\tag{1.47}
$$

The additional hyperparameter $R$ in this method constrains the maximum distance a sample can have from the decision plane in relation to the length of the classification vector $w$; $R$ is called *range* in the following. The real distance is $R \cdot \frac{1}{\|w\|}$.[12] Thus, it provides an additional *outer margin* at the hyperplanes $H_R$ and $H_{-R}$, which is dependent on the inner margin.

**Definition 3** (Relative Margin)**.** *The relative margin is the combination of the inner and the outer margin.*

The range has to be either chosen manually or automatically, and we always assume $R \geq 1$, as by definition $\pm 1$ are the borders of the inner margin. The classifier scheme is depicted in Figure 1.4. Further details on motivation and variants of this classifier are the content of Section 1.3.

---

[11] The C-SVM regularization term with kernels is: $\sum_{i,j} \alpha_i \alpha_j y_i y_j k(x_i, x_j)$.

[12] Note, $\frac{2}{\|w\|}$ is the distance between the aforementioned maximum margin hyperplanes.

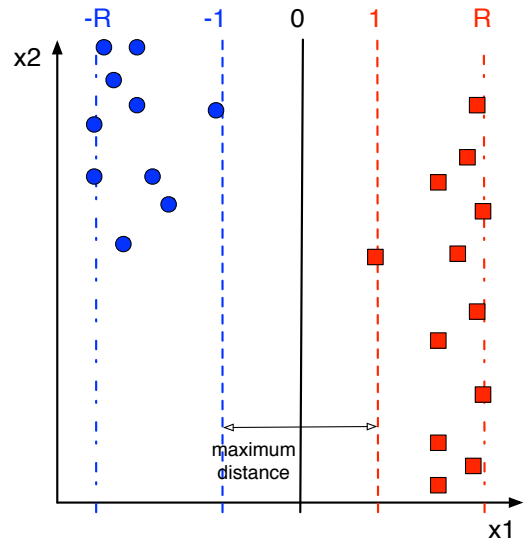

Figure 1.4: **Relative margin machine scheme.** There are two new hyperplanes, $H_R$ and $H_{-R}$, to define the outer margin in contrast to Figure 1.2.

### 1.1.5 Online Passive-Aggressive Algorithm

The passive-aggressive algorithm (PAA) was motivated by the loss functions of the C-SVM and the use of a regularization term [Crammer et al., 2006]. All three versions of the loss term were considered: hard margin, hinge, and squared hinge loss. The resulting algorithms are denoted by PA, PA-I, and PA-II respectively [Crammer et al., 2006]. In contrast to the C-SVM, the PAA is an *online learning* classifier (see also Section 1.2). It uses one single sample at a time, adapts its classification function parameter $w$ and then it forgets the sample.

In the single update step of the PAA the loss function to be minimized is a function of only one incoming training sample and instead of the norm of the classification vector $w$ the distance between the old and new classification is minimized which was an idea taken from [Helmbold et al., 1999]:

$$w_{t+1} = \operatorname{argmin}_{w \in \mathbb{R}^m} \frac{1}{2} \|w - w_t\|_2^2 + Cl(w, x_t, y_t). \tag{1.48}$$

The loss function $l$ is the same as used for the C-SVM with hard margin, hinge, or squared hinge loss. Note that no offset is used. To incorporate one, Crammer suggests to use an extra component in $w$ for the $b$ and also extend the data to homogeneous coordinates with an additional 1 which results in the classification function $f(x) = \langle (w, b), (x, 1) \rangle$. Consequently, the offset is also subject to minimization in the target function. The introduced optimization problem is always feasible even with hard margin loss and Lagrange duality can be applied. In contrast to the C-SVM, concrete solution formulas can be derived for the different losses [Crammer et al., 2006]. The detailed algorithm description is provided in Figure 1.5.

**INPUT:** aggressive parameter $C > 0$
**INITIALIZE:** $w_1 = (0, \dots, 0)$
For $t = 1, 2, \dots$
- receive instance: $x_t \in \mathbb{R}^m$
- predict: $\hat{y}_t = \text{sgn} \langle w_t, x_t \rangle$
- receive correct label: $y_t \in \{-1, +1\}$
- suffer loss: $l_t = \max \{0, \, 1 - y_t \langle w_t, x_t \rangle\}$
- update:
    1. set:

$$\alpha_t = \frac{l_t}{\|x_t\|^2} \qquad\qquad \text{(PA)}$$

$$\alpha_t = \min \left\{ C, \frac{l_t}{\|x_t\|^2} \right\} \qquad\qquad \text{(PA-I)}$$

$$\alpha_t = \frac{l_t}{\|x_t\|^2 + \frac{1}{2C}} \qquad\qquad \text{(PA-II)}$$

    2. update: $w_{t+1} = w_t + \alpha_t y_t x_t$

Figure 1.5: Online passive-aggressive Algorithm (PAA) as described in Section 1.1.5 and [Crammer et al., 2006].

The PAA is even more connected to the C-SVM as it seems at first sight. This is shown in Section 1.2.4.

### 1.1.6   Unary Classification

Instead of a classification with two classes (binary classification) some classifiers focus only on one class (unary classification) even though a second class might be present from the application point of view. The reason for omitting this second class might be the desire to model only the properties of one class and not of a second one or the lack of data as it is the case for outlier or novelty detection [Aggarwal, 2013]. A more detailed motivation for unary classification will be given in Section 1.4. In the following, we will discuss three SVM variants for unary classification algorithms

#### 1.1.6.1   Support Vector Data Description

For constructing a classifier with the data from a single class and not two classes, the support vector data description (SVDD) is a straightforward approach. Its concept is to find a hypersphere with minimal radius which encloses all samples of one class. It is assumed that samples outside this hypersphere do not belong to the class [Tax and Duin, 2004].

**Method 10** (support vector data description (SVDD))**.**

$$\begin{aligned} \min_{R',c,t'} \quad & R'^2 + C' \sum t'_j \\ s.t. \quad & \|c - x_j\|_2^2 \leq R'^2 + t'_j \text{ and } t'_j \geq 0 \; \forall j : 1 \leq j \leq n \; . \end{aligned} \tag{1.49}$$

$R'$ is the radius of the enclosing hypersphere with center $c$. The decision function is

$$F(x) = \text{sgn}\left(R'^2 - \|c - x\|_2^2\right). \tag{1.50}$$

The SVDD could be also seen as a SVM variant [Tax, 2001, Tax and Duin, 2004] and it can be used with kernels, too. In case of using kernels, the set of support vectors also tends to be small, because samples inside the hypersphere are no support vectors.

### 1.1.6.2 Unary Online Passive-Aggressive Algorithm

The concept of the SVDD to enclose the data with a hypersphere was also used to define the unary PAA [Crammer et al., 2006]. Instead of the hinge loss with its hard margin and squared version (see Section 1.1.5), the "SVDD loss" is considered:

$$l_R(c, x) = \begin{cases} 0 & \text{if } \|c - x\| \leq 0, \\ \|c - x\| - R & \text{otherwise.} \end{cases} \tag{1.51}$$

The same optimization problem is solved as for the binary PAA to determine a new center $c$ with a new incoming sample:

$$c_{t+1} = \text{argmin}_{c \in \mathbb{R}^m} \frac{1}{2} \|c - c_t\|_2^2 + C l(c_t, x_t) \tag{1.52}$$

where $l$ forces $l_R$ to be zero (hard margin, respective algorithm denoted with unary PA), or $l = l_R^q$ with $q \in \{1, 2\}$ (soft margin, unary PAq). The processing scheme is similar to the method reported in Figure 1.5. But with the different loss, the respective update factors are:

$$\alpha_t = l_R(c_t, x_t) \text{ (PA0)}, \; \alpha_t = \min\{C, l_R(c_t, x_t)\} \text{ (PA1)}, \; \alpha_t = \frac{l_R(c_t, x_t)}{1 + \frac{1}{2C}} \text{ (PA2)}, \tag{1.53}$$

and the update formula is:

$$c_{t+1} = c_t + \alpha_t \frac{x_t - c_t}{\|x_t - c_t\|}. \tag{1.54}$$

In contrast to the SVDD, the hyperparameter $R$ has to be chosen beforehand. For extending the method with an automatic tuning of $R$ an upper bound $R_{\max}$ has to be defined instead. The radius $R$ is now indirectly optimized by extending the center $c$ with an additional component $c^{(m+1)}$ which is initialized with $R_{\max}$. It is then related to the optimal $R$ by $R = \sqrt{R_{\max}^2 - (c^{(m+1)})^2}$. The respective data gets an additional component with the value zero. For further details we refer to [Crammer et al., 2006, section 6].

### 1.1.6.3   Classical One-Class Support Vector Machine

The classical one-class support vector machine ($\nu$oc-SVM) has been introduced as a tool for "estimating the support of a high-dimensional distribution" [Schölkopf et al., 2001b, title of the paper].

**Method 11** (One-Class Support Vector Machine ($\nu$oc-SVM))**.**

$$
\begin{aligned}
\min_{w,t,\rho} \quad & \tfrac{1}{2}\|w\|_2^2 - \rho + \tfrac{1}{\nu l}\sum t_j \\
\text{s.t.} \quad & \langle w, x_j\rangle \geq \rho - t_j \text{ and } t_j \geq 0 \; \forall j
\end{aligned}
\tag{1.55}
$$

*with the decision function*

$$
F(x) = \operatorname{sgn}\left(\langle w, x\rangle - \rho\right) .
\tag{1.56}
$$

Again, there is a hidden binary classification included via the decision function, namely whether a sample belongs to the one class or not. The dual of the $\nu$oc-SVM [Schölkopf et al., 2001b],

$$
\begin{aligned}
\min_{\alpha} \quad & \tfrac{1}{2}\sum_{i,j}\alpha_i\alpha_j \langle x_i, x_j\rangle \\
\text{s.t.} \quad & 0 \leq \alpha_i \leq \tfrac{1}{\nu l} \; \forall i \text{ and } \sum \alpha_i = 1 ,
\end{aligned}
\tag{1.57}
$$

is quite similar to the dual of the $\nu$-SVM after a scaling of the dual variables with $\nu$. Only the equation $\sum_{j}\alpha_j y_j = 0$ is missing. This equation cannot be fulfilled for a unary classifier, because it holds $y_j = 1 \,\forall j : 1 \leq j \leq n$. This similarity and its consequences will be analyzed in detail in Section 1.4.

# 1.2 Single Iteration:
## From Batch to Online Learning

This section contains my findings from:

Krell, M. M., Feess, D., and Straube, S. (2014a). Balanced Relative Margin Machine –
The missing piece between FDA and SVM classification. *Pattern Recognition Letters*,
41:43–52, doi:10.1016/j.patrec.2013.09.018
and

Krell, M. M. and Wöhrle, H. (2014). New one-class classifiers based
on the origin separation approach. *Pattern Recognition Letters*, 53:93–99,
doi:10.1016/j.patrec.2014.11.008.

No text parts are taken from these publications.

So far, we only defined the optimization problem for the C-SVM and its numerous
variants. To really use these models, there is still an approach required to at least
approximately solve the optimization problems which will be covered in this section.
It is not straightforward, because there is no closed form solution.[13] Furthermore, it
is important to have algorithms which scale well with the size of the dataset to make
it possible to build a model with the help of an arbitrarily large set of training data.[14]
The implementation approaches are transferred to other classifiers in the following
sections but they also provide a connection between C-SVM and PAA.

Similar to the number of SVM variants, there are also several approaches for
solving the optimization problem. In this section, we focus on a few approaches
which finally lead, with the help of the *single iteration approach*, to an algo-
rithm that operates on an arbitrary large set of training data at the price of accu-
racy. The drop of accuracy results from simplifications of the original optimization
problem. These simplifications are required to speed up the solution algorithms.
For example, the use of kernels will be finally omitted, because with increasing
size of the data an increasing size of support vectors is expected [Steinwart, 2003,
Steinwart and Christmann, 2008]. This also increases the amount of required mem-
ory and time for the prediction which in some applications might be inappropriate.

The C-SVM is categorized as a batch learning algorithm. This means that it
requires the complete set of training data to build its model. The opposite category
would be online learning classifiers like the PAA in Section 1.1.5. These classifiers
incrementally update their classification model with the incoming single training
samples and do not use all training data at once. With each sample, they perform
an update of their model parameters which have a fixed size and do not increase with an increasing number of samples.

---

[13] A single formula which allows to calculate the model parameters at once.

[14] In most cases, the performance of classifiers improves with an increasing amount of training data.

The advantage for the application is not only to have an algorithm which can be trained on arbitrarily large datasets but it also gives the possibility to adapt the model at runtime when the model is used and to update the model with new training samples (online application). In this scenario, samples are classified with the help of the current classifier model and the classification has some impact on a system. Due to resulting actions of the system or other verification mechanisms, the true label of the sample is determined a posteriori.[15] This feedback is then used for updating the online learning algorithm. Consequently, online learning algorithms are expected to work sufficiently fast in the update step and the classification step, such that both steps can be used during an application. A big advantage of online learning algorithms in such online applications is that they can adapt to changing conditions which might result in drifts in the data. Those drifts might not have occurred when acquiring the initial training data [Quionero-Candela et al., 2009].

Assume for example an algorithm running on a robot with a camera, which uses images to detect the soil type of the environment to avoid getting stuck or wet. It is impossible to have a complete training set which accounts for every situation, e.g., light condition, temperature, color of the underground, or a water drop on the camera. So the respective classification algorithm might make wrong predictions. Now as the robot is walking or driving over the ground it might detect the underground very accurately by measuring pressure on the feet and slippage. Consequently, it could adapt the image classification algorithm with the help of the afterwards detected labels.[16] For this adaption an online learning algorithm would be required with strict limitations on the resources because it has to run on the robot. If the classification or the adaptation is too slow the robot might have to stop to wait for the results to decide where to go. Furthermore, the computational resources on a robot are usually low to save space and energy and provide longterm autonomy.

Another application is the (longterm) use of EEG in embedded brain reading [Kirchner and Drechsler, 2013, Kirchner, 2014] where the operator shall not be limited in his movement space. Here, a BCI is used to infer the behavior of the human and to adapt an interface to the human. Thereby it is taking false predictions into account. For example, an exoskeleton can lessen its stiffness when the EEG classifier predicts an incoming movement [Kirchner et al., 2013, Seeland et al., 2013a], or a control scenario can repeat warnings less often if the classifier detects that the warning has been cognitively perceived. EEG data is known to be non-stationary. So online-learning can improve the system as shown in [Wöhrle et al., 2015]. Getting true labels is ensured by the concept of embedded brain reading. The real behavior of the subject can be compared with the inferred behavior and the classification process can be adapted. Furthermore, it is useful to have the complete processing on a small mobile device with low power consumption [Wöhrle et al., 2014] to ease the applicability. So here again the properties of efficient online learning are needed.

---

[15] Not in every application such a verification is possible. Sometimes unsupervised approaches are used which for example assume that the classified label was correct and can be used for the update.

[16] Note that a simultaneous localization and mapping (SLAM) algorithm is required for the matching between images, positions, and sensors.

A reason to look at an online version of the C-SVM in this context was that in our practical experience the batch learning algorithm performed well on the data in the offline evaluation due to its good generalization properties. In the application, we could show that an online classifier can have performance comparable to the original algorithm [Wöhrle et al., 2013b, Wöhrle and Kirchner, 2014]. It can even improve in the application [Tabie et al., 2014, Wöhrle et al., 2015] since the fast updates can be used for online adaptation. With the batch algorithm this is impossible if new samples come in too fast or if too much memory is consumed when all training samples are kept.

One approach to give a SVM online learning properties is not to start the learning of the model from scratch but to use a warm start by initializing an optimization algorithm with the old solution from a previous update step [Laskov et al., 2006, Steinwart et al., 2009]. This approach also works with kernels. Unfortunately, an increasing amount of time for calculating the decision function is required if the number of support vectors is increasing. Furthermore, the memory consumption increases linearly with each incoming data sample. Another approach to cope with this issue is to use the warm start approach but also include a decreasing step to the update step where the amount of data, which is kept, is reduced to keep memory consumption constant [Gretton and Desobry, 2003, Van Vaerenbergh et al., 2010]. Nevertheless, a high amount of memory and processing is still required for these approaches and an evaluation is required in the future to compare the different approaches and to analyze there properties in online applications.

The motivation of this section is to provide a more general approach to derive online learning algorithms not only for the C-SVM but also for its variants and to understand the relations between the different solvers and the different underlying classifier models. A short summary on the approaches and the respective section where they are discussed is given in Table 1.3. For a detailed analysis of the benefits of online learning in the context of the P300 dataset (Section 0.4), we refer to [Wöhrle et al., 2015]. In the experiment in Section 2.4.6 it can be seen clearly that online learning can improve classification performance, when using the online SVM introduced in Section 1.2.4.

| Approach | Samples Per Update Step | Repeated Iterations | References |
|---|---|---|---|
| Newton optimization | all | yes | [Chapelle, 2007], Section 1.2.1 |
| SMO | 2 | yes | [Platt, 1999a], Section 1.2.2 |
| successive overrelaxation, dual gradient | 1 | yes | [Mangasarian and Musicant, 1998], |
| descent, | 1 | yes | [Hsieh et al., 2008], |
| omit offset | 1&2 | yes | [Steinwart et al., 2009], Section 1.2.3 |
| PAA, | 1 | no | [Crammer et al., 2006], |
| single iteration | 1 | no | [Krell et al., 2014a], Section 1.2.4 |

Table 1.3: **Overview on SVM solution approaches grouped by similarity.** They are required because they lead to the single iteration approach. All the algorithms basically consist of an update step, where the classifier model is updated to be more optimal concerning the chosen samples, and in some cases (batch learning) they have an iteration loop over the complete set of samples with certain heuristics.

## 1.2.1   Newton Optimization

This section introduces a straightforward solution approach for the C-SVM optimization problem as a summary of [Chapelle, 2007].

For solving the C-SVM optimization problem directly, it is advantageous to directly put the side constraints into the target function to get an unconstrained optimization problem:

$$\min_{w,b} \frac{1}{2} \|w\|_2^2 + C \sum_j \left( \max\left\{0, 1 - y_j(\langle w, x_j\rangle + b)\right\}\right)^q, \ q \in \{1, 2\}. \quad (1.58)$$

This approach is slightly different to penalty methods because the hyperparameter $C$ remains fixed and is not iteratively increased. The second step by Chapelle was to introduce a kernel into the primal optimization problem:

$$\min_a \frac{1}{2} \sum_{i,j} a_i a_j k(x_i, x_j) + C \sum_j \left( \max\left\{0, 1 - y_j\left(\sum_i a_i k(x_i, x_j) + b\right)\right\}\right)^q. \quad (1.59)$$

This can be either done directly, with the representer theorem (Theorem 4), or by transforming the dual problem with kernel back to the primal problem. Note that there is no restriction on the weights $a_i$ in contrast to the dual variables $\alpha_i$ in Theo-rem 2. The classification function is: $f(x) = \sum\limits_i a_i k(x_i, x) + b$.

The third step is to repeatedly calculate the gradient ($\nabla$) and the Hessian ($H$) of the target function and perform a newton update step [Boyd and Vandenberghe, 2004]:

$$a \rightarrow a - \gamma H^{-1} \nabla \tag{1.60}$$

with step size $\gamma$. For the detailed formulas of the derivatives refer to [Chapelle, 2007]. Note that the loss functions are pieced together by other functions and there is no second derivative at the intersection points. So in this method, the one sided second derivative is used for the Hessian which makes it a *quasi*-Newton method. Furthermore, the algorithm also exploits results from the optimality conditions by setting the weights to zero if the respective sample is classified without any error (zero loss). If $\gamma \neq 1$ is chosen, this trick is required. Otherwise, all samples in the training data could become support vectors which would increase computational complexity and slow down convergence. The matrix inversion is usually replaced with the solution of linear equation and it is possible to use a sparse approximation of $H$ to save processing time. But this method might have memory problems if the number of samples is too large. Furthermore, the hinge loss has to be replaced with the approximation

$$L(y, t) = \begin{cases} 0 & \text{if } -\xi > h, \\ \frac{(\xi + h)^2}{4h} & \text{if } |\xi| \leq h \\ \xi & \text{if } \xi > h \end{cases} \quad \text{with } \xi = 1 - yt \tag{1.61}$$

and the offset $b$ is omitted [Chapelle, 2007] although it might be possible to derive the respective formulas with the offset. A special treatment of the offset is also common for other solution approaches (see Section 1.2.3).

Chapelle also states that "from a machine learning point of view there is no reason to prefer the hinge loss anyway" but does not provide a proof or reference to support this claim. A special argument for working with the hinge loss is that it tends to work on a smaller set of support vectors in contrast to using the squared hinge loss. This has not been analytically proven but there are indicators from statistical learning theory [Steinwart, 2003], and in fact Chapelle proved empirically that using his version, the SVM tends to use more support vectors and is inferior to Sequential Minimal Optimization (introduced in Section 1.2.2). Having fewer support vectors is important when working with kernels because it speeds up the processing in the classification step. Furthermore, when online learning is the goal, having fewer support vectors can speed up the update steps.

## 1.2.2 Sequential Minimal Optimization

The C-SVM optimization problem is traditionally solved with sequential minimal optimization (SMO) [Platt, 1999a] as implemented in the LibSVM library [Chang and Lin, 2011]. It is briefly described in this section.

Its principle is to reduce the dual optimization problem as good as possible and then iteratively solve the reduced problems. The dual optimization problem reads:

$$\min_{C \geq \alpha_j \geq 0, \sum \alpha_j y_j = 0} \frac{1}{2} \sum_{i,j} \alpha_i \alpha_j y_i y_j k(x_i, x_j) - \sum_j \alpha_j. \tag{1.62}$$

At the initialization all $\alpha_j$ are set to zero. The smallest optimization problem requires to choose two dual variables $\left(\text{e.g., } \alpha_1^{\text{old}}, \alpha_2^{\text{old}}\right)$ for an update to keep the equation $\sum \alpha_j y_j = 0$ valid in the update step. Now, all variables are kept fixed except these two and the respective optimization problem is solved analytically considering all side constraints. Due to the equation in the constraints, one can focus on the update of $\alpha_2^{\text{old}}$ and later on calculate

$$\alpha_1^{\text{new}} = \alpha_1^{\text{old}} + y \left(\alpha_2^{\text{old}} - \alpha_2^{\text{new}}\right) \tag{1.63}$$

where $y = y_1 y_2$. The borders for $\alpha_2^{\text{new}}$ are

$$L = \max \left\{0, \alpha_2^{\text{old}} + y \alpha_1^{\text{old}} - \frac{1+y}{2} C\right\} \text{ and } H = \min \left\{C, \alpha_2^{\text{old}} + y \alpha_1^{\text{old}} - \frac{1-y}{2} C\right\}. \tag{1.64}$$

Following [Platt, 1999a], the first step is to solve the unconstrained optimization problem which results in:

$$\alpha_2^{\text{opt}} = \alpha_2^{\text{old}} - \frac{y_2 \left(f^{\text{old}}(x_1) - y_1 - f^{\text{old}}(x_2) + y_2\right)}{2k(x_1, x_2) - k(x_1, x_1) - k(x_2, x_2)} \text{ with } f^{\text{old}}(x) = \sum_j \alpha_j^{\text{old}} y_j k(x_j, x) + b. \tag{1.65}$$

A final curve discussion shows that this unconstrained optimum has to be projected to the borders to obtain the constrained optimum:

$$\alpha_2^{\text{new}} := \begin{cases} L & \text{if} & \alpha_2^{\text{opt}} < L, \\ \alpha_2^{\text{opt}} & \text{if } L \leq & \alpha_2^{\text{opt}} \leq H, \\ H & \text{if} & \alpha_2^{\text{opt}} > H. \end{cases} \tag{1.66}$$

Now, the two variables are changed to their optimal value. Then a new pair is chosen and the optimization step is repeated until a convergence criterion is reached.

The expensive part in the calculation is to get the function values $f^{\text{old}}(x_i)$. When working with the linear kernel, this step can be simplified by tracking $w$ (initialized

with zeros):

$$w^{\text{new}} = w^{\text{old}} + y_1 \left( \alpha_1^{\text{new}} - \alpha_1^{\text{old}} \right) x_1 + y_2 \left( \alpha_2^{\text{new}} - \alpha_2^{\text{old}} \right) x_2. \qquad (1.67)$$

Now $f^{\text{old}}(x_1) - f^{\text{old}}(x_2)$ can be replaced by $\left\langle w^{\text{old}}, x_1 - x_2 \right\rangle$.

The remaining question is on how to choose the pair of dual variables for each update. Instead of repeatedly iterating over all available pairs, different heuristics can be used which rely on the error $f^{\text{old}}(x_i) - y_i$, and which try to maximize the expected benefit of an update step. Note that this method only requires to store the weights and access the training data sample wise. Nevertheless for speed up, caching strategies are used which store kernel products and error values, especially for samples with $0 < \alpha_j < C$. For further details, we refer to [Platt, 1999a, Chen et al., 2006]. The SMO principle can be also applied to other SVM variants like for example SVR [Smola and Schölkopf, 2004] or L2–SVM instead of L1–SVM which was handled in this section. A similar approach has also been applied for RMM [Shivaswamy and Jebara, 2010].

### 1.2.3 Special Offset Treatment

This section discusses simplifications of the SMO approach which only require the choice of a single index for an update and not a heuristic for choosing a pair of dual variables for an update. The approach also operates on the dual optimization problem. The simplifications are an important preparative step for the single iteration approach in Section 1.2.4. Furthermore, the same approach will be applied to other classifiers in the following sections.

When working with kernels there are simplifications where the offset $b$ in the decision function is omitted [Steinwart et al., 2009] as also mentioned in Section 1.2.1, or it is integrated in the data space using homogenous coordinates [Mangasarian and Musicant, 1998, Hsieh et al., 2008]. The approach is advantageous in case of linear separation functions as implemented in the LIBLINEAR library [Fan et al., 2008]. In this case, the solution algorithm iterates over single samples and updates the classification function parameters $w$ and $b$ of the decision function $\text{sgn}(\langle w, x \rangle + b)$ to the optimal values in relation to this single sample. We mainly follow the dual gradient descent approach from [Hsieh et al., 2008] in this section. The resulting formulas are the same as by the successive overrelaxation approach in [Mangasarian and Musicant, 1998] or the one-dimensional update step in [Steinwart et al., 2009, Cristianini and Shawe-Taylor, 2000].[17]

The reason for the simplification of the offset treatment is to get rid of the equa-tion $\sum \alpha_j y_j = 0$ in the dual optimization which resulted from the differentiation of the Lagrange function with respect to the offset $b$ (Equation (1.11)). Without this equation in the dual optimization problem, a similar approach as presented in Section 1.2.2 could be used but only one dual variable has to be chosen for one update step. If the offset is omitted ($b \equiv 0$), the dual becomes

$$\min_{C_j \geq \alpha_j \geq 0} \frac{1}{2} \sum_{i,j} \alpha_i \alpha_j y_i y_j k(x_i, x_j) - \sum_j \alpha_j \text{ for the L1 loss and} \tag{1.68}$$

$$\min_{\alpha_j \geq 0} \frac{1}{2} \sum_{i,j} \alpha_i \alpha_j y_i y_j k(x_i, x_j) - \sum_j \alpha_j + \frac{1}{4} \sum_j \frac{\alpha_j^2}{C_j} \text{ for the L2 loss.} \tag{1.69}$$

---

[17] This equivalence has not yet been reported.

To regain the offset in the simplified primal model with ($b \equiv 0$), the regularization $\frac{1}{2} \|w\|_2^2$ is replaced by $\frac{1}{2} \|w\|_2^2 + H^2 \frac{1}{2} b^2$ with an additional hyperparameter $H > 0$ which determines the influence of the offset to the target function. A calculation shows that this approach can be transformed to the previous one where the offset is omitted:

$$\|w\|_2^2 + H^2 b^2 = \|(w, Hb)\|_2^2, f(x) = \langle w, x \rangle + b = \left\langle (w, Hb), \left( x, H \frac{1}{H^2} \right) \right\rangle. \tag{1.70}$$

Only $w$ is replaced by $(w, Hb)$ and $x$ by $\left( x, \frac{1}{H} \right)$. The formula for the decision function and the optimal $w$ remain the same as in SMO. So at the end, the kernel function $k(x_i, x_j)$ has to be replaced by $k(x_i, x_j) + \frac{1}{H^2}$ in the aforementioned dual problem and $b$ can be obtained via

$$b = \frac{1}{H^2} \sum_j y_j \alpha_j \text{ which is a result of } (w, Hb) = \sum_j y_j \alpha_j \left( x_j, \frac{1}{H} \right). \tag{1.71}$$

In short, from the model perspective solving the dual when the offset $b$ is part of the regularization is equivalent to omitting it. In the following, we focus on the latter.

The optimization problem can be reduced to updates which refer only to one sample in contrast to SMO, which requires two samples but due to the additional equation was also reduced to a single variable problem. Let $f_q$ be the target function of the dual optimization problem ($q \in \{1, 2\}$) and let $e_j$ be the j-th unit vector. For updating $\alpha_j^{\text{old}}$, we first determine the quadratic function $g_q(d) = f_q(\alpha + de_j)$:

$$g_1(d) = \frac{d^2}{2} k(x_j, x_j) + d \left( -1 + \sum_i y_i y_j \alpha_i^{\text{old}} k(x_i, x_j) \right) + \text{const. and} \tag{1.72}$$

$$g_2(d) = \frac{d^2}{2} \left( k(x_j, x_j) + \frac{1}{2C_j} \right) + d \left( -1 + \frac{\alpha_j^{\text{old}}}{2C_j} + \sum_i y_i y_j \alpha_i^{\text{old}} k(x_i, x_j) \right) + \text{const.} \tag{1.73}$$

In a second step, the optimal $d$ is determined analytically $\left( d_{\text{opt}} = -\frac{g_q'(0)}{g_q''(0)} \right)$ and as in SMO the unconstrained optimum $\alpha_j^{\text{old}} + d_{\text{opt}}$ is projected to its feasible interval. This finally results in the update formula:

$$\alpha_j^{\text{new}} = \max\left\{0, \min\left\{\alpha_j^{\text{old}} - \frac{1}{k(x_j, x_j)}\left(-1 + \sum_i \alpha_i^{\text{old}} y_i y_j k(x_i, x_j)\right), C_j\right\}\right\} \quad (1.74)$$

in the L1 case and for the L2 case it is:

$$\alpha_j^{\text{new}} = \max\left\{0, \alpha_j^{\text{old}} - \frac{1}{k(x_j, x_j) + \frac{1}{2C_j}}\left(\frac{\alpha_j^{\text{old}}}{2C_j} - 1 + \sum_i \alpha_i^{\text{old}} y_i y_j k(x_i, x_j)\right)\right\}. \quad (1.75)$$

In some versions of this approach, there is an additional factor $\gamma$ on the descent step part[18] but as the formula shows, choosing a factor of one is the optimal choice.

This approach is also similar to stochastic gradient descent (SGD) [Kivinen et al., 2004] but here the regularization term of the SVM model is considered additionally to the loss term.

For choosing the sample of interest in each update, different strategies are possible. For example, different heuristics could be used again as compared in [Steinwart et al., 2009].[19] A more simple approach is to sort the weights, randomize them, or leave them unchanged and than have two loops. The first, outer loop iterates over all samples and updates them until a convergence criterion is reached like a maximum number of iterations or a too little change of the weights. After each iteration over all samples the inner loop is started. The inner loop is the same as the outer loop but iterates only over a subset of samples with positive dual weight. In case of L1 loss, the subset is sometimes restricted to weights $\alpha_j$ with $0 < \alpha_j < C_j$. Similar approaches are used for choosing the first sample in the SMO approach (Section 1.2.2) but due to the simplification presented in this section, the more complex heuristic for the second sample in the SMO algorithm is not needed anymore. The storage requirements of both approaches are the same and the update formula can again be simplified in the linear case by replacing

$$y_j \left\langle w^{\text{old}}, x_j \right\rangle = \sum_i \alpha_i^{\text{old}} y_i y_j k(x_i, x_j) \quad (1.76)$$

and also updating $w$ in every step with

$$w^{\text{new}} = w^{\text{old}} + \left(\alpha_j^{\text{new}} - \alpha_j^{\text{old}}\right) y_j x_j. \quad (1.77)$$

---

[18] For example, the term $\frac{1}{k(x_j, x_j)}$ is replaced by $\frac{\gamma}{k(x_j, x_j)}$ in Equation (1.74).

[19] Inspired by the heuristics for SMO, Steinwart mainly compares strategies for selecting pairs of samples.

### 1.2.4 Single Iteration: From Batch SVM to Online PAA

In this section, we introduce a possibility to derive online learning algorithms from SVM variants.

**Definition 4** (Single Iteration Approach). *The* single iteration approach *creates a variant of a classification algorithm with linear kernel by first deriving an optimization algorithm, which iterates over* single *samples to optimize the target function as in Section 1.2.3, and by second performing the update step only once. This directly results in an* online learning *algorithm.*

Consequently, we first plug Equation (1.76) into the update formula from Equation (1.74) or (1.75), respectively, and replace the kernel product $k(x_j, x_j)$ by $\|x_j\|_2^2$ which results in:

$$\alpha_j^{\text{new}} = \max\left\{0, \min\left\{\alpha_j^{\text{old}} - \frac{1}{\|x_j\|_2^2}\left(-1 + y_j\left\langle w^{\text{old}}, x_j\right\rangle\right), C_j\right\}\right\} \qquad (1.78)$$

$$\text{or } \alpha_j^{\text{new}} = \max\left\{0, \alpha_j^{\text{old}} - \frac{1}{\|x_j\|_2^2 + \frac{1}{2C_j}}\left(\frac{\alpha_j^{\text{old}}}{2C_j} - 1 + y_j\left\langle w^{\text{old}}, x_j\right\rangle\right)\right\}. \qquad (1.79)$$

Since the update step is performed only once, the $\alpha$ weights are always initialized with zero and do not have to be kept in memory but only $w$ has to be updated when a new sample $x^{\text{new}}$ with label $y^{\text{new}}$ and loss punishment parameter $C$ comes in:

$$\delta = \max\left\{0, \min\left\{-\frac{1}{\|x^{\text{new}}\|_2^2}\left(-1 + y^{\text{new}}\left\langle w^{\text{old}}, x^{\text{new}}\right\rangle\right), C\right\}\right\} \qquad (1.80)$$

$$\text{or } \delta = \max\left\{0, -\frac{1}{\|x^{\text{new}}\|_2^2 + \frac{1}{2C}}\left(-1 + y^{\text{new}}\left\langle w^{\text{old}}, x^{\text{new}}\right\rangle\right)\right\} \qquad (1.81)$$

$$\text{and } w^{\text{new}} = w^{\text{old}} + \delta y^{\text{new}} x^{\text{new}}. \qquad (1.82)$$

**Theorem 9** (Equivalence between passive-aggressive algorithm and online classical support vector machine). *The PAA can be derived from the respective SVM with the single iteration approach.*

*Proof.* This is a direct consequence, because the derived formulas are the same as for PA-I and PA-II (defined in Section 1.1.5). The equivalence for PA is derived by setting $C := \infty$ and $\frac{1}{2C} := 0$. □

Note that the single iteration approach can be also applied to related classifiers to derive online versions (see also Section 1.3 and 1.4). Another advantage is that now it is even possible to have a variant which combines batch and online learning. First, the classifier is trained on a larger dataset with batch learning (offline). In the second step, only the classification function parameter $w$ is stored and all other modeling parameters (but not the hyperparameters) can be removed from memory to save resources. It is even possible to transfer the classifier in this step to a mobile device with limited resources [Wöhrle et al., 2013b, Wöhrle et al., 2014] and use this device in the online application. Finally, the connected online learning algorithm can be used in the application when for every new incoming sample the online update formula is applied to update $w$.

This approach could be also applied to other combinations of batch and online learning classifiers, but can lead to unexpected behavior due to different properties of the classifiers (e.g., the online classifier has a different type of regularization or different underlying loss).

In contrast to SGD [Kivinen et al., 2004], the update formula should not be applied repeatedly on the same data samples, because it is always treated like new data. The old weights $\alpha$ cannot be considered, because they have not been stored. As a consequence, repeated iteration might put a weight of $2C$ to a sample even though $C$ should be the maximum from the modeling perspective.

When using the single iteration approach, it is also important to keep in mind that with the updates, the influence of a sample to the classification vector $w$ is permanent and that there is no decremental step to directly remove the sample. A possibility for compensation would be to introduce a forgetting factor $\gamma < 1$ in the update:

$$w^{\text{new}} = \gamma w^{\text{old}} + \delta y^{\text{new}} x^{\text{new}}. \tag{1.83}$$

This has also been suggested in [Leite and Neto, 2008] to avoid a growing of $\|w\|_2$ which occurred because a fixed margin approach was used in an online learning algorithm instead of approximating the optimal margin as in our approach.

### 1.2.5 Practice: Normalization and Threshold Optimization

When dealing with SVM variants, it is always important to normalize the features of the input data. The classifier relies on the relation between the features and without normalization one feature can easily dominate the others if it provides too large absolute values.

The presented special treatment of the offset assumes that a small offset or even no offset is a reasonable choice, which is for example the case when using the the RBF kernel, which is invariant under any translation of the data. Consequently, the approach of using no offset has shown comparable performance to the SMO approach [Steinwart et al., 2009]. When normalizing the data, the offset treatment should be considered. If the features are normalized to be in the interval $[0, 1]$, a negative offset is more expected than with a normalization to the interval $[-1, 1]$ . With increasing dimension of the data ($n$) in the linear case, the influence of the offset becomes less relevant when calculating $\|(w, b)^2\|$ because $w$ has the main influence. If there are only few dimensions (less than $10$), the offset treatment might cause problems when a linear kernel is used and a nonzero offset is required for the optimal separation of positive and negative samples. This can be partially compensated by using a small hyperparameter $H$ (e.g., $10^{-2}$). But if it is too small, there is the danger of rounding errors when its squared inverse is added to the scalar product of samples as mentioned in Section 1.2.3.

If the usage or evaluation of the classifier does not rely on the classification score but on the decision function it is often good to tune the decision threshold. This can also compensate for a poorly chosen offset. Furthermore, depending on the metric a different threshold will be the optimal choice. There are several algorithms for changing the threshold and also modifying the classification score [Platt, 1999b, Grandvalet et al., 2006, Metzen and Kirchner, 2011, Lipton et al., 2014, Kull and Flach, 2014].

To summarize, we presented the single iteration approach to derive online learning from batch learning algorithms like the PAA from the C-SVM. The benefit in memory and processing efficiency comes at the cost of accuracy and additional effort in normalizing the data appropriately and optimizing the decision threshold.

## 1.3   Relative Margin:
## From C-SVM to RFDA via SVR

In this section, we approach the class of relative margin classification algorithms from the mathematical programming perspective. We will describe and analyze our suggestions to extend the relative margin machine (RMM) concept [Shivaswamy and Jebara, 2010] introduced in Section 1.1.4 This will result in new methods, which are highly connected to other well known classification algorithms as depicted in Figure 1.6.

The main idea is that outliers at the new outer margin are treated in the same way as in the inner margin. Due to this *balanced* handling of outliers by the proposed method, it is called balanced relative margin machine (BRMM).

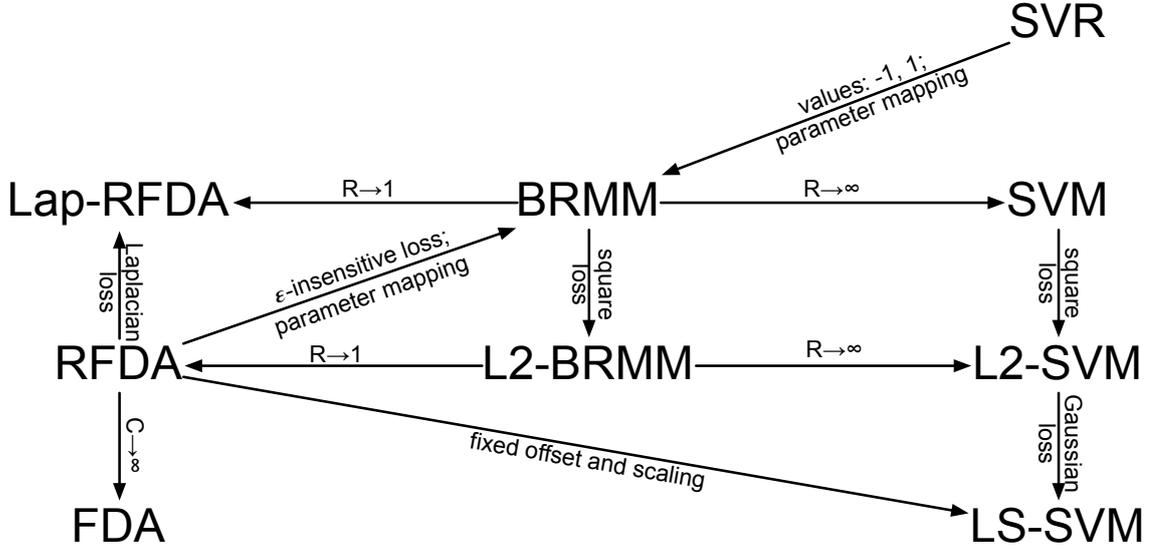

Figure 1.6: **Overview of balanced relative margin machine (BRMM) method connections.** The details can be found in Section 1.3. Visualization taken from [Krell et al., 2014a].

After further motivating the relative margin (Section 1.3.1) and introducing balanced relative margin machine (BRMM) (Section 1.3.2), we show that this model is equivalent to SVR (with the dependent variables $Y = \{-1, 1\}$) and connects C-SVM and RFDA. Though these methods are very different, they have a common rationale, and it is good to know how they are connected. Our proposed connection shows that there is a rather smooth transition between C-SVM and RFDA even though both methods are motivated completely differently. The original FDA is motivated from statistics (see Section 1.1.3) while the C-SVM is defined via a geometrical concept (see Section 1.1). Using BRMM, it is now possible to optimize the classifier type instead of choosing it beforehand. So, our suggested BRMM *interconnects* the other two methods and in that sense generalizes both of them at the same time.

Due to this relation, the way of introducing kernels, squared loss, or sparse variants is the same for this classifier as for C-SVM in Section 1.1.1.2. Additionally, we developed *a new geometric characterization of sparsity* in the number of used features for the BRMM, when used with a 1–norm regularization (Section 1.3.3.4). This finding can be transferred to RFDA and C-SVM. On the other side, the implementation techniques from Section 1.2 can be directly transferred from C-SVM to BRMM.

We finally verify our findings empirically in this section by the means of simulated and benchmark data. The goal of these evaluations is not to show the superiority of the method. This has already been mainly done in [Shivaswamy and Jebara, 2010].

The sole purpose is to show the properties of the BRMM with special focus on the transition from C-SVM to RFDA.

## 1.3.1   Motivation of the Relative Margin

There are also other motivations for using a relative margin additionally to the purpose of connecting classifiers.

### 1.3.1.1   Time Shifts in Data

The following example shows how RMM might be advantageous in comparison to C-SVM when there are drifts in particular directions in feature space. Data drifts in applications are quite common, e.g., drifts in sensor data due to noise or spatial shifts [Quionero-Candela et al., 2009]. One example of such data are EEG data, which are highly non-stationary, and often influenced by high noise levels. Another could be a changing distribution of data from a robot due to wear.

Let us assume that drifts occur mostly in directions of large spread and that the relevant information has a lower spread. In fact, drifts during the training phase increase the effective spread in the training samples themselves. Consider therefore two Gaussians in $\mathbb{R}^2$ with means $(0, -0.5)$ and $(t, 0.5)$ where $t$ changes in time. Hence, the second distribution drifts along the $x$ axis in some way. Suppose both distributions have the same variances of $\sigma_x^2 = 1$ in $x$ direction and $\sigma_y^2 = 0.1$ in $y$ direction. Figure 1.7 depicts an associated classification scenario where $t$ changes from $8$ to $6$ during the training data acquisition and from $4$ to $2$ during the test phase. It can be observed how the limitation of the outer spread of the data turns the classification plane in a direction nearly parallel to the main spread of the samples. The number of misclassifications under an ongoing drift is thus considerably smaller for RMM than for C-SVM.

We will come back to this dataset and perform an evaluation of classifiers with it in Section 1.3.4.3.

### 1.3.1.2   Affine Transformation Perspective

To give another different motivation for maximum relative margins, in [Shivaswamy and Jebara, 2010] an entirely reformulated classification problem is considered. Instead of learning an optimal classifier, it was argued that it is possible to learn an optimal affine transformation of the data such that a given classifier ($w$ and $b$ fixed) performs well and such that the transformation produces a small scatter on the data. The authors proved that such optimal transformations can be chosen to have rank one, yielding an optimization problem equivalent to a linear classification with large margin and small spread of the output at the same time. We showed that the fixation of the classifier can even be omitted and the results remain the same: choosing a suitable restricted transformation is similar to using RMM. Further details are provided in Appendix B.3.2.

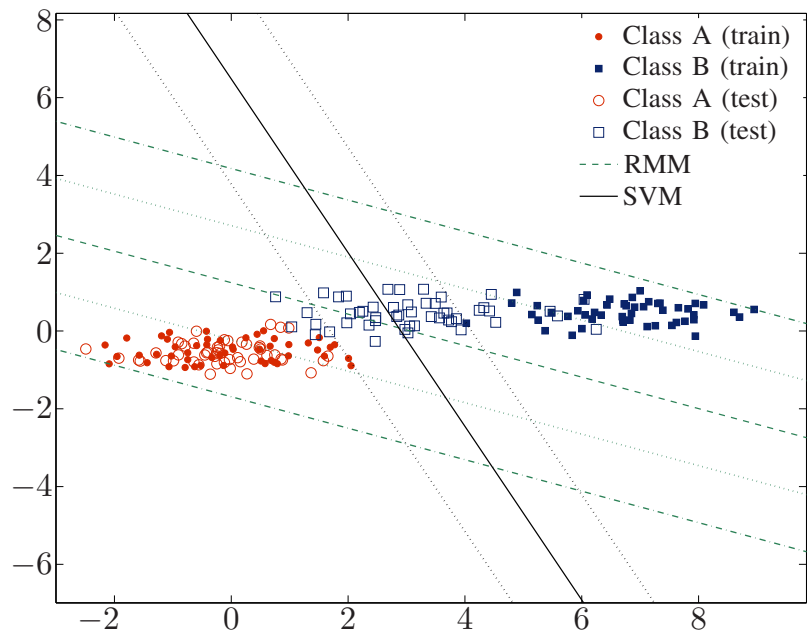

Figure 1.7: **Classification problem with drift in one component of class B.** The samples of class B are drawn from distributions with the mean of the $x$ component drifting from 8 to 6 during training and from 4 to 2 during test. The solid lines show the decision planes, the dashed lines nearby show the $\pm 1$ margins. For the RMM, the outer lines define the outer margin that limits the spread of distances to the decision plane to 2 in this case. Visualization taken from [Krell et al., 2014a].

### 1.3.2 Deriving the Balanced Relative Margin Machine

A major shortcoming of the basic RMM method is the handling of outliers at the outer margins. Such samples can in principle dominate the orientation of any separating plane, as no classification results outside the range of $\pm R$ are allowed. When working with very noisy data that might contain artifacts such outliers are very common. Two modified versions were introduced by [Shivaswamy and Jebara, 2010] to handle this insufficiency.

**Method 12** (Equation (13) from [Shivaswamy and Jebara, 2010])**.**

$$\begin{aligned}
\min_{w,b,t} \quad & \tfrac{1}{2}\|w\|_2^2 + C\sum t_j + Dr \\
s.t. \quad & r \geq y_j(\langle w, x_j\rangle + b) \quad \geq -r \qquad \forall j : 1 \leq j \leq n \\
& \quad\;\; y_j(\langle w, x_j\rangle + b) \quad \geq 1 - t_j \quad \forall j : 1 \leq j \leq n \\
& \quad\;\; t_j \quad \geq 0 \qquad\qquad \forall j : 1 \leq j \leq n.
\end{aligned}$$
(1.84)

**Method 13** (Equation (14) from [Shivaswamy and Jebara, 2010])**.**

$$\begin{aligned}
\min_{w,b,s,s',t,r} \quad & \tfrac{1}{2}\|w\|_2^2 + C\sum t_j + D(r + \tfrac{\nu}{n}\sum(s_j + s'_j)) \\
s.t. \quad & r + s_j \geq y_j(\langle w, x_j\rangle + b) \quad \geq -r - s'_j \quad \forall j : 1 \leq j \leq n \\
& \qquad\quad\; y_j(\langle w, x_j\rangle + b) \quad \geq 1 - t_j \qquad \forall j : 1 \leq j \leq n \\
& \qquad\quad\; t_j \quad \geq 0 \qquad\qquad\quad \forall j : 1 \leq j \leq n.
\end{aligned}$$
(1.85)

These methods, however, require additional variables and hyperparameters and are rather unintuitive. In the following, we propose a new variant, which is effectively similar to Shivaswamy and Jebara's variant, but at the same time considerably less complex because of fewer parameters. This makes it comparable against other classification methods and thus easier to understand. Consider at first the reformulation of Method 9:

$$\begin{aligned}
\min_{w,b,t} \quad & \tfrac{1}{2}\|w\|_2^2 + C\sum t_j \\
s.t. \quad & R \geq y_j(\langle w, x_j\rangle + b) \quad \geq -R \qquad \forall j : 1 \leq j \leq n \\
& \quad\;\; y_j(\langle w, x_j\rangle + b) \quad \geq 1 - t_j \quad \forall j : 1 \leq j \leq n \\
& \quad\;\; t_j \quad \geq 0 \qquad\qquad \forall j : 1 \leq j \leq n.
\end{aligned}$$
(1.86)

If the lowest border $-R$ is reached, $t_j$ becomes $1 + R$. As $t_j$ is subject to the minimization, this lowest border should normally not be reached. Such a high error is quite uncommon. *Therefore we drop it.* If without this border a $t_j$ became larger than $1 + R$ it either has to be considered an outlier from the modeling perspective—it has to be deleted from the data—or $R$ has been chosen too low. Both cases should not be part of the method.

After this consideration, we can introduce an outer soft margin without new variables or restrictions:

**Method 14** (L1–Balanced Relative Margin Machine (BRMM))**.**

$$\begin{aligned}
\min_{w,b,t} \quad & \tfrac{1}{2}\|w\|_2^2 + C\sum t_j \\
s.t. \quad & R + t_j \geq y_j(\langle w, x_j\rangle + b) \quad \geq 1 - t_j \quad \forall j : 1 \leq j \leq n \\
& \qquad\qquad\qquad\; t_j \quad \geq 0 \qquad\quad \forall j : 1 \leq j \leq n.
\end{aligned}$$
(1.87)

Notice that this method has one restriction less and no additional method variables or hyperparameters.[20] At the same time, it provides the same capabilities as the original method and a consistent handling of outliers. The simplicity of the method yields a high comparability to other large margin classifiers and makes it easier to implement. The name *balanced* follows from the idea to treat outliers in the outer margin in the same way as outliers in the inner margin. From our perspective, this approximation is reasonable. Depending on the application, however, this might not be appropriate. If there are reasons for different inner and outer loss, e.g., more expected outliers in the outer margin or if they are less important, the method can be adapted as follows with an additional hyperparameter but without more method variables or constraints:

$$
\begin{aligned}
\min_{w,b,t} \quad & \tfrac{1}{2}\left\|w\right\|_2^2 + \sum t_j \\
\text{s.t.} \quad & C(y_j(\langle w, x_j\rangle + b) - 1) && \geq -t_j && \forall j : 1 \leq j \leq n \\
& C'(y_j(\langle w, x_j\rangle + b) - R) && \leq t_j && \forall j : 1 \leq j \leq n \\
& t_j && \geq 0 && \forall j : 1 \leq j \leq n.
\end{aligned}
\tag{1.88}
$$

The proposed balanced version can be seen as a reasonable first approach. It is also possible to use squared loss (L2–BRMM) or a hard margin for the inner *and* the outer margin. It might be useful to use different ranges for the two classes if there are different intrinsic spreads. The only modification to the BRMM method is to replace the range by a class-specific hyperparameter $R\left(y_j\right)$. Furthermore, it is possible use the range as a variable with a new weight as hyperparameter in the target function. Both changes lead to additional hyperparameters, which complicates the hyperparameter optimization and makes the method less intuitive and less comparable to other methods.

With Method 13 Shivaswamy and Jebara introduced a variant of the $\nu$-SVM to provide a lower limit on the support vectors for the "outer margin" but it is much more reasonable to use the $\nu$ for the total number of support vectors. This can be achieved by exploiting the relation between the proposed BRMM and SVR (see Section 1.3.3.3).

### 1.3.3 Classifier Connections with the BRMM

#### 1.3.3.1 Connection between BRMM and C-SVM

The difference between C-SVM and BRMM (Methods 3 and 14) is the restriction on the classification by the range. For large values of $R$, however, this constraint becomes inactive. Hence, one can always find an $R_{\max}$ such that BRMM and C-SVM become identical for all $R \geq R_{\max}$. One approach to find this upper bound on the useful ranges from a set of training examples is to train a C-SVM on the training set. $R_{\max}$ is the highest occurring absolute value of the classification function applied on samples in the training set; every $R$ above $R_{\max}$ has no influence whatsoever. So for hyperparameter optimization, only the interval $[1, R_{\max}]$ has to be observed.

---

[20] When applying duality theory, it is more convenient to use different variables for outer and inner margin. this change has no effect on the optimal $w$ and $b$.

**Theorem 10** (BRMM generalizes C-SVM)**.** *A BRMM with $R \geq R_{\max}$ is equivalent to the C-SVM.*

As a direct consequence, values of $R$ always exist for which the BRMM, by definition, performs at least as well as the C-SVM. Depending on the available amount of training data, a good choice of $R$ might nevertheless be troublesome. The same connection to the C-SVM has already been shown for the RMM (Method 9) but not connections to the RFDA and SVR, because they do not exist. Therefore, the BRMM is necessary as discussed in the following sections.

### 1.3.3.2   Connection between BRMM and RFDA

The RFDA model (Method 8) has been introduced in Section 1.1.3. Let us focus on regularization functions of $\frac{1}{2}\|w\|_2$ and $\|w\|_1$ since we have the same regularization in BRMM and SVM approaches. Nevertheless, other regularization functions can be considered without loss of generality.

Consider now the BRMM (Method 14) with hyperparameter $R = 1$, the smallest range allowed. In this case, the inequalities of the method can be fused:

$$
\begin{aligned}
& & R + t_j \geq & & y_j(\langle w, x_j \rangle + b) & & \geq 1 - t_j & \quad\bigg|\quad R = 1 \\
\Leftrightarrow & & 1 + t_j \geq & & y_j(\langle w, x_j \rangle + b) & & \geq 1 - t_j & \quad\bigg|\quad -1 \\
\Leftrightarrow & & t_j \geq & & y_j(\langle w, x_j \rangle + b) - 1 & & \geq -t_j & \\
\Leftrightarrow & & t_j \geq & & |y_j(\langle w, x_j \rangle + b) - 1| & & & \quad\bigg|\quad |y_j| = 1 \\
\Leftrightarrow & & t_j \geq & & |(\langle w, x_j \rangle + b) - y_j|. & & &
\end{aligned}
\tag{1.89}
$$

As $t_j$ is subject to minimization, we can assume that equality holds in the last inequality:

$$
t_j = |(\langle w, x_j \rangle + b) - y_j|.
\tag{1.90}
$$

Hence, the resulting method is the same as the RFDA, except for the quadratic term $\sum t_j^2$ in the loss function of the soft margin. This difference, however, is equivalent to different noise models—linear loss functions in a RFDA correspond to a Laplacian noise model instead of a Gaussian one [Mika et al., 2001]. Conversely, a L2–BRMM can be derived from the L2–SVM.

**Theorem 11** (BRMM generalizes RFDA and LS-SVM)**.** *A BRMM with $R = 1$ is equivalent to the RFDA with Laplacian noise model (Laplacian loss, see also Table 1.1).*

*A BRMM with $R = 1$ and squared loss is equivalent to the RFDA with Gaussian noise model (Gaussian loss, see also Table 1.1). Consequently, it is also equivalent to the LS-SVM.*

In summary, both C-SVM and RFDA variants can be considered special cases of the BRMM variants, or, from a different perspective, BRMM methods interconnect the more well-established C-SVMs and RFDA, as depicted in Figure 1.6.

In [Shivaswamy and Jebara, 2010], there was a broad benchmarking of classifiers to show that in many cases the RMM performs better. The comparison also included the C-SVM and the RFDA with kernel, called regularized kernel linear discriminant analysis in this paper. With this relation it becomes now clear, why the RMM always showed comparable or better performance.

### 1.3.3.3 Connection between BRMM, $\epsilon$-insensitive loss RFDA, and SVR

As already mentioned at the end of Section 1.1.3, depending on certain assumptions on the distribution of the data, one may want to replace the loss term of the RFDA $\left( \sum t_j^2 = \|t\|_2^2 \right)$ with a different one. We already had a look at the case of assuming Laplacian noise, which results in the loss term $\|t\|_1$. We will now consider a RFDA with $\epsilon$-insensitive loss function [Mika et al., 2001]

$$\begin{aligned} \min_{w,b,t} \quad & \tfrac{1}{2} \|w\|_2^2 + C \|t\|_\epsilon \\ \text{s.t.} \quad & y_j(\langle w, x_j \rangle + b) = 1 - t_j \quad \forall j : 1 \le j \le n. \end{aligned} \tag{1.91}$$

to compare it with the BRMM. Here, $\|.\|_\epsilon$ means no penalty for components smaller than a predefined $\epsilon \in (0,1)$ and $\|.\|_1$ penalty for everything outside this region: $\|t\|_\epsilon = \sum \max\{|t_j| - \epsilon, 0\}$. This loss term is well known from support vector regression (SVR). In fact, applying SVR to data with binary labels $\{-1, 1\}$ exactly results in the $\epsilon$-insensitive RFDA. We argue that this version of RFDA or SVR is effectively equivalent to the BRMM. *This also shows that **not** the C-SVM but rather the BRMM is the binary version of the SVR.*

**Theorem 12** (Equivalence between RFDA, SVR, and BRMM). *RFDA with $\epsilon$-insensitive loss function and 2–norm regularization (or SVR reduced to the values 1 and −1) and BRMM result in an identical classification with a corresponding function, mapping RFDA (SVR) hyperparameters $(C, \epsilon)$ to BRMM hyperparameters $(C', R')$ and vice versa.*

*Proof.* By use of the mappings

$$(C', R') = \left( \frac{C}{1-\epsilon}, \frac{1+\epsilon}{1-\epsilon} \right) \text{ and } (\epsilon, C) = \left( \frac{R'-1}{R'+1}, \frac{2C'}{R'+1} \right) \tag{1.92}$$

the method definitions become equal. The mappings effectively only scale the optimization problems. The calculation is straightforward and can be found in Appendix B.2.2. So every $\epsilon$-insensitive RFDA can be expressed as BRMM and vice versa. $\qquad\square$

A direct consequence of Theorem 12 and Theorem 10 is Theorem 7 ($C = (1-\epsilon)C'$). In fact, we can directly calculate the respective border for $\epsilon$ in Theorem 7:

$$a = \frac{R_{\max} - 1}{R_{\max} + 1} = 1 - \frac{2}{R_{\max} + 1}. \tag{1.93}$$

Another positive effect is, that the $\nu$-SVR (Method 6) can be used to define a $\nu$-BRMM:

**Method 15** ($\nu$-Balanced Relative Margin Machine ($\nu$-BRMM))**.**

$$
\begin{aligned}
\min_{w,b,t} \quad & \tfrac{1}{2}\|w\|_2^2 + C\left(n\nu\epsilon + \sum s_j + \sum t_j\right) \\
s.t. \quad & \epsilon + s_j \geq y_j(\langle w, x_j\rangle + b) - 1 \quad \geq -\epsilon - t_j \quad \forall j : 1 \leq j \leq n \\
& \qquad\qquad\qquad\qquad s_j, t_j \;\geq 0 \qquad\qquad \forall j : 1 \leq j \leq n.
\end{aligned}
\tag{1.94}
$$

The replacement of $\epsilon$ in this case by $R$ is not possible because $\epsilon$ is subject to minimization and not a hyperparameter anymore. When looking at the (rescaled) dual optimization problem (derived in Appendix B.3.4), it becomes immediately clear, that $\nu$ is now a lower border on the total number of support vectors in the same way as it was the case for the $\nu$-SVM:

$$
\begin{aligned}
\min_{\alpha} \quad & \tfrac{1}{2}\sum(\alpha_i - \beta_i)(\alpha_j - \beta_j)\langle x_i, x_j\rangle y_i y_j - \tfrac{1}{Cn}\sum_j(\alpha_j - \beta_j) \\
s.t. \quad & \tfrac{1}{n} \geq \alpha_j \geq 0, \forall j : 1 \leq j \leq n, \\
& \tfrac{1}{n} \geq \beta_j \geq 0, \forall j : 1 \leq j \leq n, \\
& \sum_j \alpha_j y_j = \sum_j \beta_j y_j, \\
& \sum_j \alpha_j + \beta_j = \nu.
\end{aligned}
\tag{1.95}
$$

#### 1.3.3.4  Sparsity

So far, all considered methods shared the 2–norm in the regularization term. Particularly for C-SVM, a 1–norm regularization has been proposed [Bradley and Mangasarian, 1998]. In comparison to their 2–norm counterpart, a C-SVM with 1–norm regularization is known to operate on a reduced set of features. It can thus be regarded as a classifier with intrinsic feature selection mechanism. Omitting unimportant features can in turn render a classifier more robust. From a more practical point of view, less features might imply less sensors in the application and thus simplify the data acquisition. To achieve the same sparsity properties for BRMM, we propose to adapt the 1–norm approach to it. The resulting mathematical program can be casted to a linear one and so be solved by the Simplex algorithm [Nocedal and Wright, 2006]. For implementation, we used the GNU Linear Programming Kit [Makhorin, 2010] and directly inserted the raw model of the classifier.

Therefore, the classification function parameters are split into positive and negative components ($w = w^+ - w^-$ and $b = b^+ - b^-$) and inequality constraints are eliminated by introducing additional slack variables $g_j$ and $h_j$.

**Method 16** (1–norm Balanced Relative Margin Machine).

$$
\begin{aligned}
\min_{w^\pm, b^\pm, t, g, h \in \mathbb{R}^m_+} \quad & \sum (w_i^+ + w_i^-) + C \sum t_j \\
s.t. \quad & y_j(\langle w^+ - w^-, x_j \rangle + b^+ - b^-) & = 1 - t_j + h_j \quad \forall j : 1 \leq j \leq n \\
& y_j(\langle w^+ - w^-, x_j \rangle + b^+ - b^-) & = R + t_j - g_j \quad \forall j : 1 \leq j \leq n
\end{aligned}
\tag{1.96}
$$

Interestingly, with this method description and the properties of the Simplex algorithm [Nocedal and Wright, 2006], we proved the following:

**Theorem 13** (Feature Reduction of 1–norm BRMM). *A solution of 1–norm BRMM with the Simplex algorithms always uses a number of features smaller than the number of support vectors lying on the four margins:*

$$
\{x | \langle w, x \rangle + b \in \{1, -1, R, -R\}\}.
\tag{1.97}
$$

The formula explicitly excludes support vectors in the soft margin. This theorem is of special interest, when the dimension of the data largely exceeds the number of given samples. In this case, it can be derived that the maximum number of used features is bounded by the number of training samples.

The property of 1–norm C-SVM to work on a reduced set of features has so far only been shown empirically [Bradley and Mangasarian, 1998]. In fact, it is not possible to provide a general proof which is independent from the properties of the dataset. This can be illustrated with the help of a toy example: Consider $m$ orthogonal unit vectors in $\mathbb{R}^m$ with randomly distributed class labels. For the resulting parameters $w$ and $b$ of the 1–norm BRMM classification function we get

$$
|w_i| = 1 \; \forall 1 \leq i \leq m \text{ and } b = 0,
\tag{1.98}
$$

with a sufficiently large $C$ and arbitrary $R$. So each feature is used. Without further assumptions, better boundaries on the number of used features cannot be given.

The application of Theorem 13 to SVM and RFDA and a detailed proof are shown in Appendix B.3.1.

### 1.3.3.5   Kernels

If the common 2–norm regularization is used, the introduction of kernels is exactly the same as for C-SVM (refer to Section 1.1.1.2). The required dual optimization problem is given in Section 1.3.4.1. Interestingly, the relation between linear and RBF kernel is the same as for C-SVM in Theorem 5.

**Theorem 14** (RBF kernel generalizes linear kernel for BRMM and SVR)**.** *The linear BRMM and SVR with the regularization parameter $C'$ are the limit of the respective BRMM and SVR with RBF kernel and hyperparameters $\sigma^2 \to \infty$ and $C = C'\sigma^2$. In both cases the same range $R$ or tolerance parameter $\epsilon$ are used.*

*Proof.* The proof is the same as in [Keerthi and Lin, 2003] for Theorem 5 but mainly $\alpha - \beta$ instead of $\alpha$ and the respective dual optimization problems (see Appendix B.1.5 and Section 1.1.1.4) are used. Note that the proof highly relies on the additional equation in the dual constraints and as such cannot be applied to the algorithms versions with special offset treatment as suggested in Section 1.2.3.          □

For the 1–norm approach, the restrictions are treated the same way as in the 2–norm case, but the target function has to be changed to preserve the sparsity effect. $\langle w, x \rangle$ is replaced by $\sum \alpha_j k(x_j, x)$, where $k(.,.)$ is the kernel function, and the 1–norm of $w$ is replaced by the 1–norm of the weights $\alpha_i$. This results in a sparse solution, though sparse does not mean few features in this context but fewer kernel evaluations [Mangasarian and Kou, 2007]:

**Method 17** (1–norm Kernelized BRMM)**.**

$$
\begin{aligned}
\min_{w,b,t} \quad & \|\alpha\|_1 + C\,\|t\|_1 \\
s.t. \quad & R + t_j \geq y_j(b + \sum_{i=1}^{m} \alpha_i k(x_i, x_j)) \;\; \geq 1 - t_j \quad \forall j : 1 \leq j \leq n \\
& \qquad\qquad t_j \;\; \geq 0 \qquad \forall j : 1 \leq j \leq n,
\end{aligned}
\tag{1.99}
$$

*where $k : \mathbb{R}^n \times \mathbb{R}^n \to \mathbb{R}$ is the kernel function.*

For the special case of $R = 1$, Method 17 is equivalent to the "linear sparse kernelized Fisher's discriminant" [Mika et al., 2001].

## 1.3.4   Practice: Implementation and Applications

In this section, we will discuss the choice of the hyperparameters of BRMM, some implementations issues by also using results from Section 1.2 to derive online versions, and finally show some properties of the related classifier variants in some applications.

The BRMM has two hyperparameters: the range $R$ and the C-SVM regularization parameter $C$. Both hyperparameters are highly connected and need to be optimized.

When reducing the range $R$ from $R_{max}$ to $1$ to transfer the classifier from C-SVM over the BRMM to the RFDA, it can be observed that the number of support vectors is increasing, because in the extreme case, every sample becomes a support vector. This slows down the convergence of the optimization problem solver. To speed up the optimization in this case it is better to stick to special optimization algorithms tailored to the respective RFDA models or choose $R$ slightly larger than $1$. Furthermore, it might be a good approach to start with a large $R$ and decrease it stepwise, e.g., with a pattern search algorithm [Eitrich and Lang, 2006]. For too small $C$ the number of support vectors also becomes very large and the solution algorithm is slow. Furthermore, the performance of the respective classifier usually is not that good. Hence it is always good to start with a large $C$, e.g., $1$.

Normally, cross-validation[21] is used for hyperparameter optimization to save time. For an improved automatic optimization, it is efficient to start with high values and iteratively decrease the values with a pattern search algorithm [Eitrich and Lang, 2006]. To save resources, this could be combined with *warm start* principle to adapt the batch learning algorithms to the changed parameters [Steinwart et al., 2009]. Here the old solution is reused. With such a hyperparameter optimization, it is no longer necessary to choose between C-SVM and RFDA, because this is automatically done. Note that for the original RFDA a squared loss is required where for the C-SVM the non squared hinge loss is more common.

Using the Simplex algorithm [Nocedal and Wright, 2006] from the GNU Linear Programming Kit [Makhorin, 2010] for the sparse version of the BRMM is only possible if the problem matrix is not too large. It is possible to use other optimization algorithms but here a problem might be that these algorithms might not converge to the optimal solution or might not provide the most sparse solution. Here, some more research is needed to find a good optimization algorithm specifically tailored to the classifier model to also handle large datasets. This seems to be not that easy, because even for the 1–norm regularized, hinge loss SVM there is no implementation in the established LIBLINEAR package [Fan et al., 2008] which implements all the other linear SVM methods (with the special offset treatment trick from Section 1.2.3) and several variants. Maybe it is possible to modify the Simplex algorithm and tailor it to the sparse BRMM, use a decomposition technique as suggested in [Torii and Abe, 2009], or apply one of the many suggested algorithms for "Optimization with Sparsity-Inducing Penalties" [Bach, 2011].

---

[21] For a $k$ fold cross-validation, a dataset is divided into $k$ equal sized sets (folds), and then iteratively ($k$ times) one set is chosen as testing data and the remaining $k-1$ folds are used for training the algorithm.

### 1.3.4.1   Implementation of BRMM with 2–norm regularization

A straightforward way to use a BRMM with 2–norm without implementation is directly given by the constructive proof of Theorem 12. Using the formula

$$(\epsilon, C) = \left( \frac{R' - 1}{R' + 1}, \frac{2C'}{R' + 1} \right) \tag{1.100}$$

the SVR implementation of the LIBSVM can be directly interfaced as 2–norm BRMM algorithm. This implementation is following the SMO concept (Section 1.2.2). For implementing BRMM one can also follow the concepts from Section 1.2.3 and 1.2.4, as in the following. This will finally result in an online classifier.

For implementing the algorithm directly for BRMM models with 2–norm regularization, the dual optimization problems are used. After reintroducing separate loss variables for inner and outer loss and multiplication with $-1$, the dual problem of Method 14 reads:

$$
\begin{aligned}
\min_{\alpha, \beta} \quad & \tfrac{1}{2}(\alpha - \beta)^T Q (\alpha - \beta) - \sum \alpha_j + R \sum \beta_j \\
\text{s.t.} \quad & 0 \leq \alpha_j \leq C, 0 \leq \beta_j \leq C \quad \forall j : 1 \leq j \leq n \\
& \sum (\alpha_j - \beta_j) = 0, \\
\text{with} \quad & Q_{kl} = y_k y_l \langle x_k, x_l \rangle \quad \forall k, l : 1 \leq k \leq n, 1 \leq l \leq n.
\end{aligned} \tag{1.101}
$$

The respective dual optimization problem of L2–BRMM (squared loss) is:

$$
\begin{aligned}
\min_{\alpha, \beta} \quad & \tfrac{1}{2}(\alpha - \beta)^T Q (\alpha - \beta) - \sum \alpha_j + R \sum \beta_j + \tfrac{1}{4} \sum \frac{\alpha_j^2}{C} + \tfrac{1}{4} \sum \frac{\beta_j^2}{C} \\
\text{s.t.} \quad & 0 \leq \alpha_j, 0 \leq \beta_j \quad \forall j : 1 \leq j \leq n \\
& \sum (\alpha_j - \beta_j) = 0, \\
\text{with} \quad & Q_{kl} = y_k y_l \langle x_k, x_l \rangle \quad \forall k, l : 1 \leq k \leq n, 1 \leq l \leq n.
\end{aligned} \tag{1.102}
$$

Class dependent ranges ($R_j$) and cost parameters ($C_j$) or different regularization constants for inner and outer margin ($C, C'$) can be applied to this formulation correspondingly. For using kernels, only the scalar product in $Q$ has to be replaced with the kernel function.

As the calculation is similar to the C-SVM calculation, a similar solution approach can be used, e.g., sequential minimal optimization [Platt, 1999a, Shivaswamy and Jebara, 2010]. To follow the concept from Section 1.2.3, a classifier without the offset can be generated by dropping equation $\sum (\alpha_j - \beta_j) = 0$. For having an offset in the target function, additionally to skipping this equation, $y_k y_l$ has to be added to $Q_{kl}$.

The following algorithm now uses update formulas for $\alpha_j$ and $\beta_j$, though after each update at least one of them will be zero. Following the similar calculations in

Section 1.2.3 the update formulas are:

$$\alpha_j^{i+1} = P_j\left(\alpha_j^i - \frac{1}{Q_{jj}}\left(Q_{j\cdot}\cdot(\alpha^i - \beta^i) - 1\right)\right)$$

$$P_j(x) = \max\left\{0, \min\left\{x, C_j\right\}\right\}$$

$$\beta_j^{i+1} = P_j'\left(\beta_j^i - \frac{1}{Q_{jj}}\left(R_j - Q_{j\cdot}\cdot(\alpha^i - \beta^i)\right)\right)$$

$$P_j'(x) = \max\left\{0, \min\left\{x, C_j'\right\}\right\}. \tag{1.103}$$

To get these formulas, the hyperparameters $C$ and $R$ are replaced by the aforementioned class dependent variants $C_j$ and $R_j$. For the L2 variant, the same approach leads to:

$$\alpha_j^{i+1} = P\left(\alpha_j^i - \frac{1}{Q_{jj} + \frac{1}{2C_j}}\left(Q_{j\cdot}\cdot(\alpha^i - \beta^i) - 1 + \frac{\alpha_j}{2C_j}\right)\right)$$

$$\beta_j^{i+1} = P\left(\beta_j^i - \frac{1}{Q_{jj} + \frac{1}{2C_j'}}\left(R_j - Q_{j\cdot}\cdot(\alpha^i - \beta^i) + \frac{\beta}{2C_j'}\right)\right)$$

$$P(x) = \max\left\{0, x\right\}. \tag{1.104}$$

Independent of the chosen variant, the resulting classification function is:

$$f(x) = \sum y_j(\alpha_j - \beta_j)(\langle x_j, x\rangle + 1). \tag{1.105}$$

In the linear case, the formulas for optimal $w$ and $b$ ($w = \sum y_j\left(\alpha_j - \beta_j\right)x_j, b = \sum y_j\left(\alpha_j - \beta_j\right)x_j$) can be plugged into the update formulas [Hsieh et al., 2008]:

$$\alpha_j^{i+1} = P_j\left(\alpha_j^i - \frac{1}{Q_{jj}}\left(y_j\left(\left\langle x_j, w^i\right\rangle + b^i\right) - 1\right)\right)$$

$$\beta_j^{i+1} = P_j'\left(\beta_j^i - \frac{1}{Q_{jj}}\left(R_j - y_j\left(\left\langle x_j, w^i\right\rangle + b^i\right)\right)\right)$$

$$\left(w^{i+1}, b^{i+1}\right) = \left(w^i, b^i\right) + \left(\alpha_j^{i+1} - \alpha_j^i\right)y_j\left(x_j, 1\right) - \left(\beta_j^{i+1} - \beta_j^i\right)y_j\left(x_j, 1\right), \tag{1.106}$$

and for L2–BRMM correspondingly. Now, only the diagonal of $Q$ and the samples have to be stored/used and not the complete matrix, which makes this formula particularly useful for large scale applications.

For choosing the index $j$, there are several possibilities [Steinwart et al., 2009]. For implementation, we chose a simple one [Mangasarian and Musicant, 1998]: in an outer loop we iterate over all indices in random order and in an inner loop we just repeatedly iterate over the active indices. An index $j$ is active, when either $\alpha_j$ or $\beta_j$ is greater than zero. The iteration stops after some maximum number of iterations, or when the maximum change in an iteration loop falls below some predefined threshold. For initialization all variables $(w^0, b^0, \alpha^0, \beta^0)$ are set to zero. In the linear case, the "single iteration" approach could be used here, too. To simulate RMM we used a simplification by setting $C'_j = \infty \; \forall j$. Further details on deriving the formulas and solvability are given in Appendix B.3.3.

### 1.3.4.2   Synthetic Data: Visualization of the Relations

To illustrate the relations between BRMM, C-SVM, and RFDA by means of classification performance and to analyze the influence of the range $R$, we apply all classifiers to a synthetic dataset. Additionally, the performance difference between the original RMM with hard outer margin and BRMM with soft outer margin is investigated. For comparability, the data model is the same as employed in [Shivaswamy and Jebara, 2010]. Data are sampled from two Gaussian distributions representing two classes. The distributions have different means, but identical covariance: $\mu_1 = (1,1), \mu_2 = (19,13), \Sigma = \left( \begin{smallmatrix} 17 & 15 \\ 15 & 17 \end{smallmatrix} \right)$. It shall be noted that the Gaussian nature of the sample distributions clearly favors RFDA-like classification techniques.

Evaluation was done using a $5$–fold cross-validation on a total of $3000$ samples per class. For simplicity we used $1$–norm RMM/BRMM and fixed the regularization parameter $C$ at $0.003$. Using $2$–norm RMMs or other values for $C$ results in similar graphics.

Figure 1.8a shows the classification performance as a function of the range $R$. The first observation is that the performance does not change for $R \geq 8$. For these values of $R$, no sample lies inside the outer margin. Hence, they already have a distance less than $R$ from the separating plane, and a further increase of $R$ has no influence anymore—RMM and BRMM effectively become a C-SVM. With decreasing range the error rate drops because the classifier can better adapt to the distributions. Without outer soft margin the data are pressed into the inner soft margin. This results in worse performance for small ranges and the regularization term looses importance. In the BRMM case, i.e., with soft outer margin, the classifier gets closer and closer to a RFDA-type classifier when the range decreases. Note: The RFDA variant presented here uses a Laplacian noise model whereas the RFDA normally uses a Gaussian one. Nevertheless, we can see that the BRMM can mimic both C-SVM and RFDA. In some cases a well-chosen range can in fact constitute better classifiers *somewhere between* the two popular methods. The following section provides an example for this case.

### 1.3.4.3  Synthetic Data with Drift

We now have a look at the behavior on data with drift and the feature reduction ability. For this, we used synthetic data from the same model as used to motivate the RMM principle in Section 1.3.1.1. The data consist of samples from two two-dimensional, Gaussian distributions $\mu_1 = (0, -0.5)$, $\mu_2 = (t, 0.5)$, $\Sigma = \left( \begin{smallmatrix} 1 & 0 \\ 0 & 0.1 \end{smallmatrix} \right)$ with the same variance but different means. For the second distribution, the mean $t$ of the $x$ component changes linearly over time: from $8$ to $6$ during training and from $4$ to $2$ during the test phase. A total of $1000$ samples were computed per class in the training phase, and another $1000$ as test case. To additionally investigate how the different classifiers handle meaningless noise features, the dataset was extended by $50$ additional noise components. Each sample of these components was drawn from a uniform distribution on the unit interval. The first two components of the data, however, still resemble what is shown in Figure 1.7, only with more samples. Lastly, we generated some additional variation and outliers in the data by adding Cauchy-distributed noise ($x_0 = 0, \gamma = 0.1$) to each component. To omit too large outliers, we replaced noise amplitudes larger than $10$ by $10$.

For the classification, we used the RMM and BRMM implementations with regularization (1–norm, 2–norm) and loss (L1, L2) variants as introduced in Section 1.3.3.4 and 1.3.4.1. RFDA ($R = 1$) and SVM ($R = 8$) variants appear in the results as special cases of the BRMM methods as previously discussed.

The range $R$ was varied between $1$ and $8$. Due to the high noise, the hyperparameter $C$ had a big influence on the error and its optimal choice was highly dependent on the chosen range. Since we wanted to show the effect of the range, we kept $C$ fixed over all ranges ($0.03$ for the 2–norm and $0.002$ for the 1–norm approaches). Figure 1.8b shows the classification performance in terms of the error rate on the testing data as a function of $R$. The relatively high error rates (cf. Figure 1.8a) are due to the drift in the data and the high noise.

The 2–norm approaches operate on the complete set of features and therefore perform worse than the 1–norm approaches (e.g., minimum error of $22\%$ for 2-norm L1–BRMM). Even lower performances (not shown) were observed using a RBF kernel. The systematic drift can only be observed in two feature dimensions, but for 2–norm regularized, RBF kernel, or polynomial kernel approaches every feature has an impact on the classification function, due to the model properties. So these models are worse in handling the drift and building a good classifier, because the given classification problem clearly favors strategies, which ignore the irrelevant features. Since they do not reduce features, these approaches are very sensitive to the noise in the data.

Generally, RMMs perform worse than BRMMs because of the bad treatment of

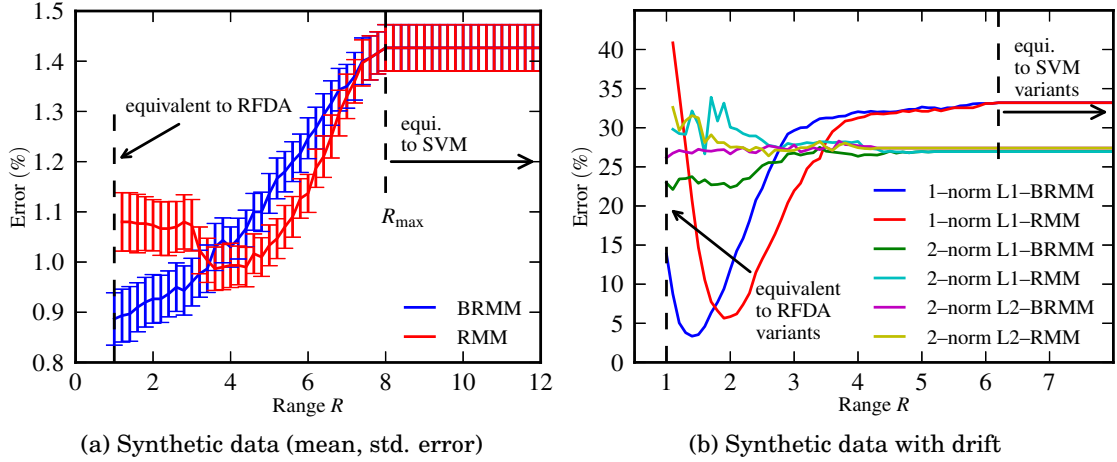

(a) Synthetic data (mean, std. error)          (b) Synthetic data with drift

Figure 1.8: **Classifier performance as function of $R$ on synthetic data.** RMM and BRMM (1–norm approaches in (a)) are compared and the transitions to corresponding RFDA and SVM variants are highlighted at the respective values of $R$ ($R = 1$ at the lower end; $R \geq 8$ (a) and $R \geq 6.2$ (b) at the upper end). Visualizations taken from [Krell et al., 2014a].

outliers at the outer margin. For different choices of the hyperparameter $C$, the results look similar.

With changing range, the errors of the 1–norm approaches show a smooth transition with clear minima around $5\%$ error rate at a range of $1.5$ for BRMM and $2.0$ for RMM. As expected, the number of features used by the 1–norm approaches is notably reduced. With ranges larger than $3$ for BRMM and $4$ for RMM, only one feature is retained. The number monotonically increases with decreasing range: while RMM uses five features for ranges lower than $2.4$, BRMM uses only the two relevant features for the lower ranges. For higher values of the hyperparameter $C$ this relation remains the same but especially for the RFDA case with $R = 1$ the numbers increase up to the total number of $52$ features. So the feature reduction ability might get lost and the classifier apparently tends to more and more overfitting and less generalization on the path from C-SVM via BRMM to RFDA.

### 1.3.4.4   MNIST: Handwritten Digit Classification

In this section, we describe a dataset which will not only be used in the following section for an evaluation but also in several other parts of this thesis.

The MNIST dataset consists of pictures of handwritten digits (0-9) of different persons with predefined train and test sets (around $60000$ and $10000$ samples) [LeCun et al., 1998]. The images in the dataset are normalized to have the numbers centered and with same size and intensity. It is an established benchmark dataset where the meaning of the data is directly comprehensible. Since it is freely available, we use it to enable reproducibility of our evaluations. Other arguments for its usage are, that it enables simple, intuitive visualizations (for the backtransformation) and provides a large set of training samples. The currently best classification result with $0.23\%$ test error rate is with a multi-column deep neural networks [Schmidhuber, 2012]. Note that this algorithm is tuned to this type of data and it cannot be seen as a pure classifier anymore because it intrinsically also learns a good representation of the data which corresponds to preprocessing, feature generation, and normalization. Hence, this algorithm tries to learn all ingredients of the decision process at once and is not comparable to classical classification algorithms which rely on a good preprocessing. Despite that, for our evaluations we are not interested in the absolute performance values but in the differences between the SVM variants.

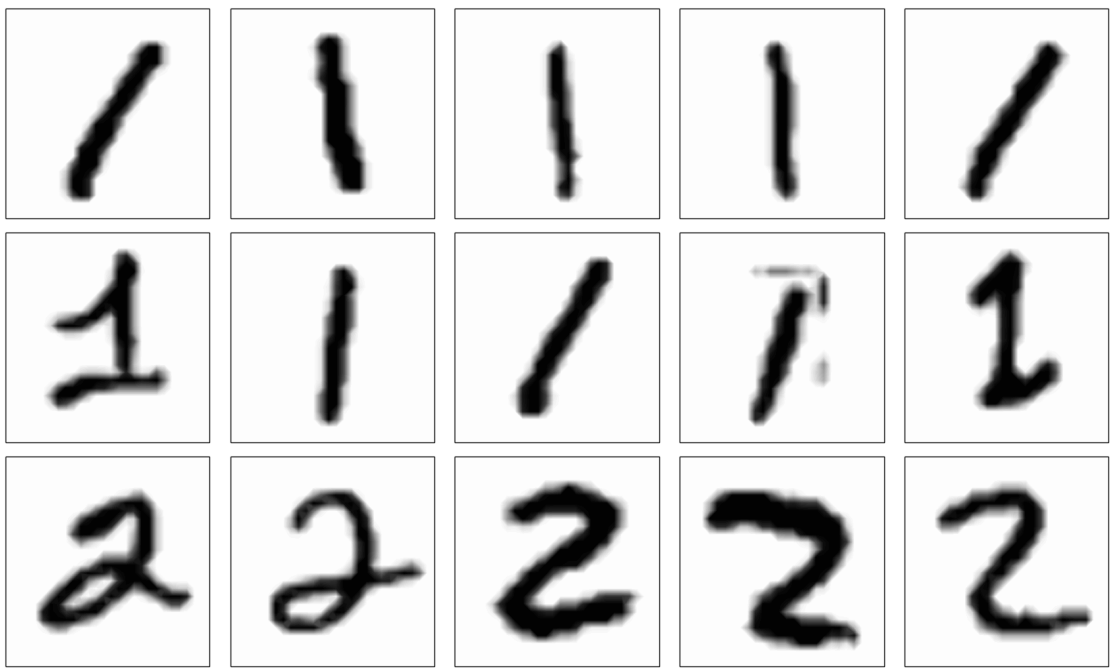

Figure 1.9: **Examples of normalized digits (1 and 2).** The original feature vector data has been mapped to the respective image format.

### 1.3.4.5  Benchmark Data: Visualization of the RFDA–SVM Relations

In this section, we verify that BRMM behaves as expected also on real world data. For this, we use the MNIST data (see Section 1.3.4.4) and a selection of IDA benchmark datasets described by [Rätsch et al., 2001]. The selection has the sole purpose of generating a comprehensible figure: we show a selection with similar error levels, so that the curve shapes are distinct.

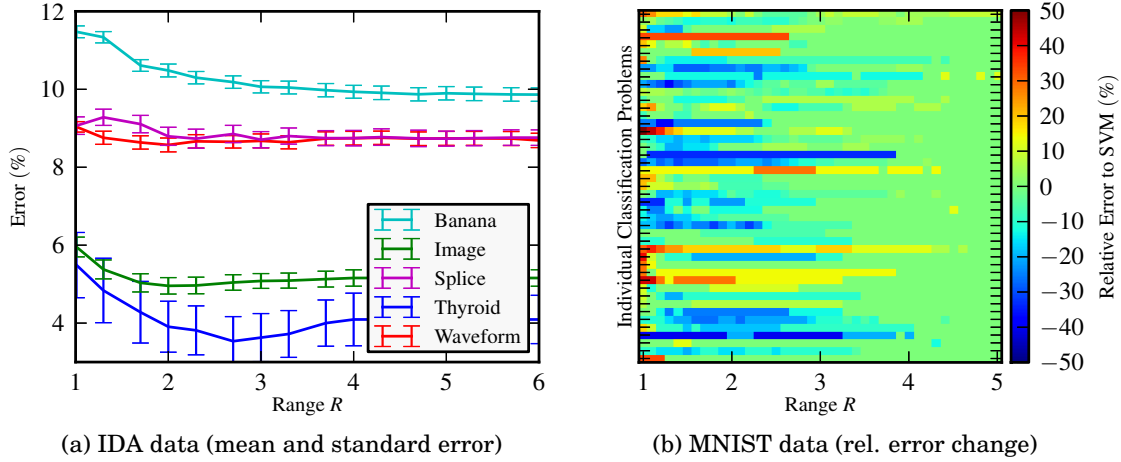

(a) IDA data (mean and standard error)   (b) MNIST data (rel. error change)

Figure 1.10: **Classifier performance as function of $R$ on benchmark data.** For the MNIST data (b) the individual results (0 vs. 1, 0 vs. 2, ... 8 vs. 9) are displayed with the percentage change of the error relative to the error of the corresponding C-SVM classifier. Visualizations taken from [Krell et al., 2014a].

We used RBF kernels and determined its hyperparameter $\gamma$ as proposed by [Varewyck and Martens, 2011]. For the IDA data classification, the regularization parameter $C$ was chosen using a 5–fold cross-validation tested with the three complexities suggested by Varewyck ($0.5, 2, 8$). On the MNIST data we fixed the $C$ to $2$ due to high computational load. As before, we visualize the performance as a function of the range parameter $R$.

For the IDA data evaluation we did a 5–fold cross-validation with five repetitions. The results are shown in Figure 1.10a. For the MNIST data, train and test data are predefined. Since BRMM is a binary classifier, we performed separate evaluations for each possible combination of two different digits, resulting in 45 classification problems for which the results are shown individually in Figure 1.10b. The results are illustrated as relative error changes compared to a C-SVM classifier to obtain comparable values for the effect of the range. This relative change in performance shown in Figure 1.10b is given by

$$\frac{\text{Error}(\text{BRMM}) - \text{Error}(\text{SVM})}{\text{Error}(\text{SVM})} \cdot 100. \qquad (1.107)$$

When looking at the performance on both the IDA results and the individual MNIST comparisons reveal that the influence of the range is highly dataset specific. For the IDA datasets the improvement using BRMM is marginal. For the MNIST data, all classifiers with a range larger than $7$ were equivalent to C-SVM. A performance improvement using the appropriate $R$ can be observed in many cases.

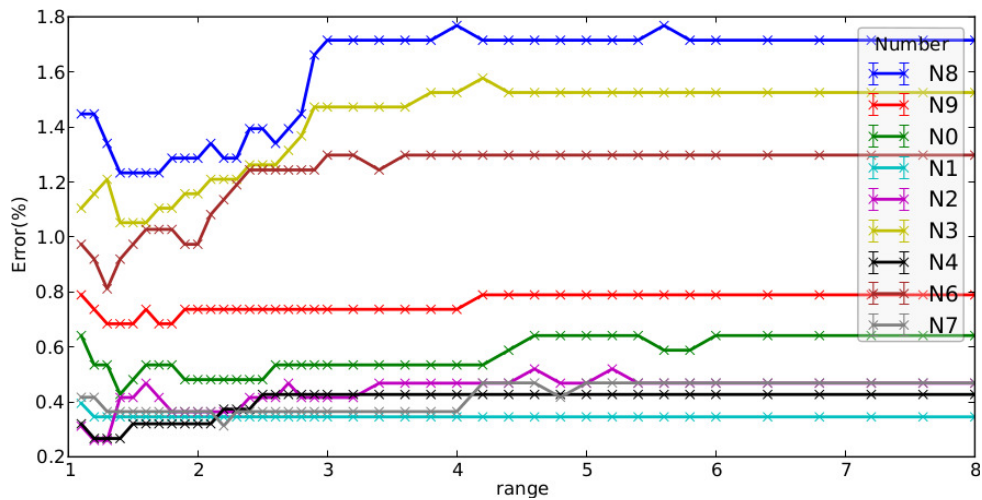

(a) MNIST data with number 5

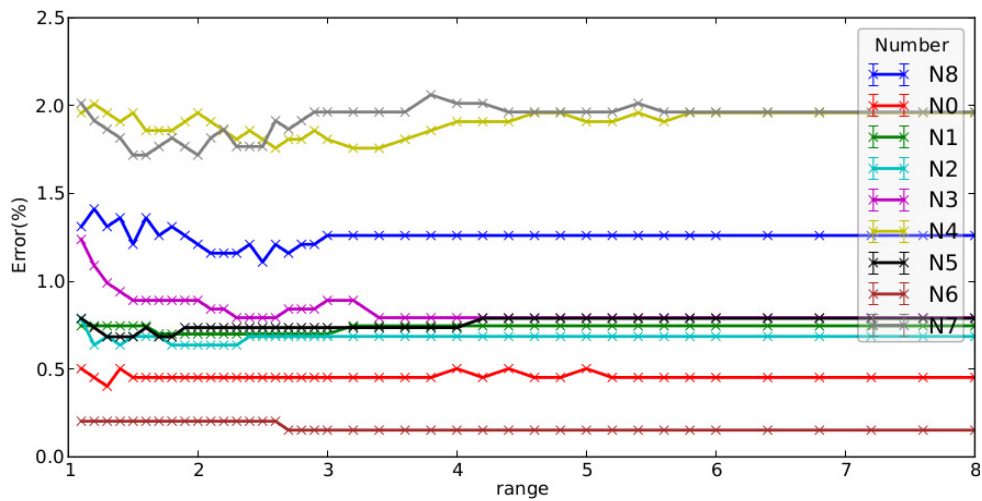

(b) MNIST data with number 9

Figure 1.11: **Classifier performance as function of $R$ on MNIST data for two special numbers.** "NX" stands for the binary classification of X with 5 or 9 respectively.

However, there are cases where the performance does not change or even decreases.

Figure 1.11 displays the single results with the real error for the binary comparison with the digit 9 and the digit 5. For the digit 9 there is mostly no change in performance but for the digit 5 there is great potential for performance improvement using BRMM.

### 1.3.4.6 Application of the BRMM to EEG Datasets

EEG data is known to be highly non-stationary due to constantly changing processes in brain and changing sensor electrode impedances [Sanei and Chambers, 2007]. To investigate the usability and the feature reduction properties of 1–norm BRMM approach in this context, we used five preprocessed EEG datasets from the P300 experiment as described in Section 0.4. No spatial filtering was used to really let the classifier do the dimensionality reduction and make this task more challenging. The signal amplitudes for each time point at each electrode were used as features, which resulted in 1612 features, which we normalized to have zero mean and variance one. For each of the remaining 5 subjects, we had two recording days with 5 repetitions of the experiment. For each of the 5 subjects and for each of the two recording days, we took one of the 5 sets for training and the remaining 4 of one day for testing. This procedure was repeated for each dataset. Each set having between 700 and 800 data samples.

For comparison, we used the classical 2–norm SVM, a 1–norm SVM and 1–norm BRMM as classifiers. Since the datasets contain an unbalanced number of samples per class (ratio $6 : 1$), we assigned the weight 8 to the underrepresented class which was good on average (cross-validation on training data). This weighting was achieved by using class specific $C_j$. The classification performance is measured by means of balanced accuracy (BA) (Figure 3.5). The BRMM range was fixed at $R = 1.5$. This value was found to be adequate for these datasets in a separate optimization on the training data. $C$ was optimized by first using a 5–fold cross validation to find a rough range of values for each classifier. The optimal hyperparameters were then automatically chosen on each individual training set, and the trained classifier was evaluated on the corresponding test set.

Table 1.4: Classification performance on EEG data

|                    | 1–norm BRMM | 1–norm SVM | 2–norm SVM |
|--------------------|-------------|------------|------------|
| balanced accuracy  | 0.872       | 0.854      | 0.857      |
| standard error     | 0.006       | 0.008      | 0.007      |
| standard deviation | 0.028       | 0.036      | 0.032      |

The results (mean of balanced accuracy) are shown in Table 1.4. Our suggested 1–norm BRMM outperforms the other classifiers significantly ($p < 0.05$, paired t-test corrected for 3 comparisons), the SVMs in turn perform on par. This indicates that a relative margin which accounts for the drifts in the data might be a better choice on EEG data. As expected, the number of features the 1–norm approaches used was notably smaller than for 2–norm SVM. 1–norm SVM used only 66–102 features, 1–norm BRMM used 101–255. This corresponds to less than 16% of the available features used by 2–norm SVM, and less than 30% of the number of examples. The increased number of features by 1–norm BRMM is expected for two reasons. The relative margin and the respective needs to be modeled with more variables. Furthermore, it is possi-ble to have a larger number of training samples at the hyperplane of the outer margin which increases the possibility of more features being used due to Theorem 13.

### 1.3.4.7 Summary

In the applications, we could show that the BRMM is a reasonable classifier which generalizes RFDA and C-SVM and provides a smooth transition between both classifiers, which can be easily fetched from the geometrical perspective. The increased performance comes with the price of the additional hyperparameter $R$ which needs to be optimized.

If a 2–norm regularization is used, the implementation of this new algorithm is straightforward by using the approaches from Section 1.2 or interfacing the existing highly efficient implementation of the equivalent SVR in the LIBSVM package.

## 1.4 Origin Separation: From Binary to Unary Classification

Focusing the classification on one class is a common approach if there are not enough examples for a second class (e.g., novelty and outlier detection [Aggarwal, 2013]), or if the goal is to describe a single target class and its distribution [Schölkopf et al., 2001b]. Some unary (one-class) classifiers are modifications of binary ones like k-nearest-neighbours [Aggarwal, 2013, Mazhelis, 2006], decision trees [Comité et al., 1999], and SVMs [Schölkopf et al., 2000, Tax and Duin, 2004, Crammer et al., 2006]. This section focuses on the connections between SVM variants, and their unary counterparts.

The $\nu$oc-SVM (see Section 1.1.6.3) was presented in [Schölkopf et al., 2001b] as a model for "Estimating the support of a high-dimensional distribution" just one year after the publication of the $\nu$-SVM [Schölkopf et al., 2000]. In both cases, the algorithms are mainly motivated by their theoretical properties and a hyperparameter $\nu$ is introduced which is a lower bound on the fraction of support vectors. It is shown in [Schölkopf et al., 2001b] that the $\nu$oc-SVM is a generalization of the Parzen windows estimator [Duda et al., 2001]. Furthermore, in the motivation of the $\nu$oc-SVM [Schölkopf et al., 2001b] the authors state that their "strategy is to map the data into the feature space corresponding to the kernel and to separate them from the origin with maximum margin". The important answer of how this strategy leads to the final model description and if there is a direct connection to the existing C-SVM or $\nu$-SVM is not given, despite similarities in the model formulations. A more concrete geometric motivation is published in [Mahadevan and Shah, 2009, p. 1628] as a side remark. They argue that "the objectives of 1-class SVMs are 2-fold:" "Develop a classifier or hyperplane in the feature space which returns a positive value for all samples that fall inside the normal cluster and a negative value for all values outside this cluster." and "Maximize the perpendicular distance of this hyperplane from the origin. This is because of the inherent assumption that the origin is a member of the faulty class." However, they did not provide a proof that the $\nu$oc-SVM fulfills these objectives and indicate that the C-SVM is the basis of this model, which is wrong. It turns out that this concept can be used as a generic approach to turn binary classifiers into unary classifiers, which is the basis of this section.

**Definition 5** (Origin Separation Approach). *In the origin separation approach, the origin is added as a negative training example to a unary classification problem with only positive training samples. With this modified data, classical binary classifiers are trained.*[22]

In Figure 1.12 the concept is visualized in the context of SVM classification. We will prove that, when applying this generic concept to the $\nu$-SVM, solutions can be mapped one-to-one to the $\nu$oc-SVM (Section 1.4.1).

Additionally to figuring out the relations between already existing unary classifiers, it is also possible to combine the origin separation with the previously introduced relative margin (see Section 1.4.2) and/or the single iteration approach (see Section 1.4.4) and generate entirely new unary classifiers.

The geometric view of the SVDD, where a hypersphere with minimal radius is constructed to include the data, is inherently different from the origin separation approach, which creates a separating hyperplane instead. Nevertheless, we will show and visualize a relation between SVM and SVDD with the help of the origin separation approach (see Section 1.4.3).

The connection between C-SVM and PAA via the single iteration approach is of special interest here. The original *unary* PAA (see Section 1.1.6.2) was motivated from the SVDD and *not* from the C-SVM which was the original motivation of the binary PAA. Based on the connection of the PAA to the C-SVM (and thus the BRMM), we apply the origin separation approach to derive new unary classifiers from C-SVM, BRMM, and RFDA for *online learning* which can be used to apply the algorithms when resources are limited. This completes the picture on PAAs given in [Crammer et al., 2006].

---

[22] A strict (hard margin) separation of the origin is required to avoid a degeneration of the classifier.

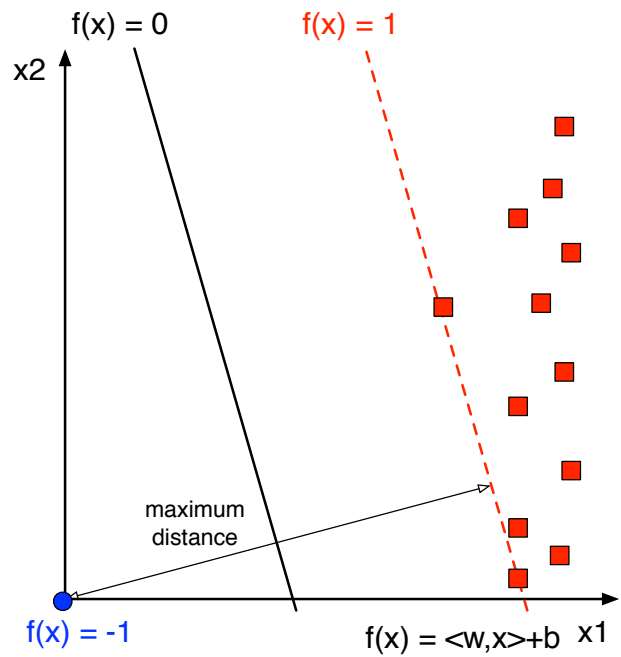

Figure 1.12: **Origin separation scheme.** An artificial sample (blue dot) for a second class ($y = -1$) is added to the origin.

Figure 1.13 visualizes the variety of resulting classifiers and their relations, which will be explained in detail in the following sections. We will focus on the main methods and give the details on further models in Appendix B.4.

In Section 1.4.5, the properties of the classifiers will be analyzed at the example of handwritten digit recognition. Another application, where unary classification might be useful, is EEG data analysis as explained in Section 1.4.6.

### 1.4.1 Connection between $\nu$-SVM and $\nu$oc-SVM

It has been proven under the assumption of separability and hard margin separation that the $\nu$oc-SVM defines the hyperplane with maximum distance for separating the data from the origin [Schölkopf et al., 2001b, Proposition 1]. This concept is similar to the well known maximum margin principle in binary classification. In the following, we will generalize this proposition to arbitrary data and maximum margin separation with a soft margin, e.g., as specified for the $\nu$-SVM.

**Theorem 15** (Equivalence between $\nu$-SVM and $\nu$oc-SVM via origin separation)**.** *Applying the origin separation approach to the $\nu$-SVM results in the $\nu$oc-SVM.*

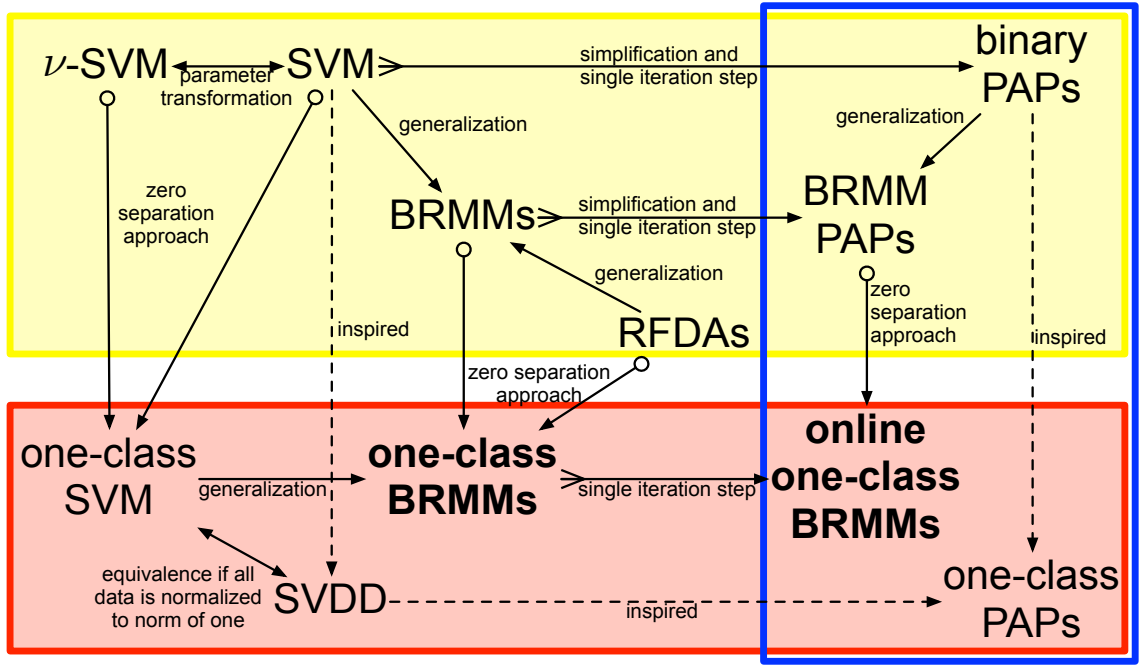

Figure 1.13: **Scheme of relations between binary classifiers (yellow) and their one-class (red) and online (blue) variants.** The new variants introduced are in bold. The details are explained in Section 1.4. Visualization taken from [Krell and Wöhrle, 2014].

*Proof.* The $\nu$-SVM (Method 4) is defined by the optimization problem

$$
\begin{aligned}
\min_{w', t', \rho', b'} \quad & \tfrac{1}{2} \|w'\|_2^2 - \nu \rho' + \tfrac{1}{n'} \sum t'_j \\
\text{s.t.} \quad & y_i \left( \left\langle w', x'_j \right\rangle + b' \right) \geq \rho' - t'_j \text{ and } t'_j \geq 0 \ \forall j \ .
\end{aligned}
\tag{1.108}
$$

$n'$ is the number of training samples. $w'$ and $b'$ define the classification function $f(x) = \operatorname{sgn}(\langle w', x \rangle + b')$. The slack variables $t'_j$ are used to handle outliers which do not fit the model of linear separation.

In the origin separation approach, only the origin (zero) is taken as the negative class ($y_0 = -1$). In this case, the origin must *not* be an outlier ($t_0 = 0$), because it is the only sample of the negative class.[23] Consequently, the respective inequality becomes: $-(\langle w', 0 \rangle + b') = \rho'$. Accordingly, $b'$ can be automatically set to $-\rho'$. To achieve class balance, as many samples as we have original samples of the positive class are added to the origin for the negative class. This step only affects the total number of samples which is doubled ($n' = 2n$), such that $n$ only represents the number of real positive training samples and not the artificially added ones. Putting everything together

---

[23] Also known as hard margin separation.

$(y_j = 1, b' = -\rho', n' = 2n)$, results in

$$\min_{w',t',\rho'} \quad \tfrac{1}{2}\|w'\|_2^2 - \nu\rho' + \tfrac{1}{2n}\sum t_j'$$
$$\text{s.t.} \quad \langle w', x_i \rangle \geq 2\rho' - t_j' \text{ and } t_j' \geq 0 \; \forall j \; . \tag{1.109}$$

By applying the substitutions:

$$w' \to \frac{\nu}{2}w, \; \rho' \to \frac{\nu}{4}\rho, \; \text{and } t_j' \to \frac{\nu}{2}t_j \tag{1.110}$$

in Equation (1.109) and by multiplying its inequalities with $\frac{2}{\nu}$ and its target function with $\frac{4}{\nu^2}$, this model is shown to be equivalent to:

$$\min_{w,t,\rho} \quad \tfrac{1}{2}\|w\|_2^2 - \rho + \tfrac{1}{\nu l}\sum t_j$$
$$\text{s.t.} \quad \langle w, x_j \rangle \geq \rho - t_j \text{ and } t_j \geq 0 \; \forall j \tag{1.111}$$

with the decision function $f(x) = \text{sgn}\left(\langle w, x \rangle - \frac{\rho}{2}\right)$. This is the model of the $\nu oc\text{-}SVM$ (Method 11). There is only a difference in the offset of the decision function which should be $-\rho$ instead of $-\frac{\rho}{2}$. This difference can be geometrically justified as explained in the following and the function can be changed accordingly. Additionally to the decision hyperplane, a SVM is identified with its margin, additional hyperplanes for the positive and the negative class. The difference in the offsets corresponds to choosing the hyperplane of the positive class as the decision criterion instead. This is reasonable, because for the SVM models the training data is assumed to also include outliers which are on the opposite side of the respective hyperplane. Furthermore, the decision criterion might be changed in a postprocessing step or varied in the evaluation step [Bradley, 1997]. □

## 1.4.2 Novel One-Class Variants of C-SVM, BRMM, and RFDA

Since the BRMM generalizes the C-SVM and RFDA, it is sufficient to apply the origin separation approach directly to the BRMM (Method 14).

With the same argumentation as for the $\nu$-SVM in Theorem 1.4.1 we insert the origin (zero sample) into the inequality

$$y_0 \left( \langle w, x_0 \rangle + b \right) \geq 1 - t_0 \tag{1.112}$$

and enforce $t_0 = 0$ which results in $-\left(\langle w, 0\rangle + b\right) = 1$ and consequently $b = -1$. Subtracting the inequality with 1 finally results in the **Method 18** (One-Class Balanced Relative Margin Machine).

$$\min_{w,t} \quad \frac{1}{2} \|w\|_2^2 + C \sum t_j$$
$$\text{s.t.} \quad 1 + R + t_j \geq \langle w, x_j \rangle \geq 2 - t_j \text{ and } t_j \geq 0 \; \forall i \; . \tag{1.113}$$

Modifying the decision function $f(x) = \text{sgn} \left( \langle w, x \rangle - 1 \right)$ in the same way as we did for the $\nu$-SVM results in $f(x) = \text{sgn} \left( \langle w, x \rangle - 2 \right)$. Note that the offset is now fixed, which enables the application of the single iteration approach from Section 1.2.4 without any changes to the offset treatment. With the extreme case, $R = \infty$, we obtain a new one-class SVM (Coc-SVM) It is expected to be very similar to the $\nu$oc-SVM because they were derived from C-SVM and $\nu$-SVM which are equivalent according to Theorem 6. However, the new model provides a better geometric interpretation and a simplified implementation.

Due to the single iteration approach it would be possible to use the implementations of the binary counterparts with linear kernel. Only the hard margin separation for the artificially added sample has to be realized. Nevertheless, it helps to take a deeper look into implementation strategies to adapt them to the special setting of unary classification with the origin separation approach. Furthermore, for the use of nonlinear kernels special care has to be taken, because the origin of the underlying RHKS might not have a corresponding sample in the original data space anymore. Especially the zero sample is not at the origin.

For solving the optimization problem in Equation (1.113), it is no longer required to update pairs of samples [Platt, 1999a]. Because of the special offset treatment the approach from Section 1.2.3 can be directly applied. Considering the fixed offset ($b = -1$), the respective update formulas can be derived (see also Appendix B.4.2):

$$\alpha_j^{(k+1)} = P \left( \alpha_j^{(k)} - \frac{1}{\|x_j\|^2} \left( \left\langle w^{(k)}, x_j \right\rangle - 2 \right) \right)$$
$$\beta_j^{(k+1)} = P \left( \beta_j^{(k)} + \frac{1}{\|x_j\|^2} \left( \left\langle w^{(k)}, x_j \right\rangle - (R+1) \right) \right)$$
$$w^{(k+1)} = w^{(k)} + ((\alpha_j^{(k+1)} - \alpha_j^{(k)}) - (\beta_j^{(k+1)} - \beta_j^{(k)})) \, x_j \tag{1.114}$$
$$\text{with } P(z) = \max \{0, \min \{z, C\}\} \; .$$

Comparing these formulas with the formulas of the binary classifier in Section 1.3.4.1 shows that the implementations of binary classifiers require only minor modifications to be also used for unary classification: The offset has to be fixed to $-1$ and its update has to be suppressed.

**Squared Loss and Kernels**

The origin separation approach can also be used for variants of the discussed algorithms if squared loss variables ($t_j^2$) in the target function or kernels are used as introduced in Section 1.1.1.2. The formulas can be derived in the same way as for the binary classifiers (see Appendix B.4.2).

Kernels are motivated by an implicit mapping of the data to a higher dimensional space (RHKS). Consequently, the separation from the origin is applied in the RHKS and not in the originally data space. For example, using a Gaussian kernel ($k(x, y) = e^{\gamma \|x-y\|_2^2}$) results in a separation of points on an infinite dimensional unit hypersphere from its center at the origin in the RHKS, because

$$\|x\|_k := k(x, x) = 1 \ \forall x \in \mathbb{R}^m \ . \tag{1.115}$$

**Strict Separation from the Origin for SVM:** $C = \infty$

For a different geometric view on the new one-class SVM, consider the extreme case of hard margin separation ($C = \infty$), which enforces the slack variables to be zero. Let $X$ denote the set of training instances $x_j$ with the convex hull $\text{conv}(X)$. The origin separation approach reveals that the optimal hyperplane (for the positive class) is tangent to $\text{conv}(X)$ in its point of minimal norm $x'$ (Theorem 23). The hyperplane is orthogonal to the vector identified with $x'$ and $w = x' \frac{2}{\|x'\|_2^2}$ .

### 1.4.3 Equivalence of SVDD and One-Class SVMs on the Unit Hypersphere

The approach of SVDD (Method 10) is different from the origin separation. Here, the goal is to find a hypersphere with minimal radius $R$ around a center $c$ such that the data is inside this hypersphere and the outliers are outside

$$\begin{aligned} \min_{R,c,t'} \quad & R^2 + C' \sum t_i' \\ \text{s.t.} \quad & \|c - x_i\|_2^2 \le R^2 + t_i' \text{ and } t_i' \ge 0 \ \forall i \ . \end{aligned} \tag{1.116}$$

**Theorem 16** (Equivalence of SVDD and $\nu$oc-SVM on the Unit Hypersphere). *If the data is on the unit hypersphere (normalized to a norm of one),*

$$w = c, \ t_i = \frac{t_i'}{2}, \ \rho = \frac{\|c\|_2^2 + 1 - R^2}{2}, \ C' = \frac{1}{\nu l} \tag{1.117}$$

*gives a one-to-one mapping between the SVDD and the $\nu$oc-SVM model.*

For a proof refer to [Tax, 2001] or Appendix B.2.3 but not to [Tax and Duin, 2004], where the proof is incomplete. The equivalence of the models is also reasonable from our new geometric perspective as visualized in Figure 1.14. Intersecting the data space (unit hypersphere) with a SVM hyperplane separates the data space into the same two parts as when cutting it with the SVDD hypersphere. From the geometric view, $R^2 + d^2 = 1$ should also hold true, where $d$ is the distance of the origin to the separating hyperplane of the SVM. So maximizing this distance in the SVM model is equivalent to minimizing the radius of the hypersphere in the SVDD model. If the data is not normalized to a norm of one, the models differ. Note that when using Gaussian kernels, data is internally normalized to unit norm (see Equation (1.115) or [Tax, 2001]).

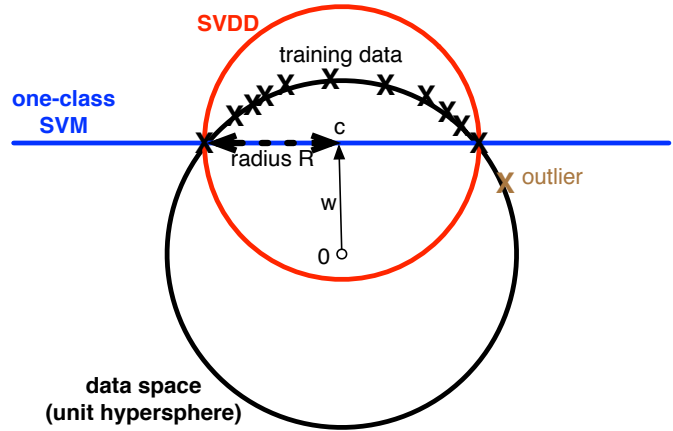

Figure 1.14: **Geometric relation between SVDD with a separating hypersphere (red) with radius** $R$ **and center** $c$ **and *one-class SVM* with its hyperplane (blue) and classification vector** $w$ **when the data lies on a unit hypersphere (black).** Samples outside the red hypersphere are outliers in the SVDD model and samples below the blue hyperplane are outliers for $\nu$oc-SVM. The remaining data belongs to the class of interest. Visualization taken from [Krell and Wöhrle, 2014].

**Theorem 17** (From $\nu$oc-SVM to the New One-Class SVM)**.** *Let* $\rho(\nu)$ *denote the optimal value of the* $\nu$*oc-SVM model. If* $\rho(\nu) > 0$*,* $\nu$*oc-SVM is equivalent to our new one-class SVM by substituting*

$$\bar{w} = \frac{2w}{\rho(\nu)}, \ \bar{t}_i = \frac{2t_i}{\rho(\nu)}, \ \bar{C} = \frac{2}{\nu l \rho(\nu)} \tag{1.118}$$

*even if the data is not normalized. So both models are similar, too.*

The proof can be found in Appendix B.2.3.

### 1.4.4  Novel Online Unary Classifier Variants of the C-SVM

In Section 1.4.2, formulas for the weight update belonging to a single sample have been derived. According to Section 1.2.3 and Section 1.2.4 the application of the single iteration approach is straightforward and leads to the update formulas for an online classifier version:

**Method 19** (Online One-Class BRMM)**.**

$$
\begin{aligned}
\alpha &= \max\left\{0, \min\left\{\frac{1}{\|x^{new}\|_2^2}\left(2 - \left\langle w^{old}, x^{new}\right\rangle\right), C\right\}\right\} \\
\beta &= \max\left\{0, \min\left\{\frac{1}{\|x^{new}\|_2^2}\left(\left\langle w^{old}, x^{new}\right\rangle - (R+1)\right), C\right\}\right\} \\
w^{new} &= w^{old} + (\alpha - \beta)\,x_j\,.
\end{aligned}
\tag{1.119}
$$

This model combines, the single iteration, the relative margin, and the origin separation approach. For an online one-class SVM variant, only $\alpha$ is used ($\beta \equiv 0$). Update formulas for variants with different loss can be defined respectively (see Appendix B.4.3).

This direct transfer of the introduced unary classifiers to online classification completes the picture on the binary PAA, which are connected to the C-SVM by the single iteration approach. It results in *online versions* for the unary variants of the batch algorithms: C-SVM ($R = \infty$), BRMM, and RFDA ($R = 1$).

### 1.4.5  Comparison of Unary Classifiers on the MNIST Dataset

To get a first impression of the new unary classifiers, a comparison on the MNIST dataset (see Section 1.3.4.4) was performed. We chose a one vs. rest evaluation scheme, where classifiers were trained only on one digit (target class) and tested on all other digits (rest, outliers). Using unary classifiers on this data has three advantages: First, the classifiers describe how to detect a single digit and not how to detect the difference to all the other digits. Second, the classifiers do not have to handle class imbalance. And third, the classifiers can better detect new outliers like bad handwriting or letters (which are not part of this dataset).

#### 1.4.5.1  Comparison of Classifiers with different Range or Radius

For dimensionality reduction, a principal component analysis [Lagerlund et al., 1997, Rivet et al., 2009, Abdi and Williams, 2010] (PCA) was applied (with training on the given training data), keeping the 40 most important principal components.[24] Furthermore, all resulting feature vectors were normalized to unit norm. The regular-ization parameter $C$ was individually optimized using a grid search with 5-fold cross-validation on the training data. As performance metric, the average of the area under the ROC curve [Bradley, 1997] (AUC) over all possible digits was used, to account for class imbalance (ratio $1 : 9$) and sensitivity on the decision boundary that was not optimized [Swets, 1988, Bradley, 1997, Straube and Krell, 2014]. The pySPACE configuration file is provided in the appendix in Figure C.2. The results are depicted in Figure 1.15.

---

[24] The PCA implementation of scikit-learn was used here [Pedregosa et al., 2011].

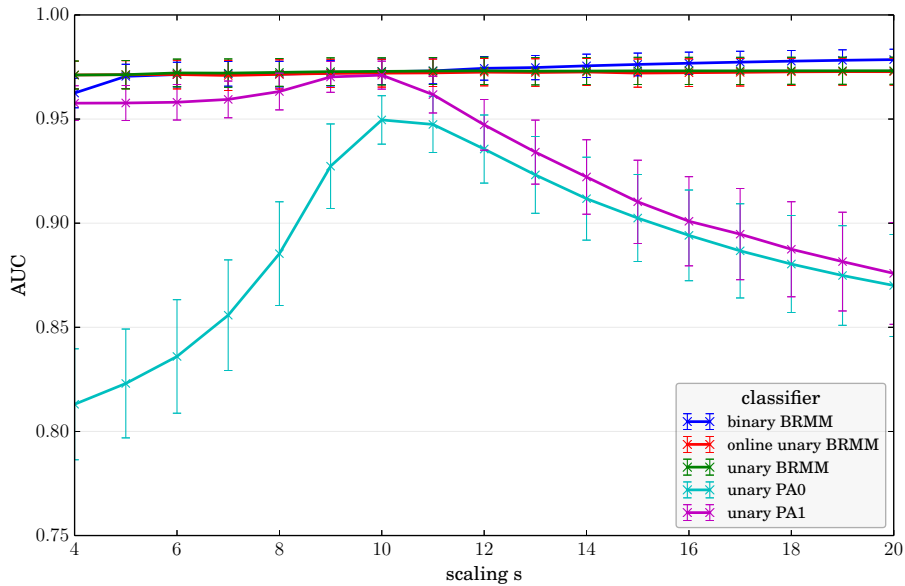

Figure 1.15: **Comparison of different unary classifiers on the MNIST dataset with varying radius (unary PAAs PA0 and PA1, $R_{\max} = \frac{s}{10}$) or range (BRMM variants, $R = \frac{s}{4}$)) parameter.** Both hyperparameters are calculated with the help of the *scaling* parameter $s$. The average AUC with standard error is displayed in a scenario, where the classifier has been trained on one digit out of ten. The binary BRMM is displayed for comparison, too. For the BRMM variants, the border at $R = 1$ ($s = 4$) corresponds to a RFDA variant and the upper border ($s = 20$) is equivalent to the respective C-SVM variant. Visualization taken from [Krell and Wöhrle, 2014].

For the unary BRMM variants, different range parameters were tested, but no influence on the performance can be observed. In this application, the online unary BRMM performs as well as the batch algorithm (unary BRMM), although it requires less training time (600ms instead of 30min average time) and memory ($O(n)$ instead of $O(n \cdot l)$). This is a clear advantage of the online classifier, since it allows to train the algorithm on large datasets and potentially increase performance with constantly low processing resources.

For the binary BRMM, there is a performance increase on the way from RFDA to C-SVM. C-SVM performs better than the unary classifiers, but requires all digits for the training (nine times more samples). Consequently, using the C-SVM instead of a unary classifier requires more computing resources.

For the unary PAAs (online classifiers) PA0 and PA1, which were motivated by SVDD as described in Section 1.1.6.2, different values for the maximal radius were tested. The PAAs optimize their radius parameter, but need a predefined maximum value. By increasing the maximum radius, first performance increases and then monotonically decreases. PA1 clearly outperforms PA0, because it allows for misclassifications in its model, which improves its generalization capability. This effect is quite common and has been observed also for numerous other classifiers. With the optimal choice of the radius, PA1 performs as well as the BRMMs. This is possibly the same effect as the equivalence of the SVDD and the one-class SVMs on data on a unit hypersphere, as used in this example. Unfortunately, the intrinsic optimization of the radius is not working sufficiently well and the maximum radius parameter needs additional tuning. This is a clear disadvantage compared to the other classifiers, especially since hyperparameter tuning is often difficult in one-class scenarios like outlier detection.

To summarize, despite the slightly worse performance value in comparison to C-SVM ($R = 20$), in this application unary classifiers are useful due to reduced computing resources and because they might better generalize on unseen data like handwritten letters. For online learning, the new online unary SVM is better than the already existing online one-class algorithms (PA0 and PA1), because it does not require the optimization of the hyperparameter $R_{\max}$.

### 1.4.5.2 Equivalence of $\nu$oc-SVM and the novel one-class C-SVM

To visualize the equivalence between $\nu$oc-SVM and our new version of a one-class SVM, which is directly derived from C-SVM, an additional evaluation was conducted by varying the hyperparameters $\nu$ and $C$ (see Figure 1.16). This results in one performance curve for each digit, the classifier has been trained on. Everything else was kept as in the previous evaluation in Section 1.4.5.1 (e.g., testing on all remaining digits). The pySPACE configuration file is provided in the appendix in Figure C.1.

The performance of the new one-class SVM is constant when the regularization parameter $C$ is chosen to be very high or very low. Both performance curves show the same increase in performance, the same maximum performance, and then the same decrease. Only the scaling between the hyperparameters is different and consequently the curves look differently. The similarity of the curves indicates an equivalence of the solutions. This equivalent behavior is also expected from the theory (see Section 1.4.1) and was the motivation to derive the other classifiers, because the derivation requires the C-SVM modeling and is not applicable to $\nu$-SVM. For very low $\nu$, the performance of $\nu$oc-SVM decreases drastically. Reasons for this might be differences in implementation, rounding errors, or a degeneration of the optimization problem.

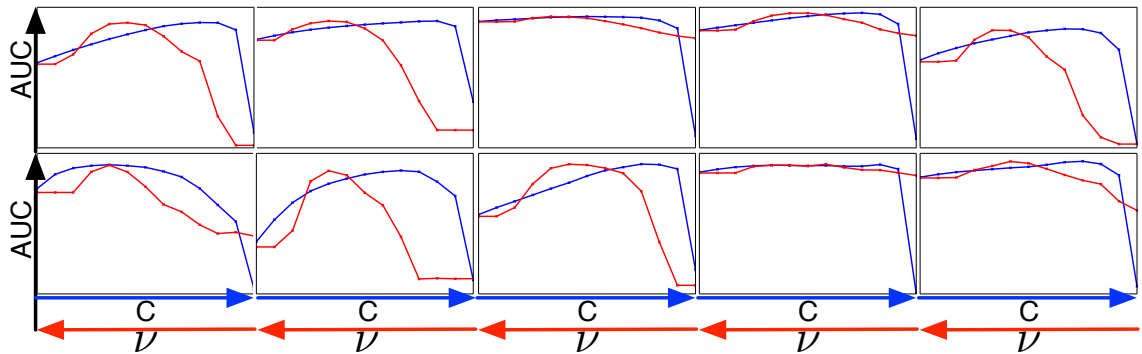

Figure 1.16: **Performance comparison of** $\nu$*oc-SVM* **(blue) and** ***new one-class SVM*** **(red)** trained on different *digits* (0-9) with varying hyperparameters $\nu$ and $C$. Visualization taken from [Krell and Wöhrle, 2014].

### 1.4.5.3  Generalization on Unseen Data and Sensitivity to Normalization

In the following, the effect of different normalization techniques and the generalization on unseen data are analyzed. Normalization techniques change the position of the data in relation to the origin. Consequently, an effect on the origin separation is expected. One-class classifiers are not dependent on the second class and should better handle changes in this class.

In comparison to the evaluation in Section 1.4.5.1, the PCA is omitted and only $0.25\%$ of the training data are used for pretraining and calibration of the algorithms. In the testing phase, each sample is first classified and if it is the label of interest, an external verification is assumed, which allows to have an incremental training for the unary online algorithms (unary PA1 and online unary SVM). To show the generalization capability, only the label/digit of interest (one of $1-9$) was used as positive class and $0$ as opposing class for calibration and second class for the binary classifier (new one-class SVM). For testing, all digits were used. For normalization, three approaches are compared: no normalization (No), normalization of the feature vector to unit norm (Euclidean), and finally determining mean and variance of each feature on the training data and normalizing the data to zero mean and variance of one (Gaussian). For the unary PA1, the optimization of the hyperparameter $R_{\max}$ was included in the $5$-fold cross-validation (which optimizes the hyperparameter $C$) using the same range of values for the hyperparameter as in Section 1.4.5.1. To account for the random selection of training samples and the splitting in the cross-validation step, the experiment was repeated $100$ times. The results are shown in Figure 1.17. Three conclusions can be drawn from the results:

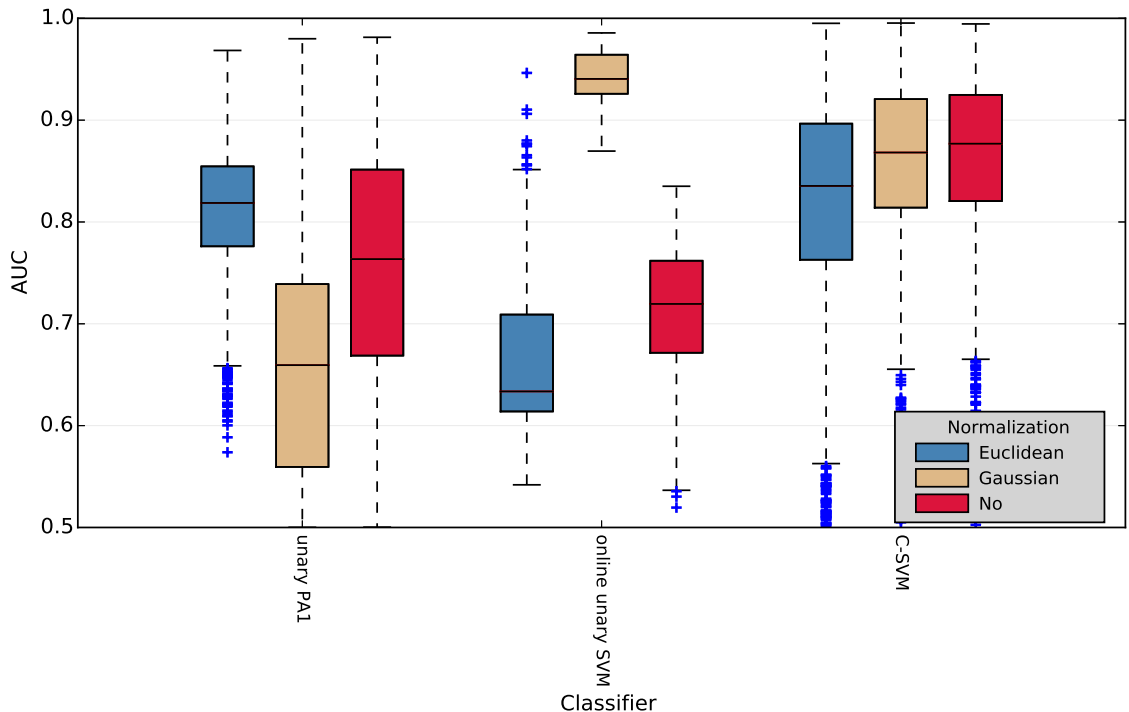

Figure 1.17: **Comparison of different normalization techniques and online classifiers** on the MNIST dataset. The box plots show the median of the performance values (AUC) with the 25% and 75% quantile.

1. *The one-class classifiers are highly dependent on the chosen normalization.*

   As already mentioned, this was expected, because the normalization largely changes the position of the data to the origin in this example. For the binary C-SVM classifier there are no large differences in performance between the different normalization techniques but the unary classifiers largely differ. For the online unary SVM, the Gaussian feature normalization is best and for the unary PA1 Euclidean feature normalization is best.

2. *The unary PA1 shows the worst performance.*

   One reason for this might be that the small calibration dataset was not sufficient for a tuning of $R_{\max}$. If it were chosen to small, the incremental learning would change the center of the circle to much. Furthermore, the approach of just putting a circle around the samples of interest might not be the right approach in this example, because it does not generalize enough.

3. *The online unary SVM clearly outperforms the other classifiers.*

   This was expected due to the incremental training and because of the focusing on the class of interest. Hence, it does not overfit on the "outliers" (irrelevant class) and performs better when other types of "outliers" come in.

### 1.4.6   P300 Detection as Unary Classification Problem

In this Section, we evaluate the application of unary classification on the P300 dataset which is described in Section 0.4.

Normally, P300 detection is treated as a binary classification problem [Krusienski et al., 2006], and sometimes even as a multi-class problem [Hohne et al., 2010]. The important class is the P300 ERP. As the second class, the brain signal which corresponds to the unimportant more frequent stimulus is taken or other noise samples, which are not related to the important stimulus. In such a classification task, the classifier might therefore not learn the characteristics of the P300 signal but how to differentiate the important class from the unimportant class. To simplify the use in the application and from the modeling perspective, we suggest to focus on the important class and use a unary classification. This reduces the training effort and the classifier models the signal of interest. Furthermore, the problem of class imbalance in P300 detection in a two class approach can be avoided. It is caused by the fact that the important stimulus is rare and has to be treated from the algorithm and evaluation point of view [Straube and Krell, 2014].

**Processing**   In the following, we introduces the methods used for processing and classifying the EEG data.

The preprocessing was as described in [Feess et al., 2013] and displayed in Figure 3.4. For the normalization we again compared Euclidean, Gaussian, and no ("Noop") feature normalization. For classification, we compared binary and unary, online and batch BRMM including the special cases of $R = 1$ (RFDA) and $R = \infty$ (C-SVM). Furthermore, we included the unary PAAs, PA1 and PA2. The online classifiers were kept fixed on the testing data. For our investigation, the threshold was optimized on the training data [Metzen and Kirchner, 2011] because unary classifiers are very sensitive to it.

For all classifiers, the regularization parameter $C$ has to be optimized. We tested with the following range: $[10^{-4}, 10^{-3.5}, \ldots, 10^2]$. A second hyperparameter is only relevant for the BRMM (with a tested range of $1 + 10^{-1}, 1 + 10^{-0.8}, \ldots, 1 + 10^1$) and the unary PAAs, PA1 and PA2 (with a maximum radius of $10^{-0.5}, 10^{-0.7}, \ldots, 10^{1.5}$). To determine the optimal hyperparameters, a grid search with a 5-fold cross validation was performed on the training data.

In each session, 5 datasets were recorded. For evaluation, we trained the algorithms on one of five sets and tested on the remaining 4 datasets. This is a typical cross-validation scheme. Although not given to the unary classifiers, the data of the second class (frequent irrelevant standard stimuli) were used for evaluation and training the other supervised algorithm: the xDAWN algorithm uses data which does not belong to the ERP of interest to determine the noise in the data. For Gaussian feature normalization, the label of the data is irrelevant. Only the threshold optimization really needs the second class. As discussed in Section 3.3, the BA was taken as performance metric [Straube and Krell, 2014].

**Results and Discussion**   The results of the evaluation are depicted in Figure 1.18. The above mentioned classifiers and normalization methods are compared.

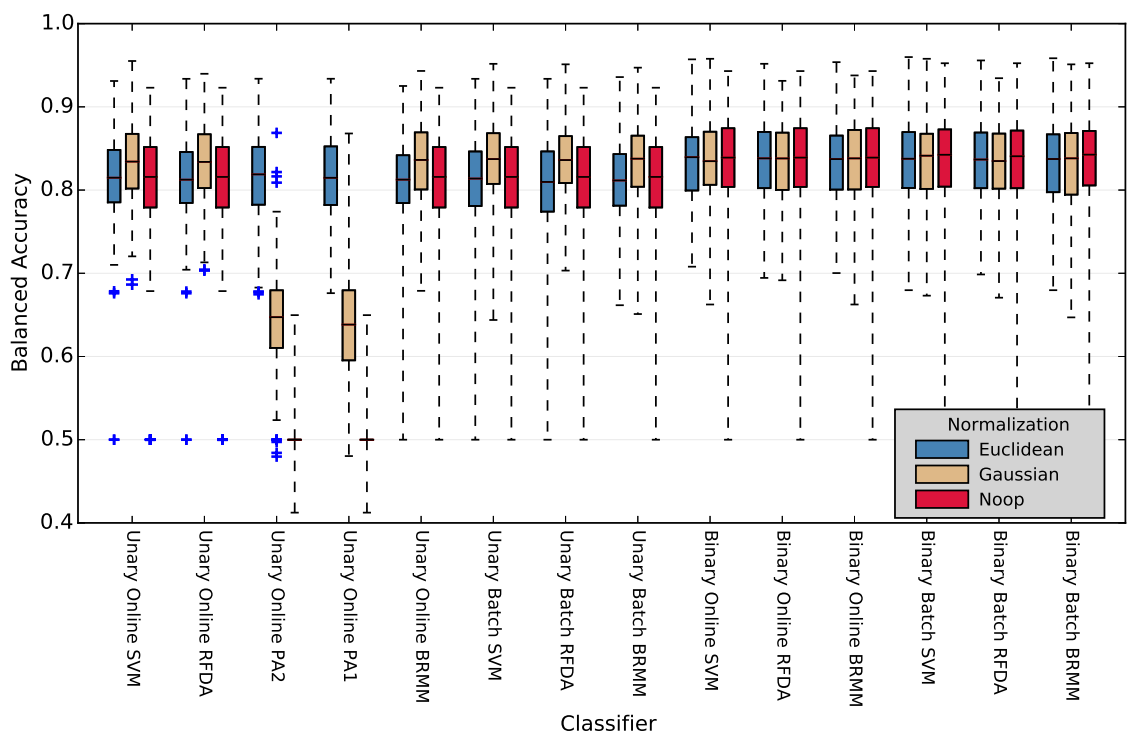

Figure 1.18: **Comparison of the different classifiers and normalization techniques.** Unary and binary classifier variants are compared as well as online and batch learning variants. In the box plots, median and 75% quantiles are displayed.

When using Euclidean feature normalization, the performance of PA1 and PA2 is comparable to the performance of the unary online SVM. This is reasonable, because the respective batch algorithms (one-class SVM and SVDD) are equivalent, when applied on data with a norm of one. Nevertheless, the performance of PA1 and PA2 with Euclidean feature normalization is inferior to the performance of the other classifiers with Gaussian feature normalization. This shows, that for the observed data, the approach of linear separation with hyperplanes is superior to the approach of separation with the help of surrounding hyperspheres. Another problem might be

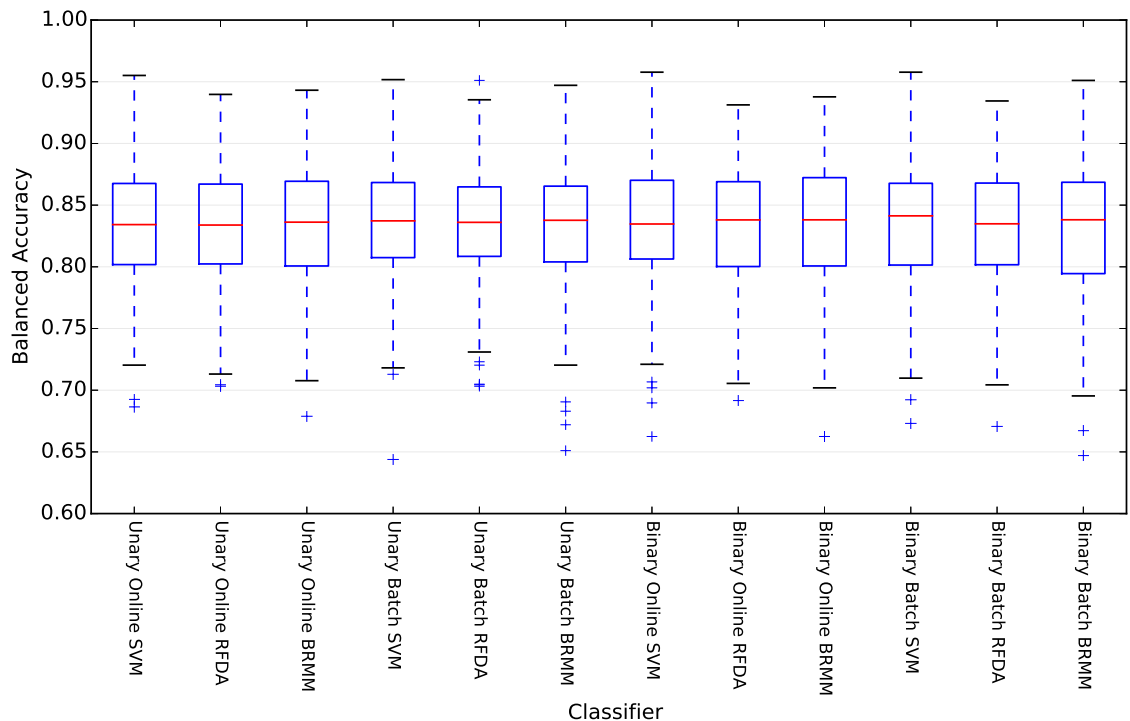

Figure 1.19: **Comparison of classifiers (except PA1 and PA2) after Gaussian feature normalization.** In the box plots, median and $75\%$ quantiles are displayed.

the choice of the optimal maximum range of PA1 and PA2 as in the experiment in Section 1.4.5.1

The processing with Gaussian feature normalization always performs slightly better or equal to the other normalization techniques (except for PA1 and PA2). Again, the unary classifiers are more sensitive to the type of normalization, which is reasonable due to the origin separation approach. The results for using the Gaussian feature normalization only are displayed in Figure 1.19. It can be observed that when using this normalization, all (other) classifiers show comparable performance results. This holds for the comparison of online and batch learning algorithms but most importantly the binary classifiers do not outperform the variants of the investigated unary classifiers in Figure 1.19. A reason for this behavior is, that the xDAWN algorithm already reduced dimensionality a lot and has the main influence. If it were left out, the performance would drop especially for the unary classifiers (results here not reported).

## 1.4.7   Practice: Normalization and Evaluation

As the experiments show, the choice of normalization is crucial. This is similar to the considerations as in Section 1.2.5. One has to consider, if the approach of sep-arating the data from the origin is reasonable. For example, separating the data $\{(0, 1), (1, 0), (0, -1), (-1, 0)\}$ from $(0, 0)$ would not make any sense and is not even possible with a hard margin separating unary one-class SVM or SVDD. In this case, it is always good to reflect, if the origin can be considered as an outlier. For P300 detection, a zero sample (without Gaussian normalization) corresponds to no relevant signal in the data and is the perfect opposite class. In fact, if the preprocessing were perfect it would map all the other data to zero. For unnormalized MNIST data, a zero vector can identified be with an empty image which corresponds to no digit, which is definitely an outlier or can be seen as *the* opposing class.

With increasing dimensionality of the data it is easier to separate the training dataset from the origin but this might also decrease the capability of the unary classifier to describe the data.

From the intuition the geometric idea behind SVDD seems more appropriate than the origin separation approach but the experiments indicated the opposite.

Taking everything together, the origin separation requires careful consideration before application. This is probably the reason, why it is used seldom in the direct way. On the other side, when using the RBF kernel, which is quite common, most problems disappear. First of all, still data should be normalized but the separability to the origin is not relevant any more. Second, in this case, the data is lying in an infinite dimensional sphere and the positive orthant and consequently the data is always separable from the origin, which is the center of the sphere. Third, SVDD and the application of the origin separation to C-SVM result in the same classification and it does not matter anymore which approach is considered more reasonable. Last but not least, the $\nu$oc-SVM generalizes nicely the Parzen windows estimator as shown in [Schölkopf et al., 2001b], which is a reasonable approach to approximate probabilities. Consequently, using the RBF kernel together with the origin separation approach is a good choice and according to Theorem 5 it also generalizes the linear version.

The usage of unary classifiers is difficult to evaluate and hyperparameters are difficult to optimize. In some cases, a visualization might be useful but will not give a quantification. From our point of view, the best way out is to use another class. Since the introduced unary classifiers all come with a decision function which determines, if new incoming data belongs to the given data or not, this approach is reasonable. The second class can be:

- the large number of irrelevant samples (e.g., unrelated EEG data in P300 detection or other digits or letters in case of the MNIST data),

- a small number of outliers (e.g., data from a (simulated) crashed robot or data from missed targets in P300 classification), or

- synthetic data by adding noise to the given training samples, which is often used in the literature but which might not be representative for future incoming data.

In any case, class imbalance should be considered in the evaluation (see Section 3.3). Furthermore, the offset should be carefully optimized or an evaluation should be used, which is not dependent on a decision criterion.

To summarize, we could show that the origin separation approach is an intuitive way to derive unary classifiers from binary classifiers like the numerous SVM variants in Section 1.1. The respective implementations from the binary classifiers can be used. On the other side, unary classification comes with difficulties of offset tuning, data normalization, and appropriate evaluation. With our presented geometric intuition it becomes immediately clear that the origin separation approach is only working when it is reasonable to have a linear hyperplane (for modeling the data), and to consider the origin as the opposite class. Knowing that the approach is equivalent to the possibly more intuitive SVDD concept when using a fitting kernel or normalization technique even improves our geometric concept. It is now easy to understand, why different models perform quite similar and why it is important to also have a look at evaluation techniques, decision criterion optimization, feature normalization in the preprocessing, and the use of kernels which is probably most important.

## 1.5  Discussion

In this chapter, numerous classifiers were introduced and their new and old relations were summarized for the first time. For the experts, most knowledge might be already known or trivial but for the normal user of these algorithms, the given relations remain mostly unknown because they are not reported or just distributed in the literature. But how does this summary of classifier connections help to answer the question of "which" classifier to use? This will be discussed in the following with three different perspectives/use cases.

**Learning and Teaching Perspective**   The first requirement to answer this question is to know the classifiers. Hence, summarizing them is a first approach. But still getting to know them might be difficult. Here, our set of relations can probably help more than just learning about regularization and loss functions. It is not required to learn the single models but to understand the concepts on how the models are derived. This can be directly used in teaching as outlined in the following.

Assuming the concept of a squared loss, kernel, SVR, and the related ridge regression are already known, it is very intuitive and straightforward to look at the simplification of binary classification by considering only two possible values for regression: $\{-1, +1\}$. This directly results in BRMM and RFDA/LS-SVM. Since, RFDA is the special case of BRMM with $R = 1$ a good next step is to look for $R = \infty$ and get C-SVM. This can be supported by respective visualizations and formal descriptions of the algorithms. So with the help of the *relative margin concept* (Section 1.3) a first set of classifiers can be derived without much effort. With C-SVM and relative margin, one should give a short introduction to the geometric background of maximum margin separation.

The next step in teaching would be to answer the question of how to implement the algorithms as done in Section 1.2. This can be connected to practical questions as in robotics, where limited memory and processing power have to be considered. Here, one answer can be the online algorithms, derived by the *single iteration approach*.

Finally, unary classification can be seen as a tool to handle multi-class classification, large class imbalance, or simply just to describe one class. The origin separation approach from Section 1.4 can then be used to derive the unary classifiers again geometrically. As a "better" justification, the relation of the $\nu$oc-SVM to the probability modeling Parzen windows estimator and the maybe geometrically more intuitive SVDD can be used.

This teaching approach can be supported by several visualizations of the classifiers as already given in the previous sections but also with a more general overview graphic as provided by Figure 1.20 to highlight, how the different approaches are connected.

**Application Perspective**  Another interesting point of view is the application. Assuming, a (linear) C-SVM turned out to be a very good choice due to its generalization capabilities even on a small number of samples in some preliminary data analysis. If the data shall now be processed on a robot or an embedded device with limited resources, one could directly transfer the linear decision function. If later on a verification of new data becomes possible and drifts in the data are expected, the single iteration approach provides a direct way to adapt the classifier with low effort of implementation, low processing power, and no additional memory usage.

If more data is acquired, it makes sense to model statistical properties of the data and drifts. Here, the relative margin concept is a first direct and smooth approach which transfers C-SVM to RFDA. The transfer can be automatically achieved by using BRMM and tuning its hyperparameter $R$. If the amount of available data becomes very large and hyperparameter optimization showed, that $R = 1$ is a reasonable choice and that the regularization parameter $C$ can be chosen very large than it might be a good step to switch to the limit, which is the FDA. An advantage of this step is, that this classifier allows for very fast online updates [Schlögl et al., 2010].

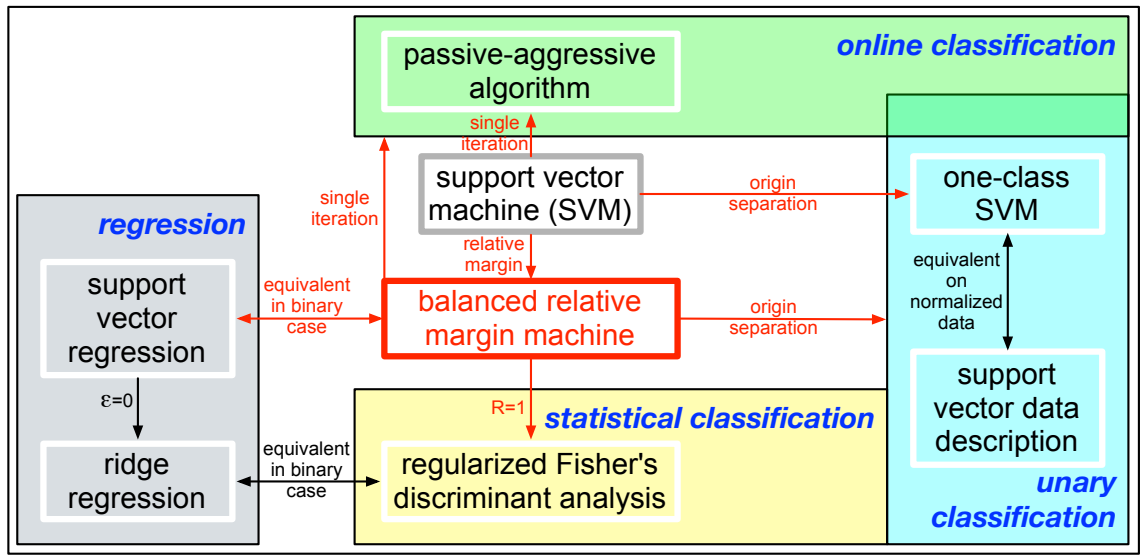

Figure 1.20: **Simplified overview of connections of support vector machine (SVM) variants.** The details and several further connections can be found distributed in Chapter 1. The red color highlights the new connections, provided by the new generalized model. For every classifier it is possible to use squared or a non-squared loss. Except for the online classifiers (green box), a kernel function can be always used. Last but not least, the three introduced approaches can be combined as depicted in Figure 1.1.

On the other hand, one might realize, that only one class is relevant in the data and so uses the *zero separation approach* to only work with one classifier as was suggested for the P300 detection in Section 1.4.6. Alternatively, if the application might request the capability to work and many classes and might even require to be extensible for new classes. This is for example the case when first only the goal is to predict movements, where data with no movement planing can be taken as ("artificial") second class but later this goal is changed to also distinguish different movement types. Another example might be soil detection for a robot by images and verification over sensors. During runtime, an arbitrary number of new underground types could occur. A set of already defined classifiers says, that a new image does not seems to belong to already observed soil types and this is verified by other sensors. Hence, a new unary classifier could be generated to determine this soil type for future occurrence.

Such automatic behavior will be required for longterm autonomy and for constructing intelligent robots. For completeness, it should be mentioned that such a problem could be also tackled with unsupervised algorithms (clustering) or even better with a mixture of unsupervised and supervised algorithm. Last but not least, it is important to mention, that all these considerations from the application point of view most often do not occur separately but this is not a problem, because the introduced approaches can be easily combined.

**Implementation and Optimization Perspective**  Luckily, with the implementation of the BRMM with special offset treatment, all the mentioned approaches and variants can be handled within just one implementation. For the single iteration approach, the number of maximum iterations could be set to the total number of trainings samples and in the online learning case, the update formulas can be directly reused. For getting the border cases for the relative margin approach, $R$ can be set to the respective values. And for integrating the origin separation approach it is only required to keep the offset fixed at $-1$ after an update step.

A similar view can be taken, when optimizing the classifier using the generalized BRMM model with its variants. The number of iterations can be taken as a hyperparameter which is tuned and if it gets close to the number of samples, the online learning version should be used instead. When switching between L1 and L2 loss the old support vectors are a first good guess for the new classifier and can be reused. Especially if the linear kernel is used, sparsity of the number of support vector is less relevant and with an increasing number of samples it might make sense to switch from L1 to L2 loss.

If the range parameter $R$ is optimized it is good to start with high values, especially when only few samples are available. When optimized, it might turn out that the maximum *Range* $R_{\max}$ should be used, to avoid an outer margin or $R$ should be taken very small to model drifts in the data and so the respective C-SVM or RFDA variants should be used. In case of few data it is probably impossible to determine a good $R$ for the beginning the maximum value is a good choice, which could be later on adapted.

The one-class approach cannot be directly part of the optimization because it is more a conceptual question of how to model the data. Nevertheless, when starting with unary classification, samples of the opposite class (e.g., outliers) might occur, which raise the desire to be integrated into the classifier. This is in fact possible in the model. Furthermore, there is even the possibility to remove the zero separation if enough data of real outliers is available but still use the old model and adapt it.

From our practical experience, the models often perform very similar. This is now reasonable due to their strong connections. Consequently, the less complex version is the best choice for the application.

**Summary**  This chapter showed that all the SVM variants are strongly connected and it is possible to somehow move between them. This collection of connections draws a more general picture of the different models and can be also seen as a very general classifier model. It can be used for different views on data classification from the teaching, learning, application, implementation, and the optimization perspective. Hence, these views should (hopefully) simplify the choice of the classifier and increase the understanding by not looking at a single variant but considering the complete graph/net of classifiers.

In future, the benefits of the new algorithms need to be investigated in detail in further applications. It would be good, to also have a smooth transition between solution algorithms of BRMM with $R$ equal or close to 1 and the larger values. (Maybe there is a solution for the SVR which can be transferred or vice versa.) Last but not least, algorithms for improved (online) tuning of the hyperparameters and the decision boundary need to be developed and analyzed.

**Related Publications**

# Chapter 2

# Decoding: **Backtransformation**

This chapter is based on:

Krell, M. M. and Straube, S. (2015). Backtransformation: A new representation of data processing chains with a scalar decision function. *Advances in Data Analysis and Classification*. submitted.

Dr. Sirko Straube contributed by reviewing my text and by several discussions which also led to my discovery of the backtransformation. I wrote the paper.

## Contents

This chapter presents our backtransformation approach to decode complex data processing chains.

The basis of machine learning is understanding the data [Chen et al., 2008], and generating meaningful features [Domingos, 2012, "Feature Engineering Is The Key", p. 84]. Looking at the pure values of data and the implementation and parameters of algorithms does usually provide no insights. Consequently, for numerous data types and processing algorithms, visualization approaches have been developed [Rieger et al., 2004, Rivet et al., 2009, Le et al., 2012, Haufe et al., 2014, Szegedy et al., 2014] as a better abstraction to enhance the understanding of the behavior of the applied algorithms and of the data. Here, the visualization of an algorithm is often realized in a similar way as for the input data. Sometimes knowledge about the algorithm or the data is used to provide a visualization which is easier to interpret or which provides further insights. For example, for frequency filters, the frequency response is a much more helpful representation than the pure weights of the filter. Furthermore, internal parts of an algorithm can give additional helpful information, too, like the support vectors of a SVM, the signal template matrix $A$ of the xDAWN[1], the covariance matrices used by the FDA, or the characteristics of a single neuron in an artificial neural network [Szegedy et al., 2014].

To come up with a representation/visualization gets way more complicated when algorithms are combined for a more sophisticated preprocessing before applying a final decision algorithm [Verhoeye and de Wulf, 1999, Rivet et al., 2009, Krell et al., 2013b, Kirchner et al., 2013, Feess et al., 2013], i.e., for processing chains. Under these circumstances, understanding and visualization of single algorithms does only explain single steps in the processing chain that are typically not independent from each other as outlined in the following examples.

- If the data of intermediate processing steps is visualized, the ordering of two filters will change the visualization but might have no effect on the result.
- If a dimensionality reduction algorithm like the PCA is used, visualizations will differ when different numbers of dimensions are retained. But reducing the feature dimension to 75% or 25% will make no difference on the whole decision process, if the classifier uses only 10% of the highest ranked principal components (e.g., a C-SVM with 1-norm regularization).
- Two completely different dimensionality algorithms are used, and exactly the same or completely different classifiers are added to the processing chain, but the effect on the data might be the same.

---

[1] The xDAWN is a dimensionality reduction algorithm and spatial filter for time series data, where the goal is to enhance an underlying signal, which occurs time locked [Rivet et al., 2009]. The common dimensionality algorithms linearly combine features to create a new set of reduced and more meaningful features. A spatial filter, like the xDAWN, combines the data of sensors/channels to new pseudochannels but applies the same processing at every time point. A typical temporal filter is a decimator which combines low-pass filtering and downsampling.

- In the worst case, a dimensionality reduction algorithm is applied, but leaving it out does not change the overall picture of the algorithm, because the classifier or the data does not require this reduction.

Hence, one is often interested in knowledge about the whole data transformation in the processing chain but a general approach for solving this problem is missing. This situation gets even worse the more complex the data and the associated processing chains become. If dimensionality reduction algorithms are used for example to reduce the complexity of the data and to get rid of the noise, the structure of the output data is usually very different from the original input after the reduction step. In such a case, it is very difficult to understand the connection between decision algorithm, preprocessing, and original data even if single parts can be visualized. Consequently, a concept for representing the complete processing chain in the domain and format of the original input data is required.

**State of the Art**  Several approaches are described in the literature to visualize the outcome and transformation of classification algorithms, but again, taking the perspective of a single processing step neglecting the processing history (i.e., the preceding algorithms).

A very simple approach for data in a two-dimensional space is given in the scikit-learn documentation[2] [Pedregosa et al., 2011]. We adapted the provided script (see Figure C.5) to a visualization of SVM variants in Figure 2.1.

If the classifier provides a probability fit as classification function, the approach from [Baehrens et al., 2010] can be applied. Its main principle is to determine the derivative of the probability fit to give information about the classifier dependent on a chosen sample. The result is the local importance of the data components concerning the sample of interest. Unfortunately, this calculation of the derivative is quite complex, difficult to automatize, computationally expensive, and does not consider any processing before the classification and is restricted to a small subset of classifiers. Nevertheless, in [Baehrens et al., 2010] a very good overview about existing methods is given and the benefits of their suggested approach but also the limitations are shown, which will also hold for our (general) approach.

The visualization of the FDA is discussed in [Blankertz et al., 2011] in the context of an EEG based BCI application with different views on the temporal, spatial, and spatio-temporal domain. Here, the classifier is applied on spatial features and visualized as a spatial filter together with an interpretation in relation to the original data and other spatial filters. For other visualizations, the classifier weights are not

---

[2]  http://scikit-learn.org/stable/auto_examples/plot_classifier_comparison.html

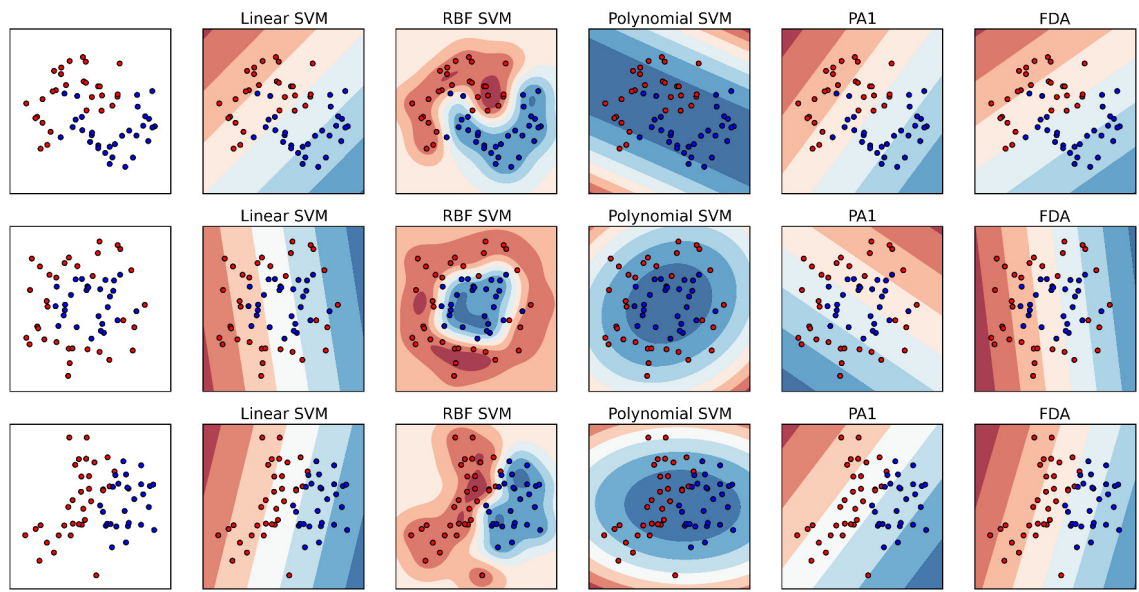

Figure 2.1: **Visualization of different classifiers trained on different datasets.**
Displayed are (from left to right) the training data with positive (red) and negative
(blue) samples, the C-SVM with linear, RBF, and polynomial kernel, the PAA with
hinge loss (PA1), and the FDA. The contour plots show the values of the classification
function, where dark red is used for high positive values which correspond to the
positive class and dark blue means very low negative values for the opposite class.
From top to bottom three different datasets are used.

directly used. Furthermore, no complex signal processing chain is used, even though
spatial filters are very common for the preprocessing of this type of data. The FDA
was applied to the raw data and largely improved with a shrinkage criterion. As a
side remark, they mention the possibility to visualize the FDA weights directly, when
applied to spatio-temporal features [Blankertz et al., 2011, paragraph before section
6, p. 18].

   This direct visualization of weights of a linear C-SVM has already been suggested
in [LaConte et al., 2005].[3]  This approach is intuitive, easy to calculate, and enables
a combination with the preprocessing. Furthermore, it can be generalized to other
data and other classifiers [Blankertz et al., 2011].

**Contribution**   Our concept, denoted as *backtransformation*, incorporates the afore-
mentioned approaches, but with the fundamental difference that it takes all prepro-
cessing steps in the respective chain into account. With this approach, we are able to
extract the complete transformation of the data from the chain, so that, e.g., changes
in the order of algorithms or the effect of insertions/deletions of single algorithms become immediately visible. Backtransformation also considers processing chains, where the original (e.g., spatio-temporal) structure of the data is hidden. The data processing chain is identified with a (composed) function, mapping the input data to a scalar. In its core, backtransformation is only the derivative of this function, calculated with the chain rule or numerically. The derivative is either calculated locally for each sample of interest (general backtransformation) or globally when the processing chain consists of affine transformations only (affine backtransformation). While the general backtransformation gives information on which components in the data have a large (local) influence on the decision process and which components are rather unimportant, the affine backtransformation is independent from the single sample.[4]

---

[3] Further methods are presented but they are tailored to fMRI data.

Numerous established data processing algorithms are affine transformations and it is often possible to combine them to process the data. In Section 2.2, a closer look is taken at this type of algorithms and it is shown that it is possible to retrieve the information on how the data is transformed by the complete decision process, even if a dimensionality reduction algorithm or a temporal filter hide information. The affine backtransformation iteratively goes back from the decision algorithm through all processing steps to determine a parameterization of the composed processing function and to enable a semantic interpretation. This results in a helpful representation of the processing chain, where each component in the source domain of the data gets a weight assigned showing its impact in the decision process. In fact, summing up the products of weights and respective data parts is equivalent to applying the single algorithms on the data step-by-step.

**General Setting** The requirement to apply the proposed backtransformation as outlined in the following is that the data processing is a concatenation of differentiable transformations (e.g., affine mappings) and that the last algorithm in the chain is a (decision) function which maps the data to a *single scalar*. The mapping to the label ($F(x)$) is not relevant, here.

For each processing stage, the key steps of the backtransformation are to first choose a mathematical representation of input and output data and then to determine a parameterization of the algorithm which has to be mapped to fit to the chosen data representations. Finally, the derivatives of the resulting transformations have to be calculated and iteratively combined. In its core it is the application of the chain rule for derivatives (see Section 2.1). For the case of using only affine mappings, it is just the multiplication of the transformation matrices, as shown in Section 2.2. Details on the implementation are given in Section 2.3. For an example of a processing chain of windowed time series data with a two-dimensional representation of the data see Figure 2.2 and Section 2.2.1.

---

[4] The respective derivatives are constant for every sample and as such not depending on it.

The backward modeling begins with the parametrization of the final decision function and continues by iteratively combining it backwards with the preceding algorithms in a processing chain. With each iteration, weights are calculated, which correspond to the components of the input data of the last observed algorithm.

For the abstract formulation of the backtransformation approach, data with a one-dimensional representation before and after each processing step is used. The output of each processing step is fed into the next processing algorithm.

## 2.1   Backtransformation using the Derivative

This section shortly introduces the general backtransformation. Let the input data be denoted with $x^{(0)} = x^{\text{in}} \in \mathbb{R}^{n_0}$ and let the series of processing algorithms be represented by differentiable mappings

$$F_0 : \mathbb{R}^{n_0} \to \mathbb{R}^{n_1}, \dots, F_k : \mathbb{R}^{n_k} \to \mathbb{R} \tag{2.1}$$

which are applied to the data consecutively.[5] Then, the application of the processing chain can be summarized to:

$$x^{\text{out}} = x^{(k+1)} = F(x^{(0)}) = (F_k \circ \dots \circ F_0)(x^{(0)}) \ . \tag{2.2}$$

With this notation, the derivative can be calculated with the chain rule:

$$\frac{\partial F}{\partial y}\left(x^{(0)}\right) = \frac{\partial F_k}{\partial y^{(k)}}\left(x^{(k)}\right) \cdot \frac{\partial F_{k-1}}{\partial y^{(k-1)}}\left(x^{(k-1)}\right) \cdot \dots \cdot \frac{\partial F_1}{\partial y^{(1)}}\left(x^{(1)}\right) \cdot \frac{\partial F_0}{\partial y^{(0)}}\left(x^{(0)}\right), \tag{2.3}$$

where $x^{(l)} \in \mathbb{R}^{n_l}$ is the respective input of the $l$-th algorithm in the processing chain with the mapping $F_l$ and $x^{(l+1)}$ is the output. The terms $\frac{\partial F_l}{\partial y^{(l)}}$ and $\frac{\partial F}{\partial y}$ represent the total differentials of the differentiable mappings and not the partial derivatives. Equation (2.3) is a matrix product. It can be calculated iteratively using the backtransformation matrices $B_l$ and the derivatives $\frac{\partial F_{l-1}}{\partial y^{(l-1)}}(x^{(l-1)})$:

$$B_k = \frac{\partial F_k}{\partial y^{(k)}}\left(x^{(k)}\right) \text{ and } B_{l-1} = \frac{\partial F_{l-1}}{\partial y^{(l-1)}}\left(x^{(l-1)}\right) \cdot B_l \text{ with } l = 1, \dots, k \ . \tag{2.4}$$

Now each matrix $B_l \in \mathbb{R}^{n_l \times 1}$ has the same dimensions as the respective $x^{(l)}$ and tells which change in the components of $x^{(l)}$ will increase (positive entry in $B_l$), decrease (negative entry), or will have no effect (zero entry) on the decision function. The higher the absolute value of an entry (multiplied with the esti-mated variance of the respective input), the larger is the influence of the respective data component on the decision function. Consequently, not only the backtransformation of the complete processing chain ($B_0$) but also the intermediate results ($B_l$; $l > 0$) might be used for analyzing the processing chain. $B_k$ is the matrix used in the existing approaches, which do not consider the preprocessing [LaConte et al., 2005, Baehrens et al., 2010, Blankertz et al., 2011]. Note that the $B_l$ are dependent on the input of the processing chain and are expected to change with changing input. So the information about the influence of certain parts in the data is only a *local* information. A *global* representation is only possible when using affine transformations instead of arbitrary differentiable mappings $F_l$.

---

[5] The notation of data and its components differs in this chapter in relation to Chapter 1, because instead of looking at training data, we look at one data sample $x^{(0)}$ with its different processing stages $x^{(l)}$ and the respective changes in each component of the data ($x^{(l)}_{gh}$).

## 2.2  Affine Backtransformation

For handling affine transformations like translations, the data vectors are augmented by adding a coordinate with value $1$ to have homogenous coordinates. Every affine transformation $F$ can be identified with a tuple $(A, T)$, where $A$ is a linear mapping matrix and $T$ a translation vector and the corresponding mapping of the processing algorithm applied on data $x^{\text{in}}$ reads as

$$x^{\text{out}} = F(x^{\text{in}}) = Ax^{\text{in}} + T = (A \ \ T) \begin{pmatrix} x^{\text{in}} \\ 1 \end{pmatrix} . \qquad (2.5)$$

So by extending the matrix $(A \ \ T)$ to a Matrix $A'$ with an additional row of zeros with a $1$ at the translational component, the mapping on the augmented data $x'^{\text{in}} = \begin{pmatrix} x^{\text{in}} \\ 1 \end{pmatrix}$ can be written in the simple notation: $x'^{\text{out}} = A' x'^{\text{in}}$. With a processing chain with corresponding matrices $A'_0, \ldots, A'_k$ the transformation of the input data $x'^{\text{in}}$ can be summarized to

$$x'^{\text{out}} = A'_k \cdot \ldots \cdot A'_1 \cdot A'_0 \cdot x'^{\text{in}} . \qquad (2.6)$$

With this notation, the backtransformation concept now boils down to iteratively determine the matrices

$$B_k = A'_k , B_{k-1} = A'_k \cdot A'_{k-1} , \ldots , \text{ and } B_0 = A'_k \cdot A'_{k-1} \cdot \ldots \cdot A'_1 \cdot A'_0 . \qquad (2.7)$$

This corresponds to a convolution of affine mappings.[6] Each $B_l \in \mathbb{R}^{n_k \times 2}$ defines the mapping of the data from the respective point in the processing chain (after $l$ previous processing steps) to the final decision value. So *each product* $B_l$ consists of a weighting vector $w^{(l)}$ and an offset $b^{(l)}$ (and the artificial second row with zero en-tries and 1 in the last column). The term $w^{(l)}$ can now be used for interpretation and understanding the respective sub-processing chain or the complete chain with $w^{(0)}$ (see Section 2.4). The following section renders possible (and impossible) algorithms which can be used for the affine backtransformation and how the weights from the backtransformation are determined in detail for a data processing chain applied on two-dimensional data.

---

[6] Note that no matrix inversion is required even though one might expect that, because the goal is to find out what the original mapping was doing with the data which sounds like an inverse approach.

### 2.2.1   Affine Backtransformation Modeling Example

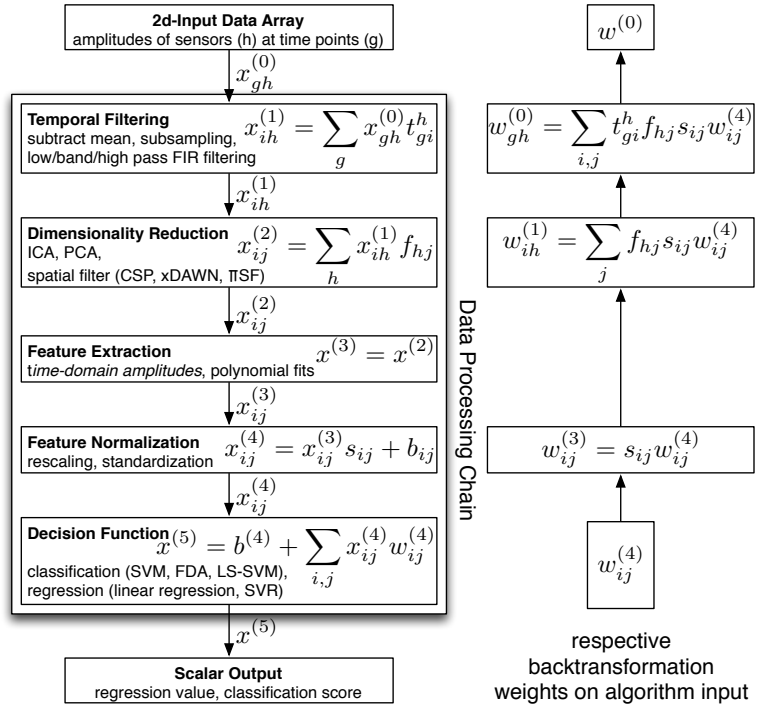

Figure 2.2: **Illustrative data processing chain scheme with examples of linear algorithms and the formulas for the backtransformation in short.** Spatio-temporal data $x_{gh}^{(0)}$ are processed from top to bottom ($x^{(5)}$). Every component of the scheme is optional. Backtransformation takes the classifier parametrization $w^{(4)}$ and projects it iteratively back ($w^{(k)}$) through the processing chain and results in a representation $w^{(0)}$ corresponding to the input domain. For more details refer to Section 2.2.1.

In this section, a more concrete example of applying the backtransformation principle is given for processing time series epochs of fixed length of several sensors with the same sampling frequency. Examples for affine transformations are given to show that there is a large number of available algorithms to construct a good processing chain. Some cases will be highlighted which are not affine. A possible processing chain is depicted in Figure 2.2. Note that all components of this chain are optional and the presented scheme can be applied to an arbitrary data processing chain of affine maps even if dimensions like time and space are replaced by others or left out (see Section 2.2 and 2.4.2).

An intuitive way of handling such data is to represent it as two-dimensional arrays with the time on one axis and space (e.g., sensors) on the other axis, since important preprocessing steps like temporal and spatial filters just operate on one axis. So this type of representation eases the use and the parameterization of these algorithms compared to the aforementioned mathematically equivalent one-dimensional representation. Furthermore, a two-dimensional representation of the data helps for its visualization and interpretation. For parametrization of the two-dimensional arrays, the common double index notation is used, where the data $x^{(0)}$ is represented by its components $x_{gh}^{(0)}$ with temporal index $g$ and spatial index $h$. This index scheme will be kept for all processing stages even if the data could be represented as one-dimensional feature vectors for some stages. The same indexing scheme can be applied for the parametrization of the affine data processing algorithms in the chain as will be shown in the following. As before, the input of the i-th algorithm is denoted with $x^{(i-1)}$ and the output with $x^{(i)}$ respectively. To fit to the concept of backtransformation, first the parametrization of the decision algorithm will be introduced and then the preceding algorithms step-by-step . An overview of the processing chain, the chosen parameterizations, and the resulting weights from the backtransformation is depicted in Figure 2.2.

### 2.2.1.1 Linear Decision Function

A linear decision function can be parameterized using a decision vector/matrix $w_{ij}^{(4)} \in \mathbb{R}^{m_i \times n_j}$ and an offset $b^{(4)} \in \mathbb{R}$. The transformation of the input $x^{(4)} \in \mathbb{R}^{m_i \times n_j}$ to the decision value $x^{(5)} \in \mathbb{R}$ is then defined as

$$x^{(5)} = b^{(4)} + \sum_{i=1}^{m_i} \sum_{j=1}^{n_j} x_{ij}^{(4)} w_{ij}^{(4)} \ , \tag{2.8}$$

with $m_i$ time points and $n_j$ sensors. Examples for machine learning algorithms with linear decision function are all the algorithms introduced in Chapter 1 without kernel or linear kernel. Using a RBF kernel would result in a smooth but not linear decision function. Even worse, working with a decision tree [Comité et al., 1999] as classifier would result in a non-differentiable decision function such that even the general backtransformation could not be applied.

### 2.2.1.2   Feature Normalization

With a scaling $s \in \mathbb{R}^{m_i \times n_j}$ and transition $b \in \mathbb{R}^{m_i \times n_j}$ and the same indexes as for the linear decision function, an affine feature normalization can be written as

$$x_{ij}^{(4)} = x_{ij}^{(3)} s_{ij} + b_{ij} \text{ with } i \in \{1, \dots m_i\} \text{ and } j \in \{1, \dots n_j\} \ . \tag{2.9}$$

This covers most standard feature normalization algorithms like rescaling or standardization [Aksoy and Haralick, 2001]. Nonlinear scalings, e.g., using absolute values as in $\min\left\{10, \left|x_{ij}^{(3)}\right|\right\}$, or sample dependent scalings, e.g., division by the Euclidean norm $s_{ij} = \frac{1}{\left\|x^{(3)}\right\|_2}$, are not affine mappings and could not be used here. For the affine backtransformation the formula of the feature normalization needs do be inserted into the formula of the decision function:

$$x^{(5)} = b^{(4)} + \sum_{i,j} \left( x_{ij}^{(3)} s_{ij} + b_{ij} \right) w_{ij}^{(4)} = b^{(3)} + \sum_{i,j} x_{ij}^{(3)} s_{ij} w_{ij}^{(4)} \ . \tag{2.10}$$

Here, $b^{(3)} = b^{(4)} + \sum_{i,j} b_{ij}$ summarizes the offset. As denoted in Figure 2.2, $s_{ij} w_{ij}^{(4)}$ is the weight to the input data part $x_{ij}^{(3)}$.

### 2.2.1.3   Feature Generation

For simplicity, the data amplitudes at different sensors have been directly taken as features and nothing needs to be changed in this step $\left( x^{(3)} = x^{(2)} \right)$. Other linear features like polynomial fits would be possible, too [Straube and Feess, 2013]. Nonlinear features (e.g., standard deviation, sum of squares, or sum of absolute values of each sensor) would not work for the affine backtransformation but for the general one. Symbolic features, mapped to natural numbers will be even impossible to analyze with the general backtransformation.

### 2.2.1.4   Dimensionality Reduction on the Spatial Component

A spatial filter transforms real sensors to new pseudo sensors by linear combination of the signal of the original sensors. To use well known dimensionality reduction algorithms like PCA, and independent component analysis [Jutten and Herault, 1991, Hyvärinen, 1999, Rivet et al., 2009] (ICA) for spatial filtering, the space component of the data is taken as feature component for these algorithms and the time component for the samples. Examples for typical spatial filters are common spatial patterns [Blankertz et al., 2008] (CSP), xDAWN [Rivet et al., 2009, Wöhrle et al., 2015], and $\pi$SF [Ghaderi and Straube, 2013].

The backtransformation with the spatial filtering is the most important part of the concept, because spatial filtering hides the spatial information needed for visualization or getting true spatial information into the classifier.[7]

The number of virtual sensors ranges between the number of real sensors and one. The spatial filter for the j-th virtual sensor is a tuple of weights $f_{1j}, ..., f_{n_h j}$ defining the linear weighting of the $n_h$ real channels. The transformation for the i-th time point is written as

$$x_{ij}^{(3)} = \sum_{h=1}^{n_h} x_{ih}^{(1)} f_{hj} \; ,$$

(2.11)

where the time component could be ignored, because the transformation is independent of time. The transformation formula can be substituted into formula (2.11):

$$x^{(5)} = b^{(3)} + \sum_{i,j} \sum_{h=1}^{n_h} x_{ih}^{(1)} f_{hj} s_{ij} w_{ij}^{(4)}$$

(2.12)

$$= b^{(3)} + \sum_{i,h} x_{ih}^{(1)} \cdot \left( \sum_j f_{hj} s_{ij} w_{ij}^{(4)} \right) \; .$$

(2.13)

Equation (2.13) shows, that the weight $\sum_j f_{hj} s_{ij} w_{ij}^{(4)}$ is assigned to the input data component $x_{ih}^{(1)}$. If there is no time component, a spatial filter is just a linear dimensionality reduction algorithm. It is also possible to combine different reduction methods or to do a dimensionality reduction after the feature generation.

### 2.2.1.5  Detrending, Temporal Filtering, and Decimation

There are numerous discrete-time signal processing algorithms [Oppenheim and Schafer, 2009]. Detrending the mean from a time series can be done in several ways. Having a time window, a direct approach would be to subtract the mean of the time window, or to use some time before the relevant time frame to calculate a guess for the mean (baseline correction). Often, such algorithms can be seen as finite impulse response (FIR) filters, which eliminate very low frequencies. Filtering the variance is a quadratic filter [Krell et al., 2013c] and infinite impulse response (IIR) filters have a feedback part. Both filters are not applicable for the affine backtransformation, because they have no respective affine transformations. One can either use uniform temporal filtering, which is similar to spatial filtering with changed axis, or introduce different filters for every sensor. As parametrization, $t_{gi}^h$ is chosen for the weight at sensor $h$ for the source $g$ and the

---

[7] This was also the original motivation to develop this concept.

resulting time point $i$ with a number of $m_g$ time points in the source domain:

$$x_{ih}^{(1)} = \sum_{g=1}^{m_g} x_{gh}^{(0)} t_{gi}^h . \tag{2.14}$$

Starting with the more common filter formulation as convolution (filter of length $N$):

$$x_{ih}^{(1)} = \sum_{l=0}^{N} a_l \cdot x_{(n-l)h}^{(0)} \overset{g:=n-l}{=} \sum_{g=n-N}^{n} a_{(n-g)} \cdot x_{gh}^{(0)} , \tag{2.15}$$

the filter coefficients $a_i$ can be directly mapped to the $t_{gi}^h$ and the other coefficients can be set to zero.

Reducing the sampling frequency of the data by downsampling is a combination of a low-pass filter and systematically leaving out several time points after the filtering (decimation). When using a FIR filter, the given parameterization of a temporal filter can be used here, too. For leaving out samples, the matrix $t_{gi}$ for channel $h$ can be obtained from an identity matrix by only keeping the rows, where samples are taken from.

The final step is similar to the spatial filtering part:

$$x^{(5)} = b^{(3)} + \sum_{i,h} \left( \sum_{g=1}^{m_g} x_{gh}^{(0)} t_{gi}^h \right) \cdot \left( \sum_j f_{hj} s_{ij} w_{ij}^{(4)} \right) \tag{2.16}$$

$$= b^{(3)} + \sum_{g,h} x_{gh}^{(0)} \cdot \left( \sum_{i,j} t_{gi}^h f_{hj} s_{ij} w_{ij}^{(4)} \right) \tag{2.17}$$

$$= b^{(3)} + \sum_{g=1}^{m_g} \sum_{h=1}^{n_h} x_{gh}^{(0)} w_{gh}^{(0)} . \tag{2.18}$$

The input component of the original data $x_{gh}^{(0)}$ finally gets assigned the weight $w_{gh}^{(0)} = \sum_{i,j} t_{gi}^h f_{hj} s_{ij} w_{ij}^{(4)}$. Note that for some applications it is good to work on normalized and filtered data for interpreting data and the behavior of the data processing. In that case, the backtransformation is stopped before the temporal filtering and the respective weights are used.

### 2.2.1.6 Others

The aforementioned algorithms can be combined and repeated (e.g., concatenations of FIR filters or PCA and xDAWN). Having a different feature generator, multiple filters, decimation, or skipping a filter or normalization the same calculation scheme could be used resulting in different $b^{(3)}$ and $w^{(0)}$. Nevertheless, $w^{(0)}$ has the same indexes as the original data $x^{(0)}$. After the final mapping to a scalar by the deci-sion function, a shift of the decision criterion (e.g., using threshold adaptation as suggested in [Metzen and Kirchner, 2011]) is possible but has no impact on the backtransformation because it only requires $w^{(0)}$ and not the offset. If a probability fit [Platt, 1999b, Lin et al., 2007, Baehrens et al., 2010] was used, this step has to be either ignored or the general approach (Section 2.1) has to be applied. Since the probability fit is a mostly sigmoid function which maps $\mathbb{R} \to [0, 1]$, it is also possible to visualize its derivative separately. For the interpretation concerning a sample, the function value is determined and the respective (positive) derivative is multiplied with the affine transformation part to get the local importance. Hence, the relations between the weights remain the same but the absolute values only change. This approach of mixing the calculations is much easier to implement.

If nonlinear preprocessing is used to normalize the data (e.g., to have variance of one), the normalized data can be used as input for the backtransformation and the respective processing chain. This might be even advantageous for the interpretation when the visualization of the original data is not helpful due to artifacts and outliers. An example for such a case is to work with normalized image data like the MNIST dataset (see Section 1.3.4.4) instead of the original data, where the size of the images and the position of the digits varied a lot (see also Section 2.4.2 and Section 2.4.3).

## 2.3 Generic Implementation of the Backtransformation

This section gives information on how to apply the backtransformation concept in practice especially when the aforementioned calculations are difficult or impossible to perform and a "generic" implementation is required to handle arbitrary processing chains.

The backtransformation has been implemented in pySPACE (see also Section 3) and can be directly used. This modular Python software gives simple access to more than 200 classification and preprocessing algorithms and so it provides a reasonable interface for a generic implementation. It provides data visualization tools for the different processing stages and largely supports the handling of complex processing chains.

In practice, accessing the single parameterizations for the transformation matrices $A_i$ for the affine backtransformation might be impossible (e.g., because external libraries are used without access to the internal algorithm parameters) or too difficult (e.g., code of numerous algorithms needs to be written to extract these parameters). In this case, the backtransformation approach cannot be applied directly in the way it is described in Section 2.2. Instead, the respective products and weights for the affine backtransformation can be reconstructed with the following trick which only requires the algorithms to be affine. No access to any parameters is needed. First, the offset of the transformation product is obtained by processing a zero data sample with the complete processing chain. The processing function is denoted by $F$. The resulting scalar output is the offset

$$b^{(0)} = F(0). \tag{2.19}$$

Second, a basis $\{e_1, \ldots, e_n\}$ of the original space (e.g., the canonical basis) needs to be chosen. In the last step, the weights $w_i^{(0)}$, which directly correspond to the base elements, are determined by also processing the respective base element $e_i$ with the processing chain and subtracting the offset $b^{(0)}$ from the scalar output:

$$w_i^{(0)} = F(e_i) - F(0). \tag{2.20}$$

The calculation of the derivative for the general backtransformation approach is more complicated. Deriving and implementing the derivative function for each algorithm used in a processing chain and combining the derivatives can be very difficult, especially if the goal is to implement it for a large number of relevant algorithms, e.g., as provided in pySPACE. The variety of possible derivatives even of classification functions can be very diverse [Baehrens et al., 2010]. A generic approach would be to use automatic differentiation tools [Griewank and Walther, 2008]. These tools generate a program which calculates the derivative directly from the program code. They can also consider the concatenation of algorithms by applying the chain rule. For most standard implementations, open source automatic differentiation tools could be applied. For existing frameworks, it is required to modify each algorithm implementation such that the existing differentiation tools know all derivatives of used elemental functions used in the code, which might be a lot of work. Furthermore, this approach would be impossible if black box algorithms were used. So for simplicity, a different approach, which is similar to the previous one for the affine case can be chosen. This is the numerical calculation of the derivative of the complete decision function via differential quotients for directional derivatives:

$$\frac{\partial F}{\partial e_i}(x_0) \approx \frac{F(x_0 + he_i) - F(x_0)}{h} \ . \tag{2.21}$$

Here, $e_i$ is the $i$-th unit vector, and $h$ is the step size. It is difficult to choose the optimal $h$ for the best approximation, but for the backtransformation a rough approximation should be sufficient. A good first guess is to choose $h = 1.5 \cdot 10^{-8} \langle x_0, e_i \rangle$ if $\langle x_0, e_i \rangle \neq 0$ and in the other case $h = 1.5 \cdot 10^{-8}$ [Press, 2007]. In the backtransformation implementation in pySPACE, the value of $1.5 \cdot 10^{-8}$ can be exchanged easily by the user. It is additionally possible to use more accurate formulas for the differential quotient at the cost additional function evaluations like

$$\frac{\partial F}{\partial e_i}(x_0) \approx \frac{F(x_0 - he_i) - 8F(x_0 - \frac{h}{2}e_i) + 8F(x_0 + \frac{h}{2}e_i) - F(x_0 - he_i)}{6h} \,. \tag{2.22}$$

## 2.4 Applications of the Backtransformation

Having a transformation of the decision algorithm back through different data representation spaces to the original data space might help for the understanding and interpretation of processing chains in several applications (e.g., image detection, classification of neuroscientific data, robot sensor regression) as explained in the following. First, some general remarks will be given on visualization techniques. Afterwards, the affine and the general backtransformation will be applied on handwritten digit classification (Section 2.4.2, Section 2.4.3, and Section 2.4.4) because it is a relatively simple problem which can be understood without expert knowledge. A more complex example on EEG data classification is given in Section 2.4.5. Finally, an outlook on the possibility of more sophisticated usage is given with processing chain manipulation. The affine backtransformation can be additionally used for ranking and regularization of sensors (see Section 3.4.3)

### 2.4.1 Visualization in General

As suggested in [LaConte et al., 2005] for fMRI data, the backtransformation weights could be visualized in the same way as the respective input data is visualized. This works only if there is a possibility to visualize the data and if this visualization displays the "strength" of the values of the input data. Otherwise, additional effort has to be put into the visualization, or the weights have to be analyzed as raw numbers. For interpreting the weights, it is usually required to also have the original data visualized for comparison (as averaged data or single samples) because higher weights in the backtransformation could be rendered meaningless if the corresponding absolute data values are low or even zero. Additionally to the backtransformation visualization of *one* data processing chain, different chains (with different hyperparameters, training data, or algorithms) can be compared (see Section 2.4.4). Differences in the weights directly correspond to the differences in the processing. Normally, weights with high absolute values correspond to important components for the processing and weights close to zero are less important and might be even omitted. This very general interpretation scheme does not work for all applications. In some cases, the weights have to be set in relation to the values of the respective data components: If data values are close to zero, high weights might still be irrelevant, and vice versa. To avoid such problems, it is better to take normalized data, which is very often also a good choice for pure data visualization. Another variant to partially compensate for this issue is to also look at the products of weights and the respective data values.

According to [Haufe et al., 2014], the backtransformation model is a *backward model* of the original data and as such mixes the reduction of noise with the emphasis of the relevant data pattern. To derive the respective *forward model* they suggest to multiply the respective weighting vector with the covariance matrix of the data. From a different perspective, this approach sounds reasonable, too: If backtransformation reveals that a feature gets a very high weight by the processing chain, but this feature is zero for all except one outlier sample a modified backtransformation would reveal this effect. Furthermore, if a feature is highly correlated with other features, a sparse classifier might just use this one feature and skip the other features which might lead to the wrong assumption, that the other features are useless even though they provide the same information. On the other hand, if features are highly correlated as it holds for EEG data this approach might be also disadvantageous. The processing chain might give a very high weight to the feature, where the best distinction is possible, but the covariance transformation will blur this important information over all sensors and time points. Using such a blurred version for feature selection would be a bad choice. Another current drawback of the method from [Haufe et al., 2014] is that it puts some assumptions on the data which often do not hold: The expectancy values of noise, data, and signal of interest are assumed to be zero "w.l.o.g." (without loss of generality). Hence, more realistic assumptions are necessary for better applicability.

Note that in Figure 2.2, Section 2.2, and Section 2.2.1 it has been shown that every iteration step in the backtransformation results in weightings $w^{(i)}$ which correspond to the data $x^{(i)}$. This data is obtained by applying the first $i$ algorithms of the processing chain on the original input data $x^{(0)}$. So depending on the application, it is even possible to visualize data and weights of intermediate processing steps. This can be used to further improve the overall picture of what happens in the processing chain.

### 2.4.2   Handwritten Digit Classification: Affine Processing Chain

For a simple application example of the affine backtransformation approach, the MNIST dataset is used (see Section 1.3.4.4). These normalized greyscale images have an inherent structure due to $28 \times 28$ used pixels. but they are stored as one-dimensional feature vectors (784 features). For processing, we first applied a PCA on the feature vectors and reduced the dimension of the data to $4$ (or $64$). As a second step, the resulting features were normalized to have zero mean and standard deviation of one on the training data. Finally, a linear C-SVM (LIBSVM) with a fixed regularization parameter (value: 1) is trained on the normalized PCA features. Without backtransformation, the filter weights for the 4 (or 64) principal components could be visualized in the domain of the original data and the single (4 or 64) weights assigned by C-SVM could be given, but the interplay between C-SVM and PCA would remain unknown, especially if all 784 principal components would be used. This information can only be given with backtransformation and is displayed in Figure 2.3 for the distinction of digit pairs (from 0, 1, and 2). The generic implementation of the affine backtransformation was used, since only affine algorithms were used in the processing chain (PCA, feature standardization, linear classifier). The forward model to the backtransformation, obtained by multiplication with the covariance matrix, is also visualized in Figure 2.3. Note that the original data is not normalized (zero mean), although this was an assumption on the data for the covariance transformation approach from [Haufe et al., 2014]. Nevertheless, the resulting graphics look reasonable.

Generally, it can be seen that the classifier focuses on the digit parts, where there is no overlay between the digits on average. For one class there are high positive values and for the other there are high negative weights. For the classification with 64 principal components, the covariance correction smoothes the weight usage and results in a visualization which is similar to the visualization of the backtransformation for the classification with 4 principal components. Hence, the 60 additional components are mainly used for canceling out "noise".

### 2.4.3 Handwritten Digit Classification: Nonlinear Classifier

To show the effect of the generic backtransformation for a nonlinear processing chain, the evaluation of Section 2.4.2 is repeated with a RBF kernel for C-SVM instead of a linear one. The hyperparameter of the kernel, $\gamma$, has been determined according to [Varewyck and Martens, 2011]. Everything else remained unchanged. Again the generic implementation was used. Note that every sample requires its own backtransformation. So for the visualization of the backtransformation, only the first four single samples were taken.

It can be clearly seen in Figure 2.4 that there is a different backtransformation for each sample. Similar to the results in Section 2.4.2 (Figure 2.3), the backtransformation with covariance correction (when 64 principal components are taken as features) seems to be more useful in contrast to the raw visualization which also contains the noise cancellation part. This is surprising because this approach has been originally developed for linear models and not for nonlinear ones [Haufe et al., 2014]. Using a correction with a "local" covariance would be more appropriate in this case but more demanding from the computation and implementation point of view. A large number of principal components seems to be a bad choice for the nonlinear kernel, because it does not seem to generalize that well and is using a lot of small components instead of focusing on the big shape of the digits.

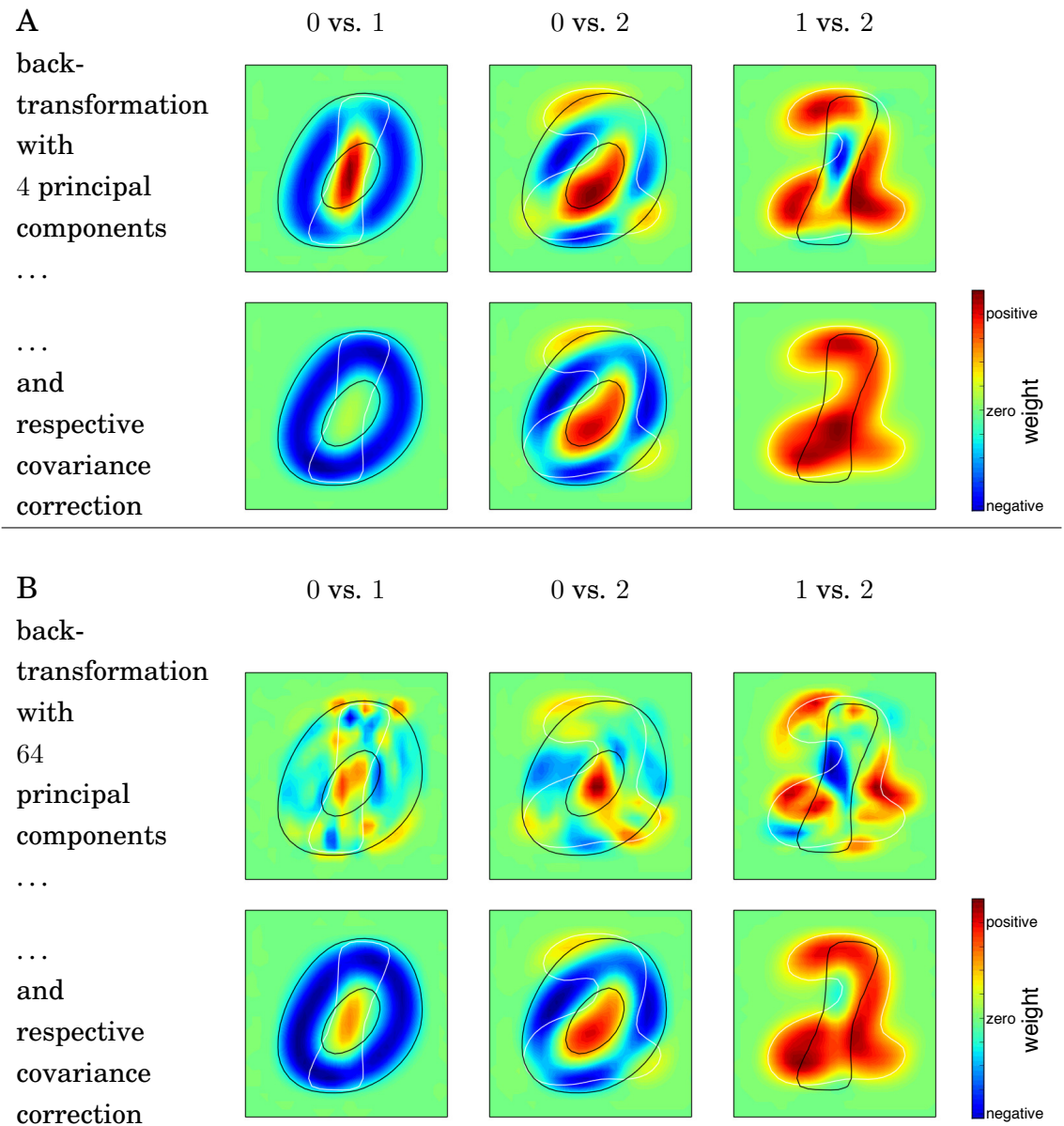

Figure 2.3: **Contour plots of backtransformation weights for handwritten digit classification:** The white and black silhouettes display an average contour of the original data (digits 0 vs. 1, 0 vs. 2, and 1 vs. 2). The colored contour plots show the respective weights in the classification process before and after covariance correction with a different number of used principal components (case A and B). Negative weights (blue) are important for the classification of the first class (black silhouette) and positive weights (red) for the second class (white silhouette). Green weights are close to zero and do only contribute weakly to the classification process.

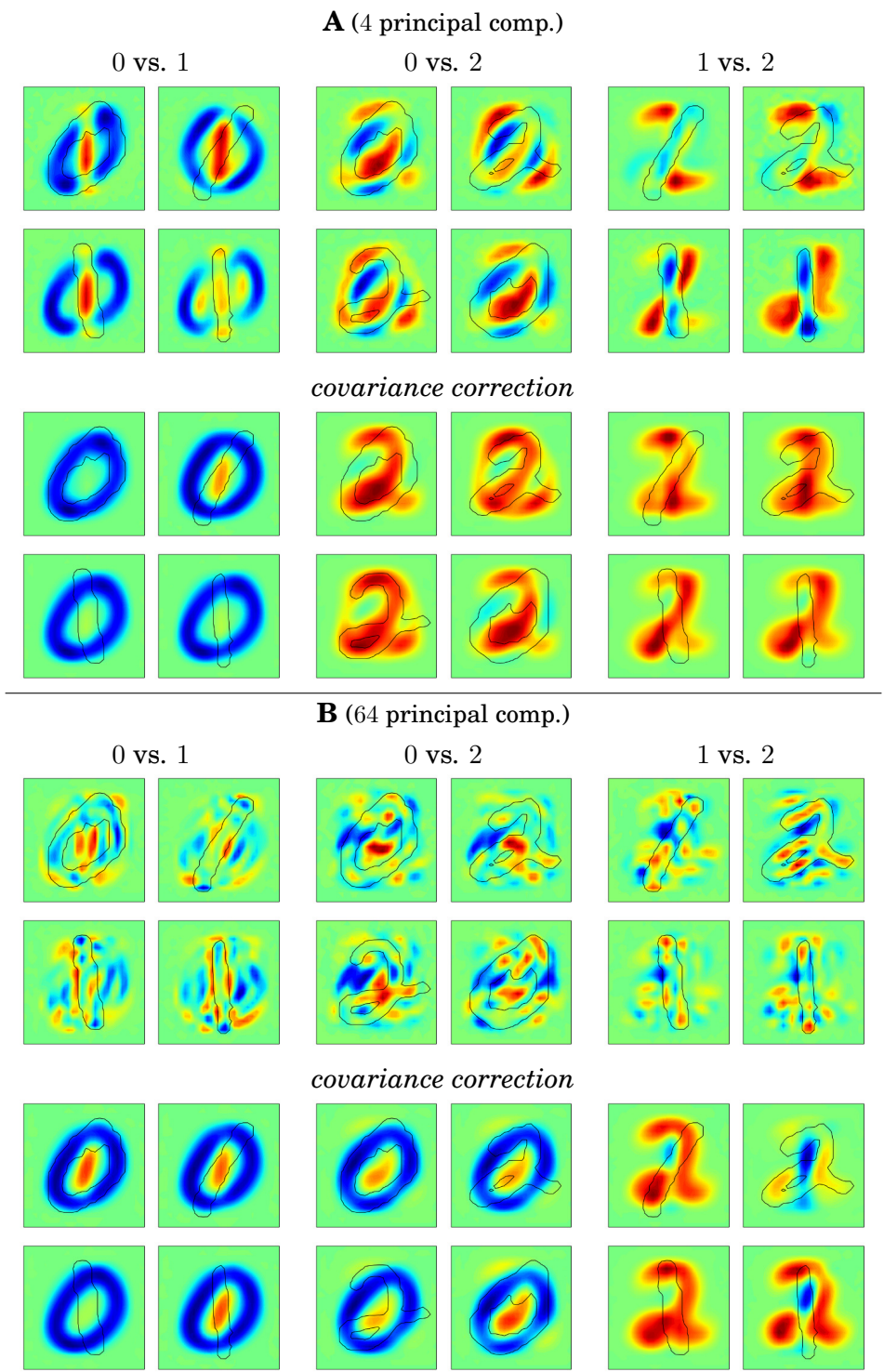

Figure 2.4: **Contour plots of backtransformation weights for handwritten digit classification with nonlinear classifier:** The setting is the same as in Figure 2.3 except that no average shapes are displayed but the shape of the sample of interest where the backtransformation is calculated for.

In case of using only 4 principal components, the approach mainly shows the shape of the digit 2 (or 0 for the first column). In contrast, the visualizations without covariance correction clearly indicate with a blue color which parts are relevant for classifying it as the first class and with the red color which parts are important for the second class. An interesting effect occurs for the first classifier at the fourth digit (1). Here a closer look could be taken at the classifier and the data to find out why there are yellow weights outside the regular shape of the digit 1. This might be the result of some artifacts in the data (e.g., a sample with very bad handwriting near to the observed sample) or an artifact in the processing.

In the nonlinear and the linear case with 64 principal components the backtransformation reveals that the decision process is not capable of deriving real shape features for the digits. This might be a reason, why a specially tuned deep neural network performs better in this classification task [Schmidhuber, 2012].

### 2.4.4  Handwritten Digit Classification: Classifier Comparison

This section is based on an evaluation in:

Krell, M. M., Straube, S., Wöhrle, H., and Kirchner, F. (2014c). Generalizing, Optimizing, and Decoding Support Vector Machine Classification. In *ECML/PKDD-2014 PhD Session Proceedings, September 15-19, Nancy, France*.

I wrote this paper completely on my own to have a first, very short summary of this thesis and reused some text parts. My coauthors helped me with reviews and discussions about the paper and my thesis in general.

Again, the MNIST dataset was used with the classification of the digits 0, 1, and 2; the data was reduced in dimensionality with PCA from 784 to 40; and then it was normalized with a standardization (zero mean and variance of one on the given training data). For classification, a squared loss penalization of misclassifications was used to obtain the more common Gaussian loss for RFDA and to be better comparable. RFDA, L2–SVM, the respective online SVM using the single iteration approach (see Section 1.2.4), and the $\nu$oc-SVM were compared. The classifiers were chosen as good representatives of the algorithms introduced in Chapter 1 and to compare their behavior on a visual level. Backtransformation can summarize all three processing steps and provides the respective weights belonging to the input data. This is visualized in Fig. 2.5.

The linear classifiers itself do only determine the 40 weights of the normalized principal components. These weights would be difficult to interpret, but with the given affine backtransformation the weighting and its correspondence to the average shapes can be observed. As expected due to the model similarities (single iteration ap-proach, Section 1.2) similar weight distributions were obtained for the L2–SVM and its online learning variant (PA2 PAA). The visualizations of L2–SVM and RFDA look similar due to the connection with BRMM (relative margin approach, Section 1.3). However, for the distinction between the two digits 0 and 2 some larger differences can be observed. The unary classifier is different to the other classifiers as expected because it has been trained on a single digit only (origin separation approach, Section 1.4). Nevertheless, characteristics of the other class can be marginally observed due to the use of PCA which has been trained on both classes. This can be seen in the second and third row: although trained on the digit 2 in both cases, the classification results look different.

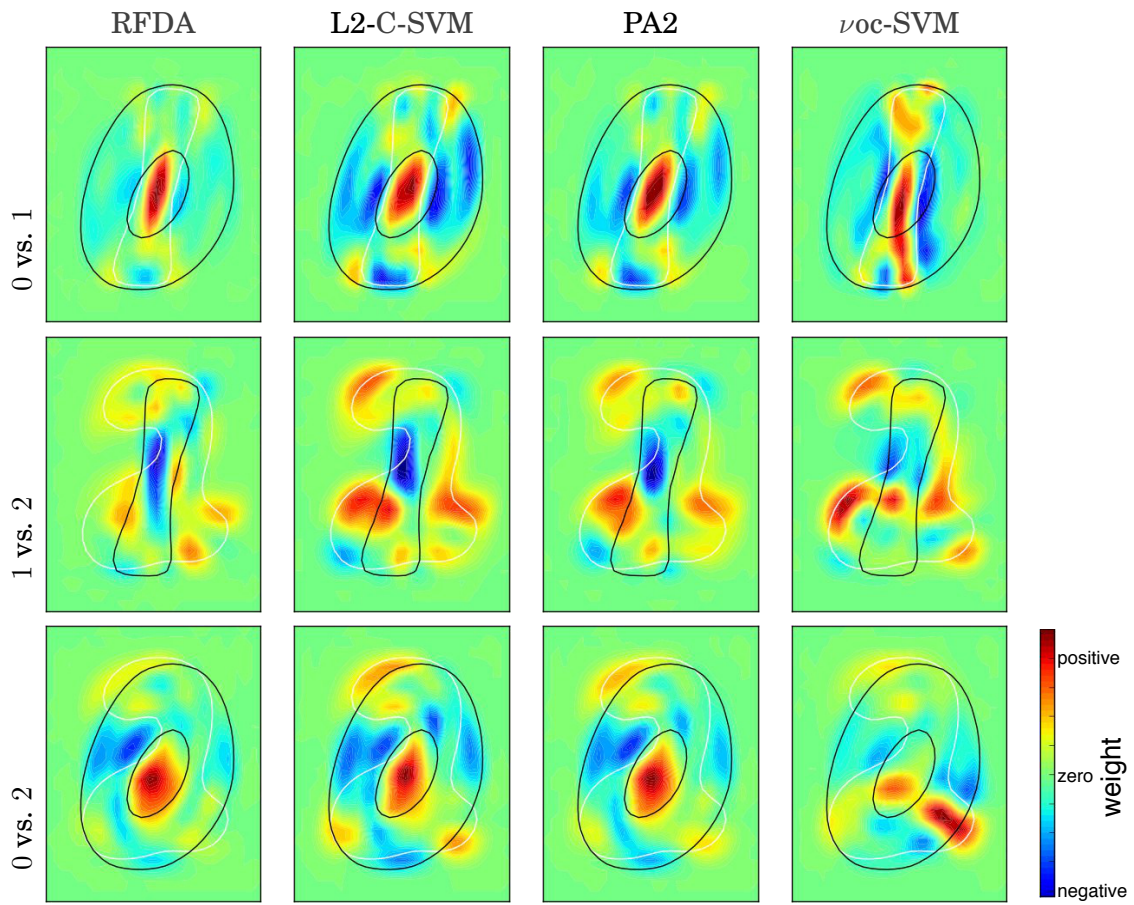

Figure 2.5: **Contour plots of backtransformation weights for handwritten digit classification with different classifiers:** The white and black silhouettes display an average contour of the original data (digits 0, 1, and 2). The colored contour plots show the respective weights in the classification process. Negative weights (blue) are important for the classification of the first class (black silhouette) and positive weights (red) for the second class (white silhouette). Green weights are close to zero and do not contribute to the classification process. For the unary classification, the second class (white) was used. Visualization taken from [Krell et al., 2014c].

## 2.4.5   Movement Prediction from EEG Data

The EEG is a very complex signal, measuring electrical activity on the scalp with a very high temporal resolution and more than 100 sensors. Several visualization techniques exist for this type of signal, which are used in neuroscience for analysis. When processing EEG data for BCIs, there is a growing interest in understanding the properties of processing chains and the dynamics of the data, to avoid relying on artifacts and to get information on the original signal back for further interpretation [Kirchner, 2014]. Here, very often spatial filtering is used for dimensionality reduction to linearly combine the signals from the numerous electrodes to a largely reduced number of new virtual sensors with much less noise (see Section 2.2.1.4). These spatial filters and much more importantly the data patterns they are enhancing are visualized with similar methods as used for visualizing data. If the spatial filter is the main part of the processing (e.g., only two filters are used), this approach is sufficient to understand the data processing. However, often more filters and other, additional preprocessing algorithms are used. Hence, the original spatial information cannot be determined for the input of the classifier. This disables a good visualization of the classifier and an understanding of what the classifier learned from the training data. So here, backtransformation can be very helpful.

To illustrate this, a dataset from an EEG experiment was taken [Tabie and Kirchner, 2013]. In this experiment, subjects were instructed to move their right arm as fast as possible from a flat board to a buzzer in approximately 30 cm distance. The classification task was to predict upcoming movements by detecting movement-related cortical potentials [Johanshahi and Hallett, 2003] in the EEG single trials. Before applying the backtransformation and visualizing the data as depicted in Figure 2.6, the data has been normalized with a standardization, a decimation, and temporal filtering. Only the last part of the signal closed to the movement was visualized. The processing chain was similar to the one in Section 2.2.1. The details are described in [Seeland et al., 2013b].

The averaged input data in Figure 2.6 shows a very strong negative activation at the motor cortex mainly at the left hemisphere over the electrodes C1, Cz, and FCC1h.[8] This activation is consistent with the occurrence of movement related cortical potentials and is expected from the EEG literature [Johanshahi and Hallett, 2003]. The region of the activation (blue circle on the left hemisphere at the motor cortex region) is associated with right arm movements, which the subjects had to perform in the experiment.

---

[8] A standard extended 10 − 20 electrode layout has been chosen with 128 electrodes (see Figure C.6).

time before movement onset:

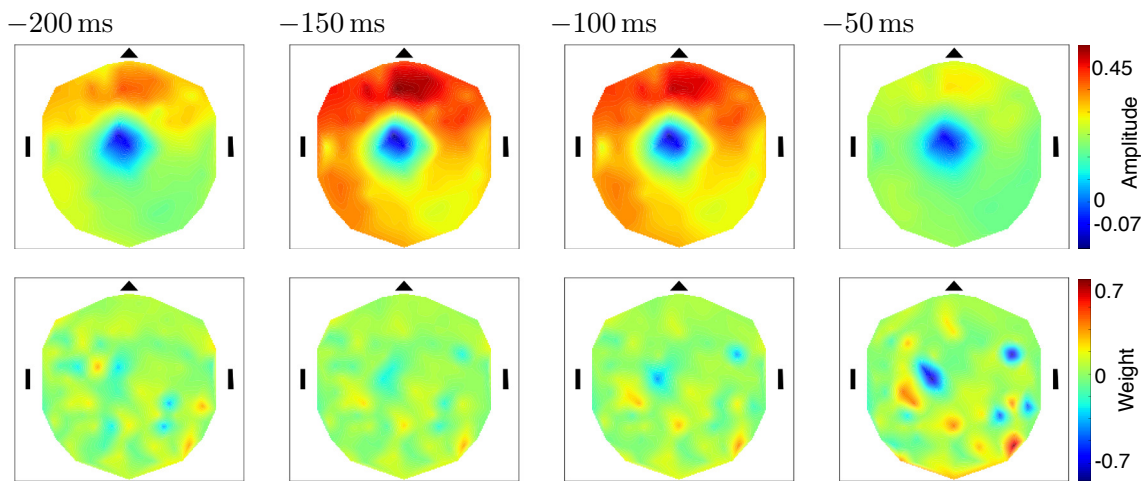

Figure 2.6: **Visualization of data for movement prediction and the corresponding processing chain:** In the first row the average of the data before a movement is displayed as topography plots and in the second row the backtransformation weights are displayed, respectively. The data values from the different sensors were mapped to the respective position on the head, displayed as an ellipse with the nose at the top and the ears on the sides.

The backtransformation weights are much more spread over the head compared to the averaged data. There is a major activation at the left motor cortex at electrodes C1 and CP3, but also a large activation at the back of the head at the right hemisphere around the electrode P8. On the time scale, the most important weights can be found at the last time point, $50\,\mathrm{ms}$ before movement onset.

This is reasonable, because the most important movement related information is expected to be just before the movement starts, although movement intention can be detected above chance level on average $460\,\mathrm{ms}$ before the movement onset [Lew et al., 2012]. Note that the analysis has been performed on single trials and not on averaged data and that for a good classification the largest difference is of interest and not the minimal one. The high weights at C1 and CP3 clearly fit to the high negative activation found in the averaged data and as such highlight the signal of interest. For interpreting the other weights, two things have to be kept in mind. First, EEG data usually contains numerous artifacts and second, due to the conductivity of the skin it is possible to measure every electric signal at a certain electrode also on the other electrodes. Keeping that in mind, the activation around P8 could be interpreted as a noise filter for the more important class related signal at C1 and CP3. This required filtering effect on EEG data is closely related to spatial filtering, which emphasizes a certain spatial pattern [Blankertz et al., 2011, section 4.2]. It could be also a relevant signal which cannot be observed in the plot of the averaged data. These observations are now a good starting point for domain experts to take a closer look at the raw data to determine which interpretation fits better.

### 2.4.6 Reinitialization of Linear Classifier with Affine Preprocessing

There could be several reasons for exchanging the preprocessing in a signal processing chain. For example, first some initial preprocessing is loaded but in parallel a new better fitting data specific processing is trained or tuned on new incoming data (e.g., a new spatial filter [Wöhrle et al., 2015]). If dimensionality would not be fitting after changing the preprocessing chain, a new classifier would also be needed. But even if dimensions of old and new preprocessing were the same it might be good to *adapt* the classifier to that change to have a better initialization. Here, the affine backtransformation can be used as described in the following.

For this application, a processing chain of affine transformations is assumed which ends with a sample weighting online learning algorithm like PAA. Since the classification function is a weighted sum of samples, it enables following calculation:

$$w = \sum_i \alpha_i y_i \hat{x}_i = \sum_i \alpha_i y_i (A x_i + T) = A \sum_i \alpha_i y_i x_i + T \sum_i \alpha_i y_i \tag{2.23}$$

$$= A w^{(0)} + T b \text{ with } w^{(0)} = \sum_i \alpha_i y_i x_i \text{ and } b = \sum_i \alpha_i y_i \ . \tag{2.24}$$

Here, $x_i$ is the training data with the training samples $y_i$ and $\hat{x}_1$ is the preprocessed training data given to the classifier. The weights $\alpha_i$ are calculated by update formulas of the classifier. During the update step, $w^{(0)}$ must be calculated additionally but neither $x_i$, $y_i$, nor $\alpha_i$ are stored. When changing the preprocessing from $(A, T)$ to $(A', T')$

$$w' = A' w^{(0)} \tag{2.25}$$

is a straightforward estimate for the new classifier. The advantage of this formula is, that it just requires additionally calculating and storing $w^{(0)}$. So the resulting classifier can be still used for memory efficient online learning. Especially, even if neither $(A', T')$ nor $(A, T)$ is known, $w'$ can be calculated using the new signal processing function $\hat{F}(x) = A'x + T'$:

$$w' = A' w^{(0)} = \hat{F}(w^{(0)}) - T' b = \hat{F}(w^{(0)}) - 0 A' w^{(0)} b - T' b = \hat{F}(w^{(0)}) - \hat{F}(0 w^{(0)}) b \ . \tag{2.26}$$

So, $w'$ can be computed by processing $w^{(0)}$ and a sample of zero entries in the signal processing chain. This only requires some minor processing time but no additional resources. Usually the processing chain is very fast and so the additional processing time should not be a problem.

For giving a proof of concept, the data introduced in Section 0.4 was used. We concatenated the 5 recordings of each subject and obtained 10 datasets with more than 4000 samples each. In a preceding preparation the data was standardized, decimated, and bandpass filtered (see first 4 nodes in Figure 3.4). As modular preprocessing, a chain was trained on each of the datasets consisting of the xDAWN filter (8 retained pseudo channels), a simple feature generator, which used the amplitudes of the signal as features, and a feature normalization linearly mapping each feature to the interval $[0, 1]$, assuming 5% of the data to be outliers (see Figure C.3). This modular processing chain was then *randomly* loaded[9] in a simulated incremental learning scenario, where a sample was first classified and then the classifier (PA1, see Section 1.1.6.2) directly got the right label and performed an update step. The classifier has not been trained before. After a fixed number of iterations, the preprocessing was again randomly changed, to analyze the effect of changing the preprocessing (for the specification file see Figure C.3). Due to the randomization, the preprocessing does not fit to the data. Consequently, with every change of the preprocessing a drop in performance is expected. In contrast, the incremental learning should increase the performance over time, because, the classifier adapts to the data and the preprocessing. For simplicity, the regularization parameter $C$ was fixed to 1 for the overrepresented target class and 5 for the other class. The BA was used as performance metric to account for class imbalance. The evaluation is repeated 10 times to have different randomizations. It is clear, that this setting is artificial, but for showing the problem of the classifier to deal with changing preprocessing and how our approach can can overcome this issue it is helpful.

In Figure 2.7 the positive effect of the backtransformation on the performance is shown, when the preprocessing is randomly changed after a varying amount of processed data samples. The new approach using backtransformation is not negatively affected by changing the preprocessing in contrast to the simple approach of not adapting the classifier to the different processing. There is even a slight improvement in performance. When changing the processing too often, the simple classifier without the backtransformation adaptation would be as good as a guessing classifier (performance of 0.5).

For Figure 2.8 the preprocessing is randomly changed every 1000 samples and the change of performance in time is displayed during incremental training. It can be clearly seen that performance dramatically drops when the preprocessing is changed after 1000 samples.

For the experiment, $w = 0$ and $b = 0$ was used for initialization and hyperparameters were not optimized but fixed. A different initialization or other hyperparameters might show better or worse performance in total, but the clear positive effect of the backtransformation as a good initialization after changing the preprocessing will be the same. The effect might get lost when also using incremental learning for the preprocessing and compensating the changing preprocessing in the classifier, because incremental learning in the preprocessing could generate stationary features from non-stationary data and backtransformation would undo this positive effect.

---

[9] The randomly chosen processing chain was trained on one of the 9 other datasets but not the one of the current evaluation.

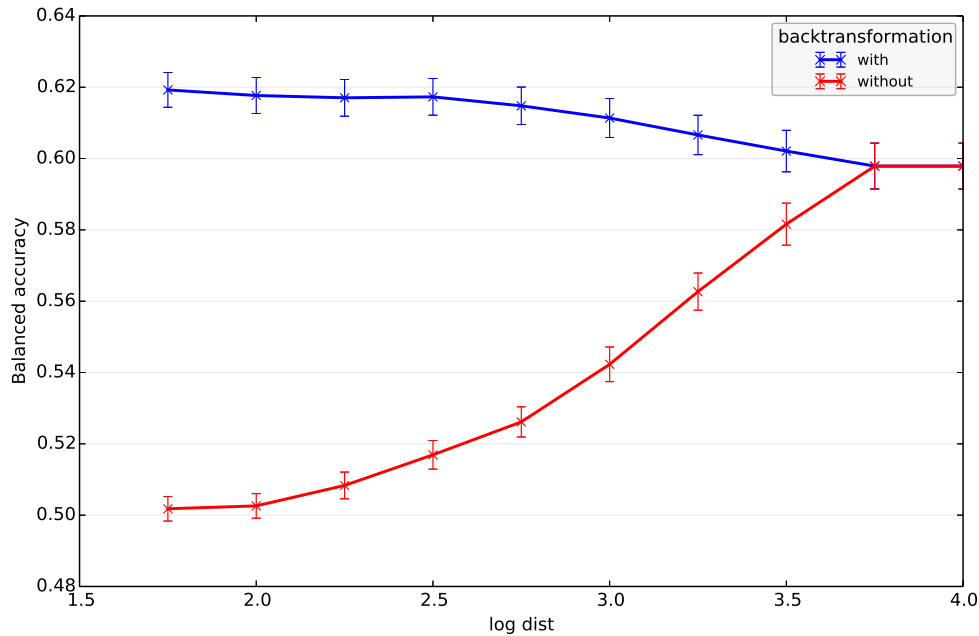

Figure 2.7: **Adaption to random preprocessing:** Performance (and standard error) of an online classifier which gets an incremental update after each incoming sample. After every $10^{\log \text{dist}}$ incoming samples the preprocessing is changed by randomly loading a new preprocessing.

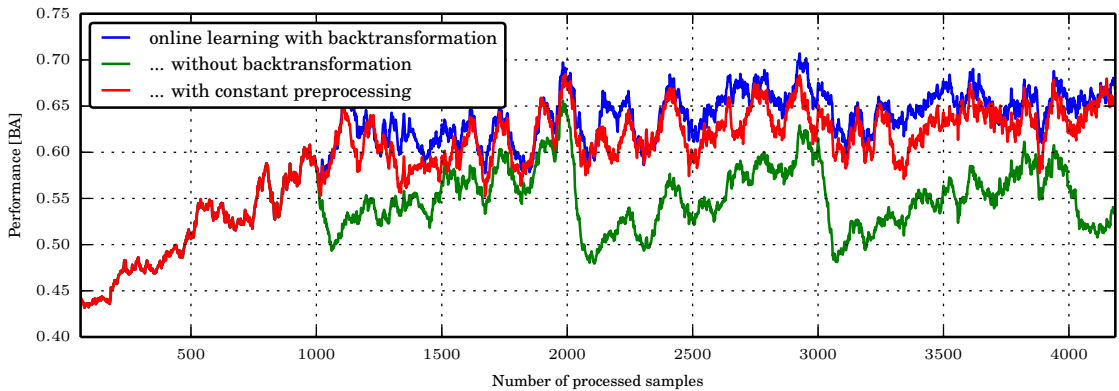

Figure 2.8: **Performance trace to random preprocessing after every** $1000$ **samples:** Performance of an online classifier which gets an incremental update after each incoming sample is displayed were the metric is the average over all evaluations. The metric BA is calculated with a moving window of $60$ samples as described by [Wöhrle et al., 2015].

## 2.5 Discussion

With the affine backtransformation, we introduced a direct approach to look at the complete data processing chain (in contrast to separate handling of its components) and to transform it to a representation in the same format as the data. We generalized the concept to arbitrary differentiable processing chains. We showed, that it is necessary and possible to break up the black box of classifier and preprocessing. The approach could be used to improve the understanding of complex processing chains and might enable several applications in future. It was shown that backtransformation can be used for visualization of the decision process and a direct comparison with a visualization of the data is possible and enables an interpretation of the processing.

Our approach extends existing algorithms by also considering the preprocessing, by putting no restrictions on the decision algorithm, by providing the implementation details, and by integrating the backtransformation in the pySPACE framework which already comes with a large number of available algorithms. The framework is required and very useful for the suggested generic implementation. A big advantage is, that our generic approach enables the usage of arbitrary (differentiable) processing algorithms and their combination. Due to the integration into a high-level framework, the backtransformation can be applied to different data types and applications and it can benefit from future extensions of pySPACE to new applications and new data types.

Backtransformation can be used for interpreting the behavior of the decision process, but it remains an open question of how the further analysis is performed, because additional investigations and expert knowledge might still be required. A related problem occurs when using temporal and spatial filters. Here the solution is to visualize the frequency response and the spatial pattern instead of the pure weights of the transformation. The frequency response gives information on how frequencies are filtered out and spatial patterns give information on which signal in space is emphasized by the respective spatial filter. It is important for the future to develop new methods, which improve the interpretability of the decision process. This could be achieved for example by extending the method of covariance multiplication with a more sophisticated calculation of the covariance matrix or by deriving a different for-mula for getting a forward model, which describes how the data is generated.[10] This might enable the backtransformation to reveal new signals or connections in the data which can then be used to improve the observed data processing chain. This improvement is especially important for longterm learning. If a robot shall generate its own expert knowledge from a self-defined decision process, the process of interpretations needs to be more automized.

In future, it would be also interesting to analyze the application of the backtransformation further, e.g., by using other data, processing chains, or decision algorithms like regression.

**Related Publications**

---

[10] In the context of EEG data processing, especially source localization methods might be very helpful because they enable an interpretation of the processing in relation to parts of the brain and not the raw sensors which accumulate the signals from different parts of the brain.

# Chapter 3

# Optimizing: pySPACE

This chapter is based on:

Krell, M. M., Straube, S., Seeland, A., Wöhrle, H., Teiwes, J., Metzen, J. H., Kirchner, E. A., and Kirchner, F. (2013b). pySPACE a signal processing and classification environment in Python. *Frontiers in Neuroinformatics*, 7(40):1–11, doi:10.3389/fninf.2013.00040.

For a clarification of my contribution, I refer to Section 3.5.2.

## Contents

This chapter presents the software pySPACE which allows to use the contributions from the previous chapters and furthermore design, optimize, and evaluate data processing chains. In the following, we will discuss typical data analysis problems illustrated by examples from neuroscientific and robotic data.

**Motivation**   Most data in neuroscience and robotics are not feature vector data but time series data from the different sensors which were used. Consequently, for this data a classifier (as for example introduced in Chapter 1) usually cannot be directly applied and a sophisticated preprocessing is required as for example explained in Section 2.2.1. There are also other areas where such data are used but we will use these two examples to show some problems where our approach might provide help.

Time series are recorded in various fields of neuroscience to infer information about neural processing. Although the direct communication between most parts of the nervous system is based on spikes as unique and discrete events, graded potentials are seen as reflections of neural population activity in both, invasive and non-invasive techniques. Examples for such time series come from recordings of local field potentials (LFPs), EEG, or even fMRI.

Common characteristics of time series data reflecting neural activity are: (i) a high noise level (caused by external signal sources, muscle activity, or overlapping uncorrelated brain activity) and (ii) a large amount of data that is often recorded with many sensors (electrodes) and with a high sampling rate. To reduce noise and size the data are preprocessed, e.g., by filtering in the frequency domain or by averaging over trials and/or sensors. These approaches have been very successful in the past, but the solutions were often chosen manually, guided by the literature, visual inspection and in-house written scripts, so that possible drawbacks remain. It is still not straightforward to compare or reproduce analyses across laboratories and the investigator has to face many choices (e.g., filter type, desired frequency band, and respective hyperparameters) that cannot be evaluated systematically without investing a large amount of time. Another critical issue is that the data might contain so far undiscovered or unexpected signal components that might be overseen by the choice of the applied data analysis. *False or incomplete hypotheses can be a consequence*. An automatic optimization of the processing chain might avoid such effects. On the other hand, the success of applications using automatically processed and classified neurophysiological data has been widely demonstrated, e.g., for usage of BCIs [Lemm et al., 2004, Bashashati et al., 2007, Hoffmann et al., 2008, Seeland et al., 2013b, Kirchner et al., 2013] and classification of epileptic spikes [Meier et al., 2008, Yadav et al., 2012]. These applications demonstrate that automated signal processing and classification can indeed be used to directly extract relevant information from such time series recordings.

Similar problems also apply for data in robotics. The noise level is usually lower but it might still cause problems. A big problem is the amount of data, which can/could be recorded with a robot in contrast to its limited processing power and memory. For example, deep see grippers are constructed with multimodal sensor processing to fulfill complex manipulation tasks [Aggarwal et al., 2015, Kampmann and Kirchner, 2015]. There are sensors in the numerous motors of more and more complex robots [Lemburg et al., 2011, Manz et al., 2013, Bartsch, 2014]. Sometimes internal sensors are used to enable or improve localization [Schwendner et al., 2014] but usually several other sensors are added to enable SLAM [Hildebrandt et al., 2014]. Often video image data is used for SLAM but also for object manipulation and terrain classification [Manduchi et al., 2005, Müller et al., 2014]. For the processing of this data, usually expert knowledge is used (as in neuroscience too) for constructing a feasible signal processing chain. But taking the expert out of the loop is necessary for real longterm autonomy of robots.

Solving all facets of the aforementioned problems, which are all connected to the problem of *optimizing* the signal processing chain, is probably impossible. Nevertheless, we will show that it is at least possible to improve the situation from the interface/framework perspective which can be used as the basis for further approaches. Here, recent tools can help to tackle the data processing problem, especially when made available open source, by providing a common ground that everyone can use. As a side effect, there is the chance to enhance the reproducibility of the conducted research, since researchers can directly exchange how they processed their data based on the respective specification or script files. A short overview of the variety of existing approaches is given in the related work (Section 3.5.1). There is an increasing number and complexity of signal processing and classification algorithms that enable more sophisticated processing of the data. However, this is also considered as a problem, since it also demands (i) tools where the signal processing algorithms can be directly compared [Sonnenburg et al., 2007, Domingos, 2012] and (ii) to close the still existing large gap between developer and user, i.e., make the tools usable for a larger group of people with no or few experience in programming or data analysis.

**Contribution** With the software pySPACE, we introduce a modular framework that can help scientists to process and analyze time series data in an automated and parallel fashion. The software supports the complete process of data analysis, including processing, storage and evaluation. No individual execution scripts are needed, instead users can control pySPACE via text files in YAML Ain't Markup Language [Ben-Kiki et al., 2008] (YAML) format, specifying what data operation should be executed. The software was particularly designed to process windowed (segmented) time series and feature vector data, typically with classifiers at the end of the processing chain. For such supervised algorithms the data can be separated into training and testing data. pySPACE is, however, not limited to this application case: data can be preprocessed without classification, reorganized (e.g., shuffled, merged), or manipulated using own operations. The framework offers automatic parallelization of independent (not communicating) processes by means of different execution backends, from serial over multicore to distributed cluster systems. Finally, processing can be executed in an offline or in an online fashion. While the normal use case is concerned with recorded data saved to a hard disk (and therefore offline), the online mode, called *pySPACE live*, offers the application-directed possibility to process data directly when it is recorded without storing it to hard disk. We refer to this processing here as *online* due to the direct access in contrast to *offline* processing where the input data is loaded from a hard disk.

To tackle the challenge of an increasing number of signal processing algorithms, additional effort was put into the goal of keeping pySPACE modular and easy-to-extend. Further algorithms can be added by the advanced user; the algorithms will be automatically included in the collection of available algorithms and into the documentation. Furthermore, the software is capable of using existing signal processing libraries, preferably implemented in Python and of using existing wrappers to other languages like C++. So far, interfaces are implemented to external classifiers (from scikit-learn [Pedregosa et al., 2011] and LibSVM [Chang and Lin, 2011]), modular toolkit for data processing [Zito et al., 2008] (MDP), WEKA [Hall et al., 2009], and MMLF (http://mmlf.sourceforge.net/). Core functionality of pySPACE uses the Python libraries NumPy [Dubois, 1999] and SciPy [Jones et al., 2001].

pySPACE was implemented as a comprehensive tool that covers all aspects a user needs to perform the intended operations. The software has a central configuration where the user can optionally specify global input and output parameters and make settings for individual paths to external packages as well as setting computational parameters. The processing is then defined in individual specification files (using YAML) and the framework can be executed with the respective operation on several datasets at once. This functionality is not only provided for internal algorithms, but can also be used with external frameworks like WEKA and MMLF. For the basic signal processing algorithms implemented in pySPACE, we adopted the node and flow concept of the MDP software together with basic principles that were introduced together with it. Currently, more than 200 of such signal processing nodes are integrated into pySPACE. These nodes can be combined and result in numerous different processing flows. Different evaluation schemes (e.g., cross-validation) and performance metrics are provided, and different evaluation results can be combined to one output file. This output can be explored using external software or by using a graphical user interface (GUI) provided within pySPACE.

A drawback of most frameworks is that they focus on the preprocessing and a machine learning part is often missing or vice versa. Furthermore, they do not enable a simple configuration and parallel execution of processing chains. To enable an interfacing to existing tools, pySPACE supports a variety of data types. As soon as several datasets have to be processed automatically with a set of different processing algorithms (including classification) and numerous different hyperparameter values, pySPACE is probably the better choice in comparison to the other tools. Additionally, the capability to operate on feature vector data makes pySPACE useful for a lot of other applications, where the feature generation has been done with other tools. To the best of our knowledge, pySPACE is unique in its way of processing data with special support of neurophysiological data and with its number of available algorithms.

**Outline**  The structural concepts of pySPACE will be presented in Section 3.1. In Section 3.2 we will describe how the software is interfaced including the requirements for running it. This is followed by a short description of optimization aspects in pySPACE (Section 3.3). Several examples and application cases will be highlighted in Section 3.4 including a more complex analysis, using the power of pySPACE and the content of the previous chapters (Section 3.4.3). Finally, we discuss the related work, the connection between this thesis and the pySPACE framework, and summarize with a more personal view.

## 3.1 Structure and Principles

The software package structure of pySPACE was designed in order to be self-explanatory for the user and to correspond to the inherent problem structure. Core components in the main directory are `run` containing everything that can be executed, `resources` where external and internal data formats and types are defined, `missions` with existing processing algorithms the user can specify, and `environments` containing infrastructure components for execution. How to run the software is described in Sections 3.2 and 3.4. The other packages and their connections are described in the following.

### 3.1.1 Data

When analyzing data, the first difficulty is getting it into a framework or into a format, one can continue working with. So as a good starting point, one can look at the way the data are organized and handled within the software, including ways to load data into the framework and how the outcome is stored. Data are distinguished in pySPACE by granularity: from single data samples to datasets and complete summaries (defined in the `resources` package), as explained in the following. They require at the same time different types of processing which are subsequently described in Sections 3.1.2 and 3.1.3 and depicted in Figure 3.1.

Four types of *data samples* can occur in pySPACE: the raw data stream, the windowed time series, feature vector, and prediction vector. A data sample comes with some metadata for additional description, e.g., specifying sensor names, sampling frequency, feature names, or classifier information. When loading a *raw data stream* it is first of all segmented into a *windowed time series*. Windowed time series have the form of two-dimensional arrays with amplitudes sorted according to sensors on the one axis and time points on the other. *Feature vectors* are one-dimensional arrays of feature values. In a *prediction vector* the data sample is reduced to the classification outcome and the assigned label or regression value.

For analysis, data samples are combined to *datasets*. In pySPACE, a dataset is defined as a recording of one single experimental run, either as streamed data or already preprocessed as a set of the corresponding time series windows, or as a loose collection of feature vectors. It also has metadata specifying the type, the storage format, and information about the original data and preceding processing steps. For each type of dataset, various loading and saving procedures are defined. Currently supported data formats for loading streaming datasets are the comma separated values (.csv), the European Data Format (.edf), and two formats specifically used for EEG data which are the one from Brain Products GmbH (Gilching, Germany) (.eeg) and the EEGLAB [Delorme and Makeig, 2004] format (.set). With the help of the EEGLAB format several other EEG data formats can be converted to be used in pySPACE. For cutting out the windows from the data stream, either certain markers can be used or stream snippets with equal distance are created automatically. For supervised learning, cutting rules can be specified to label these windows. Feature vector datasets can be loaded and stored in .csv files or the "attribute-relation file format" (ARFF), which is, e.g., useful for the interface to WEKA [Hall et al., 2009].

Groups of datasets, e.g., experimental repetitions with the same subject or different subjects, can be combined to be analyzed and compared jointly. Such dataset collections are called *summary* in pySPACE. Summaries are organized in folder structures. To enable simple evaluations, all single performance results in a summary are combined to one .csv file, which contains various metrics, observed parameters and classifier information.

### 3.1.2 Algorithms

Nodes and operations are the low and high-level algorithms in pySPACE (see Figure 3.1). They are organized in the missions package. New implementations have to be placed in the missions package and can then be used like the already implemented ones. Here, the type and granularity of input (as depicted in Figure 3.1) have to be considered, the algorithms need to inherit from the base class, and implement some basic processing function(s).

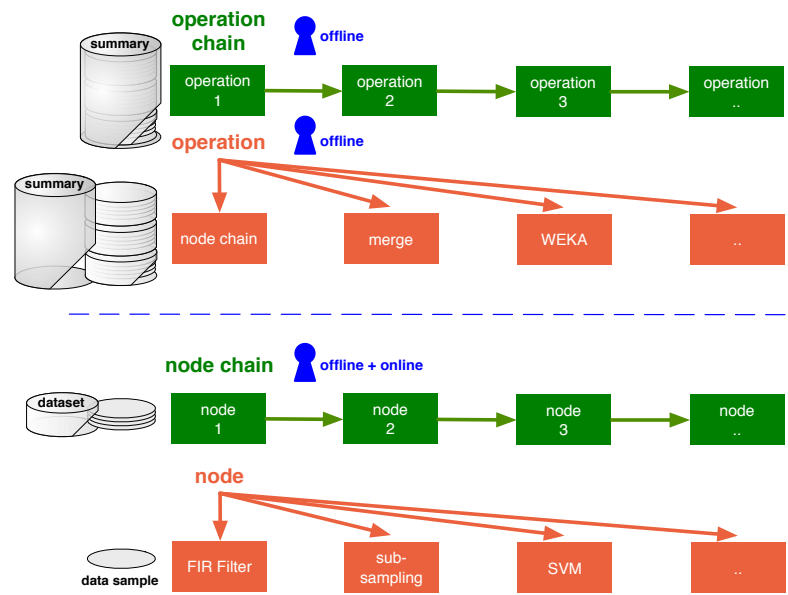

Figure 3.1: **High-level and low-level processing types (upper and lower part) and their connection to the data granularity (summary, dataset, sample).** Access levels for the user are depicted in blue and can be specified with YAML files (Section 3.2.2). Only low-level processing can be performed online. For offline analysis, it is accessed by the node chain operation. For the operations and nodes several different algorithms can be chosen. Algorithms are depicted in orange (Section 3.1.2) and respective infrastructure components concatenating these in green (Section 3.1.3). Visualization taken from [Krell et al., 2013b].

#### 3.1.2.1 Nodes

The signal processing algorithms in pySPACE which operate on data samples (e.g., single feature vectors) are called nodes. Some nodes are trainable, i.e., they define their output based on the training data provided. The concept of nodes was inspired by MDP as well as the concept of their concatenation, which is presented in Section 3.1.3.1. In contrast to frameworks like MDP and scikit-learn, the processing in the nodes is purely sample based[1] to ease implementation and online application of the algorithms. Nodes are grouped depending on their functionality as depicted in Figure 3.2. Currently, there are more than 100 nodes available in pySPACE plus some wrappers for other libraries (MDP, LibSVM, scikit-learn). A new node inherits from the base node and at least defines an execute function which maps the input (time series, feature vector, or prediction vector) to a new object of one of these types. Furthermore, it has a unique name ending with "Node" and its code is placed into the respective nodes folder. Templates are given to support the implementation of new nodes. For a complete processing of data from time series windows over feature vectors to the final predictions and their evaluation, several processing steps are needed as outlined in the following and in Figure 3.2.

---

[1] There is no special handling of batches of data.

*Preprocessing* comprises denoising time series data and reducing dimensionality in the temporal and frequency domain. By contrast, the *spatial filters* operate in the spatial domain to reduce noise. This can be done by combining the signals of different sensors to new virtual sensors or by applying sensor selection mechanisms. *Classification* algorithms typically operate on feature vector data, i.e, before classification the time series have to be transformed with at least one *feature generator* to a feature vector. A classifier is then transforming feature vectors to predictions. In *postprocessing*, feature vectors can be normalized and score mappings can be applied to prediction scores. For every data type a *visualization* is possible. Furthermore, there are *meta* nodes, which internally call other nodes or node chains. Thus, they can combine results of nodes or optimize node parameters. If training and testing data are not predefined, the data must be *split* to enable supervised learning. By default, data are processed as testing data.

*Source* nodes are necessary to request data samples from the datasets, *sink* nodes are required for gathering data together to get new datasets or to evaluate classification performance. They establish the connection from datasets to data samples which is required for processing datasets with concatenations of nodes.

#### 3.1.2.2 Operations

An operation automatically processes one data summary[2] and creates a new one. It is also responsible for the mapping between summaries and datasets. Several operations exist for reorganizing data (e.g., shuffling or merging), interfacing to WEKA and MMLF, visualizing results, or to access external code. The most important operation is, however, the node chain operation that enables automatic parallel processing of the modular node chain (see Section 3.1.3.1). An operation has to implement two main functions. The first function creates independent processes for specified parameter ranges and combinations, as well as different datasets. This functionality is the basis for the parallelization property of pySPACE (see Section 3.1.3.3). The process itself defines the mapping of one or more datasets from the input summary to a dataset of the output summary and its call function is the important part. The second function of an operation is called "consolidate" and implements the clean up part after all its processes finished. This is especially useful to store some meta information and to check and compress the results. Operations and their concatenations are used for offline analysis (see Section 3.2.3). In Section 3.4.1 an example of an operation will be given and explained.

---

[2] Note that a summary can also consist of just a single dataset.

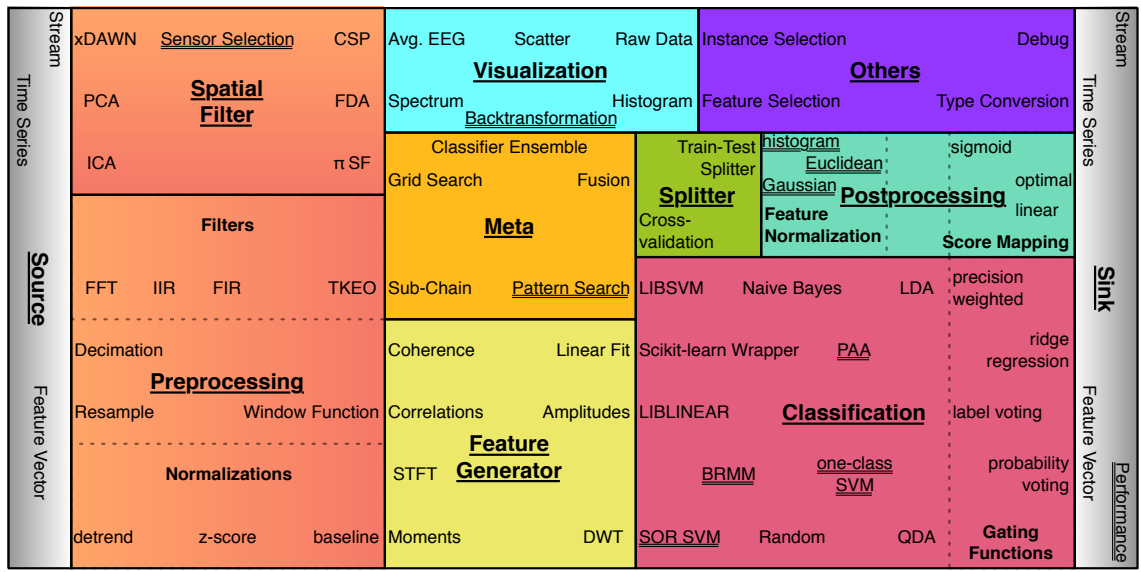

Figure 3.2: **Overview of the more than** 100 **processing nodes in pySPACE**. The given examples are arranged according to processing categories (package names) and subcategories. The size of the boxes indicates the respective number of currently available algorithms. My contributions in the context of this thesis (in terms of implemented nodes) are highlighted with double rule.

### 3.1.3 Infrastructure

So far we have discussed what to process (data) and which algorithms to use (nodes, operations). The infrastructure of pySPACE now defines how the processing is done. This core part is mainly defined in the environment package and usually not modified. It comprises the online execution (see Section 3.2.4), the concatenation of nodes and operations (as depicted in Figure 3.1), and the parallel execution of processing tasks.

### 3.1.3.1   Node Chains

Nodes can be concatenated to a node chain to get a desired signal processing flow. The only restriction here is what a particular node needs as input format (raw stream data, time series, feature vector, or prediction vector). The input of a node chain is a dataset (possibly in an online fashion), which is accessed by a source node at the beginning of the node chain. For offline analysis, a sink node is at the end of the node chain to gather the result and return a dataset as output. In the online analysis, incoming data samples are processed immediately and the result is forwarded to the application. Between the nodes, the processed data samples are directly forwarded, and if needed cached for speed-up. Additional information can be transferred between nodes where this is necessary. To automatically execute a node chain on several datasets or to compare different node chains, a higher level processing is used: the node chain operation as depicted in Figure 3.3.

### 3.1.3.2   Operation Chains

Similar to concatenating nodes to node chains, operations can be concatenated to operation chains. Then, the first operation takes the general input summary and the others take the result summary of the preceding operation as input. At the end, the operation chain produces a series of consecutive summaries. Additionally to combining different operations, a benefit of the operation chain in combination with node chain operations is that a long node chain can be split into smaller parts and intermediate results can be saved and reused. In an operation chain, operations are performed sequentially so that parallelization is only possible within each operation.

### 3.1.3.3   Parallelization

An offline analysis of data processing often requires a comparison of multiple different processing schemes on various datasets. This can and should be done in parallel to get a reduction of processing time by using all available central processing units (CPUs). Otherwise, exhaustive evaluations might not be possible as they require too much time. *Operations* in pySPACE provide the possibility to create independent processes, which can be launched in a so-called "embarrassingly parallel" mode. This can be used for investigations where various different algorithms and hyperparameters are compared (e.g., spatial filters, filter frequencies, feature generators). As another application example, data from different experimental sessions or different subjects might be processed in parallel. The degree of process distribution is determined in pySPACE by usage of the appropriate *back-end* for multicore and cluster systems. Figure 3.3 schematically shows how a data summary of two datasets is processed automatically with different node chains in parallel.

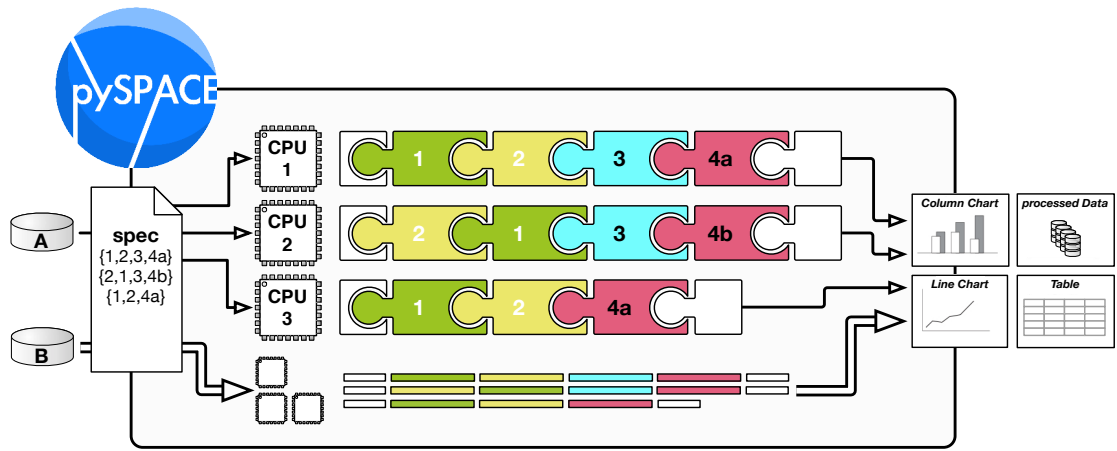

Figure 3.3: **Processing scheme of a node chain operation in pySPACE.** A and B are two different *datasets* (Section 3.1.1), which shall be processed as specified in a simple *spec* file (Section 3.2.2). The processing is then performed automatically. As a *result*, it can produce new data but also visualizations and performance charts. To speed up processing the different processing tasks can be *distributed* over several CPUs (Section 3.1.3.3). The puzzle symbols illustrate different *modular nodes* (Section 3.1.2.1), e.g., a cross-validation splitter (1), a feature generator (2), a visualization node (3), and two different classifiers (4a, 4b). They are concatenated to a *node chain* (Section 3.1.3.1). Visualization taken from [Krell et al., 2013b].

Additionally, some nodes of the meta package can distribute their internal evaluations by requesting own subprocesses from the back-end.[3] This results in a two-level parallelization.

For further speed-up, process creation and process execution are parallelized. For the online application, different processing chains are executed in parallel if the same data is used for different signal processing chains, e.g., to predict upcoming movements and to detect warning perception (P300) from the EEG.

## 3.2   User and Developer Interfaces

pySPACE was designed as a complete software environment[4] without requiring individual hand-written scripts for interaction. Users and developers have clearly defined access points to pySPACE that are briefly described in this section. Most of these are files in the YAML format. Still, major parts of pySPACE can also be used as a library,[5] e.g., the included signal processing algorithms.

---

[3] This feature has been mainly developed by Anett Seeland.
[4] in contrast to libraries
[5] This requires adding the pySPACE folder to the PYTHONPATH variable.

### 3.2.1   System and Storage Interface

The main configuration of pySPACE on the system is done with a small setup script
that creates a folder, by default called *pySPACEcenter*, containing everything in one
place the user needs to get started. This includes the global configuration file, links
to main scripts to start pySPACE (see Sections 3.2.3 and  3.2.4), a sub-folder for files
containing the mission specification files (see Section 3.2.2), and the data storage
(input and output). Examples can be found in the respective folders. The global con-
figuration file is also written in YAML and has default settings that can be changed
or extended by the user.

### 3.2.2   Processing Interface

No matter if node chains, operations, or operation chains are defined (Figure 3.1),
the specifications for processing in pySPACE are written in YAML. Examples are
the node chain illustrated in Figure 3.4 or the operation illustrated in Figure 3.8.
In addition to this file, the user has to make sure that the data are described with a
short metadata file where information like data type and storage format are specified.
If the data have been processed with pySPACE before, this metadata file is already
present.

   The types of (most) parameters in the YAML files are detected automatically
and do not require specific syntax rules as can be inferred from the illustrated
node chain (Figure 3.4), i.e., entries do not have to be tagged as being of type in-
teger, floating point, or string.  On the highest level, parameters can consist of
lists (introduced with minus on separate lines like the node list) and dictionar-
ies (denoted by "key: value" pairs on separate lines, or in the Python syntax, like
`{key1: value1, key2: value2}`).  During processing, these values are directly
passed to the initialization of the respective object.

   Figure 3.4 shows an example of a node chain specification that can be used to
process EEG data.  It illustrates the concatenation of different node categories (in-
troduced in Section 3.1.2.1).[6]  Data samples for this node chain could, e.g., consist
of multiple EEG channels and multiple time points, so that after loading one would
obtain windowed time series. Each data sample is then processed as specified: each
channel is standardized, reduced in sampling rate, and lowpass filtered. Then, the
data are equally split into training and testing data to train the supervised learning
algorithms, which are, in this example, the spatial filter xDAWN [Rivet et al., 2009],
the feature normalization and the classifier later on (here, the LibSVM Support Vec-
tor Machine as implemented by [Chang and Lin, 2011]). Included in this node chain
is a hyperparameter optimization (grid search) of the regularization parameter of

---

   [6] For simplicity, most default parameters were not displayed.

```
# supply node chain with data
− node : TimeSeriesSource
# three preprocessing algorithms
− # standardize each sensor : mean 0, variance 1
  node : Standardization
− # reduce sampling frequency to 25 Hz
  node : Decimation
  parameters :
    target_frequency : 25
− # filtering with fast Fourier transform
  node : FFTBandPassFilter
  parameters :
    pass_band : [0.0, 4.0]
# split data to have 50% training data
− node : TrainTestSplitter
  parameters :
    train_ratio : 0.5
    random : True
# linear combination of sensors to get
# reduced number of (pseudo) channels (here 8)
− node : xDAWN
  parameters :
    retained_channels : 8
# take all single amplitudes as features
− node : TimeDomainFeatures
# mean 0 and variance 1 for each feature
# (determined on training data)
− node : GaussianFeatureNormalization
# meta node, calling classifier for
# optimizing one parameter (complexity ``C``)
− node : GridSearch
  parameters :
    optimization : # define the grid
      ranges : {``C``: [0.1,0.01,0.001,0.0001]}
    evaluation : # which metric to optimize
      metric : Balanced_accuracy
    validation_set : # how to split training data
      splits : 5   # 5−fold cross−validation
    nodes :
      # classifier wrapper around external SVM
      − node : LibSVMClassifier
        parameters :
          complexity : ``C``
          kernel : LINEAR
# Optimize the decision boundary for BA
− node : ThresholdOptimization
# calculate various performance metrics
− node : PerformanceSink
```

Figure 3.4: **Node chain example file.** Comments are denoted by a "#". For further explanation see Section 3.2.2.

the classifier. This is done with five-fold cross-validation on the training data. Finally, performance metrics are calculated respectively for training and testing data. In a real application, the example in Figure 3.4 can be used to classify P300 data as described in Section 0.4.

### 3.2.3  Offline Analysis

Stored data can be analyzed in pySPACE using the `launch.py` script. This script is used for operations and operation chains. The user only needs the respective specification file in YAML. The file name is a mandatory parameter of `launch.py`. For having non-serial execution but a distribution of processing, the parallelization mode parameter (e.g., "mcore" for multicore) is required. The operation specified in a file called `my_operation.yaml` can be executed from the command line, e.g., as

```
./launch.py -o my_operation.yaml --mcore
```
.

GUIs exist for th construction of node chains and especially for the exploration of the results. With the latter (example given in Figure 3.9), different metrics can be displayed, parameters compared, and the observation can be reduced to sub-parts of the complete results output, e.g., explore only results of one classifier type, though several different were processed. In Section 3.4.1 an example of an offline analysis is given and explained.

### 3.2.4  Online Analysis

For processing data from a recording device in an application, it is required to define a specific node chain, train it (if necessary), and then use it directly on incoming data. This is possible using the *pySPACE live* mode.[7] It allows to define a certain application setup (such as involved components, communication parameters, acquisition hardware, number and type of node chains) by using additional parameter files that reference other pySPACE specification files (like in the offline analysis).

Several node chains can be used concurrently to enable simultaneous and parallel processing of different chains. For this, data are distributed to all node chains and the results are collected and stored or sent to the configured recipient (e.g., a remote computer). The data can be acquired from a custom IP-based network protocol or directly from a local file for testing purposes and simulation. Data from supported acquisition-hardware[8] can be converted to the custom network protocol using a dedicated software tool, that comes bundled with pySPACE.

---

[7] I did *not* contribute to pySPACE live except some debugging, tuning, and enabling the incremental learning.

[8] e.g., the BrainAmp USB Adapter by Brain Products GmbH (Gilching, Germany)

### 3.2.5 Extensibility, Documentation and Testing

Integration of new nodes, operations, and dataset definitions is straightforward due to the modular structure of pySPACE. Once written and included in the software structure, they automatically appear in the documentation and can be used with the general YAML specification described above.[9] All operations and nodes come with a parameter description and a usage example. If necessary, single nodes can be defined externally of pySPACE and they will still be included likewise, if they are specified via the global configuration file (Section 3.2.1).

If pySPACE shall be used for more complex evaluation schemes, pure YAML syntax would not be sufficient anymore for our domain-specific language (DSL) (e.g., 5000 testing values from $10^{-5}$ to $10^0$ with logarithmic scaling). Consequently, we allow for Python code injections via strings in a YAML file which are later on replaced by the real values. The injections can be used for defining parameter ranges and when modifying the parameters in the node definitions. Some examples are given in Figure C.2, indicated by the "eval(...)" string. This combines the simplicity of the YAML format with the power of Python to describe more complex evaluations in a readable and compressed format.

These configuration files are experiment descriptions which can be directly exchanged between the users to discuss problems, to standardize processing schemes, or just to easily communicate reproducible approaches. The comparison between our experiment descriptions is much easier than comparing scripts because of

- better structure of the description,
- standardized keywords, and
- less required information/commands and consequently very good compressed representation.

The documentation of pySPACE is designed for both users and developers. We followed a top down approach with smooth transition from high-level to low-level documentation and final linking to the source code for the developers. The documentation is automatically compiled with the documentation generator Sphinx.[10] We largely customized the generator of the documentation structure which creates overviews of existing packages, modules and classes (API documentation). Some properties of nodes are automatically determined and integrated into their documentation like input data types and possible names for usage in the YAML specification. Furthermore, a list of all available nodes, and lists of usage examples for operations and operation chain are generated automatically by parsing the software structure. In contrast to other famous projects using Sphinx like scikit-learn, or Python, we also programmed the main page with the restructured text format, which is the basis of Sphinx documentation, and did not use specific commands for webpage design (html).

---

[9] Class names and YAML strings are automatically matched.
[10] http://sphinx-doc.org/

Additionally, test scripts and unit tests are available in the `test` component of pySPACE. During the software development process, the infrastructure mostly remains untouched but very often new nodes are implemented or existing nodes are extended. To improve test coverage, we developed a *generic test concept* for the nodes:

1. The node documentation is checked for an example to create a node.
2. A node is created using this example.
3. Predefined data is used to be processed by the node, including the training procedure.

Furthermore, an interface is provided to use this generic testing concept to easily generate self-defined tests by providing the respective input and output data. This concept will be used in future to define a test suite, which just checks for changes in the processing to support a change management.

The documentation is generated and unit tests are automatically executed on an everyday basis. For bug fixing, bug reports are possible via email to the pySPACE developer list or via issue reports on `https://github.com/pyspace/pyspace`.

### 3.2.6   Availability and Requirements

pySPACE can be downloaded from `https://github.com/pyspace` and is distributed under GNU General Public License. The documentation can be found there, too. The software can be currently used on Linux, OS X, and Windows. For parallelization, off-the-shelf multi-core PCs as well as cluster architectures using message passing interface (MPI) or the IBM LoadLeveler system can be interfaced. The software requires Python2.6 or 2.7, NumPy, SciPy, and YAML. Further optional dependencies exist, e.g., Matplotlib [Hunter, 2007] is required for plotting. Computational efficiency is achieved by using C/C++-Code libraries where necessary, e.g., NumPy is working with C-arrays and implementations and SVM classification can be performed using the Python wrapper of the LIBSVM C++ package.

## 3.3   Optimization Problems and Solution Strategies

The optimization of data processing chains is very complex. Hence, some separation into subproblems and respective solution strategies is required. In this section, we will highlight some subproblems and solution approaches. pySPACE can be seen as an interface to implement existing approaches and also to explore new approaches.

**Performance Evaluation and the Class Imbalance Problem**  This large paragraph introduces several metrics integrated into pySPACE and analyzes their sensitivity to the class ratio.

> It includes text parts and figures from Dr. Sirko Straube and is based on:
> Straube, S. and Krell, M. M. (2014). How to evaluate an agent's behaviour to infrequent events? – Reliable performance estimation insensitive to class distribution. *Frontiers in Computational Neuroscience*, 8(43):1–6, doi:10.3389/fncom.2014.00043.
> I contributed a few text parts to this paper. My contributions were the (re-)discovery of the class imbalance problem in the machine learning context, the evaluation in this paper which pictures the class imbalance problem, and discussions about the paper and about performance metrics in general.

For optimizing the processing chain, a performance measure is required to quantify which algorithm is better than another. The basis of defining performance metrics is the confusion matrix, which is introduced in Figure 3.5. The figure also shortly summarizes the most important metrics.[11] They are separated into two groups, because when keeping the decision algorithm but changing the ratio of positive and negative samples, some metrics are sensitive to this change and some are not.[12] Another issue, when looking at these metrics is the lack of common naming conventions.

The true positive rate (TPR) is also called *Sensitivity* or *Recall*. The true negative rate (TNR) is equal to the *Specificity*. When the two classes are balanced, the accuracy (ACC) and the balanced accuracy (BA) are equal. The weighted accuracy (WA) is a more general version introducing a class weight $w$ (for BA: *w=0.5*). The BA is sometimes also referred to as the *balanced classification rate* [Lannoy et al., 2011], *classwise balanced binary classification accuracy* [Hohne and Tangermann, 2012], or as a simplified version of the AUC [Sokolova et al., 2006, Sokolova and Lapalme, 2009]. Another simplification of the AUC is to assume standard normal distributions so that each value of the AUC corresponds to a particular shape of the receiver operating characteristic [Green and Swets, 1988, Macmillan and Creelman, 2005] (ROC) curve. This simplification is denoted $AUC_z$ and it is the shape of the AUC that is assumed when using the performance measure $d'$. This measure is the distance between the means of signal and noise distributions in standard deviation units given by the z-score. The two are related by $AUC_z = \Theta(d'/\sqrt{2})$ where $\Theta$ is the normal distribution function. A formula for calculating the general AUC is given by

---

[11] All mentioned performance metrics and many more are integrated into pySPACE and calculated for every (binary) classifier evaluation.

[12] Explained later in this section in more detail and depicted in Figure 3.6.

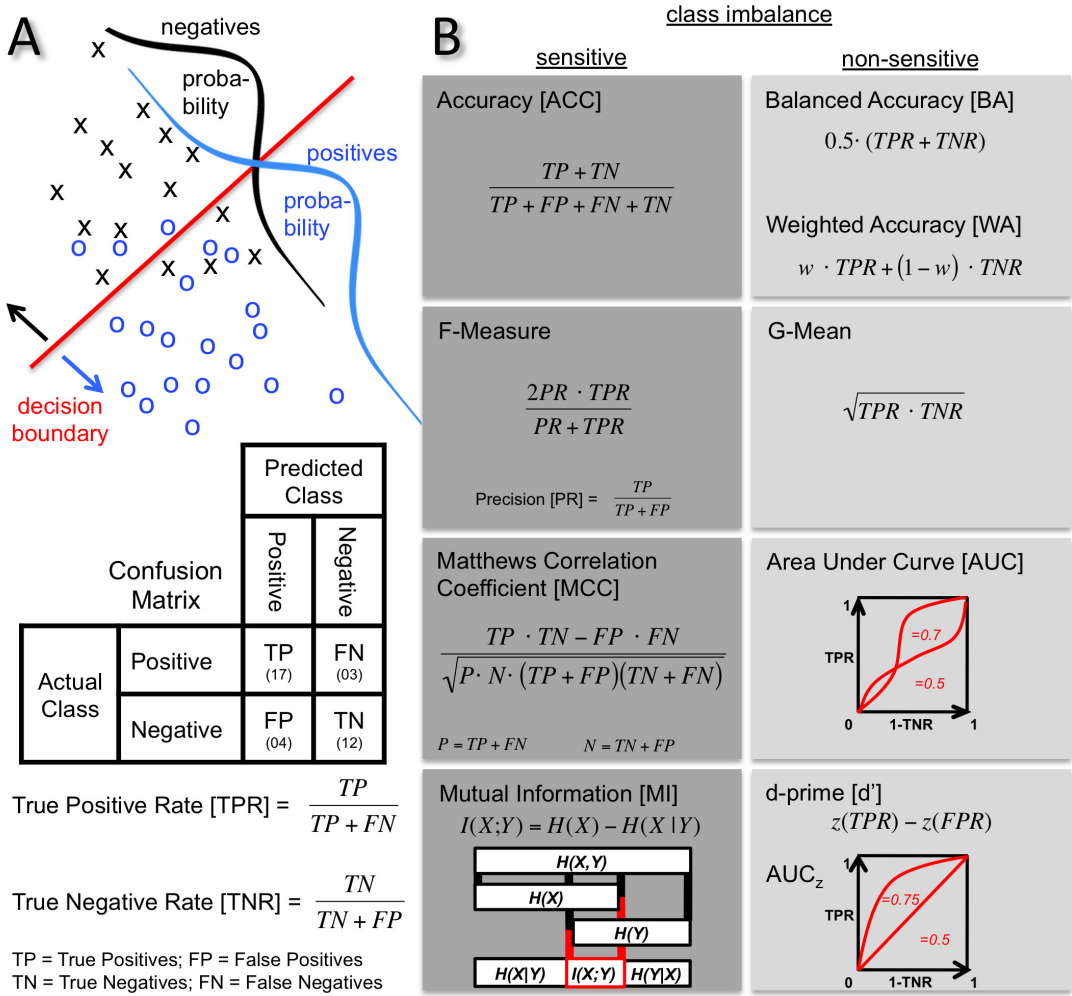

Figure 3.5: **Confusion matrix and metrics. (A)** The performance of an agent discriminating between two classes (positives and negatives) is described by a confusion matrix. Top: The probabilities of the two classes are overlapping in the discrimination space as illustrated by class distributions. The agent deals with this using a decision boundary to make a prediction. Middle: The resulting confusion matrix shows how the prediction by the agent (columns) is related to the actual class (rows). Bottom: The true positive rate (TPR) and the true negative rate (TNR) quantify the proportion of correctly predicted elements of the respective class. **(B)** Metrics based on the confusion matrix grouped into sensitive and non-sensitive metrics for class imbalance when both classes are considered. Visualization and shortened description taken from [Straube and Krell, 2014].

[Keerthi et al., 2007]:

$$\frac{1}{\sum_{y_i=1} 1 \cdot \sum_{y_j=-1} 1} \sum_{y_i=1} \sum_{y_j=-1} \frac{1 - \mathrm{sgn}(f(x_j) - f(x_1))}{2} \qquad (3.1)$$

with the testing data samples $x_j$ with label $y_j$.

Matthews correlation coefficient (MCC) is also known as phi-coefficient, or Pearson correlation coefficient from statistics and can be also straightforwardly used for regression problems in contrast to the other metrics. The F-measure [Powers, 2011] is also referred to as the F-score. A version as weighted harmonic mean is named F$\beta$-score, where $\beta$ denotes the weighting factor. It has been for example used in the aforementioned optimization approaches by Keerthi et al. and Eitrich et al.[13] Despite its strong sensitivity to the ratio of positive and negative samples and its lack of interpretability it is still very often used, especially in text classification where largely unbalanced settings occur [Lipton et al., 2014].

The problem of metrics being sensitive to class imbalance is quite old [Kubat et al., 1998] but still seems to be no common knowledge. In [Straube and Krell, 2014], the authors argument that class imbalance is very common for realistic experiments. Prominent examples can be also found at the evaluation of unary classification (see Section 1.4). For multi-class evaluations, it gets even worse [Lipton et al., 2014]. Lipton et al. report, that it is crucial to optimize the decision criterion (threshold) when using the F-measure and using the default threshold (zero) for the SVM is usually not a good choice. This effect is also partially related to class imbalance. Generally, it is always good choose a threshold which optimizes the performance measure of interest [Metzen and Kirchner, 2011]. This threshold optimization algorithm is integrated into pySPACE.

Using metrics which are sensitive to the class ration makes them incomparable, when class ratios change. Furthermore, their absolute values are fairly meaningless if the class ratio is not reported. Figure 3.6 visualizes the effect of changing the class ratio in an evaluation and how it effects the sensitive metrics. It also visualizes a related effect of comparability between true classifiers and different "guessing" classifiers. The graphic clearly shows, that at least from the perspective of class imbalance, no metric should be used which is sensitive to the class ratio, because the values change very much. Even the normalization of the mutual information (MI) does not make the metric insensitive. For more details on metrics and the imbalance problem refer to [Straube and Krell, 2014].

**Classifier Problem**  Usually, classifiers are defined as an optimization problem and not as a concrete algorithm. Except for the sparse approaches, solving SVM variants is not so difficult because the models of interest are defined as convex optimization problems which can be simplified by duality theory (see Section 1). The only difficulty is that the algorithms need to be able to tackle large amounts of data, e.g., by online learning, iteration over samples, or reduction of the training data (see also Section 1.2). Different implementations of classifiers are available in pySPACE (not limited to SVM variants).

---

[13] It was also primarily used by the developers of pySPACE but later on replaced by the BA due to sensitivity to class imbalance.

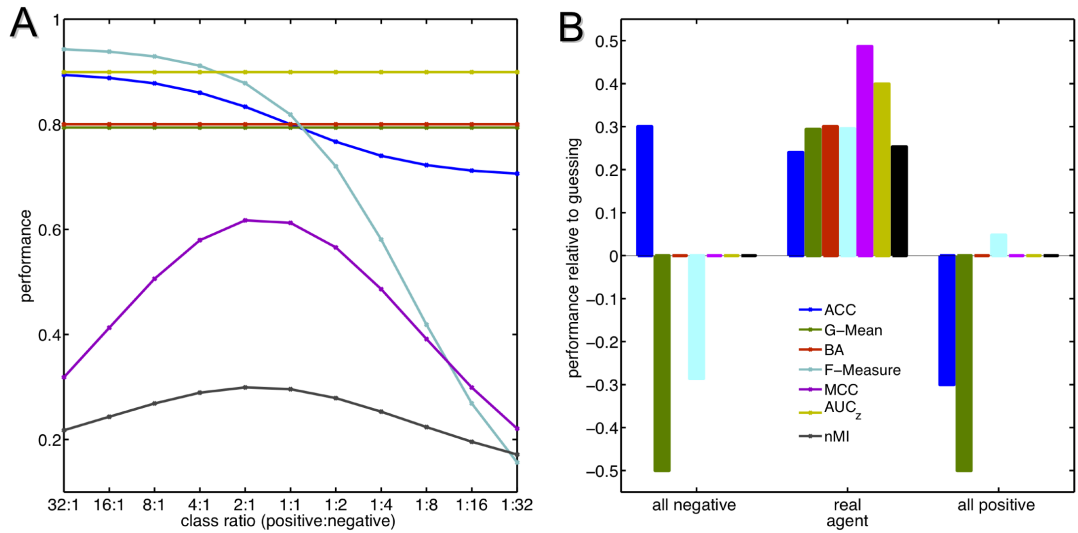

Figure 3.6: **Performance, class ratios and guessing.** Examples of metric sensitivities to class ratios (A) and agents that guess (B). Effect of the metrics AUC and d′ are represented by AUC$_z$ using the simplification of assumed underlying normal distributions. The value for $d'$ in this scenario is 0.81. Similarly, the BA also represents the effect on the WA. **(A)** The agent responds with the same proportion of correct and incorrect responses, no matter how frequent positive and negative targets are. For the balanced case (ratio 1:1) the obtained confusion matrix is [TP 90; FN 10; TN 70; FP 30]. **(B)** Hypothetical agent that guesses either all instances as positive (right) or as negative (left) in comparison to the true agent used in (A). Class ratio is 1:4, colors are the same as in (A). The performance values are reported as difference to the performance obtained from a classifier guessing each class with probability 0.5, i.e., respective performances for guessing are: [ACC 0.5; G-Mean 0.5; BA 0.5; F-Measure 0.29; MCC 0; AUC$_z$ 0.5; nMI 0]. Visualization and description taken from [Straube and Krell, 2014]. nMI denotes the normalized MI. The respective normalization factor is the inverse of the maximum possible MI due to class imbalance.

**Over- and Underfitting**   Note that the overall goal is to find the perfect processing which finally detects/classifies the signal of interest as well as possible. This especially includes the generalization to unseen data and situations. The main dilemma is that we build a model of our data on the given training data but then the model is expected to perform well on unseen data.

A direct approach to tackle the problem with unseen data is to later integrate this new data into the decision algorithm with online learning in an online application (see also Section 1.2).

If a classifier model does not fit the data well enough, it is said to be *underfitting*. This problem is often already considered in the classifier design, especially by the kernel, loss, and regularization approaches mentioned in Section 1.1.1.2.

The kernel enables complex models that can fit to the data even in cases where a linear model is not appropriate for the data at hand. The loss term is usually motivated by an assumption on the noise in the data which inhibits a perfect matching of the chosen model to the data. Both concepts enable a good fitting of the model to the data. Only if either loss or kernel is not chosen well, the model will be underfitting.

Unfortunately, more often data cannot be provided sufficiently enough and consequently the model fitting might be too exact and does not generalize well on unseen data. This effect is called overfitting.

Here, regularization (in combination with the loss term) is an approach to avoid this effect and to obtain more general models (e.g., because the margin between the two classes is maximized or sparse solutions are enforced).

The prize to pay is that the resulting regularization parameter has to be optimized additionally to potential hyperparameters of the chosen loss function or the kernel. Unfortunately, optimizing hyperparameters can again result in over- or underfitting especially if too many hyperparameter are used and optimized. So the problem of over- and underfitting might be just lifted to a higher level.

**Hyperparameter Optimization**  Hyperparameters of the classifier considered in this thesis are

- the regularization parameter $C$,
- the extension of this parameter with class weighting (i.e., $C(y_j)$),
- the range parameter $R$ of the BRMM or the radius $R$ of the unary PAA, and
- specific kernel parameters (see Table 1.1).

Sometimes, even more hyperparameters are introduced for additional tuning, like sample dependent weightings $C_j$ or feature weightings. Furthermore, hyperparameters of the solution algorithms like the number of iterations and the stopping tolerance could be optimized. The type of loss and regularization could be changed, too, which is not considered in the following. The optimization of these hyperparameters can not yet be considered as sufficiently well "solved".

Even the most basic step of choosing an appropriate evaluation metric is not always straightforward as previously discussed. For evaluating an algorithm, there are several evaluation schemes (like k-fold cross-validation) which are quite well studied. Most common evaluation approaches result in a function which is not even continuous and might have several local optima. Another difficulty is, that function evaluations are very expensive because they require to repeatedly train the classifier and evaluate it on testing data.

A straightforward approach to handle function evaluations and reduce processing time is to use parallelization as done in pySPACE. To really speed up the repeated classifier training with different hyperparameters, warm starts [Steinwart et al., 2009] can be used to initialize the optimization algorithms which construct the classifier.[14] For $(n-1)$-fold cross-validation (leave-one-out error) there are special schemes for additional speed up [Lee et al., 2004, Loosli et al., 2007, Franc et al., 2008]. Another approach for saving processing resources is to use heuristics for the hyperparameter optimization and to focus on finding a "quasi-optimal" solution [Varewyck and Martens, 2011], which is often sufficient. This approach is specifically designed for the C-SVM with RBF kernel. First data is normalized, then the hyperparameter $\gamma$ from the kernel is calculated directly, and finally for the regularization parameter $C$ only at most $3$ values have to be tested. This scheme can be used in pySPACE and is very helpful because finding a good $\gamma$ by hand is difficult. A similar (but more complex) approach can be found in [Keerthi and Lin, 2003], which uses Theorem $5$ for the C-SVM and can be generalized to BRMM and SVR using Theorem $14$. First, the $C$ for the linear case is optimized and then a line search is performed with a fixed ratio between the hyperparameters $\gamma$ and $C$ of the respective classifier with RBF kernel.

The hyperparameter optimization and the C-SVM classifier problem can be also seen as a bilevel optimization problem. Hence, one approach is to tackle both problems at once [Keerthi et al., 2007, Moore et al., 2011]. In [Moore et al., 2011] only SVR was handled with a simple validation function. In [Keerthi et al., 2007] the validation function is smoothed which results in a difference to the targeted validation function. The evaluation is not broad (only $4$ datasets) but it is promising. Unfortunately, the code is not provided and their implementation does not scale well with the number of training samples.[15] Is is surprising that the authors did not continue their work on this algorithm. It would be interesting to investigate this approach more in detail in future (e.g., using a large scale optimizer like WORHP [Büskens and Wassel, 2013]).

For analyzing the aforementioned change of the validation function, we integrated smooth versions of existing metrics into pySPACE for further analysis. Here we realized, that some smoothing techniques will not work because the resulting metrics are too different from the target metric. For the metrics related to [Keerthi et al., 2007], it is important to look at the parameter of the smoothing function or even adapt it during the optimization. Too low values of this parameter result in a large difference to the target metric and too high values might result in numerical problems with too high values of the derivative. We also integrated the smoothing approach from [Eitrich and Lang, 2006, Eitrich, 2007].[16]

---

[14] We implemented and tested a pattern search (Figure 3.7) using a warm start which resulted in a large speed up but also required a lot of memory resources, because several processing chains had to be used in parallel for different choices of the hyperparameter values and randomization of the data splitting.

[15] Maybe this could be handled using parallelization techniques.

**Pattern Search** Eitrich et al. use a smoothed metric to optimize several SVM hyperparameters with a pattern search method (Figure 3.7). An important aspect of their approach is the large speed up due to parallelization of the pattern search, the function evaluation, and the C-SVM solving strategies [Eitrich, 2006]. Due to the use of the pattern search, the method is derivative free and it can be applied to a very large class of optimization problems in contrast to the previous bilevel optimization. Unfortunately, the pattern search comes with additional hyperparameters.

We also integrated the pattern search into pySPACE. We only used a parallelization of the pattern search and the validation cycle, but not for the solution of the C-SVM problem in contrast to [Eitrich, 2006]. Implementing such a concept in Python is not straightforward, because communication and other overhead due to the parallelization has to be kept low and when using the standard parallelization package in Python (multiprocessing) an additional second level of parallelization is not possible anymore.

When exploring performance plots of BRMMs with pySPACE (not reported) several observations can be made as listed in the following

- Rather high values for $C$ and $R$ (e.g., $1$ and $10$, respectively) provide better results and faster convergence of the solution algorithms compared to very low values (e.g., $10^{-5}$ and $1.1$).
- If the evaluation metric is not smoothed, there will always be plateaus (for mathematical reasons) but they are not relevant if the number of testing samples is sufficiently high.
- There is a maximum value of $R$ and $C$ which should be considered. It should be acknowledged that there is always a maximum meaningful value for $R$ and $C$. Choosing higher values will result in the same performance and also in a plateau in the hyperparameter landscape.
- Combining all three observations it is good to start with rather high values. Furthermore at least at the beginning of the pattern search, the hyperparameters should be reduced, when performance is not decreasing instead of requesting a performance improvement. So the algorithm does not get stuck on a plateau.
- It is often more efficient to work with logarithmic steps in the hy-

---

[16] Smoothing the validation function is support by the fact, that the classification function can be sometimes chosen partially smooth, with the regularization parameter as variable (see Theorem 24 in the appendix). Consequently, the composed function is expected to be at least partially smooth.

perparameter   landscape   as   also   suggested   in   [Keerthi and Lin, 2003, Varewyck and Martens, 2011].

These approaches of customizing the pattern search are possible with our implementation.

1. Take sequence of direction sets $D_k$ (e.g., $D_k = \{e_i | i = 1, .., n\} \cup \{-e_i | i = 1, .., n\}$), initial step size $s_0$, initial starting point $x_0$, $f_0 := f(x_0)$ (current minimal value), contraction parameter $c$, step tolerance $t$, and a decreasing sequence $p(s_k)$ to define the minimal improvement (e.g. constantly zero) and iterate over $k$

2. Evaluate the points $x_k + s_0 \cdot d$ for $d \in D_k$

3. If $f(x_k + s_0 \cdot d) < f_k - p(s_k)$:
   - $s_{k+1} = s_k$ or increased
   - $x_{k+1} = x_k + s_0 \cdot d$
   - Continue with Step 2

4. Otherwise: $s_{k+1} = c \cdot s_k$ and $x_{k+1} = x_k$

5. If $s_{k+1} < t$: **STOP**

6. Continue with Step 2

Figure 3.7: **General scheme of the pattern search [Nocedal and Wright, 2006].** There are numerous variants/extensions of this method like restricting the number of iterations or performing the evaluations asynchronously [Gray and Kolda, 2006].

**Grid Search**   Despite the previously mentioned promising approaches for hyperparameter optimization, in most cases the grid search is used (or even no hyperparameter optimization is performed or reported at all). In this case, the algorithms are evaluated on a predefined grid of values for the hyperparameters and the best one is chosen. This approach is also implemented in pySPACE with support of parallelization. It is inefficient for two reasons. First, it does not exploit the knowledge about the topography of the landscape of function values, to derive good regions to expand. And second, if the optimal point is outside of the grid region, the performance result can be much worse than in the other hyperparameter optimization approaches. Nevertheless, it usually provides sufficiently good results as for example shown by the variant in [Varewyck and Martens, 2011]. The large dependency of the performance on the chosen grid makes this famous algorithm difficult to compare to real optimization algorithms, because it is always possible to choose a grid which performs at least equally good or better, or a grid which performs worse.

**Preprocessing Optimization** So far, we only discussed methods for optimizing the hyperparameters of the classifier. What is missing in the literature are approaches to additionally optimize the preprocessing. Even though, the generation of meaningful feature in the preprocessing is expected to have a large impact [Domingos, 2012], it is mostly done by hand and using expert knowledge.

In [Flamary et al., 2012] raw data from a time series was used and the optimization of the filter in the preprocessing was combined with the classifier construction. The target function of C-SVM is extended with a regularization term of the filter (including an additional regularization constant). For optimization, a two-stage algorithm is suggested, which switches between between updates of C-SVM and filter.

The optimization of a multi-column deep neural network [Schmidhuber, 2012] can also be seen as a joint optimization of feature generation and classifier. Here, the different layers of the neural network can be identified with different types of preprocessing or feature generation. In the context of pure feature learning without classification, neural networks are also used [Ranzato et al., 2007].

**Discussion** For optimizing the complete processing chain, pySPACE shows a great advantage to the previously mentioned approaches. Grid search and pattern search can be applied to complete processing chains without much additional effort. It is even possible to have hybrid approaches, where the grid defines different types of algorithms and the pattern search optimizes the respective algorithm hyperparameters. Furthermore, arbitrary node chains, evaluation schemes, and performance metrics which are available in pySPACE can be combined to define the optimization procedure. Even without the optimization algorithms, pySPACE largely supports the comparison of algorithms as shown in the examples in this thesis. In future, we plan to use this interface to implement a complete automatic optimization process which will be called autoSPACE and which will work on a database of datasets.

## 3.4 pySPACE Usage Examples

pySPACE is applicable in various situations, from simple data processing over comprehensive algorithm comparisons to online execution. In this section an example for an offline analysis is given that comprises most of the key features of pySPACE. Thereby it is shown how the intended analysis can be easily realized without the need for programming skills. Published work and related projects are named where pySPACE has been used, most often with such an offline analysis. Finally, a more complex example is given which incorporates content from the previous main chapters.

### 3.4.1   Example: Algorithm Comparison

In the following, an exemplary and yet realistic research question for processing neu­rophysiological data serves to explain how a node chain can be parameterized and thus different algorithms and hyperparameters can be tested. To show that for such a comparison of algorithms and/or algorithm hyperparameters pySPACE can be a perfect choice, the whole procedure from data preparation to final evaluation of the results is described.

#### Data and Research Question

We take the data described in Section 0.4. Our aim, besides the distinction of the two classes *Standard* and *Target*, is to investigate the effect of different spatial filters, i.e., ICA, PCA, xDAWN, and CSP (see also Section 2.2.1.4), on the classification per­formance, or whether one should not use any spatial filter at all (denoted by "Noop"). Spatial filters aim to increase the signal-to-noise ratio by combining the data of the original electrodes to pseudo-channels. Thereby, not only performance can be in­creased, but also information is condensed into few channels, enabling reduction of dimensionality and thereby reducing the processing effort. Thus, a second research question here is to evaluate the influence of the number of pseudo-channels on the classification performance.

#### Data Preparation

In our example, each recording session consists of five datasets. To have a suffi­cient amount of data, they were all are concatenated. This is an available *operation* in pySPACE after the data were transferred from stream (raw EEG format) to the pySPACE time series format. Therefore, after data preparation, all merged record­ings that should be processed are present in the input path (see below), each in a separate sub-directory with its own meta file.

#### Processing Configuration

The algorithm comparison has to be specified in a file as depicted in Figure 3.8. The *type* keyword declares the intended *operation*, i.e., node chains will be executed. The data, which can be found in the directory `P300_data` (*input_path*) will be processed according to the specifications in the file `P300.yaml`. This file is identical to the one presented in Figure 3.4, except that it is parameterized to serve as a *template* for all node chains that should be executed. The parameterization is done by inserting unique words for all variables that need to be analyzed. This means, in this example that the specification of the xDAWN node is replaced by

```
─ node : ␣␣alg␣␣
  parameters :
      retained_channels : ␣␣channels␣␣
```

introducing ␣␣alg␣␣ as parameter for the different spatial filters and ␣␣channels␣␣ for the varying number of pseudo-channels. All values that should be tested for these two parameters are specified in the operation file (Figure 3.8) below the keyword *parameter_ranges*. pySPACE will create all possible node chains of this operation using the Cartesian product of the value sets (grid). The value of the parameter ␣␣alg␣␣ is the corresponding node name, with Noop (meaning *no op*tion) telling pySPACE that in this condition nothing should be done with the data. In the example Noop could serve as a baseline showing what happens when no spatial filter is used.

```
type: node_chain  # operation type
input_path: "P300_data"  # location of data in storage folder
templates: ["P300.yaml"]  # specification of node chain(s)
parameter_ranges:  # Cartesian product of parameters to be tested
  __alg__: ['CSP', 'xDAWN', 'ICA', 'PCA', 'Noop']  # nodes tested
  __channels__: [2, 4, 6, 8, 10, 20, 30, 40, 50, 62]
# number of pseudo-channels
runs: 10  # number of repetitions
```

Figure 3.8: **Operation specification example file for spatial filter comparison.** For more details see discussion in Section 3.4.1.

In this example, varying the number of retained channels will lead to equal results for each value in the case of using Noop. Therefore, an additional constraint could ensure that Noop is only combined with one value of ␣␣channels␣␣ which would reduce computational effort. Furthermore, instead of a grid of parameters, a list of parameter settings could be specified or Python commands could simplify the writing of spec files for users with basic Python knowledge. For example, the command range(2, 63, 2) could be used to define a list of even numbers from 2 to 62 instead of defining the number of retained pseudo-channels individually.

Finally, the *runs* keyword declares the number of repeated executions of each node chain. Repetitions can be used to compensate for random effects in the results due to components in the node chain that use randomness, like the *TrainTestSplitter*. Using different data splitting strategies when processing the same data with different parameterizations (e.g., spatial filters or number of retained pseudo-channels) would make the results incomparable. To avoid such behavior and to *ensure reproducibility* of the results, randomness in pySPACE is realized by using the random package of Python with a fixed seed that is set to the index of the repeated execution. In other words, the same value of *runs* returns the same results for a given dataset and operation. For obtaining different results, this number has to be changed.

**Execution and Evaluation**

The execution of the operation works as described in Section 3.2.3. The result is stored in a folder in the data storage, named by the time-stamp of execution. For replicability, it contains a zipped version of the software stack and the processing specification files. For each single processing result there is a subfolder named after the processed data, the specified parameters and their corresponding values. For evaluation, performance results are not stored separately in these single folders, but the respective metrics are summarized in a .csv tabular. Furthermore, by default the result folders are also compressed and only one is kept as an example.

The result visualization with the evaluation GUI of pySPACE can be seen in Figure 3.9. Here, the varied parameters (compare test parameters in Figure 3.8 with selection in upper left of Figure 3.9) as well as the data can be selected and individually compared with respect to the desired metric.

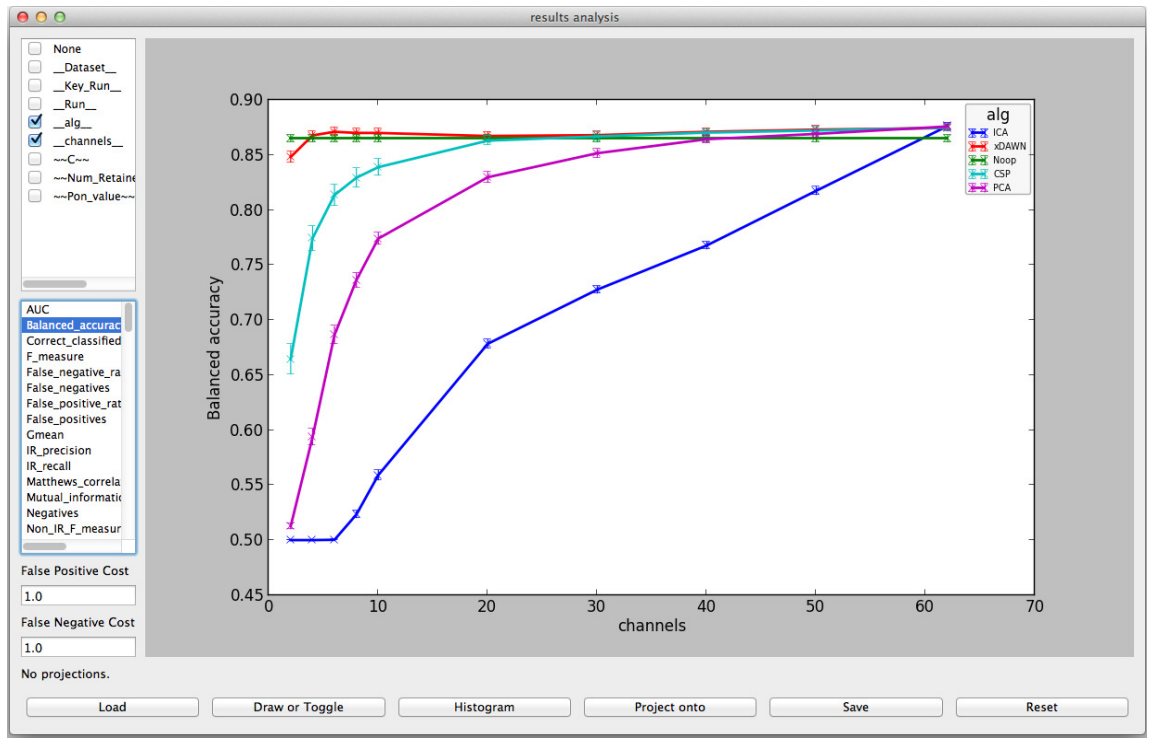

Figure 3.9: **Visualization from the evaluation GUI** for the result of the spatial filter comparison, explained in Section 3.4.1. Visualization taken from [Krell et al., 2013b].

It is not surprising, that the xDAWN is superior to the other algorithms because it was specifically designed for the type of data used in this analysis. The well performance of the CSP is interesting, because it is normally only used for the detection of changes in EEG frequency bands connected to muscle movement. The bad performance of the ICA shows, that the pseudo-channels have no ordering in importance in contrast to the other filters. A correct reduction step would be here to reduce dimensionality internally in the algorithm in its whitening step. This error in interfacing the implementation from the MDP library was fixed due to this result. Normally, the ICA should perform better than the PCA.

### 3.4.2 Usage of the Software and Published Work

This section shortly highlights the use of pySPACE in the community, in different projects, and in several publications.

Since pySPACE became open source software in August 2013, there is not yet a public user community. Usage statistics from the repository are unfortunately not yet available. The software was announced at the machine learning open source software webpage (`http://mloss.org/software/view/490/`) in the context of a presentation at a workshop [Krell et al., 2013a] which resulted in 2753 views and 575 downloads. The publication which first presented the software to the community in a special issue about Python tools for neuroscience [Krell et al., 2013b] resulted in 1654 views, 192 paper downloads, 10 citations, and 118 mentions in public networks. Furthermore, pySPACE has been presented at 3 conferences [Krell et al., 2013a, Krell et al., 2014b, Krell, 2014], where the last presentation resulted in a video tutorial (`http://youtu.be/KobSyPceR6I`, 345 views). In 2015, pySPACE was presented at the CeBIT.

pySPACE has been developed, tested and used since 2008 at the Robotics Innovation Center of the German Research Center for Artificial Intelligence in Bremen and by the Robotics Research Group at the University of Bremen:

- project **VI-Bot** (`http://robotik.dfki-bremen.de/en/research/projects/vi-bot.html`): EEG data analysis for movement prediction and detection of warning perception during robot control with an exoskeleton,

- project **IMMI** (`http://robotik.dfki-bremen.de/en/research/projects/immi.html`): EEG and EMG data analysis for movement prediction, detection of warning perception, and detection of the perception of errors applied in embedded brain reading [Kirchner, 2014],

- direct control of a robot with different types of EEG signals,

- **project Recupera** (`http://robotik.dfki-bremen.de/en/research/projects/recupera.html`): EEG and EMG data analysis for movement detection to support rehabilitation,

- **project ACTIVE** (`http://robotik.dfki-bremen.de/en/research/projects/active.html`): analysis of epileptic seizure EEG data,

- **project TransTerrA** (`http://robotik.dfki-bremen.de/en/research/projects/transterra.html`): transfer of results from the project IMMI,

- **project VirGo4** (`http://robotik.dfki-bremen.de/en/research/projects/virgo4.html`): tuning of regression algorithm for robot sensors [Rauch et al., 2013, Köhler et al., 2014],

- **project City2.e 2.0** (`http://robotik.dfki-bremen.de/de/forschung/projekte/city2e-20.html`): comparison of different methods for parking space occupancy prediction (in future),

- **project LIMES** (`http://robotik.dfki-bremen.de/en/research/projects/limes.html`): parallelization of robot simulations,

- classification of iterative closest point (ICP) matches into good and bad ones,

- soil detection from sensor values of a robot, and

- every evaluation in this thesis and the visualizations of the backtransformation.

The existing publications are mainly results from the projects VI-Bot and its follower project IMMI. They only show a small subset of possible applications of the software, documenting its applicability to EEG and EMG data (e.g., [Kirchner and Tabie, 2013, Kirchner et al., 2014b, Kirchner, 2014]).

In [Kirchner et al., 2010, Wöhrle et al., 2013a, Seeland et al., 2013b, Kirchner et al., 2013, Kim and Kirchner, 2013, Kirchner, 2014, Wöhrle et al., 2014, Seeland et al., 2015] pySPACE was used for evaluations on EEG data in the context of real applications. P300 data as described in Section 0.4 is used to customize complex control environments because warnings do not have to be repeated if they were perceived by the operator. Another application is to predict/detect movements to use it in rehabilitation and/or to adapt an exoskeleton/orthosis due to the predicted/detected movement. In [Kim and Kirchner, 2013], human brain signals are analyzed which are related to perception of errors, like interaction error and observation error. Special formulas for a moving variance filter are used in pySPACE for EMG data preprocessing [Krell et al., 2013c]. In [Metzen et al., 2011a, Ghaderi and Straube, 2013, Ghaderi and Kirchner, 2013, Wöhrle et al., 2015] the framework is used for the evaluation of spatial filters as also done in Section 3.4.1. An example for a large-scale comparison of sensor selection algorithms can be found in [Feess et al., 2013] and Section 3.4.3. Here, the parallelization in pySPACE for a high performance cluster was required, due to high computational load coming from the compared algorithms and the amount of data used for this evaluation. In [Ghaderi et al., 2014], the effect of eye artifact removal from the EEG was analyzed. There are also several publication, looking at the adaptation of EEG processing chains [Metzen and Kirchner, 2011, Metzen et al., 2011b, Wöhrle et al., 2015, Ghaderi and Straube, 2013, Wöhrle and Kirchner, 2014, Tabie et al., 2014]. Some machine learning evaluations on EEG data were performed [Metzen and Kirchner, 2011, Metzen et al., 2011b, Kassahun et al., 2012].

In the context of this thesis, pySPACE was for example used for evaluations of new classifiers on synthetic and benchmarking data [Krell et al., 2014a, Krell and Wöhrle, 2014] and for visualizing data processing chains with the backtransformation [Krell et al., 2014c, Krell and Straube, 2015].

### 3.4.3 Comparison of Sensor Selection Mechanisms

This section is based on:

Feess, D., Krell, M. M., and Metzen, J. H. (2013). Comparison of Sensor Selection Mechanisms for an ERP-Based Brain-Computer Interface. *PloS ONE*, 8(7):e67543, doi:10.1371/journal.pone.0067543.

It was largely reduced to the parts relevant for this thesis and some additional observations and algorithms were added. This includes some text parts that are written by David Feess. David Feess and I equally contributed to this paper. David's focus was the state of the art, the sensor selection with the "performance" ranking, and writing most parts of the paper. The main contribution of Dr. Jan Hendrik Metzen was the probability interpretation of the results (not reported in this section) and the evaluations with the two SSNR approaches. There were several discussion between the authors about the paper and the evaluation. My main contribution was in the implementation and in design of the other ranking algorithms like the ranking with spatial filters or SVMs.

In this section, we will highlight a more complex application/evaluation, which touches all aspects of this thesis. The analysis will be applied to data from a passive BCI application (P300 data, see Section 0.4).

A major barrier for a broad applicability of BCIs based on EEG is the large number of EEG sensors (electrodes) typically used (up to more than 100).[17] The necessity for this results from the fact that the relevant information for the BCI is often spread over the scalp in complex patterns that differ depending on subjects and application scenarios. Since passive BCIs aim at minimizing nuisance of their users, it is important to look at sensor selection algorithms in this context. The less sensors need to be applied, the less preparation time is required and the more mobile the system might become. So the users will probably be less aware of the fact that their EEG is recorded.

---

[17] The cap has to be placed on the user's scalp and for each electrode a conductive gel has to be applied.

Recently, a number of methods have been proposed to determine an individual optimal sensor selection. In [Feess et al., 2013] a selection of approaches has been compared against each other and most importantly against several baselines (for the first time). The following baselines were analyzed:

- Use the complete set of sensors.

- Use two electrode constellations corresponding to commercialized EEG systems: one 32 electrode 10–10 layout as used in the actiCAP EEG system (Figure C.6) and the original 10–20 layout with 19 sensors.

- Use random selections of sensors (100 repetitions).

- Use the normal evaluation scheme on the data and recursively eliminate the sensor which is least decreasing the *performance*.[18]

Note that the latter might be computational expensive but given an evaluation scheme it is the most direct intuitive way, because when reducing sensors, the goal is always not to loose performance or even increase it due to reduced noise from irrelevant sensors.

For a realistic estimation of the reduced system's performance sensor constellations found on one experimental session were transferred to a different session for evaluation. Notable (and unanticipated) differences among the methods were identified and could demonstrate that the best method in this setup is able to reduce the required number of sensors considerably. Even though the final best approach was tailored to the given type of data, the presented algorithms and evaluation schemes can be transferred to any binary classification task on sensor arrays. The results will be also reported in this section.

Even though, the analysis is performed on EEG data, sensor selection algorithms are also relevant in other applications In robotics for example, reducing the number of relevant sensors for a certain classification task can help to save resources (material, time, money, electricity) and it can improve the understanding because with fewer sensors the interpretation (e.g., with the backtransformation from Chapter 2) becomes easier. In contrast to the EEG application, there are two minor differences. Some sensors are normally divided into sub-sensors, but for really removing the sensor, all sub-sensors need to be removed. For example, an inertial measurement unit (IMU) provides "sub-sensors" for movement in $x$, $y$, and $z$ direction or a camera could be divided into its pixel components as sub-sensors. For EEG data sensors could be also grouped, but this is not so relevant. The second difference is in the evaluation. In EEG data processing, the setting of the electrodes (sensors) between different recording sessions is never the same, because electrode conductivity, electrode positions, and the head (e.g., hair length) are always slightly different. For robots, this should normally not be such an important issue, even though the robotic system itself might be subject to wear.[19]

---

[18] The BA was used as performance metric as discussed in Section 3.3.

For a detailed description of the state of the art and methodology we refer to [Feess et al., 2013]. In this section, we will focus on some aspects in context of this thesis.

**Sensor Ranking for Recursive Backwards Elimination**

This section describes the used methods. Motivated by the processing chain, used in the final evaluation, different ranking algorithms are suggested. The ranking of the sensors is then used to recursively eliminate one sensor after the other. After each removal, a new ranking is determined and the sensor with the lowest rank is removed.

The standard processing chain in this experimental paradigm is given in Figure 3.4 with the only exception in the feature generation to be consistent with [Feess et al., 2013]. Features are extracted from the filtered signal by fitting straight lines to short segments of each channel's data that are cut out every $120\,\mathrm{ms}$ and have a duration of $400\,\mathrm{ms}$. The slopes of the fitted lines are then used as features [Straube and Feess, 2013].

Similar when the goal is to decode the decision process, a ranking of sensors can be based on the different stages of a processing chain for the related decision process as shown in the following. The respective algorithm short names for the evaluation are denoted in brackets in the title.

**Spatial Filter Ranking (*xDAWN, CSP, PCA*)**  When a spatial filter has been trained, its filter weights can be used for a ranking, by for example adding up the absolute coefficient of the first four spatial filters (xDAWN, CSP, PCA, see also Sec-

---

[19] One approach to handle these changes is online learning (see Section 1.4).

tion 2.2.1.4):

$$W_h = \sum_{j=1}^{4} |f_{hj}|. \tag{3.2}$$

Here, $W_h$ provides the weight associated with the $h$-th real sensor. The weight with the lowest value $W_h$ is iteratively removed. (The filter is than trained on the reduced set of sensors.)

**Signal to Signal-Plus-Noise Ratio (*SSNR*$_{AS}$, *SSNR*$_{VS}$)**   There are two additional methods connected to the xDAWN [Rivet et al., 2012], which was specifically designed for P300 data. The first calculates the signal to signal-plus noise ratio in the actual sensor space (SSNR$_{AS}$). The second, calculates the same ration in the virtual space after the application of the xDAWN (SSNR$_{VS}$). The value is calculated with every sensor removed and the sensor with the lowest increase (or highest decrease) of the ratio is selected for recursive removal.

**Support Vector Machine-Recursive Feature Elimination (*1SVM, 2SVM, 1SVMO, 2SVMO*)**   If no spatial filter is applied in the processing chain, the weights in the classifier still have a one-to-one correspondence to the original sensors. This can be used for a ranking. Given a classification vector $w$ with components $w_{ij}$ where the $j$ component is related to the $j$-th sensor, we can again define the ranking

$$W_h = \sum_i |w_{ih}|. \tag{3.3}$$

Again recursively the sensor $h$ with the lowest $W_h$ is removed as in cursive feature elimination [Lal et al., 2004]. To simulate the hard margin case, the regularization parameter $C$ was fixed to $100$.[20] Using real hard margin separation would not be feasible, because with small electrode numbers the two classes become inseparable. Additionally to the ranking with the C-SVM (2SVM), the variant with 1–norm regularization (1SVM) was used due to its property to induce sparsity in the feature space.

This view on the classifier was the original motivation to look at the sparsity properties presented in Section 1.3.3.4 and the backtransformation (Chapter 2) and to extend the original analysis from [Feess et al., 2013].

Most importantly, this ranking turned out to be a very good example of the necessity to optimize the hyperparameter $C$ at least roughly. We will show, that $C$ should be optimized and not chosen very high. Therefore, we additionally performed a grid search ($C \in \{10^{-2}, 10^{-1.5}, \ldots, 10^{2}\}$) with 5-fold cross validation optimizing the BA.

---

[20] This was not reported in [Feess et al., 2013] but could be reproduced with the configuration file.

This resulted in additional rankings, denoted with 1SVMO and 2SVMO respectively.

A variant would be to use a sum of squares, to be more close to the 2–norm regularization of the C-SVM [Tam et al., 2011].

Instead of reducing the processing chain such that it is possible to gain sensor weights from the linear classifier, it would be also possible to use the affine backtransformation (see Chapter 2) after the first preprocessing right before the application of the xDAWN filter:

$$W_h = \sum_i \left| w_{ih}^{(1)} \right| . \tag{3.4}$$

Note that this way of ranking sensors could be applied to any affine processing chain.

**Ranking in Regularization (SSVMO)**    A disadvantage of the 1–norm regularized C-SVM is that it is only inducing sparsity in the feature space but not directly in the number of sensors. There are several approaches to induce grouped sparsity [Bach et al., 2012]. An intuitive approach would be to use

$$\|w\|_{1,\infty} = \sum_h \max_i |w_{ih}| \tag{3.5}$$

where the second index $h$ again corresponds to the sensor. The advantage of this regularization is, that the resulting classifier can be still defined as a linear optimization problem and it might be possible to derive a proof of sparsity similar to Theorem 13. Unfortunately, this way of regularization solely focuses on sparsity and not on generalization. In fact, a short analysis showed that often equally high weights referring to one sensor are assigned. To compensate for this, we choose a mixed regularization:

**Method 20** (Sensor-Selecting Support Vector Machine (SSVM))**.**

$$
\begin{aligned}
\min_{w,b,t} \quad & \sum_{i,h} |w_{ih}| + C_s \sum_h \max_i |w_{ih}| + C \sum_k t_k \\
s.t. \quad & y_k \left( \sum_{i,h} w_{ih} x_{ih}^k + b \right) \geq \quad 1 - t_k \quad \forall k \\
& \qquad\qquad\qquad\quad t_k \geq \quad 0 \qquad \forall k .
\end{aligned}
\tag{3.6}
$$

Here, the additional regularization constant $C_s$ weights between sparsity in sensor space and the original 1–norm regularization. The final classifier weights can again be used for ranking. We used the same approach for optimizing $C$ as for the 1SVMO but $C_S$ had to be fixed to 1000 because an optimization was computationally too expensive.

If the first index is related to the time, sparsity in time can be induced, accordingly. If a smaller time interval is needed, the final decision can be accelerated.

**Evaluation Schemes**

For evaluating the performance of a sensor selection method, three datasets are required: one on which the actual sensor selection is performed, one where the system (spatial filter, classifier, etc.) is trained based on the selected sensor constellation, and one where the system's performance is evaluated. From an EEG-application point of view, sensor selection should be performed on data from a prior usage session of the subject and not on data from the current one, on which the system is trained and evaluated (one would not demount sensors that are already in position after a training run). Since the selected sensor constellations are transferred from one usage session to another, this evaluation scheme is denoted as *inter-session* (see also Figure 3.10). The sensor constellations are thus evaluated on data from a different usage session with potentially different positioning of EEG sensors, different electrode impedances, etc. For the selected sensor constellation, the system is trained on data from one run of the session and evaluated on the remaining 4 runs. Thus, the inter-session scheme does not imply that classifiers are transferred between sessions but only that sensor constellations are transferred. If the sensor properties (e.g., impedance, position) between different recordings are not expected to change, this evaluation part should be omitted.

An alternative evaluation scheme, which is used frequently in related work, is the *intra-session* scheme (as depicted in Figure 3.10): in this scheme, the sensor selection is performed on data from the usage session itself; namely on the same run's data on which the system is trained later on. Thus, sensor constellations are not transferred to a different session and the influence of changes in EEG sensor positions and impedances is not captured. While this scheme is not sensible in the context of an actual application, it is nevertheless used often for evaluation of sensor selection methods because data of multiple usage sessions from the same subject may not be available. We perform the intra-session evaluation mainly to investigate to which extent its results generalize to the inter-session evaluation scheme.

To mimic an application case with a training period prior to an actual operation period, the evaluation is performed by applying an "inverse cross-validation like" schema on basis of the runs from one session. In the intra-session scheme, one run is used for sensor selection and training of the classification flow, and the remaining four runs from that session are used as test cases. This is repeated so that each of the 5 runs is used for sensor selection/training once and results for our dataset (consisting of 5 subjects with 2 sessions each) in a total of $5 \cdot 2 \cdot 5 = 200$ performance scores per selection method and sensor set size. In the inter-session scheme we can perform the sensor selection on each of the five runs of the other session of the subject, and thus we obtain $5 \cdot 200 = 1000$ performance scores.

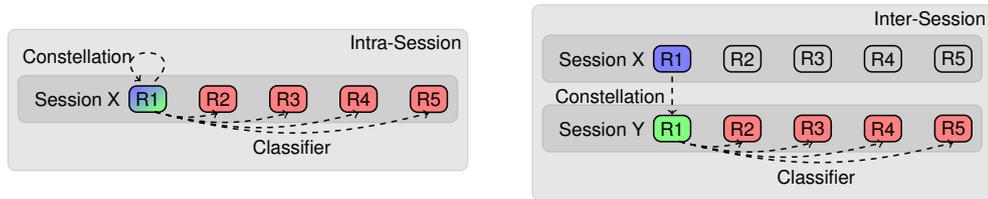

Figure 3.10: **Intra-session and inter-session scheme.** R1–R5 denote the runs from each experimental session. In the *intra-session* scheme (left), the sensor selection (blue) is performed in the same run in which the system is trained (green), and the evaluation (red) is performed on the remaining runs from that session. In the *inter-session* scheme (right), the sensor constellations are transferred to a different session of the same subject. Note that run and session numbering were permuted during the experiment so that in each condition, each run was used for sensor selection and training. Visualization taken from [Feess et al., 2013].

Another possible evaluation scheme, which we will not follow in this thesis is to look at the transfer between subjects or to be more general the transfer between different systems, the sensors are attached to.

### Standard Signal Processing and Classification

During the training phase, the regularization parameter $C$ of the C-SVM is optimized using a grid search ($C \in \{10^0, 10^{-1}, \dots, 10^{-6}\}$) with a 5-fold cross-validation.

The individual sensor selection methods require the training and evaluation of different parts of the signal processing chain: the *SSNR* and *Spatial Filter* sensor selection algorithms can be applied for each run based on the signals after the low-pass filter and require no separate evaluation based on validation data. For the SVM-based methods, the entire signal processing chain has to be trained. In this case, the xDAWN filter is not used during the sensor selection in order to retain a straight-forward mapping from SVM weights to sensor space. Again, no evaluation on validation data is required. The *Performance* method requires to train the entire signal processing chain, too; however, additionally, a validation of the trained system's performance is required. For this, the data from a run is split using an internal 5-fold cross-validation. Each of the methods yields one sensor constellation per run for each session of a subject.

### Processing with pySPACE

For this evaluation, pySPACE is very helpful. Even though, the evaluation was split into several parts, the standardized configuration files ensured consistency between the experiments. Without the parallelization capabilities, the evaluation would probably last too long. Adding ranking capabilities to spatial filters and linear classifiers (even combined with hyperparameter optimization) was straightforward due to the software structure and only the *SSNR* and *Performance* methods needed some extra implementation. This was combined into one electrode selection algorithm, which was able to interface the different methods and to store and load electrode ranking results. For the evaluation the resulting rankings (coming from the recursive backward elimination) only had to be loaded and evaluated depending on the chosen number of electrodes and the chosen evaluation scheme.

**Results**

Figure 3.11 shows the results for the intra-session scheme. At first it can be noticed that all standard caps perform essentially on chance level. The same is true for the $SSNR_{AS}$ and *2SVM* selection heuristics: for more than 5 sensors, both curves lie close to the center of the random selection patches. The PCA filter method performs even worse than random for a large range of constellation sizes. The $SSNR_{VS}$ method, the *xDAWN* filter, the *Performance* ranking, and the *1SVM* ranking deliver a performance considerably better than chance level for 30 or less sensors. The latter three perform nearly identically for the whole range and they are better than chance level. The CSP method performs slightly worse than these methods for less than 20 sensors. The *Performance* ranking performs slightly worse than these methods in the range between 30 and 40 sensors. For $SSNR_{VS}$, the mean performance remains on the baseline level of using all sensors down to around 18 sensors and is remarkably better than any of the other heuristics.

It can be clearly seen, that ranking using the classifiers with optimized complexity (1SVMO, 2SVMO, SSVMO) perform comparable or even slightly better than the not optimized 1SVM ranking. Especially, the 2SVMO ranking shows a large improvement in comparison to the ranking with no optimization (2SVM).

In the inter-session results shown in Figure 3.12, all sensor selection methods drop in absolute performance compared to the intra-session scheme. Random constellations and standard caps are not effected by the type of transfer since they are not adapted to a specific session anyway. The relative order of the curves remains identical to the intra-session results. The performance of the best methods is still above or in the upper range of the random constellations, and $SSNR_{VS}$ still outperforms all random constellations in the relevant range.

**Discussion**

As the sensor selection of the $SSNR_{AS}$ and *2SVM* methods performs essentially equivalent to random selection, apparently these methods are not able to extract any useful information from the data. It can be clearly seen, that the *2SVM* requires a hyperparameter optimization (2SVMO) to be able to generalize well, which also holds a bit for the 1SVM. A potential reason for the failure of the PCA could be that the sources with highest variance, which are preferred by PCA, might be dominated by EEG artifacts rather than task-related activities.

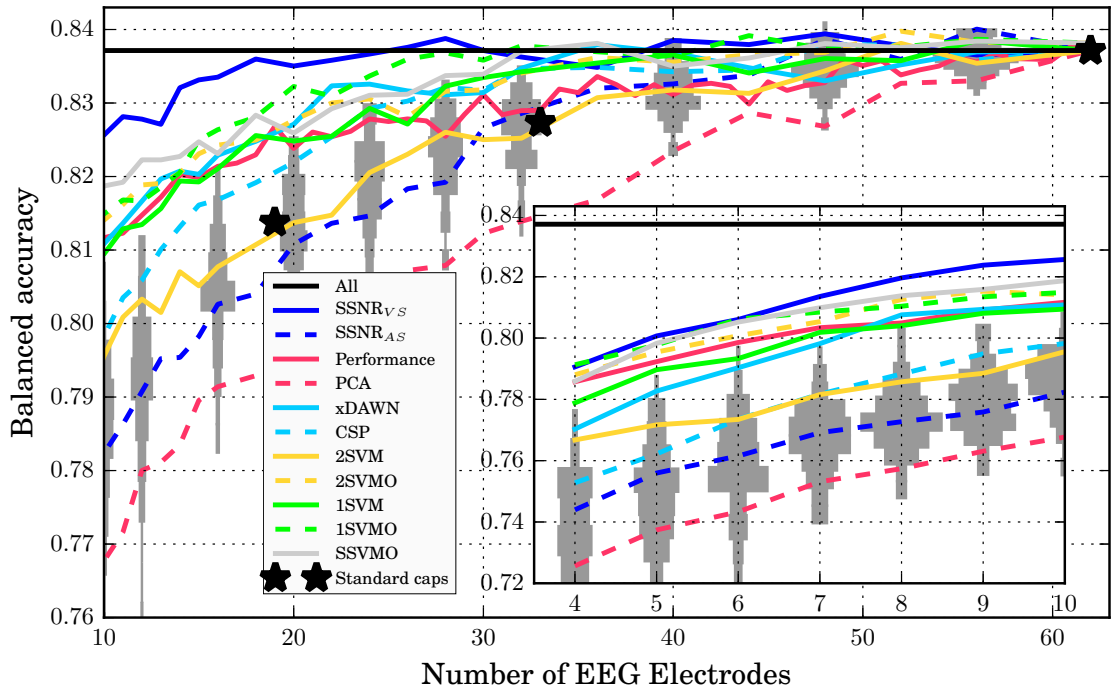

Figure 3.11: **Intra-session evaluation** of the classification performance versus the number of EEG electrodes for different sensor selection approaches. The horizontal line *All* is a reference showing the performance using all available 62 electrodes. The grey patches correspond to histograms of performances of 100 randomly sampled electrode constellations. The elongation in $y$-direction spans the range of the occurring performances and the width of the patches in $x$-direction corresponds to the quantity of results in that particular range. The three black stars represent widely accepted sensor placements for 19, 32, and 62 EEG electrodes. All other curves depict the mean classification performance over all subjects and cross-validation splits. The results for 4–10 sensors are shown separately in the inset. By using an inset the curves in the main graphic appear less compressed. Description taken from [Feess et al., 2013].

In accordance with the results of [Rivet et al., 2012], *SSNR$_{VS}$* performs considerably better than the relatively similar *SSNR$_{AS}$* ranker. This is most likely due to the fact that *SSNR$_{AS}$* cannot take redundancy between channels into account. *SSNR$_{VS}$* accomplishes this by aggregating redundant information from different channels into a single surrogate channel via spatial filtering.

It is perhaps surprising that *Performance* is not the best ranking and *SSNR$_{VS}$* performs much better; we suspect that this might be caused by an overfitting of the sensor selection by *Performance* to the selection session. This effect might be reduced by using a performance estimate which is more robust than the mean, such as the median or the mean minus one standard deviation (to favor constellations with smaller variances in performance and less outliers). However, this issue requires further investigation.

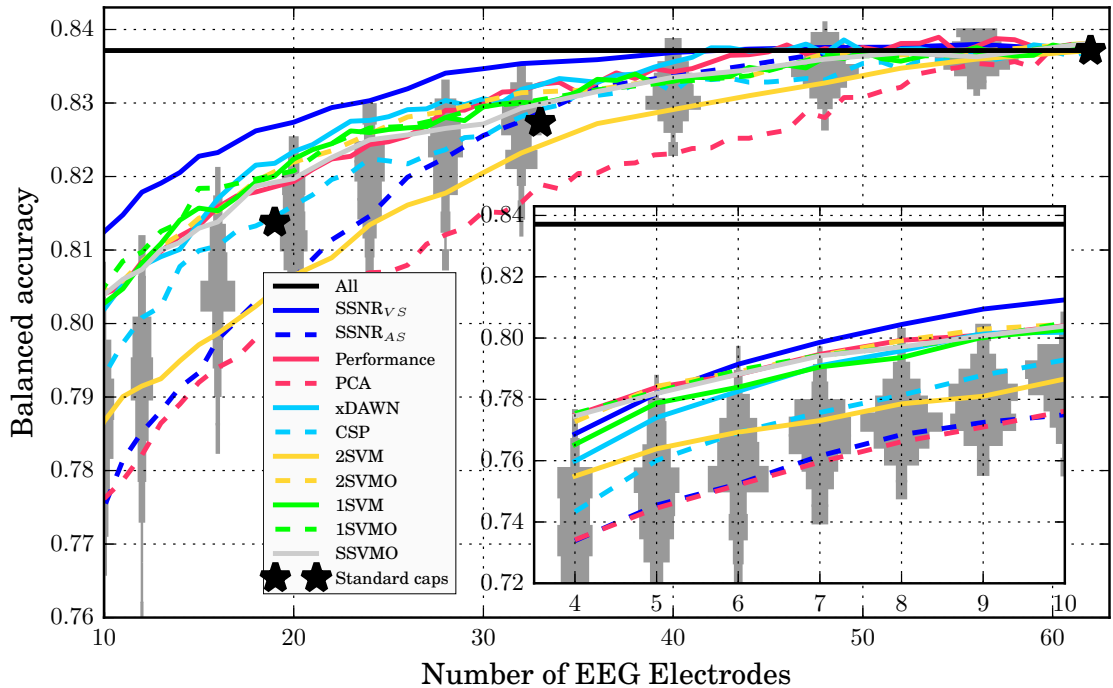

Figure 3.12: **Inter-session evaluation** of the classification performance versus the number of EEG electrodes for different sensor selection approaches. For more details, please see Figure 3.11

The sensor selection capabilities of the SSVMO ranking are reasonable, since the performance is comparable to the other good rankings (xDAWN, 1SVMO, 2SVMO). Maybe, with an improved hyperparameter tuning, this algorithm is able to outperform these algorithms. Therefore, a more efficient implementation would be required. Furthermore, the integration of warm starts for speeding up the hyperparameter optimization might be helpful for future investigations.

For the inter-session scheme, the loss in performance of all methods in comparison to the intra-session scheme is expected. It results from the fact that due to day-to-day changes in brain patterns and differences in the exact sensor placement, different constellations may be optimal on different days—even for the same subject.

The fact, that the relative order of the results remains unchanged, however, indicates that a comparison of electrode selection approaches can in principle be performed without the effort of acquiring a second set of data for each subject. This facilitates the process of deciding for a particular sensor selection approach substantially. For obtaining a realistic estimate of the classification performance in future recordings with less sensors one needs a second, independent data recording session for each subject, however.

All in all, we showed that a reduction of the number of sensors is possible from 62 to at least 40 sensors and that sensor selection approaches using the backtransformation concept show a good performance (SSVMO, 1SVMO, 2SVMO), even though there is still room for improvement. Furthermore, we demonstrated a more complex use case of pySPACE and the necessity to compare algorithms and to optimize hyperparameters.

## 3.5 Discussion

### 3.5.1 Related Work

Based on the commercial software package Matlab, there are open source toolboxes existing for processing data from neuroscience, like EEGLAB [Delorme and Makeig, 2004] and FieldTrip [Oostenveld et al., 2011] for magnetoencephalography (MEG) and EEG, and SPM (`http://www.fil.ion.ucl.ac.uk/spm/`) especially for fMRI data. Respective Python libraries are for example PyMVPA [Hanke et al., 2009], OpenElectrophy [Garcia and Fourcaud-Trocmé, 2009], and the NIPY software projects (`http://nipy.org/`). These tools, are very much tailored to their special type of data and application and are not appropriate for more general signal processing and classification. In scientific computing in general, Python is probably the programming language mostly used, because it is easy to learn/use/read, because it can be made efficient by using C/C++ interfaces, and because it provides high quality libraries which already define a lot of required functionality.

The Python machine learning stack is organized roughly, starting from core libraries for numerical and scientific computation such as NumPy [Dubois, 1999] and SciPy [Jones et al., 2001], over libraries containing implementations of core machine learning algorithms such as scikit-learn [Pedregosa et al., 2011], to higher level frameworks such as MDP, which allow to combine several methods and evaluate their performance empirically. Besides that, there are non-standardized ways of interfacing with machine learning tools that are not implemented in Python such as LibSVM [Chang and Lin, 2011] and WEKA [Hall et al., 2009].

The distinction between libraries and frameworks is typically not strict; frameworks often contain some implementations of basic processing algorithms as libraries do and libraries typically include some basic framework-like tools for configuration and evaluation. pySPACE can be considered as a high-level framework which contains a large set of built-in machine learning algorithms as well as wrappers for external software such as scikit-learn, MDP, WEKA, and LibSVM.

In contrast to libraries like scikit-learn, the focus of pySPACE is much more on configuration, automation, and evaluation of large-scale empirical evaluations of signal processing and machine learning algorithms. Thus, we do not see pySPACE as an alternative to libraries but rather as a high-level framework, which can easily wrap libraries (and does so already for several ones), and which makes it easier to use and compare the algorithms contained in these libraries.

In contrast to frameworks like MDP, pySPACE requires less programming skills since a multitude of different data processing and evaluation procedures can be completely specified using configuration files in YAML-syntax without requiring the user to write scripts, which would be a "show-stopper" for users without programming experience. Similarly, frameworks based on GUIs are not easily used in distributed computing contexts on remote machines without graphical interface. Thus, we consider pySPACE's YAML-based configuration files a good compromise between simplicity and flexibility.

Additionally, pySPACE allows to execute the specified experiments on different computational modalities in a fully automated manner using different back-ends: starting from a serial computation on a single machine, over symmetric multiprocessing on shared-memory multi-core machines, to distributed execution on high-performance clusters based on MPI or IBM's job scheduler LoadLeveler. Further back-ends like one integrating IPython parallel [Pérez and Granger, 2007] could easily be integrated in the future. Other tools for parallel execution are either restricted to the symmetric multiprocessing scenario like joblib [Varoquaux, 2013] or by themselves not directly usable in machine learning without some "glue" scripts such as IPython parallel. Recently, the framework SciKit-Learn Laboratory (skll, https://skll.readthedocs.org/) became open source. This framework also uses the command line for distributing different data processing operations on a summary of datasets but in contrast to pySPACE it only interfaces scikit-learn, which largely limits its capabilities.

A further advantage of pySPACE is that it allows to easily transfer methods from the offline benchmarking mode to the processing in real application scenarios. The user can use the same YAML-based data processing specifications in both modes.

For loading EEG and related data, pySPACE is already quite powerful. But in the data handling of more arbitrary data, it would greatly benefit from interfacing to the Python library pandas [McKinney, 2010]. This library provides a large range of efficient, large scale data handling methods, which could increase the performance of pySPACE and enlarge the number of available formats, data handling algorithms, and data cleaning methods.

There are several further open source signal processing toolboxes which could be interesting to be interfaced with pySPACE like OpenVibe [Renard et al., 2010], BCI2000 [Schalk et al., 2004], EEGLAB [Delorme and Makeig, 2004], Oger [Verstraeten et al., 2012], pyMVPA [Hanke et al., 2009], Shogun [Sonnenburg et al., 2010], and many more, including frameworks which would only use the automatic processing and parallelization capabilities of pySPACE. These interfaces might help to overcome some limitations of the software like the focus on feature vector and segmented time series data or the missing interactive data visualization.

### 3.5.2 My Contribution to pySPACE for this Thesis

pySPACE was not exclusively my own work.[21] For good software development always a team is required and especially major changes to an existing software require a discussion between developers and users. The original benchmarking framework was written by Dr. Jan Hendrik Metzen and Timo Duchrow and the code for the signal processing chains was adapted from MDP.

Nevertheless, for the goal to *implement a framework for better automatizing the process of optimizing the construction of an appropriate signal processing chain including a classifier* and to make this framework open source large changes had to be made. This also includes usability and documentation issues. In context of these changes, I see my major contribution to the framework and to my thesis.

For comparing classifiers, originally the WEKA framework was used. To also evaluate classifiers in signal processing chains (node chains), which was required for the application and the work on Chapter 1, I implemented the concept of classifiers and their evaluation including numerous different performance measures. This work was followed by implementing algorithms for the hyperparameter optimization. For increasing the usability, I suggested, discussed, and implemented the major restructuring of the software, I eased the setup of the software, and largely improved the documentation and the testing suite. The basic concepts introduced in this chapter already existed right from the beginning of the software in 2008 without my contribution because they are required by the problem of tuning signal processing chains itself. My contribution is to make this structure visible in the code and in its documentation for users and developers.

Last but not least, I implemented all algorithms used for evaluations and visualizations in this thesis like the BRMM and the generic backtransformation.

---

[21] This can be seen at http://pyspace.github.io/pyspace/history.html and http://pyspace.github.io/pyspace/credits.html.

All the authors of [Krell et al., 2013b] had an important contribution to the framework and helped making it open source. Due to my aforementioned contribution to pySPACE, I was the main author of this publication. I defined the structure and wrote most text parts of the paper but also the other authors contributed a few parts. For the introduction (which is also used in this chapter) especially Dr. Sirko Straube contributed some text parts. He also mainly implemented Figure 3.1. The text parts about online processing are mainly the work of Johannes Teiwes and Hendrik Wöhrle. Figure 3.3 and the pySPACE logo and some more graphics in the pySPACE documentation are joint work with Johannes Teiwes. The evaluation example (see Section 3.4.1) was joint work with Anett Seeland. The related work (see Section 3.5.1) was mostly written by Dr. Jan Hendrik Metzen.

### 3.5.3 Summary

In this chapter a general framework was presented which supports the tuning/optimization, analysis, and comparison of signal processing chains. Even though more automation and more sophisticated algorithms for the optimization (as for example mentioned in Section 3.3) should be integrated into pySPACE, basic concepts and tools are already available and the software provides interfaces to integrate these approaches. The framework supports a wide range of data formats, platforms, and applications and provides several parallelization schemes, performance metrics, and most importantly algorithms from diverse categories. It can be used for benchmarking as well as real online applications. Numerous results and publications would not have been possible without this framework and the cluster which we use to parallelize our calculations.

In the process of developing methods two aspects of pySPACE were very helpful: the reproducibility and the simplicity of the configuration. For discussing approaches and problems the respective files were used and exchanged. So often new approaches could be tested using configuration files from other scientists. This also ensured comparability and saved a lot of time. Even though the main research results of pySPACE have been in the area of EEG data processing so far, the concepts and implementations can be transferred to other problem settings. For example, some analyses on robotics data have been performed using sensor selection capabilities, the parallelization, or a regression algorithm, and in Chapter 1 we showed an analysis of data on a more abstract level, unrelated to a direct application. Usually, algorithms are developed and tested using scripts and later on these might be exchanged and then need to be adapted to other applications or evaluations. By using pySPACE as a common ground the exchange of algorithms between team members and the transfer to other applications or evaluations was straightforward.

The generic documentation and testing in pySPACE largely ease the maintenance. Nevertheless, working on the framework with the goal to make it usable for everyone in every application is extremely demanding and would probably require some full time developers. In the future, pySPACE could benefit from additional algorithms (e.g., by using an improved wrapper to Weka, or enabling evaluations of clustering algorithms), input/storage formats (e.g., using pandas), job distribution back-ends (e.g., database access, or the distribution concept from the SciKit-Learn Laboratory), and use cases (e.g., soil detection by robots, support in rehabilitation, video and picture processing). Furthermore, there are several possibilities, to improve testing coverage, performance, usability, logging, and automation of the framework. Especially the latter is interesting to target a fully autonomous optimization of a signal processing chain. A broad scientific user community of pySPACE would provide a basis for easy exchange and discussion of signal processing and classification approaches, as well as an increased availability of new signal processing algorithms from various disciplines.

**Related Publications**

**Presentation of the Software**

# Chapter 4

# Conclusion

Optimizing the classification of complex data is a difficult task which often requires expert knowledge. To ease the optimization process especially for non-experts, three approaches are introduced in this thesis to improve the design and understanding of signal processing and classification algorithms and their combination.

**Classifier Connections**

Several connections between existing SVM variants have been shown and resulted in additional new SVM variants including unary classifiers and online learning algorithms, which were shown to be relevant for certain applications. These connections replace the loose net of SVM variants by a strongly connected one, which can be regarded as a more *general* overall model. Knowing the connections, it is easier to understand differences and similarities between the classifiers and save time when teaching, implementing, optimizing, or just applying the classifiers. Furthermore, different concepts can be transferred and existing proofs of properties can be generalized to the connected models.

**Backtransformation**

To interpret and *decode* the complete signal processing chain which ends with a classifier, the backtransformation approach was presented. Whenever the processing consists of affine transformations, it results in a representation of the processing chain, giving weights for each component in the input domain, which can be directly visualized. It replaces the handcrafted and cumbersome visualization and interpretation of single algorithms with a joint view on the complete processing which is very easy to obtain due to a generic implementation.

**pySPACE**

The pySPACE framework was presented as a tool, to process data, tune algorithms and their hyperparameters, and to enable the communication between scientists. This largely supports *optimization* of the signal processing chain. Especially for the classifiers handled in this thesis, the hyperparameter optimization and the choice of the preprocessing is important. Furthermore, the framework was required for comparing the classifiers and analyzing them, as well as for implementing the backtransformation in a generic way. pySPACE was used for all evaluations in this thesis and even in numerous other cases and so proved its usability as a tool for scientific research (see Section 3.4.2).

### Implications for the Practitioner

All three approaches are not to be taken separately[1] but jointly to tackle the question of

> *"**How** shall I use **which** classifier and*
> ***what** features of my data does it rely on?"*

So given a new problem, how can the insights and tools provided in this thesis help?

The first step is that the respective data has to be prepared such that it can be loaded into pySPACE. Usually, this step is quite simple due to the available loading routines, examples, and documentation. Now it is possible to explore several processing chains with pySPACE using different visualization, evaluation, and *optimization* techniques ("How", see Chapter 3). For a more systematic approach when choosing the classifier, the "general" view/model on SVM classifier variants in Chapter 1 can be used ("which"). As outlined in Section 1.5, the model can help in different ways:

- It can be used to understand/teach the models and their connections.
- It provides a rough guideline from the application point of view.
- It can partially be used to optimize the choice of classifier with an optimization algorithm, e.g., as provided by pySPACE. By parameterizing the number of iterations and possibly integrating it into the performance metric, the optimization could help to decide between batch and online learning using the single iteration approach (Section 1.2). Furthermore, the choice between C-SVM and RFDA is parameterized with the relative margin concept from the BRMM (Section 1.3). This provides a smooth transition, which can be used for optimization.

The resulting different processing chains can be compared and analyzed by *decoding* them with the backtransformation ("what", see Chapter 2). Therefore, the pySPACE framework can be used with its generic implementation of the backtransformation. Irrelevant data can be detected and new insights about the data or the processing can be gained. This can be used to derive new algorithms and/or improve the processing chain.

---

[1] For a separate discussion of the approaches refer to Section 1.5, 2.5, and 3.5, respectively.

Last but not least, the backtransformation can be used to support application driven dimensionality reduction, e.g., by extending the classifier model with a respective "sparsification" or by providing a ranking of input components in the source domain instead of the feature domain.

## Outlook

The aforementioned contributions can only be seen as the beginning and a lot of research has to follow.

For the generalizing part (Chapter 1), connections between further classifiers (not limited to SVM variants) should be derived and knowledge and particularities concerning one algorithm should be transferred to the connected ones, if possible.

For the decoding part (Chapter 2), different visualization techniques should be developed for different types of data, and especially new tools should be developed to ease the interpretability of the affine as well as the general backtransformation.

For the framework and optimization part (Chapter 3), the number of algorithms,[2] supported data types,[3] and optimization algorithms[4] should be increased especially with the goal of making pySPACE useful for more applications and providing more functionality, automation, and efficiency of the optimizing approaches.

Overall, the three introduced concepts should be analyzed in further applications to prove their usefulness.

By pushing all aspects further and integrating the results in pySPACE, it might be possible to achieve the longterm goal of creating a nearly fully automized algorithm for autonomous longterm learning and efficient optimization of signal processing chains (autoSPACE).

---

[2] e.g., wrappers, clustering algorithms, and preprocessing which is tailored to not yet supported types of data

[3] e.g., text or music data

[4] e.g., joint optimization of classifier and hyperparameters or joint optimization of filtering and classification

# Appendix A

# Publications
# in the Context of this Thesis

This chapter lists all my 18 publications and the chapter number (No.) they relate to. They are sorted by personal relevance for this thesis. If they are a major part of a chapter, the chapter number is highlighted.

No. Publication details

9 journal publications (1 submitted)

**1**,2,3 Krell, M. M., Feess, D., and Straube, S. (2014a). Balanced Relative Margin Machine – The missing piece between FDA and SVM classification. *Pattern Recognition Letters*, 41:43–52, doi:10.1016/j.patrec.2013.09.018.

**1**,3 Krell, M. M. and Wöhrle, H. (2014). New one-class classifiers based on the origin separation approach. *Pattern Recognition Letters*, 53:93–99, doi:10.1016/j.patrec.2014.11.008.

**2**,3 **submitted:** Krell, M. M. and Straube, S. (2015). Backtransformation: A new representation of data processing chains with a scalar decision function. *Advances in Data Analysis and Classification*. submitted.

**3** Krell, M. M., Straube, S., Seeland, A., Wöhrle, H., Teiwes, J., Metzen, J. H., Kirchner, E. A., and Kirchner, F. (2013b). pySPACE a signal processing and classification environment in Python. *Frontiers in Neuroinformatics*, 7(40):1–11, doi:10.3389/fninf.2013.00040.

1,3 Wöhrle, H., Krell, M. M., Straube, S., Kim, S. K., Kirchner, E. A., and Kirchner, F. (2015). An Adaptive Spatial Filter for User-Independent Single Trial Detection of Event-Related Potentials. *IEEE Transactions on Biomedical Engineering*, doi:10.1109/TBME.2015.2402252.

6 conferences publications

3 talks about pySPACE:

# Appendix B

# Proofs and Formulas

## B.1 Dual Optimization Problems

### B.1.1 Well Defined SVM Model

**Theorem** 1 (The SVM Model is well defined). *The SVM optimization problem has feasible points and a solution always exists if there is at least one sample for each class. Additionally when using the hard margin the sets of the two classes need to be strictly separable. Furthermore, Slater's constraint qualification is fulfilled.*

*Proof.* The first point is fairly easy, because a feasible point $P = (w, b, t)$ is described by

$$w = 0,\, b = 0,\, t_j = 10\, \forall j : 1 \leq j \leq n\,. \tag{B.1}$$

When working with a hard margin SVM the two sets $S_{+1}$ and $S_{-1}$ with

$$S_z = \mathrm{conv}\left(\{x_j | y_j = z\}\right) \tag{B.2}$$

need to be strictly separable by a hyperplane with parametrization $P = (w, b)$ which defines a feasible point. Otherwise, the optimization problem has no solution, because there is no feasible point.

The optimization problem consists of a convex target function and linear constraints. Furthermore, $P$ is a Slater point, because small changes of $P$ are still feasible. Consequently, Slater's constraint qualification can be applied to show that the problem can be locally linearized and Lagrange duality can be applied [Burges, 1998, Slater, 2014].

The argument about solvability will only be given for Method 3 but can be applied to numerous variants analogous like the hard margin separation, squared loss, or arbitrary norm of $w$. The target function $f(w, b, t) = \frac{1}{2} \|w\|_2^2 + C \sum t_j$ is bounded below by zero and $f(P)$ is an upper bound. Furthermore, the constraints are linear and define a closed set and the target function is convex. Even strong convexity holds when fixing $b$. Consequently, there can not be more than one solution vector $w$. Due to the uniqueness of the optimal $t_j$, an optimal $b$ is also unique. With the help of the upper bound $f(P)$, the set of considered feasible points can be reduced to the ones where

$$\|w\|_2 \leq \sqrt[2]{2f(P)}, \ \|t\|_1 \leq \frac{f(P)}{C}. \tag{B.3}$$

holds. All other points result in values of the target function being higher than the value obtained by the feasible point $P$. If there were also limits for $b$, the set of relevant and feasible points would be compact and consequently a minimum would be obtained, because $f$ is a continuous function. Since this is not the case, we have to work with a sequence $(w(n), b(n), t(n))$ approaching the infimum. This sequence is existing, because $f$ is bounded from below. Due to the bounds on $w$ and $t$ we can assume (at least by working with the respective subsequences) $\lim_{n\to\infty} w(n) = w'$ and $\lim_{n\to\infty} t(n) = t'$. Without loss of generality, we can furthermore assume that $\lim_{n\to\infty} b(n) = \infty$ and that $x_1$ is a sample with $y_1 = -1$. Inserting this into the feasibility constraints results in:

$$y_1(\langle w(n), x_1 \rangle + b(n)) \geq 1 - t(n)_1 \Rightarrow \lim_{n\to\infty} -\langle w(n), x_1 \rangle - b(n) \geq \lim_{n\to\infty} 1 - t(n)_1 \tag{B.4}$$

and consequently $-\langle w', x_1 \rangle - \infty \geq 1 - t'_1$ which is a contradiction. Assuming $\lim_{n\to\infty} b(n) = \infty$ also leads to a contradiction by using the sample of the second class with $y_j = +1$. Consequently, we can assume $\lim_{n\to\infty} b(n) = b'$ holds at least for a subsequence. So finally, we obtain the solution of the optimization problem: $(w', b', t')$. $\qquad \square$

## B.1.2 Dual of the Hard Margin Support Vector Machine

**Theorem 18** (Dual Hard Margin SVM). *If the samples of the two classes are strictly separable, the duality gap for the hard margin SVM is zero and the dual optimization problem reads:*

$$\min_{\alpha_j \geq 0, \sum \alpha_j y_j = 0} \frac{1}{2} \sum_{i,j} \alpha_i \alpha_j y_i y_j \langle x_i, x_j \rangle - \sum_j \alpha_j \tag{B.5}$$

*Proof.* The proof is the same as for Theorem 2 using the Lagrange function

$$L(w, b, \alpha) = \frac{1}{2} \|w\|_2^2 - \sum \alpha_j (y_j(\langle w, x_j \rangle + b) - 1) \tag{B.6}$$

and the derivatives

$$\frac{\partial L}{\partial w} = w - \sum_j \alpha_j y_j x_j, \ \frac{\partial L}{\partial b} = -\sum_j \alpha_j y_j. \tag{B.7}$$

$\square$

### B.1.3   Detailed Calculation for the Dual of the L2–SVM

Even though, the calculations for the dual formulation are straightforward, there is always in danger of small errors. To give at least one complete example, we provide the calculation for the L2–SVM from Section 1.1.1.1 Theorem 2 in detail. The Model reads:

**Method 21** (L2–Support Vector Machine ($p = p' = 2$))**.**

$$\min_{w,b,t} \quad \frac{1}{2}\left\|w\right\|_2^2 + \sum C_j t_j^2$$
$$s.t. \qquad y_j(\langle w, x_j\rangle + b) \;\; \geq 1 - t_j \quad \forall j : 1 \leq j \leq n. \tag{B.8}$$

The respective Lagrange function is

$$L_2(w,b,t,\alpha) = \frac{1}{2}\left\|w\right\|_2^2 + \sum C_j t_j^2 - \sum \alpha_j(y_j(\langle w, x_j\rangle + b) - 1 + t_j) \tag{B.9}$$

with the derivatives

$$\frac{\partial L_2}{\partial w} = w - \sum_j \alpha_j y_j x_j, \;\; \frac{\partial L_2}{\partial b} = -\sum_j \alpha_j y_j, \;\; \frac{\partial L_2}{\partial t_j} = 2t_j C_j - \alpha_j \tag{B.10}$$

as explained in Section 1.1.1.1. This results in the equations:

$$w = \sum_j \alpha_j y_j x_j \tag{B.11}$$

$$0 = \sum_j \alpha_j y_j \tag{B.12}$$

$$t_j = \frac{\alpha_j}{2C_j}. \tag{B.13}$$

When substituting the optimal $t_j$ and $w$ in $L_2$, we calculate:

$$L_2(w, b, t, \alpha) = L_2\left(\sum_j \alpha_j y_j x_j, b, \frac{\alpha_j}{2C_j}, \alpha\right) \tag{B.14}$$

$$= \frac{1}{2}\left\langle \sum_i \alpha_i y_i x_i, \sum_j \alpha_j y_j x_j \right\rangle + \sum_j C_j \left(\frac{\alpha_j}{2C_j}\right)^2 \tag{B.15}$$

$$- \sum_j \alpha_j \left(y_j \left(\left\langle \sum_i \alpha_i y_i x_i, x_j \right\rangle + b\right) - 1 + \frac{\alpha_j}{2C_j}\right) \tag{B.16}$$

$$= \frac{1}{2} \sum_{i,j} \alpha_i \alpha_j y_i y_j \left\langle x_i, x_j \right\rangle + \sum_j \frac{\alpha_j^2}{4C_j} \tag{B.17}$$

$$- \sum_{i,j} \alpha_i \alpha_j y_i y_j \left\langle x_i, x_j \right\rangle - b \sum_j \alpha_j y_j + \sum_j \alpha_j - \sum_j \frac{\alpha_j^2}{2C_j} \tag{B.18}$$

$$= -\frac{1}{2} \sum_{i,j} \alpha_i \alpha_j y_i y_j \left\langle x_i, x_j \right\rangle - \sum_j \frac{\alpha_j^2}{4C_j} - b \cdot 0 + \sum_j \alpha_j \, . \tag{B.19}$$

Only in the last step, Equation (B.12) was used to eliminate $b$. Consequently, if $b$ were omitted in the original model, the resulting function would still be the same. Only the additional restriction from Equation (B.12) would disappear.

### B.1.4 Dual of the $\nu$-SVM

**Theorem 19** (Dual of the $\nu$-SVM). *Assuming solvability of the $\nu$-SVM*

$$\begin{aligned}
\min_{w,t,\rho,b} \quad & \frac{1}{2} \|w\|_2^2 - \nu\rho + \frac{1}{n} \sum t_j \\
s.t. \quad & y_j \left(\langle w, x_j \rangle + b\right) \geq \rho - t_j \text{ and } t_j \geq 0 \; \forall j : 1 \leq j \leq n \, .
\end{aligned} \tag{B.20}$$

*the dual optimization can be formulated as:*

$$\begin{aligned}
\min_{\alpha} \quad & \frac{1}{2} \sum_{i,j} \alpha_i \alpha_j y_i y_j \left\langle x_i, x_j \right\rangle \\
s.t. \quad & \frac{1}{n} \geq \alpha_j \geq 0 \; \forall j : 1 \leq j \leq n, \sum_j \alpha_j y_j = 0, \sum_j \alpha_j = \nu \, .
\end{aligned} \tag{B.21}$$

*Proof.* As in Theorem 1 it can be shown, that there are feasible points of the $\nu$-SVM and that it fulfills Slater's constraint qualification. The therein mentioned proof for the existence of a solution cannot be applied here. The target function is bounded only from above by zero because putting all variables to zero is a feasible solution. (As a side effect it automatically holds $\rho > 0$ [Crisp and Burges, 2000].) Furthermore, the target function contains a negative component. Consequently, it cannot be used for restricting the set of feasible points to a compact set.

The Lagrange functions reads:

$$L(w, b, t, \rho, \alpha, \gamma) = \frac{1}{2} \|w\|_2^2 - \nu\rho + \frac{1}{n} \sum_j t_j - \sum_j \alpha_j(y_j(\langle w, x_j \rangle + b) - \rho + t_j) - \sum_j \gamma_j t_j \tag{B.22}$$

with the derivatives

$$\frac{\partial L}{\partial w} = w - \sum_j \alpha_j y_j x_j, \ \frac{\partial L}{\partial b} = -\sum_j \alpha_j y_j, \ \frac{\partial L}{\partial t_j} = \frac{1}{n} - \alpha_j - \gamma_j, \ \frac{\partial L}{\partial \rho} = -\nu + \sum_j \alpha_j . \tag{B.23}$$

Again, setting the derivatives to zero and substituting them in the Lagrange functions, provides the dual problem. $\rho$ is eliminated from the Lagrange function, but the additional constraint equation remains. Everything else is the same as for the dual of the L1–SVM in Theorem 2.                                                                   $\square$

## B.1.5   Dual of the Binary BRMM

The general (primal) L1–BRMM model with special offset treatment reads:

$$\begin{aligned}
\min_{w, b, s, t} \quad & \frac{1}{2} \|w\|_2^2 + \frac{H}{2} b^2 + \sum C_j t_j + \sum C_j' s_j \\
\text{s.t.} \quad & R_j + s_j \geq y_j(\langle w, x_j \rangle + b) \quad \geq 1 - t_j \quad \forall j : 1 \leq j \leq n \\
& s_j \quad \geq 0 \quad \forall j : 1 \leq j \leq n \\
& t_j \quad \geq 0 \quad \forall j : 1 \leq j \leq n.
\end{aligned} \tag{B.24}$$

The corresponding L2–BRMM model is very similar:

$$\begin{aligned}
\min_{w, b, s, t} \quad & \frac{1}{2} \|w\|_2^2 + \frac{H}{2} b^2 + \sum C_j t_j^2 + \sum C_j' s_j^2 \\
\text{s.t.} \quad & R_j + s_j \geq y_j(\langle w, x_j \rangle + b) \quad \geq 1 - t_j \quad \forall j : 1 \leq j \leq n.
\end{aligned} \tag{B.25}$$

The Lagrange functions read:

$$L_1(w, b, s, t, \alpha, \beta, \gamma, \delta) = \frac{1}{2} \|w\|_2^2 + \frac{H}{2} b^2 + \sum C_j t_j + \sum C_j' s_j \tag{B.26}$$

$$- \sum \alpha_j(y_j(\langle w, x_j \rangle + b) - 1 + t_j) \tag{B.27}$$

$$+ \sum \beta_j(y_j(\langle w, x_j \rangle + b) - R_j - s_j) \tag{B.28}$$

$$- \sum \gamma_j s_j - \sum \delta_j t_j \text{ and} \tag{B.29}$$

$$L_2(w, b, s, t, \alpha, \beta) = \frac{1}{2} \|w\|_2^2 + \frac{H}{2} b^2 + \sum C_j t_j^2 + \sum C_j' s_j^2 \tag{B.30}$$

$$- \sum \alpha_j(y_j(\langle w, x_j \rangle + b) - 1 + t_j) \tag{B.31}$$

$$+ \sum \beta_j(y_j(\langle w, x_j \rangle + b) - R_j - s_j). \tag{B.32}$$

The original problem is now equivalent to first maximize $L$ with positive dual variables $(\alpha, \beta, \gamma, \delta)$ and then minimize with respect to the primal variables $(w, b, s, t)$. The dual problem is the reverse. Hence, we first need the derivatives with respect to the primal variables:

$$\frac{\partial L}{\partial w} = w - \sum (\alpha_j - \beta_j) y_j x_j, \qquad \frac{\partial L}{\partial b} = Hb - \sum (\alpha_j - \beta_j) y_j, \qquad \text{(B.33)}$$

$$\frac{\partial L_1}{\partial s_j} = C_j' - \beta_j - \delta_j, \qquad \frac{\partial L_2}{\partial s_j} = 2 s_j C_j' - \beta_j, \qquad \text{(B.34)}$$

$$\frac{\partial L_1}{\partial t_j} = C_j - \alpha_j - \gamma_j, \qquad \frac{\partial L_2}{\partial t_j} = 2 t_j C_j - \alpha_j. \qquad \text{(B.35)}$$

For getting the dual problems, two steps are required. Setting the derivatives zero, gives equations for the primal variables, which then can be replaced in the optimization problem, such that only the dual problem remains. The other step, just for cosmetic reasons, is to multiply the problem with $-1$ and switch from maximization to minimization. Together, this results in

$$\begin{aligned} \min_{\alpha, \beta} \quad & \tfrac{1}{2}(\alpha - \beta)^T Q (\alpha - \beta) - \sum \alpha_j + \sum R_j \beta_j \\ \text{s.t.} \quad & 0 \leq \alpha_j \leq C_j \quad \forall j : 1 \leq j \leq n \\ & 0 \leq \beta_j \leq C_j' \quad \forall j : 1 \leq j \leq n \end{aligned} \qquad \text{(B.36)}$$

in the L1 case and

$$\begin{aligned} \min_{\alpha, \beta} \quad & \tfrac{1}{2}(\alpha - \beta)^T Q (\alpha - \beta) - \sum \alpha_j + \sum R_j \beta_j + \tfrac{1}{4} \sum \tfrac{\alpha_j^2}{C_j} + \tfrac{1}{4} \sum \tfrac{\beta_j^2}{C_j'} \\ \text{s.t.} \quad & 0 \leq \alpha_j \quad \forall j : 1 \leq j \leq n \\ & 0 \leq \beta_j \quad \forall j : 1 \leq j \leq n \end{aligned} \qquad \text{(B.37)}$$

in the L2 case with

$$Q_{kl} = y_k y_l \left( \langle x_k, x_l \rangle + \frac{1}{H} \right) \quad \forall k, l : 1 \leq k \leq n, 1 \leq l \leq n. \qquad \text{(B.38)}$$

Without the special offset treatment, the equation

$$\sum_j y_j (\alpha_j - \beta_j) = 0 \qquad \text{(B.39)}$$

would have to be added to the constraints in the dual optimization problem. (The proof would be nearly the same.)

### B.1.6   Dual of the One-Class BRMM Models

The dual problem formulations are required to derive update formulas to get a solution in an iterative way and to generate the online versions of the algorithms. Furthermore, they are needed to introduce kernels.

**Method 22** (Dual of the One-Class BRMM).

$$\min_{\alpha,\beta} \quad \frac{1}{2}\sum_{i,j}(\alpha_i - \beta_i)(\alpha_j - \beta_j)\langle x_i, x_j\rangle - 2\sum\alpha_j + (R+1)\sum\beta_j$$
$$\text{s.t.} \quad 0 \leq \alpha_j \leq C \text{ and } 0 \leq \beta_j \leq C \ \forall j : 1 \leq j \leq n \ . \tag{B.40}$$

**Theorem 20** (Dual of the One-Class BRMM). *Method 22 is a dual problem of Method 28 and both methods are connected via*

$$w = \sum(\alpha_j - \beta_j)x_j \ . \tag{B.41}$$

*Proof.* To simplify the calculations, we use the equivalent formulation:

$$\min_{w,t,u} \quad \frac{1}{2}\|w\|_2^2 + C\sum t_j + C\sum u_j$$
$$\text{s.t.} \quad 1 + R + u_j \geq \langle w, x_j\rangle \geq 2 - t_j \text{ and } t_j, u_j \geq 0 \ \forall j : 1 \leq j \leq n \ . \tag{B.42}$$

This is a convex optimization problem and with $w = 0$, $t_j = 10$, $u_j = 10 \ \forall i$ a slater point is defined. Hence, strong duality holds (Slater's constraint qualification) [Boyd and Vandenberghe, 2004]. From the modified problem, we can derive the Lagrange function:

$$L(w,t,u,\alpha,\beta,\gamma,\delta) = \frac{1}{2}\|w\|_2^2 \tag{B.43}$$
$$+ C\sum t_j - \sum\gamma_j t_j + \sum\alpha_j\left(2 - t_j - \langle w, x_j\rangle\right) \tag{B.44}$$
$$+ C\sum u_j - \sum\delta_j u_j + \sum\beta_j\left(\langle w, x_j\rangle - R - 1 - u_j\right) \tag{B.45}$$

and calculate the derivatives:

$$\frac{\partial L}{\partial w} = w - \sum(\alpha_j - \beta_j)x_j \ , \frac{\partial L}{\partial t_j} = C - \gamma_j - \alpha_j \ , \frac{\partial L}{\partial u_j} = C - \delta_j - \beta_j \ . \tag{B.46}$$

To get the dual problems, the derivatives have to be set to zero and substituted in the Lagrange function. Since all dual variables have to be positive, the equations $\frac{\partial L}{\partial t_j} = 0$ and $\frac{\partial L}{\partial u_j} = 0$ can be used to eliminate $\gamma_j$ and $\delta_j$ with the inequalities $\alpha_j \leq C$ and $\beta_j \leq C$. Putting everything together gives us the dual problem:

$$\max_{\alpha,\beta} \quad -\frac{1}{2}\sum_{i,j}(\alpha_i - \beta_i)(\alpha_j - \beta_j)\langle x_i, x_j\rangle + 2\sum\alpha_j - (R+1)\sum\beta_j$$
$$\text{s.t.} \quad 0 \leq \alpha_j \leq C \text{ and } 0 \leq \beta_j \leq C \ \forall j : 1 \leq j \leq n \ . \tag{B.47}$$

Multiplication of the target function with $-1$ completes the proof. □

**Method 23** (Dual of the L2–One-Class BRMM).

$$\min_{\alpha,\beta} \quad \frac{1}{2}\sum_{i,j}(\alpha_i - \beta_i)(\alpha_j - \beta_j)\langle x_i, x_j\rangle - 2\sum\alpha_j + (R+1)\sum\beta_j + \sum\frac{\alpha_j^2+\beta_j^2}{2C} \tag{B.48}$$
$$\text{s.t.} \quad 0 \leq \alpha_j \text{ and } 0 \leq \beta_j \,\forall j : 1 \leq j \leq n \,.$$

**Theorem 21** (Dual of the L2–One-Class BRMM). *Method 23 is a dual problem of Method 31 and both methods are connected via* $w = \sum(\alpha_j - \beta_j)x_j$.

*Proof.* To simplify the calculations, we use an equivalent formulation:

$$\min_{w,t,u} \quad \frac{1}{2}\|w\|_2^2 + \frac{C}{2}\sum t_j^2 + \frac{C}{2}\sum u_j^2 \tag{B.49}$$
$$\text{s.t.} \quad 1 + R + u_j \geq \langle w, x_j\rangle \geq 2 - t_j \,\forall j : 1 \leq j \leq n \,.$$

This is a convex optimization problem and with $w = 0$, $t_j = 10$, $u_j = 10 \,\forall j$ a slater point is defined. Hence, strong duality holds (Slater's constraint qualification) [Boyd and Vandenberghe, 2004]. From the modified problem formulation, we can derive the Lagrange function:

$$L(w,t,u,\alpha,\beta) = \frac{1}{2}\|w\|_2^2 \tag{B.50}$$
$$+ \frac{C}{2}\sum t_j^2 + \sum\alpha_j\left(2 - t_j - \langle w, x_j\rangle\right) \tag{B.51}$$
$$+ \frac{C}{2}\sum u_j^2 + \sum\beta_j\left(\langle w, x_j\rangle - R - 1 - u_j\right) \tag{B.52}$$

and calculate the derivatives:

$$\frac{\partial L}{\partial w} = w - \sum(\alpha_j - \beta_j)x_j \,, \frac{\partial L}{\partial t_j} = Ct_j - \alpha_j \,, \frac{\partial L}{\partial u_j} = Cu_j - \beta_j \,. \tag{B.53}$$

To get the dual problems, the derivatives have to be set to zero and substituted in the Lagrange function, to eliminate the primal variables: $w$, $t$, and $u$. This gives us the dual problem:

$$\max_{\alpha,\beta} \quad -\frac{1}{2}\sum_{i,j}(\alpha_i - \beta_i)(\alpha_j - \beta_j)\langle x_i, x_j\rangle + 2\sum\alpha_j - (R+1)\sum\beta_j - \sum\frac{\alpha_j^2+\beta_j^2}{2C} \tag{B.54}$$
$$\text{s.t.} \quad 0 \leq \alpha_j \text{ and } 0 \leq \beta_j \,\forall j : 1 \leq j \leq n \,.$$

Multiplication of the target function with $-1$ completes the proof. □

**Method 24** (Dual of the Hard-Margin One-Class BRMM).

$$\min_{\alpha,\beta} \quad \frac{1}{2}\sum_{i,j}(\alpha_i - \beta_i)(\alpha_j - \beta_j)\langle x_i, x_j\rangle - 2\sum\alpha_j + (R+1)\sum\beta_j$$
$$s.t. \quad 0 \leq \alpha_j \text{ and } 0 \leq \beta_j \; \forall j : 1 \leq j \leq n \; . \tag{B.55}$$

**Theorem 22** (Dual of the Hard-Margin One-Class BRMM). *If there exists a $w'$ with $R > \langle w', x_j\rangle > 0 \; \forall j$, Method 24 is a dual problem of Method 32 and both methods are connected via $w = \sum(\alpha_j - \beta_j)x_j$.*

*Proof.* Since $w'$ fulfills the constraints $R > \langle w', x_j\rangle > 0 \; \forall j$, it is a slater point of Method 32. Hence, strong duality holds (Slater's constraint qualification) [Boyd and Vandenberghe, 2004]. The Lagrange function for Method 32 is:

$$L(w, \alpha, \beta) = \frac{1}{2}\|w\|_2^2 + \sum\alpha_j(2 - \langle w, x_j\rangle) + \sum\beta_j(\langle w, x_j\rangle - R - 1) \tag{B.56}$$

with the derivative:

$$\frac{\partial L}{\partial w} = w - \sum(\alpha_j - \beta_j)x_j \; . \tag{B.57}$$

Hence, the dual problem reads:

$$\max_{\alpha,\beta} \quad -\frac{1}{2}\sum_{i,j}(\alpha_i - \beta_i)(\alpha_j - \beta_j)\langle x_i, x_j\rangle + 2\sum\alpha_j - (R+1)\sum\beta_j$$
$$\text{s.t.} \quad 0 \leq \alpha_j \text{ and } 0 \leq \beta_j \; \forall j : 1 \leq j \leq n \; . \tag{B.58}$$

Multiplication of the target function with $-1$ completes the proof. $\qquad\square$

**Method 25** (Dual BRMM Variants with Kernel $k$). *To introduce kernels, $\langle x_i, x_j\rangle$ is again replaced by $k(x_i, x_j)$ in the dual problems. The decision functions reads:*

$$f(x) = \text{sgn}\left(\left(\sum(\alpha_j - \beta_j)k(x_j, x)\right) - 2\right) \tag{B.59}$$

## B.2   Model Connections

### B.2.1   Least Squares SVM and Ridge Regression

The model for the ridge regression is:

**Method 26** (Ridge Regression).

$$\min_{w,b,t} \quad \frac{1}{2}\|w\|_2^2 + \frac{C}{2}\sum t_j^2$$
$$s.t. \quad y_j - (\langle w, x_j\rangle + b) = t_j \quad \forall j : 1 \leq j \leq n \tag{B.60}$$

*with $y_j \in \mathbb{R}$.*

When restricting this model to $y_j \in \{-1, +1\}$ (binary classification) it is equivalent to LS-SVM (Method 7) due to the equation:

$$t_j^2 = (1 - y_j(\langle w, x_j \rangle + b))^2 = (y_j - (\langle w, x_j \rangle + b))^2. \tag{B.61}$$

## B.2.2 Equality of $\epsilon$-RFDA and BRMM

The $\epsilon$-RFDA method from Section 1.3.3.3 reads

**Definition 6** (2–norm regularized, $\epsilon$-insensitive RFDA).

$$\begin{aligned}
\min_{w,b,t} \quad & \tfrac{1}{2} \|w\|_2^2 + C \|t\|_\epsilon \\
\text{s.t.} \quad & y_j(\langle w, x_j \rangle + b) = 1 - t_j \quad \forall j : 1 \leq j \leq n.
\end{aligned} \tag{B.62}$$

**Theorem 12** (Equivalence between RFDA, SVR, and BRMM). *The RFDA with $\epsilon$-insensitive loss function and 2–norm regularization (or the SVR reduced to the values 1 and $-1$) and BRMM result in an identical classification with a corresponding function, mapping RFDA (SVR) hyperparameters $(C, \epsilon)$ to BRMM hyperparameters $(C', R')$ and vice versa.*

*Proof.* As a first step, we want to replace the $\epsilon$–norm by a linear formulation. Using the definition of $\|.\|_\epsilon$ and replacing $|t_j| - \epsilon$ by a new variable $h_j$, the method can be written as

$$\begin{aligned}
\min_{w,b,h} \quad & \tfrac{1}{2} \|w\|_2^2 + C \sum \max \{h_j, 0\} \\
\text{s.t.} \quad & |y_j(\langle w, x_j \rangle + b) - 1| = h_j + \epsilon \quad \forall j : 1 \leq j \leq n.
\end{aligned} \tag{B.63}$$

Since $h$ is subject to minimization, the constraint can just as well be specified as inequality

$$|y_j(\langle w, x_j \rangle + b) - 1| \leq h_j + \epsilon. \tag{B.64}$$

Additionally, to omit the $\max \{h_j, 0\}$ term we introduce a new positive variable $s_j$ and define $s_j = h_j$ if $h_j > 0$ and $s_j = 0$ for $h_j \leq 0$. This results in a further reformulation of the original method:

$$\begin{aligned}
\min_{w,b,s} \quad & \tfrac{1}{2} \|w\|_2^2 + C \sum s_j \\
\text{s.t.} \quad & |y_j(\langle w, x_j \rangle + b) - 1| \leq s_j + \epsilon \quad \forall j : 1 \leq j \leq n \\
& s_j \geq 0 \quad \forall j : 1 \leq j \leq n.
\end{aligned} \tag{B.65}$$

The last step is to replace the absolute value. This is done with the help of a case-by-case analysis that results in the linear program

$$
\begin{aligned}
\min_{w,b,s} \quad & \tfrac{1}{2}\,\|w\|_2^2 + C\sum s_j \\
\text{s.t.} \quad & y_j(\langle w, x_j\rangle + b) \geq 1 - (s_j + \epsilon) \quad \forall j : 1 \leq j \leq n \\
& y_j(\langle w, x_j\rangle + b) \leq 1 + (s_j + \epsilon) \quad \forall j : 1 \leq j \leq n \\
& \qquad\qquad\qquad s_j \geq 0 \qquad\qquad\quad \forall j : 1 \leq j \leq n.
\end{aligned}
\tag{B.66}
$$

As $\epsilon < 1$ we can scale the problem by dividing by $1 - \epsilon$, yielding the scaled variables $w', b', s_j'$. We also scale the target function with $\frac{1}{(1-\epsilon)^2}$, such that the scaled problem reads

$$
\begin{aligned}
\min_{w,b,s} \quad & \tfrac{1}{2}\,\|w'\|_2^2 + \tfrac{C}{1-\epsilon}\sum s_j' \\
\text{s.t.} \quad & y_j(\langle w', x_j\rangle + b') \geq 1 - s_j' \qquad\; \forall j : 1 \leq j \leq n \\
& y_j(\langle w', x_j\rangle + b') \leq \tfrac{1+\epsilon}{1-\epsilon} + s_j' \quad \forall j : 1 \leq j \leq n \\
& \qquad\qquad\qquad s_j' \geq 0 \qquad\qquad\;\; \forall j : 1 \leq j \leq n.
\end{aligned}
\tag{B.67}
$$

The scaling also has an effect on the classification function, which is scaled in the same way as the variables. This scaling does not change the sign of the classification values and so the mapping to the classes is still the same. Renaming $\frac{C}{1-\epsilon}$ to $C'$ and $\frac{1+\epsilon}{1-\epsilon}$ to $R'$, the result is BRMM with hyperparameters $C'$ and $R'$.

To make the proof in the other direction, we first have to search for the $\epsilon$ corresponding to $R'$ and afterwards scale the $C'$ with the help of $\epsilon$. This results in

$$
\epsilon = \frac{R'-1}{R'+1}, C = (1-\epsilon)C' = \frac{2C'}{R'+1}.
\tag{B.68}
$$

For the mapping between RFDA and SVR we have to use the fact that $y_j \in \{-1, 1\}$ and consequently $|y_j| = 1$ and $y_j^2 = 1$ :

$$
|y_j(\langle w, x_j\rangle + b) - 1| = |y_j|\,|y_j(\langle w, x_j\rangle + b) - 1| = |(\langle w, x_j\rangle + b) - y_j|.
\tag{B.69}
$$

Note that it is possible to always replace the absolute value function $|a|$ with $-b \leq a \leq b$ if $b$ is subject to minimization and it automatically holds $b \geq 0$.

<div style="text-align: right">□</div>

### B.2.3　One-Class Algorithm Connections

**Theorem 16** (Equivalence of SVDD and $\nu$oc-SVM on the Unit Hypersphere)**.** *If all training samples lie on the unit hypersphere, SVDD (Method 10) is equivalent to $\nu$oc-SVM (Method 11).*

*Proof.* SVDD is defined by

$$\min_{R',a,t'} \quad R'^2 + C' \sum t'_j$$
$$\text{s.t.} \quad \|a - x_j\|_2^2 \leq R'^2 + t'_j \text{ and } t'_j \geq 0 \; \forall j : 1 \leq j \leq n \;. \tag{B.70}$$

The norm part can be rewritten:

$$\|a - x_j\|_2^2 = \|a\|_2^2 - 2 \langle a, x_j \rangle + \|x_j\|_2^2 \tag{B.71}$$

With this equation and since the samples are on the unit hypersphere $\left( \|x_j\|_2^2 = 1 \right)$ the SVDD method can be reformulated to:

$$\min_{R',a,t'} \quad R'^2 + C' \sum t'_j$$
$$\text{s.t.} \quad \langle a, x_j \rangle \geq \frac{\|a\|_2^2 + 1 - R'^2 - t'_j}{2} \text{ and } t'_j \geq 0 \; \forall j : 1 \leq j \leq n \tag{B.72}$$

with $f_a(x) = \text{sgn} \left( R'^2 - \|a\|_2^2 + 2 \langle a, x_j \rangle - 1 \right)$. Using the mapping

$$w = a, \; t_j = \frac{t'_j}{2}, \; \rho = \frac{\|a\|_2^2 + 1 - R^2}{2}, \; \nu = \frac{1}{C'l} \tag{B.73}$$

results in

$$\min_{w,\rho,t} \quad \|w\|_2^2 + 1 - 2\rho + 2\frac{1}{\nu l} \sum t_j$$
$$\text{s.t.} \quad \langle w, x_j \rangle \geq \rho - t_j \text{ and } 2t_j \geq 0 \; \forall j : 1 \leq j \leq n \;. \tag{B.74}$$

Scaling the target function and the restriction $2t_j \geq 0$ with $0.5$, shows the equivalence to the $\nu$oc-SVM model (Method 11). The decision functions are the same, too:

$$f_a(x) = \text{sgn} \left( 2 \langle w, x_j \rangle - 2\rho \right) = \text{sgn} \left( \langle w, x_j \rangle - \rho \right) \;. \tag{B.75}$$

On the other hand, to get from the $\nu$oc-SVM to the SVDD, the reverse mapping

$$a = w, \; t'_j = 2t_j, \; R^2 = \|w\|_2^2 + 1 - 2\rho, \; C' = \frac{1}{\nu l} \tag{B.76}$$

can be used. $\qquad \square$

**Theorem 17** (From $\nu$oc-SVM to the New One-Class SVM)**.** *Let $\rho(\nu)$ denote the optimal value of $\nu$oc-SVM model. If $\rho(\nu) > 0$, $\nu$oc-SVM is equivalent to our new one-class SVM.*

*Proof.* Having the optimal $\rho(\nu)$, the optimal $w$ for $\nu$oc-SVM can be determined by the optimization problem.

$$
\begin{aligned}
\min_{w,t} \quad & \tfrac{1}{2}\left\| w \right\|_2^2 + \tfrac{1}{\nu l}\sum t_j \\
\text{s.t.} \quad & \langle w, x_j \rangle \geq \rho(\nu) - t_j \text{ and } t_j \geq 0 \ \forall j \ .
\end{aligned} \tag{B.77}
$$

Now, scaling the target function with $\frac{4}{(\rho(\nu))^2}$ and the restrictions with $\frac{2}{\rho(\nu)}$ gives the equivalent optimization problem:

$$
\begin{aligned}
\min_{w,t} \quad & \tfrac{1}{2}\left\| \tfrac{2w}{\rho(\nu)} \right\|_2^2 + \tfrac{2}{\nu l \rho(\nu)}\sum \tfrac{2t_j}{\rho(\nu)} \\
\text{s.t.} \quad & \left\langle \tfrac{2w}{\rho(\nu)}, x_j \right\rangle \geq 2 - \tfrac{2t_j}{\rho(\nu)} \text{ and } \tfrac{2t_j}{\rho(\nu)} \geq 0 \ \forall j \ .
\end{aligned} \tag{B.78}
$$

This scaling is feasible, since $\rho(\nu) > 0$. Substituting

$$
\bar{w} = \frac{2w}{\rho(\nu)}, \ \bar{t}_j = \frac{2t_j}{\rho(\nu)}, \ \bar{C} = \frac{2}{\nu l \rho(\nu)} \tag{B.79}
$$

results in the new one-class SVM. Finally, for the decision function it holds:

$$
f(x) = \operatorname{sgn}\left( \langle w, x_j \rangle - \rho(\nu) \right) = \operatorname{sgn}\left( \left\langle \tfrac{\rho(\nu)}{2}\bar{w}, x_j \right\rangle - 2\tfrac{\rho(\nu)}{2} \right) = \operatorname{sgn}\left( \langle \bar{w}, x_j \rangle - 2 \right) \ . \tag{B.80}
$$

$\square$

**Theorem 23** (Hard-Margin One-Class SVM: $C = \infty$). *Let $X$ denote the set of training instances $x_j$ with the convex hull $\operatorname{conv}(X)$. For the Hard-Margin One-Class SVM, the origin separation approach reveals that the optimal hyperplane (for the positive class) is tangent to $\operatorname{conv}(X)$ in its point of minimal norm $x'$. The hyperplane is orthogonal to the vector $x'$ with $w = x'\frac{2}{\|x'\|_2^2}$ .*

*Proof.*

**Method 27** (Hard-Margin One-Class SVM).

$$
\begin{aligned}
\min_{w} \quad & \tfrac{1}{2}\left\| w \right\|_2^2 \\
\text{s.t.} \quad & \langle w, x_j \rangle \geq 2 \ \forall j : 1 \leq j \leq n \ .
\end{aligned} \tag{B.81}
$$

Via convex linear combination of $\langle w, x_j \rangle \geq 2 \ \forall i$ it holds $\langle w, x \rangle \geq 2 \ \forall x \in \operatorname{conv}(X)$. Furthermore, by the Cauchy-Schwarz inequality one gets:

$$
2 \leq \langle w, x' \rangle \leq \left\| w \right\|_2 \left\| x' \right\|_2 \Rightarrow \left\| w \right\|_2 \geq \frac{2}{\left\| x' \right\|_2} \ . \tag{B.82}
$$

So if $w' = x'\frac{2}{\|x'\|_2^2}$ would fulfill all restrictions, it would be optimal, because $\|w'\|_2 = \frac{2}{\|x'\|_2}$. The following proof is a variant from [Boyd and Vandenberghe, 2004, separat-

ing hyperplane theorem]. Assume, there exists an $x_j$ with $\left\langle \frac{2x'}{\|x'\|_2^2}, x_j \right\rangle < 2$. This can be reformulated to $\langle x', x_j \rangle < \|x'\|_2^2$. Due to convexity it holds $(1-\alpha)x' + \alpha x_j \in \text{conv}(X)$ for any $0 \leq \alpha \leq 1$. Consider the function $h : \mathbb{R} \to \mathbb{R}$, $h(\alpha) = \|(1-\alpha)x' + \alpha x_j\|_2^2$ and its derivative at zero:

$$\frac{\partial h}{\partial \alpha}(0) = \frac{\partial}{\partial \alpha} \left( (1-\alpha)^2 \|x'\|_2^2 + \alpha^2 \|x_j\|_2^2 + 2\alpha(1-\alpha) \langle x', x_j \rangle \right)(0) \tag{B.83}$$

$$= \left( -2(1-\alpha) \|x'\|_2^2 + 2\alpha \|x_j\|_2^2 + 2(1-2\alpha) \langle x', x_j \rangle \right)(0) \tag{B.84}$$

$$= -2 \|x'\|_2^2 + 2 \langle x', x_j \rangle \ . \tag{B.85}$$

With our assumption we get: $\frac{\partial h}{\partial \alpha}(0) < 0$. Consequently, there exists a small $0 < t < 1$ such that $h(t) < h(0)$. In other words, there exists an

$$x'_t = (1-t)x' + t x_j \in \text{conv}(X) \tag{B.86}$$

such that $\|x'_t\|_2 < \|x'\|_2$ which contradicts the definition of $x'$ which is the point of minimal norm in $\text{conv}(X)$. Hence, our assumption was wrong and $w' = x' \frac{2}{\|x'\|_2^2}$ fulfills all the restrictions $\langle w', x_j \rangle \geq 2 \ \forall j$ and is the solution of the hard-margin one-class SVM. □

### B.2.4 Connection of Classifiers with Different Regularization Parameter

**Theorem 24** (Linear Transition with Regularization Parameter). *There is a function of optimal dual variables $\alpha(C)$ of the $C$-SVM depending on the chosen regularization parameter $C$. It can be defined, such that except for a finite number of points it is locally linear which means that it locally has a representation: $\alpha_i(C) = C v_1 + v_2$ with $v_1, v_2 \in \mathbb{R}^n$. The same holds for the function $b(C)$. Consequently, for a linear kernel the classification vector $w$ and the offset $b$ can be chosen such that, they are partially linear functions depending on the classifier weights.*

This theorem is a side effect of the proofs given in [Chang and Lin, 2001, especially the formulas in lemma 5]. Note that this can be easily extended, e.g., to class dependent weighting. In some cases, the optimization problem might have more than one solution but the solution function can be chosen such that it only includes the choice where the behavior is locally linear. It is not yet clear, if the function can be discontinuous at the finite number of points where it is not locally linear but the example calculation in [Chang and Lin, 2001] indicates, that this is not the case.

## B.3   BRMM Properties

### B.3.1   Sparsity of the 1–norm BRMM

Having the formulation of Method 16 as a linear program, the Simplex algorithm can be applied to deliver an exact solution in a finite number of steps. Since there might be more than one solution, an advantage of the Simplex algorithm is that it prefers solutions with more variables equal to zero as shown in the following proof.

**Theorem 13** (Feature Reduction of 1–norm BRMM). *A solution of the 1–norm BRMM (Method 16) with the Simplex algorithm always uses a number $n_f$ of features smaller than the number of support vectors lying on the four margins. In other words, $n_f$ is smaller than the number of training examples $x_j$ that satisfy*

$$\left\langle w^+ - w^-, x_j \right\rangle + \left( b^+ - b^- \right) \in \{1, -1, R, -R\} . \tag{B.87}$$

*Proof.* Due to the usage of the soft margin, the convex optimization problem always has a solution.

Since we have a solvable linear optimization problem, the Simplex algorithm can be applied. The set of feasible points in this special case is a polytope. The principle of the Simplex algorithm is to take a vertex of this polytope and choose step by step a neighboring one with a higher value of the target function. In the context of the Simplex algorithm these vertices are called basic feasible points. Hence the solution found by the Simplex algorithm is always a vertex of this polytope and it is called a basic feasible solution [Nocedal and Wright, 2006].

As a first step, we introduce the mathematical description of these vertices. This results in a restriction on the number of nonzero variables in the 1–norm BRMM method. The next step is then to analyze the interconnection between the variables and to connect the found restriction with the number of used features of the linear classifier.

Method 16 has a total of $2n$ linear equations which can be formulated as one equation using a matrix multiplication with a matrix $A \in \mathbb{R}^{2n \times (2m+2+3n)}$. The parameter $n$ is the number of given data vectors and $m$ is the dimension of the data space which is also the number of available features.

**Definition 7** (Basic Feasible Point in Method 16). *A basic feasible point*

$$y = (w^+, w^-, b^+, b^-, t, g, h) \tag{B.88}$$

*has only positive components and solves the method equations. Each component of $y$ corresponds to a column of $A$. $y$ can only have nonzero components so that all corresponding columns of $A$ are linearly independent.*

So a maximum of $2n$ out of $2m + 2 + 3n$ components of a basic feasible point can be different from zero because a $2n \times (2m + 2 + 3n)$ matrix can have at most $2n$ linearly independent columns. It can be proven that the basic feasible points from this definition are exactly the vertices of the Simplex algorithms applied on Method 16 [Nocedal and Wright, 2006]. Let $y$ be a basic feasible solution. We already know that a maximum of $2n$ components of $y$ can be nonzero and we now analyze the consequences for the individual parts $w^+$, $w^-$, $b^+$, $b^-$, $t$, $g$, $h$.

We are mainly interested in the classification vector $w = w^+ - w^-$ as the *number of features used* refers to the number of components of $w$ which are different from zero. The above considerations alone deliver $n_f \leq 2n$ as upper boundary for the number of features. To get a more precise boundary, we have to analyze the dependencies between the variables $t_j$, $g_j$, and $h_j$. Hence, for each training example $x_j$ we conduct a case-by-case analysis of the classification function $f(x) = \langle w^+ - w^-, x \rangle + b^+ - b^-$:

$$\begin{array}{lll}
\text{If } |f(x_j)| < 1: & t_j \neq 0 \text{ and } g_j \neq 0. \\
\text{If } |f(x_j)| = 1: & g_j \neq 0. \\
\text{If } 1 < |f(x_j)| < R: & g_j \neq 0 \text{ and } h_j \neq 0. \\
\text{If } |f(x_j)| = R: & h_j \neq 0. \\
\text{If } |f(x_j)| > R: & h_j \neq 0 \text{ and } t_j \neq 0.
\end{array} \qquad \text{(B.89)}$$

For each of the $n$ training samples at least one of the variables $t_j$, $g_j$ and $h_j$ is nonzero. Hence, the upper bound for the number of features drops from $2n$ down to $n$. Additionally, one nonzero component is required in all cases where $|f(x_j)|$ is not equal to $1$ or $R$. The number of these cases can be written as $n - n_{1R}$, where $n_{1R}$ is the number of training samples $x_j$ for which $f(x_j) \in \{1, -1, R, -R\}$. Summing up, the maximal number $2n$ of nonzero components of any basic feasible solution $y$ is composed of

- one component if $b = b^+ - b^-$ is not zero ($1_b$),
- $n$ plus another $n - n_{1R}$ components from the case-by-case analysis,
- and finally the number of used features $n_f$.

Written as an inequality we finally have

$$2n \geq 1_b + (2n - n_{1R}) + n_f \Rightarrow n_{1R} \geq n_f, \qquad \text{(B.90)}$$

as we wanted to prove. $\qquad\qquad\square$

Note that in the special case of $R = 1$ we count each vector on the hyperplane twice, accounting for the fact that these vectors still lie on two planes at the same time in terms of the method. In the case of a 1–norm SVM, the number of used features is restricted by the number of vectors lying on the two hyperplanes with $|f(x)| = 1$. These findings are a direct consequence from Theorem 13 and the connections shown in Section 1.3.

## B.3.2   Extension of Affine Transformation Perspective

Consider a totally different view on binary classification.  We now search for a good classifier together with a good transformation.  Therefore, we want a large soft margin as defined in the SVM method together with a small spread of the data after transformation.  This approach corresponds to the one in [Shivaswamy and Jebara, 2010], but in contrast our classifier is not fixed. The corresponding optimization problem is:

**Definition 8** (Classification Transformation Problem)**.**

$$
\begin{aligned}
\min_{w,b,A,T,R} \quad & \tfrac{1}{2}\left\|w\right\|_2^2 + C\sum_{j=1}^n t_j + BR \\
s.t. \quad & y_j(\langle w, Ax_j + T\rangle + b) \ \geq 1 - t_j \quad \forall j : 1 \leq j \leq n \\
& \tfrac{1}{2}\left\|Ax_j + T\right\|_2^2 \ \leq R^2 \qquad\quad \forall j : 1 \leq j \leq n \\
& t_j \ \geq 0 \qquad\qquad\quad \forall j : 1 \leq j \leq n
\end{aligned}
\tag{B.91}
$$

*where $B$ and $C$ are positive hyperparameters.*

**Lemma 25.** *The classification transformation problem always has a solution.*

*Proof.*  First, the feasible set is not empty because we can set

$$
w = 0, b = 0, A = 0, T = 0, R = 0, t_j = 1 \ \forall j : 1 \leq j \leq n.
\tag{B.92}
$$

Now we can "reduce" the feasibility set by restricting the target function to the value reached by this feasible point:

$$
\frac{1}{2}\left\|w\right\|_2^2 + C\sum_{j=1}^n t_j + BR \leq Cn.
\tag{B.93}
$$

This results in additional restrictions of the variables:

$$
\begin{aligned}
\left\|w\right\|_2 \ & \leq \ \sqrt{2Cn}, & \left\|t\right\|_1 \ & \leq \ n, \\
0 \leq R \ & \leq \ \tfrac{Cn}{B}, & \left\|Ax_j + T\right\|_2 \ & \leq \ \tfrac{2Cn}{B}.
\end{aligned}
\tag{B.94}
$$

The last one can be reformulated and seen as a restriction of $(A\,T)$ on the space which is build by the $x_j$ with an additional last component with the values $1$ (homogeneous space).

$$
\left\|A\,T\right\|_2 \leq \frac{2Cn}{B}\frac{1}{\sqrt{1 + \min\left\|x_j\right\|_2^2}}
\tag{B.95}
$$

Outside this space, we can define $AT$ to be zero without loss of generality. So we showed the affine transformation to be bounded in the space of matrices. It can be seen that the feasible set is closed. So the only remaining unbounded variable is $b$. If it were bounded, too, we could use the existence of minima of continuous functions on compact sets or bounded, closed sets in finite dimensional $\mathbb{R}$-vector space.

Nevertheless, our target function is also bounded below by zero and so we can find a sequence $(w^m, t^m, A^m, T^m, R^m, b^m)_{m \in \mathbb{N}}$ approaching the infimum. By looking only at subsequences,

$$\lim_{m \to \infty} (w^m, t^m, A^m, T^m, R^m) = (w, t, A, T, R) \tag{B.96}$$

can be assumed. If $b$ had no converging subsequence, $\lim_{m \to \infty} b = \infty$ or $-\infty$ holds for the above subsequence. If the classifications problem is not trivial, we can assume $y_1 = 1$, $y_2 = -1$. If $\lim_{m \to \infty} b = \infty$, we can use the inequality

$$-(\langle w^m, A^m x_2 + T^m \rangle + b^m) \geq 1 - t_2^m \tag{B.97}$$

and get the contradiction $-\infty \geq 1 - t_2$. The other case is similar. So our sequence approaching the infimum can be assumed to converge. Because of closure of the feasible set and continuity of the target function, the limit is one solution of the minimization problem. $\qquad \square$

**Lemma 26.** *The classification transformation problem has a solution, with an optimal Matrix $A^* = [\bar{A}\bar{T}]$ with rank one.*

*Proof.* From the previous lemma we already know that there is a solution of the problem. We call it $(w^0, b^0, t^0, A^0, T^0, R^0)$.

Assuming $w^0 \neq 0$ without loss of generality, we can find an orthonormal base and thereby an orthonormal transformation $O$ which maps $w^0$ to $\|w^0\|_2 * e_1$. This results in the same transformation as in the affine transformation problem. Now we use this base transformation to transform the problem. We have

$$\begin{aligned} & y_j(\langle w, Ax_j + T \rangle + b) \\ =\ & y_j(\langle Ow, (OAO^T)(Ox_j) + OT \rangle + b) \quad \forall j : 1 \leq j \leq n \\ & \tfrac{1}{2} \|Ax_j + T\|_2^2 \\ =\ & \tfrac{1}{2} \left\| (OAO^T)(Ox_j) + OT \right\|_2^2 \qquad\qquad \forall j : 1 \leq j \leq n. \end{aligned} \tag{B.98}$$

So by fixing the optimal $w_0$ and by using the new orthonormal base we get a subprob-lem where each partial solution is still one of the previous problem:

$$\min_{b,t,\bar{A},\bar{T},R} \quad \frac{1}{2}\left\|w^0\right\|_2^2 + C\sum t_j + BR$$
$$\text{s.t.} \quad y_j(\left\|w^0\right\|_2 (\bar{a}_1^T \bar{x}_j + \bar{T}_1) + b) \quad \geq 1 - t_j \quad \forall j : 1 \leq j \leq n$$
$$\frac{1}{2}\sum_{i=1}^{n}(\bar{a}_i^T \bar{x}_j + \bar{T}_i)^2 \quad \leq R^2 \qquad \forall j : 1 \leq j \leq n \tag{B.99}$$
$$t_j \quad \geq 0 \qquad \forall j : 1 \leq j \leq n$$

where $\bar{a}_i$ is the $i_{th}$ row of $\bar{A}$. The bar $(\bar{\cdot})$ stands for components in the representation of the new base. Since we are trying to minimize $R$, the sum in the second inequality has to be minimal. Furthermore, $a_j$ is irrelevant for the rest of the program $\forall j \neq 1$. So we can set $(\bar{a}_i^0, \bar{t}_i^0) = 0 \; \forall i \neq 1$ and we have a rank one matrix after retransformation. This matrix is still optimal in the original problem because the change has no effect on the target function.

After demonstrating that an optimal $A$ can be chosen with rank one, we can reduce the original problem or look at a subproblem and we get a program with no fixed variables, but still we use the transformation defined by $w^0$:

$$\min_{w,b,t,\bar{A},\bar{T},R} \quad \frac{1}{2}\left\|w\right\|_2^2 + C\sum t_j + BR$$
$$\text{s.t.} \quad y_j\left(\frac{\langle w,w^0\rangle}{\left\|w^0\right\|_2}(\bar{a}_1^T \bar{x}_j + \bar{T}_1) + b\right) \quad \geq 1 - t_j \quad \forall j : 1 \leq j \leq n$$
$$\frac{1}{2}(\bar{a}_1^T \bar{x}_j + \bar{T}_1)^2 \quad \leq R^2 \qquad \forall j : 1 \leq j \leq n \tag{B.100}$$
$$t_j \quad \geq 0 \qquad \forall j : 1 \leq j \leq n.$$

This method looks similar to the Relative Margin Machine formulation.  □

### B.3.3 Implementation of the BRMM with 2–norm regularization

In the following we will give further details on the calculations, which lead to the algorithm formulas given in Section 1.3.4.1. Therefore, we use the dual problem formulations from Appendix B.1.5.

For implementing a solution algorithm, in the n-th step, all except one index $j$ are kept fixed in dual and for this index the optimal $\alpha_j^{n+1}$ and $\beta_j^{n+1}$ are determined. Let

$f_1$ and $f_2$ be the target functions of the dual problems. Now we define

$$g_1(d) = f_1(\alpha + de_j, \beta) = \frac{d^2}{2} Q_{jj} + d \left( Q_{j.} (\alpha - \beta) - 1 \right) + \mathbf{c} \tag{B.101}$$

$$g_2(d) = f_2(\alpha + de_j, \beta) = \frac{d^2}{2} \left( Q_{jj} + \frac{1}{2C_j} \right) + d \left( Q_{j.} (\alpha - \beta) - 1 + \frac{\alpha_j}{2C_j} \right) + \mathbf{c} \tag{B.102}$$

$$h_1(d) = f_1(\alpha, \beta + de_j) = \frac{d^2}{2} Q_{jj} + d \left( R_j - Q_{j.} (\alpha - \beta) \right) + \mathbf{c'} \tag{B.103}$$

$$h_2(d) = f_2(\alpha, \beta + de_j) = \frac{d^2}{2} \left( Q_{jj} + \frac{1}{2C'_j} \right) + d \left( R_j - Q_{j.} (\alpha - \beta) + \frac{\beta_j}{2C'_j} \right) + \mathbf{c'} \tag{B.104}$$

with respective constants $c$ and $c'$. The remaining step is to calculate the optimal $d$, a case by case analysis concerning the boundaries, and plugging together the solution formula with the boundary constraints to get formulas for $\alpha_j^{n+1}$ and $\beta_j^{n+1}$ depending on $\alpha_j^n$ and $\beta_j^n$.

**Solvability and Constraint Qualifications**

The following argument has the same structure as the proof in Section 1.1.1.1 but is now for the BRMM instead of the C-SVM.

For using duality theory, two requirements have to be checked, which we will do now on the concrete primal problem. First, it has to be proven that there is a solution, because applying duality theory requires an optimal point. Second a constraint qualification has to hold, such that a local linearization of the problem is possible, which is the basic concept of duality theory.

The two target functions (with $q \in \{1, 2\}$) are defined as:

$$f'_q(w, b, t, s) = \frac{1}{2} \|w\|_2^2 + \frac{H}{2} b^2 + \sum C_j t_j^q + \sum C'_j s_j^q. \tag{B.105}$$

First some important observations:

- The constraints are linear.
- $f'_q$ are convex, continuous functions.
- $\Rightarrow$ The BRMM model is defined by a convex optimization problem.
- With $e$ being a vector of ones, the point $p = (\vec{0}, 0, 2e, 2e)$ is a feasible point (fulfilling all constraints) with $u := f'_q(p) = 2^q \sum \left( C_j + C'_j \right)$.
- $\Rightarrow$ An upper bound of the optimal value of the optimization problem is $u$.
- $p$ is a Slater point, meaning that it fulfills the restrictions without equality.

Since $p$ is a Slater point of a convex optimization problem, the Slater condition is fulfilled, which is a constraint qualification. So it remains to show the existence of a solution. With the help of the upper bound $u$ we can infer further restrictions for

optimal points:

$$\|w\|_2 \leq \sqrt[2]{2u}, \; |b| \leq \sqrt[2]{\frac{2u}{H}}, \; \|t\|_q \leq \sqrt[q]{\frac{u}{\min\limits_{1 \leq j \leq n} C_j}}, \; \|s\|_q \leq \sqrt[q]{\frac{u}{\min\limits_{1 \leq j \leq n} C_j'}}. \tag{B.106}$$

Together with the normal constraints of the model, these restrictions define a compact set. Since $f_q'$ is a continuous function, it has a minimum on this set. Hence, a solution exists.

For the proof of the existence of a solution, it is very useful, that $b$ is part of the target function. Otherwise, one has to work with sequences and subsequence approaching the infimum, which exists, because $f_q'$ is bounded below by zero. Assuming, that there is no minimum, results in a subsequence with converging components except of $b^n$ going to plus or minus infinity, one gets a contradiction when taking the limits of the constraints of one example for each class:

$$\lim_{n \to \infty} y_j(\langle w^n, x_j \rangle + b^n) = y_j \lim_{n \to \infty} b^n \geq 1 - \lim_{n \to \infty} t^n. \tag{B.107}$$

This results in $\infty \geq const.$ for one class and $-\infty \geq const.$ for the other class, which is a contradiction to the assumed divergence on $b^n$.

### B.3.4   $\nu$-Balanced Relative Margin Machine

The $\nu$-BRMM was derived from the $\nu$-SVR by a sign/variable shifting (between $t_j$ and $s_j$) if $y_j = -1$:

$$\begin{aligned}
\min_{w,b,\epsilon,s,t} \quad & \tfrac{1}{2}\|w\|_2^2 + C\left(n\nu\epsilon + \sum s_j + \sum t_j\right) \\
\text{s.t.} \quad & \epsilon + s_j \geq y_j(\langle w, x_j \rangle + b) - 1 \;\; \geq -\epsilon - t_j \quad \forall j : 1 \leq j \leq n \\
& s_j, t_j \;\; \geq 0 \qquad\qquad\qquad\qquad \forall j : 1 \leq j \leq n.
\end{aligned} \tag{B.108}$$

This model is always feasible and fulfills Slater's constraint qualification. The proof is similar to the C-SVM. Consequently, it is possible to derive optimality conditions and to work with the dual optimization problem. Formulating, the respective Lagrange function and calculating the derivative leads to:

$$L(w, b, \epsilon, s, t, \alpha, \beta, \gamma, \delta) = \frac{1}{2} \|w\|_2^2 + C \left( n\nu\epsilon + \sum t_j + \sum s_j \right) \tag{B.109}$$

$$+ \sum \alpha_j \left( 1 - t_j - \epsilon - y_j(\langle w, x_j \rangle + b) \right) - \sum \gamma_j t_j \tag{B.110}$$

$$+ \sum \beta_j \left( y_j(\langle w, x_j \rangle + b) - 1 - \epsilon - s_j \right) - \sum \delta_j s_j \tag{B.111}$$

$$\frac{\partial L}{\partial w} = w - \sum (\alpha_j - \beta_j) y_j x_j \tag{B.112}$$

$$\frac{\partial L}{\partial b} = - \sum y_j (\alpha_j - \beta_j) \tag{B.113}$$

$$\frac{\partial L}{\partial \epsilon} = Cn\nu - \sum \alpha_j - \sum \beta_j \tag{B.114}$$

$$\frac{\partial L}{\partial t_j} = C - \alpha_j - \gamma_j \tag{B.115}$$

$$\frac{\partial L}{\partial s_j} = C - \beta_j - \delta_j. \tag{B.116}$$

The dual $\nu$-BRMM reads:

$$\begin{aligned}
\min_{\alpha} \quad & \tfrac{1}{2} \sum (\alpha_i - \beta_i)(\alpha_j - \beta_j) \langle x_i, x_j \rangle y_i y_j - \sum_j (\alpha_j - \beta_j) \\
\text{s.t.} \quad & C \geq \alpha_j \geq 0 \, \forall j : 1 \leq j \leq n, \\
& C \geq \beta_j \geq 0 \, \forall j : 1 \leq j \leq n, \\
& \sum_j \alpha_j y_j = \sum_j \beta_j y_j, \\
& \sum_j \alpha_j + \beta_j = \nu C n \, .
\end{aligned} \tag{B.117}$$

To show, that $\nu$ is a lower border (in percentage) on the number of support vectors it is good to rescale the dual parameters and get an equivalent rescaled dual $\nu$-BRMM problem:

$$\begin{aligned}
\min_{\alpha} \quad & \tfrac{1}{2} \sum (\alpha_i - \beta_i)(\alpha_j - \beta_j) \langle x_i, x_j \rangle y_i y_j - \tfrac{1}{Cn} \sum_j (\alpha_j - \beta_j) \\
\text{s.t.} \quad & \tfrac{1}{n} \geq \alpha_j \geq 0, \forall j : 1 \leq j \leq n, \\
& \tfrac{1}{n} \geq \beta_j \geq 0, \forall j : 1 \leq j \leq n, \\
& \sum_j \alpha_j y_j = \sum_j \beta_j y_j, \\
& \sum_j \alpha_j + \beta_j = \nu \, .
\end{aligned} \tag{B.118}$$

## B.4   Unary Classifier Variants and Implementations

### B.4.1   One-Class Balanced Relative Margin Machine and its Variants

For the following algorithms the decision function reads:

$$f(x) = \operatorname{sgn}\left(\langle w, x \rangle - 2\right) . \tag{B.119}$$

For the range parameter $R$ it holds $R \geq 1$.

**Method 28** (One-Class BRMM).

$$\begin{aligned}
\min_{w,t} \quad & \tfrac{1}{2} \|w\|_2^2 + C \sum t_j \\
\text{s.t.} \quad & 1 + R + t_j \geq \langle w, x_j \rangle \geq 2 - t_j \text{ and } t_j \geq 0 \; \forall j : 1 \leq j \leq n .
\end{aligned} \tag{B.120}$$

**Method 29** (New One-Class SVM ($R = \infty$)).

$$\begin{aligned}
\min_{w,b,t} \quad & \tfrac{1}{2} \|w\|_2^2 + C \sum t_j \\
\text{s.t.} \quad & \langle w, x_j \rangle \geq 2 - t_j \text{ and } t_j \geq 0 \; \forall j : 1 \leq j \leq n .
\end{aligned} \tag{B.121}$$

**Method 30** (One-Class RFDA ($R = 1$)).

$$\min_{w} \tfrac{1}{2} \|w\|_2^2 + C \sum |\langle w, x_j \rangle - 2| . \tag{B.122}$$

**Method 31** (L2–One-Class BRMM).

$$\begin{aligned}
\min_{w,t} \quad & \tfrac{1}{2} \|w\|_2^2 + \tfrac{C}{2} \sum t_j^2 \\
\text{s.t.} \quad & 1 + R + t_j \geq \langle w, x_j \rangle \geq 2 - t_j \; \forall j : 1 \leq j \leq n .
\end{aligned} \tag{B.123}$$

**Method 32** (Hard-Margin One-Class BRMM).

$$\begin{aligned}
\min_{w} \quad & \tfrac{1}{2} \|w\|_2^2 \\
\text{s.t.} \quad & 1 + R \geq \langle w, x_j \rangle \geq 2 \; \forall j : 1 \leq j \leq n .
\end{aligned} \tag{B.124}$$

*For the existence of a solution $R > 1$ is required.  In contrast to all other BRMM methods, this model might have no solution.*

**Method 33** (1–Norm One-Class BRMM).

$$\begin{aligned}
\min_{w,t} \quad & \|w\|_1 + C \sum t_j \\
\text{s.t.} \quad & 1 + R + t_j \geq \langle w, x_j \rangle \geq 2 - t_j \text{ and } t_j \geq 0 \; \forall j : 1 \leq j \leq n .
\end{aligned} \tag{B.125}$$

### B.4.2   Iterative Solution Formulas for One-Class BRMM Variants

This section introduces update formulas using the approaches from Section 1.2.3 and 1.2.4 Let $j$ be the index of the relevant sample for the update in the $k$-th iteration.

**Theorem 27** (Update Formulas for the One-Class BRMM). *With the projection function* $P(z) = \max\{0, \min\{z, C\}\}$, *the update formulas are:*

$$\alpha_j^{(k+1)} = P\left(\alpha_j^{(k)} - \frac{1}{k(x_j, x_j)}\left(-2 + \sum(\alpha_i - \beta_i)k(x_i, x_j)\right)\right) \tag{B.126}$$

$$\beta_j^{(k+1)} = P\left(\beta_j^{(k)} + \frac{1}{k(x_j, x_j)}\left(-(R+1) + \sum(\alpha_i - \beta_i)k(x_i, x_j)\right)\right) \tag{B.127}$$

*and in the linear case:*

$$\alpha_j^{(k+1)} = P\left(\alpha_j^{(k)} - \frac{1}{\|x_j\|_2^2}\left(\left\langle w^{(k)}, x_j\right\rangle - 2\right)\right) \tag{B.128}$$

$$\beta_j^{(k+1)} = P\left(\beta_j^{(k)} + \frac{1}{\|x_j\|_2^2}\left(\left\langle w^{(k)}, x_j\right\rangle - (R+1)\right)\right) \tag{B.129}$$

$$w^{(k+1)} = w^{(k)} + ((\alpha_j^{(k+1)} - \alpha_j^{(k)}) - (\beta_j^{(k+1)} - \beta_j^{(k)}))\, x_j\ . \tag{B.130}$$

*Proof.* With the help of

$$h(\alpha, \beta) = \frac{1}{2}\sum_{i,m}(\alpha_i - \beta_i)(\alpha_m - \beta_m)k(x_i, x_m) - 2\sum\alpha_i + (R+1)\sum\beta_i \tag{B.131}$$

we define $g_1(d) = h(\alpha^{(k)} + de_j, \beta^{(k)})$, $g_2(d) = h(\alpha^{(k)}, \beta^{(k)} + de_j)$ and calculate:

$$\frac{\partial g_1}{\partial d} = d \cdot k(x_j, x_j) + \sum(\alpha_i^{(k)} - \beta_i^{(k)})k(x_i, x_j) - 2\ , \tag{B.132}$$

$$\frac{\partial g_2}{\partial d} = d \cdot k(x_j, x_j) - \sum(\alpha_i^{(k)} - \beta_i^{(k)})k(x_i, x_j) + (R+1)\ . \tag{B.133}$$

If $k(x_j, x_j) = 0$ the index can be ignored and no update is required. With $k(x_j, x_j) > 0$ the optimal $d$ can be determined with $\frac{\partial g_1}{\partial d} = 0$ or $\frac{\partial g_2}{\partial d} = 0$ respectively. With the projection of the resulting solution to the restriction interval $[0, C]$ this gives the update formulas. Replacing $k(x_i, x_j)$ with $\langle x_i, x_j\rangle$ and substituting $w^{(m)} = \sum(\alpha_j - \beta_j)x_j$ results in the formulas for the linear case. $\qquad\square$

**Theorem 28** (Update Formulas for the L2–One-Class BRMM). *With the projection*

*function $P(z) = \max\{0, z\}$, the update formulas are:*

$$\alpha_j^{(k+1)} = P\left(\alpha_j^{(k)} - \frac{1}{k(x_j, x_j) + \frac{1}{C}}\left(-2 + \frac{\alpha_j^{(k)}}{C} + \sum(\alpha_i - \beta_i)k(x_i, x_j)\right)\right) \quad \text{(B.134)}$$

$$\beta_j^{(k+1)} = P\left(\beta_j^{(k)} + \frac{1}{k(x_j, x_j) + \frac{1}{C}}\left(-(R+1) - \frac{\beta_j^{(k)}}{C} + \sum(\alpha_i - \beta_i)k(x_i, x_j)\right)\right) \quad \text{(B.135)}$$

*and in the linear case:*

$$\alpha_j^{(k+1)} = P\left(\alpha_j^{(k)} - \frac{1}{\|x_j\|_2^2 + \frac{1}{C}}\left(\left\langle w^{(k)}, x_j\right\rangle - 2 + \frac{\alpha_j^{(k)}}{C}\right)\right) \quad \text{(B.136)}$$

$$\beta_j^{(k+1)} = P\left(\beta_j^{(k)} + \frac{1}{\|x_j\|_2^2 + \frac{1}{C}}\left(\left\langle w^{(k)}, x_j\right\rangle - (R+1) - \frac{\beta_j^{(k)}}{C}\right)\right) \quad \text{(B.137)}$$

$$w^{(k+1)} = w^{(k)} + ((\alpha_j^{(k+1)} - \alpha_j^{(k)}) - (\beta_j^{(k+1)} - \beta_j^{(k)}))\, x_j\,. \quad \text{(B.138)}$$

*Proof.* With the help of

$$h(\alpha, \beta) = \frac{1}{2}\sum_{i,m}(\alpha_i - \beta_i)(\alpha_m - \beta_m)k(x_i, x_m) - 2\sum\alpha_i + (R+1)\sum\beta_i + \sum\frac{\alpha_i^2 + \beta_i^2}{2C} \quad \text{(B.139)}$$

we define $g_1(d) = h(\alpha^{(k)} + de_j, \beta^{(k)})$, $g_2(d) = h(\alpha^{(k)}, \beta^{(k)} + de_j)$ and calculate:

$$\frac{\partial g_1}{\partial d} = d\left(k(x_j, x_j) + \frac{1}{C}\right) + \sum(\alpha_i^{(k)} - \beta_i^{(k)})k(x_i, x_j) - 2 + \frac{\alpha_j}{C}\,, \quad \text{(B.140)}$$

$$\frac{\partial g_2}{\partial d} = d\left(k(x_j, x_j) + \frac{1}{C}\right) - \sum(\alpha_i^{(k)} - \beta_i^{(k)})k(x_i, x_j) + (R+1) + \frac{\beta_j}{C}\,. \quad \text{(B.141)}$$

If $k(x_j, x_j) = 0$ the index can be ignored and no update is required. With $k(x_j, x_j) > 0$ the optimal $d$ can be determined with $\frac{\partial g_1}{\partial d} = 0$ or $\frac{\partial g_2}{\partial d} = 0$ respectively. With the projection of the resulting solution to the restriction interval $[0, \infty)$ this gives the update formulas. Replacing $k(x_i, x_j)$ with $\langle x_i, x_j\rangle$ and substituting $w^{(m)} = \sum(\alpha_j - \beta_j)x_j$ results in the formulas for the linear case. $\qquad\square$

**Theorem 29** (Update Formulas for the Hard-Margin One-Class BRMM). *With the projection function $P(z) = \max\{0, z\}$, the update formulas are:*

$$\alpha_j^{(k+1)} = P\left(\alpha_j^{(k)} - \frac{1}{k(x_j, x_j)}\left(-2 + \sum(\alpha_i - \beta_i)k(x_i, x_j)\right)\right) \quad \text{(B.142)}$$

$$\beta_j^{(k+1)} = P\left(\beta_j^{(k)} + \frac{1}{k(x_j, x_j)}\left(-(R+1) + \sum(\alpha_i - \beta_i)k(x_i, x_j)\right)\right) \quad \text{(B.143)}$$

*and in the linear case:*

$$\alpha_j^{(k+1)} = P\left(\alpha_j^{(k)} - \frac{1}{\|x_j\|_2^2}\left(\left\langle w^{(k)}, x_j \right\rangle - 2\right)\right) \tag{B.144}$$

$$\beta_j^{(k+1)} = P\left(\beta_j^{(k)} + \frac{1}{\|x_j\|_2^2}\left(\left\langle w^{(k)}, x_j \right\rangle - (R+1)\right)\right) \tag{B.145}$$

$$w^{(k+1)} = w^{(k)} + ((\alpha_j^{(k+1)} - \alpha_j^{(k)}) - (\beta_j^{(k+1)} - \beta_j^{(k)}))\, x_j\,. \tag{B.146}$$

*These formulas are the same as for the One-Class BRMM but with a different projection.*

*Proof.* The proof is the same as for the One-Class BRMM but the final projection is different because, there is no upper boundary on the variables. □

### B.4.3 Online One-Class BRMM Variants

According to the origin separation approach in Section 1.2.4, deriving the update formulas is straightforward. With a new incoming sample $x_j$, the respective weights are initialized with zero, $w$ is updated and afterwards, the update weights are not needed any longer. $w$ is usually initialized with zeros, but it be also can also randomly initialized or with a vector from a different dataset.

**Method 34** (Online One-Class BRMM).

$$\begin{aligned}
\alpha &= \max\left\{0, \min\left\{\frac{1}{\|x_j\|_2^2}\left(2 - \left\langle w^{(j)}, x_j \right\rangle\right), C\right\}\right\} \\
\beta &= \max\left\{0, \min\left\{\frac{1}{\|x_j\|_2^2}\left(\left\langle w^{(j)}, x_j \right\rangle - (R+1)\right), C\right\}\right\} \\
w^{(j+1)} &= w^{(j)} + (\alpha - \beta)\, x_j\,.
\end{aligned} \tag{B.147}$$

**Method 35** (Online L2–One-Class BRMM).

$$\begin{aligned}
\alpha &= \max\left\{0, \frac{1}{\|x_j\|_2^2 + \frac{1}{C}}\left(2 - \left\langle w^{(j)}, x_j \right\rangle\right)\right\} \\
\beta &= \max\left\{0, \frac{1}{\|x_j\|_2^2 + \frac{1}{C}}\left(\left\langle w^{(j)}, x_j \right\rangle - (R+1)\right)\right\} \\
w^{(j+1)} &= w^{(j)} + (\alpha - \beta)\, x_j\,.
\end{aligned} \tag{B.148}$$

**Method 36** (Online Hard-Margin One-Class BRMM).

$$\begin{aligned}
\alpha &= \max\left\{0, \frac{1}{\|x_j\|_2^2}\left(2 - \left\langle w^{(j)}, x_j \right\rangle\right)\right\} \\
\beta &= \max\left\{0, \frac{1}{\|x_j\|_2^2}\left(\left\langle w^{(j)}, x_j \right\rangle - (R+1)\right)\right\} \\
w^{(j+1)} &= w^{(j)} + (\alpha - \beta)\, x_j\,.
\end{aligned} \tag{B.149}$$

To get the respective SVM perceptrons, $R = \infty$ has to be used, which results in $\beta = 0$ in all cases.

**Method 37** (Online One-Class SVM).

$$w^{(j+1)} = w^{(j)} + \max\left\{0, \min\left\{\frac{1}{\|x_j\|_2^2}\left(2 - \left\langle w^{(j)}, x_j\right\rangle\right), C\right\}\right\} x_j \,. \qquad \text{(B.150)}$$

**Method 38** (Online L2–One-Class SVM).

$$w^{(j+1)} = w^{(j)} + \max\left\{0, \frac{1}{\|x_j\|_2^2 + \frac{1}{C}}\left(2 - \left\langle w^{(j)}, x_j\right\rangle\right)\right\} x_j \,. \qquad \text{(B.151)}$$

**Method 39** (Online Hard-Margin One-Class SVM).

$$w^{(j+1)} = w^{(j)} + \max\left\{0, \frac{1}{\|x_j\|_2^2}\left(2 - \left\langle w^{(j)}, x_j\right\rangle\right)\right\} x_j \,. \qquad \text{(B.152)}$$

For completeness, we also give the reduced formulas for the RFDA variants ($R = 1$), except, for the hard margin case, where no solution exists.

**Method 40** (Online One-Class RFDA).

$$w^{(j+1)} = w^{(j)} + \max\left\{-C, \min\left\{\frac{1}{\|x_j\|_2^2}\left(2 - \left\langle w^{(j)}, x_j\right\rangle\right), C\right\}\right\} x_j \,. \qquad \text{(B.153)}$$

**Method 41** (Online L2–One-Class RFDA).

$$w^{(j+1)} = w^{(j)} + \frac{1}{\|x_j\|_2^2 + \frac{1}{C}}\left(2 - \left\langle w^{(j)}, x_j\right\rangle\right) x_j \,. \qquad \text{(B.154)}$$

# B.5 Positive Upper Boundary Support Vector Estimation

This section is joint work with Alexander Fabisch and is based on:
Fabisch, A., Metzen, J. H., Krell, M. M., and Kirchner, F. (2015). Accounting for Task-Hardness in Active Multi-Task Robot Control Learning. *Künstliche Intelligenz*.
My contribution to this paper is the PUBSVE algorithm and the respective formulas for the implementation after a request by Alexander Fabisch.

This section presents the SVR variant PUBSVE. It is related to this thesis due to its relation to SVM, and its special offset treatment.

We are given a set of observations $\mathcal{D} = \{(x_j, y_j)\}_{j=1}^n$ and assume that the $y_j$ depend on the $x_j$ via $y_j = f(x_j) - e_j$, where $e_j$ is some noise term. In contrast to standard regression problems, we assume $e_j \geq 0$, i.e., we always observe values $y_j$ which are less or equal than the true function value $f(x_j)$. This model is appropriate for instance in reinforcement learning when $f(x_j)$ returns the maximal reward possible in a context $x_j$, and $y_j$ is the actual reward obtained by a learning agent, which often makes suboptimal decisions.

We are now interested in inferring the function $f$ from observations $\mathcal{D}$, i.e., learn an estimate $\hat{f}$ of $f$. One natural constraint on the estimate is that $\hat{f}(x_j) \geq y_j$, i.e., $\hat{f}$ shall be an *upper boundary* on $\mathcal{D}$. Assuming positive values, the goal is to have a low $b$ and to keep the boundary as tight as possible but also to generalize well on unseen data. This can be achieved by a regularization:

**Method 42** (Positive Upper Boundary Support Vector Estimation (PUBSVE)).

$$\min_{w,b} \quad \frac{1}{2} \|w\|_2^2 + \frac{H}{2}b^2 + C \sum t_j^q$$
$$\text{s.t.} \quad \langle w, x_j \rangle + b \geq y_j - t_j \text{ and } t_j \geq 0 \ \forall j : 1 \leq j \leq n \ . \tag{B.155}$$

$H$ is a special hyperparameter, to weight between a simple maximum using the offset $b$ or having a real curve fitted.[1] The error toleration constant $C$ should be chosen to be infinity to enforce a hard margin. It was just used here, to give a more general model and make the resemblance between our error handling and the hinge loss clear (Tabular 1.1). In this case, $q \in \{1, 2\}$ was used to also allow for squared loss.[2] The $y_j$ need to be normalized (e.g., by subtracting $\min_{j'} y_{j'}$), such that a positive value of $b$ can be expected because otherwise, $\hat{f}(x) \equiv 0$ would be the solution of our suggested model. The same approach could be used, to estimate a negative lower boundary by multiplying the $y_j$ and the resulting final boundary function $f$ from the PUBSVE with $-1$. The introduction of nonlinear kernels and sparse regularization, and the implementation is straightforward (see also Chapter 1). We typically use a non-parametric, kernelized model for $\hat{f}$, e.g., $\hat{f}(x) = b + \sum_{i=1}^{n} \alpha_j k(x_j, x)$ with RBF kernel $k$ and offset $b$ because it provides an arbitrary tight boundary and usually a linear model is not appropriate.

---

[1] Usually $H$ should be chosen high for real curve fitting.
[2] If $q = 2$, the constraint $t_j \geq 0$ can be omitted.

Thanks to the offset in the target function, the special offset treatment approach can be used for implementation as outlined in the following. First the dual optimization problems are derived.

$$L_1(w, b, t, \alpha, \gamma) = \frac{1}{2}\|w\|_2^2 + \frac{H}{2}b^2 + C\sum t_j + \sum \alpha_j\left(y_j - t_j - b - \langle w, x_j\rangle\right) - \sum \gamma_j t_j$$
$$\text{(B.156)}$$

$$L_2(w, b, t, \alpha) = \frac{1}{2}\|w\|_2^2 + \frac{H}{2}b^2 + C\sum t_j^2 + \sum \alpha_j\left(y_j - t_j - b - \langle w, x_j\rangle\right) \quad \text{(B.157)}$$

$$\frac{\partial L_q}{\partial w} = w - \sum \alpha_j x_j \Rightarrow w^{\text{opt}} = \sum \alpha_j x_j \quad \text{(B.158)}$$

$$\frac{\partial L_q}{\partial b} = Hb - \sum \alpha_j \Rightarrow b^{\text{opt}} = \frac{1}{H}\sum \alpha_j \quad \text{(B.159)}$$

$$\frac{\partial L_1}{\partial t_j} = C - \alpha_j - \gamma_j \Rightarrow 0 \le \alpha_j \le C \quad \text{(B.160)}$$

$$\frac{\partial L_2}{\partial t_j} = 2Ct_j - \alpha_j \Rightarrow t_j^{\text{opt}} = \frac{\alpha_j}{2C} \quad \text{(B.161)}$$

Consequently the dual L1–PUBSVE reads:

$$\min_{\alpha : 0 \le \alpha_j \le C \forall j} \frac{1}{2}\sum_{i,j} \alpha_i \alpha_j \langle x_i, x_j\rangle + \frac{1}{2H}\left(\sum_j \alpha_j\right)^2 - \sum_j \alpha_j y_j \quad \text{(B.162)}$$

with the respective update formula after introducing the kernel function $k$:

$$\alpha_j^{\text{new}} = \max\left\{0, \min\left\{\alpha_j^{\text{old}} - \frac{1}{k(x_j, x_j) + \frac{1}{H}}\left(-y_j + \sum_i \alpha_i^{\text{old}} k(x_i, x_j) + \frac{1}{H}\sum_i \alpha_i\right), C\right\}\right\}.$$
$$\text{(B.163)}$$

For the hard margin case, set $C = \infty$. The dual L2–PUBSVE reads:

$$\min_{\alpha : 0 \le \alpha_j \forall j} \frac{1}{2}\sum_{i,j} \alpha_i \alpha_j \langle x_i, x_j\rangle + \frac{1}{2H}\left(\sum_j \alpha_j\right)^2 - \sum_j \alpha_j y_j + \frac{1}{4C}\sum_j \alpha_j^2 \quad \text{(B.164)}$$

with the update formula:

$$\alpha_j^{\text{new}} = \max\left\{0, \alpha_j^{\text{old}} - \frac{1}{k(x_j, x_j) + \frac{1}{2C} + \frac{1}{H}}\left(\frac{\alpha_j^{\text{old}}}{2C} - y_j + \sum_i \alpha_i^{\text{old}} k(x_i, x_j) + \frac{1}{H}\sum_i \alpha_i^{\text{old}}\right)\right\}.$$
$$\text{(B.165)}$$

The target function is:

$$f(x) = \sum_i \alpha_i k(x_i, x) + b \text{ with } b = \frac{1}{H} \sum_i \alpha_i. \tag{B.166}$$

To reduce the training time and memory usage of the PUBSVE significantly, instead of training the PUBSVE on the whole set of observed pairs we can update the boundaries incrementally after each update [Syed et al., 1999]: we forget every example except the support vectors $x_i$ and the corresponding weights ($\alpha_i > 0$), collect new samples and use the new samples and the support vectors to train the model of the upper and lower boundaries. The result is illustrated in Figure B.1, where the samples are drawn from uniform random distributions with $x \in [0, 1)$ and $y$ lies between the boundaries that are marked by the gray areas. We use this method to reduce the computational complexity at the cost of a slightly higher error because some previous examples that are close the estimated boundary but are not support vectors might be outside of the boundaries after another iteration.

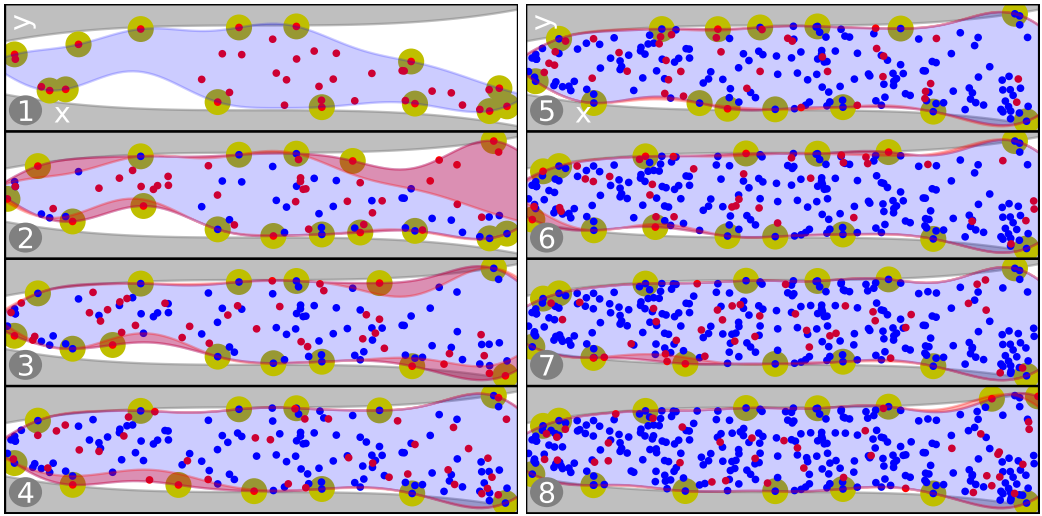

Figure B.1: **Visualization of incremental learning with PUBSVE.** 8 iterations of the incremental learning of upper and lower boundaries: for each update of the PUBSVE we take the new samples (small red dots) and the support vectors (large yellow dots) from the previous iteration as a training set. The area between the upper and the lower boundary is blue and the area that has been added in comparison to the previous iteration is red. All previous samples that will not be used for the incremental training are displayed as small blue dots. The true boundaries are marked by the gray areas.

# Appendix C

# Configuration Files

```
type: node_chain
input_path: MNIST
parameter_ranges:
  __classifier__: [LibsvmOneClass, OcRmm]
  __label__: ['0', '1', '2', '3', '4',
              '5', '6', '7', '8', '9']
  __lc__: [2, 1.5, 1, 0.5, 0, −0.5, −1, −1.5,
           −2, −2.5, −3, −3.5, −4]
node_chain:
  − node: PCASklearn
    parameters:
      n_components : 40
  − node: EuclideanFeatureNormalization
  − node: __classifier__
    parameters:
      class_labels: [__label__, REST]
      complexity: eval(10**__lc__)
      max_iterations: 100000
      nu: eval(((−__lc__+2.1)/6.2)**(1.5))
      random: true
      tolerance: eval(min(0.001*10**__lc__,0.01))
  − node: PerformanceSink
    parameters: {ir_class: __label__, sec_class: REST}
```

Figure C.1: **Operation specification file for the comparison of new one-class SVM ("OcRmm" with range=∞) and $\nu$oc-SVM ("LibsvmOneClass") on MNIST data (Section 1.4.5.2).**

```
type: node_chain
input_path: MNIST
parameter_ranges:
    __classifier__: [OcRmm, OcRmmPerceptron, 2RMM,
                     UnaryPA0, UnaryPA1]
    __label__: ['0','1','2','3','4','5','6','7','8','9']
    __lr__: eval(range(4,21))
node_chain:
    - node: PCASklearn
      parameters:
          n_components : 40
    - node: EuclideanFeatureNormalization
    - node: Grid_Search
      parameters:
          evaluation:
              metric: AUC
              performance_sink_node:
                  node: PerformanceSink
                  parameters:
                      calc_AUC: true
                      ir_class: __label__
                      sec_class: REST
          nodes:
          - node: __classifier__
            parameters:
                class_labels: [__label__, REST]
                complexity: eval(10**~~lc~~)
                max_iterations: 100
                radius: eval((__lr__)/10.0)
                range: eval(__lr__/4.0)
                tolerance: eval(0.001*10**~~lc~~)
          optimization:
              ranges:
              ~~lc~~: eval([-5.0+.5*i for i in range(15)])
          parallelization: {processing_modality: backend}
          validation_set:
              split_node:
                  node: CV_Splitter
                  parameters: {splits: 5, stratified: true}
    - node: PerformanceSink
      parameters: {ir_class: __label__, sec_class: REST}
```

Figure C.2: **Operation specification file for unary classifier comparison on MNIST data (Section 1.4.5.1).**

```
type : node_chain
input_path : P300_Data_Preprocessed_InterSession
store_node_chain : True
node_chain :
    − node : Time_Series_Source
    − node : xDAWN
      parameters :
                erp_class_label : "Target"
                retained_channels : 8
    − node : TDF
    − node : O_FN
    − node : NilSink
```

Figure C.3: **Operation specification file for storing preprocessing flows (Section 2.4.6).**

```
type: node_chain
input_path : P300_Data_Preprocessed_InterSession
parameter_ranges:
   __backtransformation__: [with, without]
   __co_adapt__: [false, double]
   __log_dist__: [1, 1.25, 1.5, 1.75, 2, 2.25, 2.5,
                  2.75, 3, 3.25, 3.5, 3.75, 4]
runs: 10
constraints:
   [("__backtransformation__"=="with" and "__co_adapt__"=="double")
   or ("__backtransformation__"=="without" and "__co_adapt__"=="False")]
node_chain:
   − node: Time_Series_Source
   − node: InstanceSelection
     parameters:
        reduce_class: false
        test_percentage_selected: 100
        train_percentage_selected: 0
   − node: Noop
     parameters: {keep_in_history: True}
   − node: RandomFlowNode
     parameters:
        dataset: __INPUT_DATASET__
        distance: eval(10**__log_dist__)
        flow_base_dir: result_folder_from_stored_preprocessing
        retrain: True
   − node: RmmPerceptron
     parameters:
        class_labels: [Target, Standard]
        co_adaptive: __co_adapt__
        co_adaptive_index: 2
        complexity: 1
        history_index: 1
        range: 100
        retrain: True
        weight: [5.0, 1.0]
        zero_training: True
   − node: Classification_Performance_Sink
     parameters: {ir_class: Standard, save_trace: True}
```

Figure C.4: **Operation specification file for reinitialization after changing the preprocssing (Section 2.4.6).**

```python
#!/usr/bin/python
# http://scikit-learn.org/stable/auto_examples/plot_classifier_comparison.html
"""Reduced script version by Mario Michael Krell taken from scikit-learn."""
# Code source: Gaël Varoquaux
#              Andreas Müller
# Modified for documentation by Jaques Grobler
# License: BSD 3 clause
import numpy as np
import matplotlib.pyplot as plt
from matplotlib.colors import ListedColormap
from sklearn.cross_validation import train_test_split
from sklearn.preprocessing import StandardScaler
from sklearn.datasets import make_moons, make_circles, make_classification
from sklearn.svm import SVC
from sklearn.lda import LDA
from sklearn.linear_model import PassiveAggressiveClassifier
h = .02  # step size in the mesh
names = ["Linear_SVM", "RBF_SVM", "Polynomial_SVM", "PA1", "FDA"]
classifiers = [SVC(kernel="linear", C=0.025), SVC(gamma=2, C=1),
               SVC(kernel="poly", degree=2, C=10),
               PassiveAggressiveClassifier(n_iter=1), LDA()]
X, y = make_classification(n_features=2, n_redundant=0, n_informative=2,
                           random_state=1, n_clusters_per_class=1)
rng = np.random.RandomState(2)
X += 2 * rng.uniform(size=X.shape)
linearly_separable = (X, y)
datasets = [make_moons(noise=0.3, random_state=0),
            make_circles(noise=0.2, factor=0.5, random_state=1),
            linearly_separable]
figure = plt.figure(figsize=(3*len(classifiers), 3*len(datasets)))
i = 1
for ds in datasets:
    X, y = ds
    X = StandardScaler().fit_transform(X)
    X_train, X_test, y_train, y_test = train_test_split(X, y, test_size=.4)
    x_min, x_max = X[:, 0].min() - .5, X[:, 0].max() + .5
    y_min, y_max = X[:, 1].min() - .5, X[:, 1].max() + .5
    xx, yy = np.meshgrid(np.arange(x_min, x_max, h),
                         np.arange(y_min, y_max, h))
    cm = plt.cm.RdBu
    cm_bright = ListedColormap(['#FF0000', '#0000FF'])
    ax = plt.subplot(len(datasets), len(classifiers) + 1, i)
    ax.scatter(X_train[:, 0], X_train[:, 1], c=y_train, cmap=cm_bright)
    ax.set_xlim(xx.min(), xx.max())
    ax.set_ylim(yy.min(), yy.max())
    ax.set_xticks(())
    ax.set_yticks(())
    i += 1
    for name, clf in zip(names, classifiers):
        ax = plt.subplot(len(datasets), len(classifiers) + 1, i)
        clf.fit(X_train, y_train)
        score = clf.score(X_test, y_test)
        if hasattr(clf, "decision_function"):
            Z = clf.decision_function(np.c_[xx.ravel(), yy.ravel()])
        else:
            Z = clf.predict_proba(np.c_[xx.ravel(), yy.ravel()])[:, 1]
        Z = Z.reshape(xx.shape)
        ax.contourf(xx, yy, Z, cmap=cm, alpha=.8)
        ax.scatter(X_train[:, 0], X_train[:, 1], c=y_train, cmap=cm_bright)
        ax.set_xlim(xx.min(), xx.max())
        ax.set_ylim(yy.min(), yy.max())
        ax.set_xticks(())
        ax.set_yticks(())
        ax.set_title(name)
        i += 1
figure.subplots_adjust(left=.02, right=.98)
figure.savefig("scikit_classifier_vis_pp.png", dpi=300, bbox_inches='tight')
```

Figure C.5: **Scikit-learn script for classifier visualization (Figure 2.1).**

Figure C.6: **Electrode positions** of a 128 channel electrode cap taken from www.brainproducts.com. For a 64 channel cap, the pink and yellow colored electrodes are not used.

# Acronyms

$\nu$**-SVM**  $\nu$ support vector machine — Section 1.1.1.3

$\nu$**oc-SVM**  classical one-class support vector machine — Section 1.1.6.3

**C-SVM**  classical support vector machine — Section 1.1

**AUC**  area under the ROC curve [Bradley, 1997]

**BA**  balanced accuracy — Figure 3.5

**BCI**  brain-computer interface

**BRMM**  balanced relative margin machine — Section 1.3.2

**CPU**  central processing unit

**CSP**  common spatial patterns [Blankertz et al., 2008]

**DSL**  domain-specific language

**EEG**  electroencephalogram

**EMG**  electromyogram

**ERP**  event-related potential

**FDA**  Fisher's discriminant — Section 1.1.3

**fMRI**  functional magnetic resonance imaging

**GUI**  graphical user interface

**ICA**  independent component analysis [Jutten and Herault, 1991, Hyvärinen, 1999, Rivet et al., 2009]

**LS-SVM**  least squares support vector machine — Section 1.1.2

**MDP**  modular toolkit for data processing [Zito et al., 2008]

**MEG**  magnetoencephalography

**MPI**  message passing interface

**PAA**  passive-aggressive algorithm— Section 1.1.5

**PCA**  principal component analysis [Lagerlund et al., 1997, Rivet et al., 2009, Abdi and Williams, 2010]

**PUBSVE**  positive upper boundary support vector estimation — Appendix B.5

**pySPACE**  Signal Processing And Classification Environment written in Python

**RBF**  radial basis function

**RFDA**  regularized Fisher's discriminant — Section 1.1.3

**RHKS**  reproducing kernel Hilbert space

**RMM**  relative margin machine — Section 1.1.4

**ROC**  receiver operating characteristic [Green and Swets, 1988, Macmillan and Creelman, 2005]

**SLAM**  simultaneous localization and mapping

**SMO**  sequential minimal optimization — Section 1.2.2

**SVDD**  support vector data description — Section 1.1.6.1

**SVM**  support vector machine — Section 1.1

**SVR**  support vector regression — Section 1.1.1.4

**YAML**  YAML Ain't Markup Language [Ben-Kiki et al., 2008]

# Symbols

| | |
|---|---|
| $b$ | offset/bias of the classification function $f$ |
| $C$ | regularization parameter of the C-SVM and its variants, also called cost parameter or complexity |
| conv | convex hull |
| $e_i$ | i-th unit vector |
| exp | exponential function |
| $f$ | classification function |
| $H_z$ | $:= \{x \in \mathbb{R}^n \,|\, \langle w, x \rangle + b = z\}$ hyper plane |
| $k$ | kernel function to replace the scalar product in the algorithm model |
| $m$ | dimensionality of the data |
| $n$ | number of samples |
| $\|.\|_p$ | p-norm |
| $\langle ., . \rangle$ | scalar product |
| $\mathrm{sgn}(t)$ | $:= \begin{cases} +1 & \text{if } t > 0, \\ -1 & \text{otherwise.} \end{cases}$ signum function |
| $t_j$ | loss value for the misclassification of $x_j$ with label $y_j$ |
| $w$ | vector $\in \mathbb{R}^m$ to describe a linear function on the data $x$ via a scalar product |
| $x$ | data sample $\in \mathbb{R}^m$ |
| $x_j$ | j-th sample of the training data $\in \mathbb{R}^m$ |
| $y_j$ | label of $x_j$ |
| $x_{ij}$ | two-dimensional data sample $x$ with $i$-th temporal and $j$-th spatial dimension (e.g., sensor) |

# List of Figures

# List of Tables